\documentclass{article}

\usepackage{acl}

\usepackage{times}
\usepackage{latexsym}
\usepackage{fancyvrb}

\usepackage[T1]{fontenc}

\usepackage[utf8]{inputenc}

\usepackage{microtype}

\usepackage{inconsolata}
\usepackage{url}            %
\usepackage{booktabs}       %
\usepackage{amsfonts}       %
\usepackage{nicefrac}       %
\usepackage{microtype}      %
\usepackage{xcolor}
\usepackage{xspace}
\usepackage{subcaption}
\usepackage{graphicx}
\usepackage{enumitem}
\usepackage{tcolorbox}      %
\usepackage{colortbl}
\usepackage{color}
\usepackage{tcolorbox}

\usepackage{placeins}

\usepackage{multirow}
\usepackage{rotating} %
\usepackage{makecell}
\usepackage{amssymb}
\usepackage{amsmath}
\usepackage{fontawesome5}
\usepackage{longtable} %
\usepackage{booktabs} %
\usepackage{arydshln}
\usepackage{wasysym} %

\usepackage{fdsymbol}

\makeatletter
\newcommand\footnoteref[1]{\protected@xdef\@thefnmark{\ref{#1}}\@footnotemark}
\makeatother

\definecolor{olmoDarkBlue}{HTML}{012e59}
\definecolor{olmoBlue}{HTML}{265ed4}
\definecolor{olmoLightBlue}{HTML}{012e59}
\definecolor{olmoTeal}{HTML}{00d5ff}
\definecolor{olmoYellow}{HTML}{ffbb00}
\definecolor{olmoOrange}{HTML}{ff9100}
\definecolor{VarnishBravoFive}{HTML}{223367}
\definecolor{VarnishBravoFour}{HTML}{1B4596}
\definecolor{VarnishBravoThree}{HTML}{265ED4}
\definecolor{VarnishBravoTwo}{HTML}{80BDFF}
\definecolor{VarnishBravoOne}{HTML}{F0F7FF}
\definecolor{VarnishAlfaFive}{HTML}{054976}
\definecolor{VarnishAlfaFour}{HTML}{01A2CA}
\definecolor{VarnishAlfaThree}{HTML}{4DE1FF}
\definecolor{VarnishAlfaTwo}{HTML}{B5F0FF}
\definecolor{VarnishAlfaOne}{HTML}{F2FCFF}
\definecolor{VarnishTangoFive}{HTML}{004752}
\definecolor{VarnishTangoFour}{HTML}{078E9E}
\definecolor{VarnishTangoThree}{HTML}{16C4CF}
\definecolor{VarnishTangoTwo}{HTML}{9AE7EC}
\definecolor{VarnishTangoOne}{HTML}{E6FDFE}
\definecolor{VarnishGolfFive}{HTML}{005340}
\definecolor{VarnishGolfFour}{HTML}{0A8F6B}
\definecolor{VarnishGolfThree}{HTML}{1EC28E}
\definecolor{VarnishGolfTwo}{HTML}{70DDBA}
\definecolor{VarnishGolfOne}{HTML}{C1F7E6}
\definecolor{VarnishOscarFive}{HTML}{DD6502}
\definecolor{VarnishOscarFour}{HTML}{FF9100}
\definecolor{VarnishOscarThree}{HTML}{FFBB00}
\definecolor{VarnishOscarTwo}{HTML}{FFD45D}
\definecolor{VarnishOscarOne}{HTML}{FFF1C4}
\definecolor{VarnishRomeoFive}{HTML}{932222}
\definecolor{VarnishRomeoFour}{HTML}{D63F3F}
\definecolor{VarnishRomeoThree}{HTML}{F7605F}
\definecolor{VarnishRomeoTwo}{HTML}{FF9F9E}
\definecolor{VarnishRomeoOne}{HTML}{FFE1E0}
\definecolor{VarnishFoxtrotFive}{HTML}{65295D}
\definecolor{VarnishFoxtrotFour}{HTML}{A44397}
\definecolor{VarnishFoxtrotThree}{HTML}{D864C9}
\definecolor{VarnishFoxtrotTwo}{HTML}{E7A2DE}
\definecolor{VarnishFoxtrotOne}{HTML}{FDF7FC}
\definecolor{VarnishPapaFive}{HTML}{271F55}
\definecolor{VarnishPapaFour}{HTML}{5C50A4}
\definecolor{VarnishPapaThree}{HTML}{7446F2}
\definecolor{VarnishPapaTwo}{HTML}{B7AFEB}
\definecolor{VarnishPapaOne}{HTML}{E6E3F7}
\definecolor{VarnishNovemberFive}{HTML}{47515C}
\definecolor{VarnishNovemberFour}{HTML}{616C7A}
\definecolor{VarnishNovemberThree}{HTML}{AEB7C4}
\definecolor{VarnishNovemberTwo}{HTML}{E8ECF2}
\definecolor{VarnishNovemberOne}{HTML}{F8F9FA}
\definecolor{VarnishLightCatRed}{HTML}{932222}
\definecolor{VarnishLightCatOrange}{HTML}{DD6502}
\definecolor{VarnishLightCatAqua}{HTML}{054976}
\definecolor{VarnishLightCatTeal}{HTML}{078E9E}
\definecolor{VarnishLightCatBlue}{HTML}{265ED4}
\definecolor{VarnishLightCatFuchsia}{HTML}{65295D}
\definecolor{VarnishLightCatPurple}{HTML}{5C50A4}
\definecolor{VarnishLightCatGreen}{HTML}{005340}
\definecolor{VarnishDarkCatRed}{HTML}{FF9F9E}
\definecolor{VarnishDarkCatOrange}{HTML}{FFD45D}
\definecolor{VarnishDarkCatAqua}{HTML}{B5F0FF}
\definecolor{VarnishDarkCatTeal}{HTML}{9AE7EC}
\definecolor{VarnishDarkCatBlue}{HTML}{80BDFF}
\definecolor{VarnishDarkCatFuchsia}{HTML}{E7A2DE}
\definecolor{VarnishDarkCatPurple}{HTML}{B7AFEB}
\definecolor{VarnishDarkCatGreen}{HTML}{70DDBA}

\newcommand{\dolma}{Dolma\xspace}
\newcommand{\olmo}{OLMo\xspace}
\newcommand*{\fnref}[1]{\textsuperscript{\ref{#1}}}

\newcommand{\dolmaWeb}{{\color{VarnishAlfaFour}\faGlobe}\xspace}
\newcommand{\dolmaRefs}{{\color{VarnishPapaThree}\faBookmark}\xspace}
\newcommand{\dolmaCode}{{\color{VarnishRomeoFour}\faCode}\xspace}
\newcommand{\dolmaBooks}{{\color{VarnishGolfThree}\faBook}\xspace}
\newcommand{\dolmaSocial}{{\color{VarnishOscarFour}\faComments}\xspace}
\newcommand{\dolmaPapers}{{\color{VarnishFoxtrotThree}\faGraduationCap}\xspace}

\newcommand{\toolkitWget}{{\color{VarnishNovemberThree}\faDownload}\xspace}
\newcommand{\toolkitFilter}{{\color{VarnishNovemberThree}\faFilter}\xspace}
\newcommand{\toolkitDedupe}{{\color{VarnishNovemberThree}\faCopy}\xspace}

\newcommand{\toolkitFilterText}{{\small\toolkitFilter}{\color{VarnishNovemberFour}\texttt{filtering}}\xspace}
\newcommand{\toolkitDedupeText}{{\small\toolkitDedupe\:}{\color{VarnishNovemberFour}\texttt{mixing}}\xspace}

\newcommand{\utf}{\texttt{UTF-8}\xspace}
\newcommand{\OlmoTiny}{\texttt{OLMo-1B}\xspace}

\newcommand{\huggingface}{\raisebox{-1.5pt}{\includegraphics[height=1.05em]{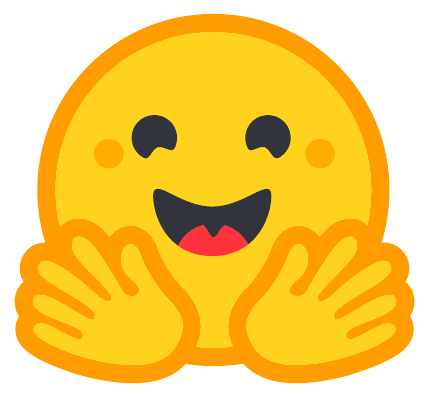}}\xspace}

\newcommand{\github}{\raisebox{-1.5pt}{\includegraphics[height=1.05em]{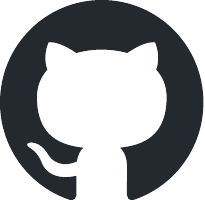}}\xspace}

\newcommand{\suppress}[1]{}

\newcommand\extrafootertext[1]{%
    \bgroup
    \renewcommand\thefootnote{\fnsymbol{footnote}}%
    \renewcommand\thempfootnote{\fnsymbol{mpfootnote}}%
    \footnotetext[0]{#1}%
    \egroup
}

\newtcolorbox{noticebox}[1][]{
  colback=white,       %
  colframe=black,      %
  fonttitle=\bfseries, %
  coltitle=black,      %
  title=Notice,        %
  #1                   %
}

\newtcolorbox{CalloutNormal}[1]{%
    colback=white,%
    colframe=olmoBlue,%
    coltitle=white,%
    center title,
    title={\vspace{.1em}\large\centering#1\vspace{.1em}},%
    width=0.8\linewidth
}

\newcommand{\dolmaLogoWithText}{\raisebox{-.3em}{\rlap{\raisebox{.3em}{\hspace{1.8em}\scriptsize Dolma}}\includegraphics[height=1.5em]{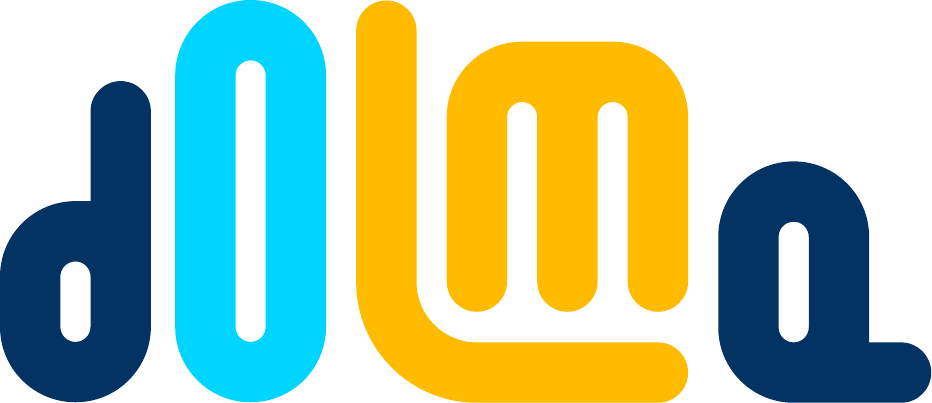}}\xspace}

\title{\dolmaLogoWithText: an Open Corpus of Three Trillion Tokens \\for Language Model Pretraining Research}

\author{%
  {\bf
    Luca Soldaini
    $^{{\color{olmoOrange}\boldsymbol{\varheartsuit}}\hspace{.1em}{\color{olmoBlue}\boldsymbol{\alpha}}}$
    \quad
    Rodney Kinney
    \hspace{-0.5em}
    $^{{\color{olmoOrange}\boldsymbol{\varheartsuit}}\hspace{.1em}{\color{olmoBlue}\boldsymbol{\alpha}}}$
    \quad
    Akshita Bhagia
    \hspace{-0.5em}
    $^{{\color{olmoOrange}\boldsymbol{\varheartsuit}}\hspace{.1em}{\color{olmoBlue}\boldsymbol{\alpha}}}$
    \quad
    Dustin Schwenk
    \hspace{-0.5em}
    $^{{\color{olmoOrange}\boldsymbol{\varheartsuit}}\hspace{.1em}{\color{olmoBlue}\boldsymbol{\alpha}}}$
    \vspace{8pt}
  } \\
  {\bf
    David Atkinson
    \hspace{-0.5em}
    $^{{\color{olmoBlue}\boldsymbol{\alpha}}}$
    \enskip
    Russell Authur
    \hspace{-0.5em}
    $^{{\color{olmoBlue}\boldsymbol{\alpha}}}$
    \enskip
    Ben Bogin
    \hspace{-0.5em}
    $^{{\color{olmoBlue}\boldsymbol{\alpha}\hspace{.1em}\boldsymbol{\omega}}}$
    \enskip 
    Khyathi Chandu
    \hspace{-0.5em}
    $^{{\color{olmoBlue}\boldsymbol{\alpha}}}$
    \vspace{1.5pt}
  } \\
  {\bf
    Jennifer Dumas
    \hspace{-0.5em}
    $^{{\color{olmoBlue}\boldsymbol{\alpha}}}$
    \enskip
    Yanai Elazar
    \hspace{-0.5em}
    $^{{\color{olmoBlue}\boldsymbol{\alpha}\hspace{.1em}\boldsymbol{\omega}}}$
    \enskip 
    Valentin Hofmann
    \hspace{-0.5em}
    $^{{\color{olmoBlue}\boldsymbol{\alpha}}}$
    \enskip
    Ananya Harsh Jha
    \hspace{-0.5em}
    $^{{\color{olmoBlue}\boldsymbol{\alpha}}}$
    \vspace{1.5pt}
    } \\
  {\bf
    Sachin Kumar
    \hspace{-0.5em}
    $^{{\color{olmoBlue}\boldsymbol{\alpha}}}$    
    \enskip
    Li Lucy
    \hspace{-0.5em}
    $^{{\color{olmoBlue}\boldsymbol{\beta}}}$
    \enskip
    Xinxi Lyu
    \hspace{-0.5em}
    $^{{\color{olmoBlue}\boldsymbol{\omega}}}$
    \enskip 
    Nathan Lambert
    \hspace{-0.5em}
    $^{{\color{olmoBlue}\boldsymbol{\alpha}}}$    
    \enskip 
    Ian Magnusson
    \hspace{-0.5em}
    $^{{\color{olmoBlue}\boldsymbol{\alpha}}}$    
    \vspace{1.5pt}
  } \\
  {\bf
    Jacob Morrison
    \hspace{-0.5em}
    $^{{\color{olmoBlue}\boldsymbol{\alpha}}}$    
    \enskip
    Niklas Muennighoff
    \hspace{-0.4em}
    \enskip
    Aakanksha Naik
    \hspace{-0.5em}
    $^{{\color{olmoBlue}\boldsymbol{\alpha}}}$
    \enskip
    Crystal Nam
    \hspace{-0.5em}
    $^{{\color{olmoBlue}\boldsymbol{\alpha}}}$
    \vspace{1.5pt}
  } \\
  {\bf
    Matthew E. Peters
    \hspace{-0.5em}
    $^{{\color{olmoBlue}\boldsymbol{\sigma}}}$
    \enskip
    Abhilasha Ravichander
    \hspace{-0.5em}
    $^{{\color{olmoBlue}\boldsymbol{\alpha}}}$
    \enskip
    Kyle Richardson
    \hspace{-0.5em}
    $^{{\color{olmoBlue}\boldsymbol{\alpha}}}$
    \enskip
    Zejiang Shen
    \hspace{-0.5em}
    $^{{\color{olmoBlue}\boldsymbol{\tau}}}$
    \vspace{1.5pt}
  } \\
{\bf
    Emma Strubell
    \hspace{-0.5em}
    $^{{\color{olmoBlue}\boldsymbol{\chi}\hspace{.1em}\boldsymbol{\alpha}}}$
    \enskip
    Nishant Subramani
    \hspace{-0.5em}
    $^{{\color{olmoBlue}\boldsymbol{\chi}\hspace{.1em}\boldsymbol{\alpha}}}$
    \enskip
    Oyvind Tafjord
    \hspace{-0.5em}
    $^{{\color{olmoBlue}\boldsymbol{\alpha}}}$
    \enskip
    Pete Walsh
    \hspace{-0.5em}
    $^{{\color{olmoBlue}\boldsymbol{\alpha}}}$
    \vspace{1.5pt}
  } \\
  {\bf
    Luke Zettlemoyer
    \hspace{-0.5em}
    $^{{\color{olmoBlue}\boldsymbol{\omega}}}$
     \enskip 
    Noah A. Smith
    \hspace{-0.5em}
    $^{{\color{olmoBlue}\boldsymbol{\alpha}\hspace{.1em}\boldsymbol{\omega}}}$
    \enskip
    Hannaneh Hajishirzi
    \hspace{-0.5em}
    $^{{\color{olmoBlue}\boldsymbol{\alpha}\hspace{.1em}\boldsymbol{\omega}}}$
    \vspace{1.5pt}
  } \\
  {\bf
    Iz Beltagy
    \hspace{-0.5em}
    $^{{\color{olmoBlue}\boldsymbol{\alpha}}}$
    \enskip
    Dirk Groeneveld
    \hspace{-0.5em}
    $^{{\color{olmoBlue}\boldsymbol{\alpha}}}$
    \enskip
    Jesse Dodge
    \hspace{-0.5em}
    $^{{\color{olmoBlue}\boldsymbol{\alpha}}}$
    \vspace{8pt}
  } \\
  {\bf
    Kyle Lo
    \hspace{-0.5em}
    $^{{\color{olmoOrange}\boldsymbol{\varheartsuit}}\hspace{.1em}{\color{olmoBlue}\boldsymbol{\alpha}}}$
    \vspace{8pt}
  } \\
  {
    $^\alpha$Allen Institute for AI \quad
    $^{\beta}$University of California, Berkeley \quad
    $^\chi$Carnegie Mellon University
  }\\
  {
    $^\sigma$Spiffy AI\quad
    $^\tau$Massachusetts Institute of Technology \quad
    $^\omega$University of Washington
  }\vspace{4pt}\\
  \texttt{\{lucas,kylel\}@allenai.org}
}

\begin{document}

\maketitle

\begin{abstract}
    Information about pretraining corpora used to train the current best-performing language models is seldom discussed:
    commercial models rarely detail their data, and
    even open models are often released without accompanying training data or recipes to reproduce them. 
    As a result, it is challenging to conduct and advance scientific research on language modeling, such as understanding how training data impacts model capabilities and limitations.
    To facilitate scientific research on language model pretraining, we curate and release \textbf{\dolma}, a three-trillion-token English corpus, built from a diverse mixture of web content, scientific papers, code, public-domain books, social media, and encyclopedic materials. 
    We extensively document
    \dolma, including its design principles, details about its construction, and a summary of its contents.
    We present analyses and experimental results on intermediate states of \dolma to share what we have learned about important data curation practices.
    Finally, we open-source our data curation toolkit
    to enable reproduction of our work as well as support further research in large-scale data curation.\footnotemark[1]

    \renewcommand{\arraystretch}{1.2}
    \begin{tabular}{rl}
         \huggingface & \href{https://huggingface.co/datasets/allenai/dolma}{\path{hf.co/datasets/allenai/dolma}}\\
         \github & \href{https://github.com/allenai/dolma}{\path{github.com/allenai/dolma}} \\
    \end{tabular}

\end{abstract}

\begin{table*}[t]
\centering
\small
\renewcommand{\arraystretch}{1.4}
\begin{tabular}{lc@{\hskip .3em}lcccc}
\toprule
\textbf{ Source } & \multicolumn{2}{c}{\textbf{ Doc Type }} & {\renewcommand{\arraystretch}{1}\begin{tabular}[c]{@{}c@{}}\textbf{\texttt{UTF-8}  bytes}\\\textit{(GB)}\vspace{.0em}\end{tabular}} & {\renewcommand{\arraystretch}{1}\begin{tabular}[c]{@{}c@{}}\textbf{ Documents }\\\textit{(millions) }\vspace{.0em}\end{tabular}} & {\renewcommand{\arraystretch}{1}\begin{tabular}[c]{@{}c@{}}\textbf{Unicode}\\\textbf{words}\\\textit{(billions) }\vspace{.0em}\end{tabular}} 
& {\renewcommand{\arraystretch}{1}\begin{tabular}[c]{@{}c@{}}\textbf{Llama}\\\textbf{tokens}\\\textit{(billions) }\vspace{.0em}\end{tabular}} \\
\midrule
{\renewcommand{\arraystretch}{1}\begin{tabular}[c]{@{}c@{}}Common Crawl\end{tabular}} & \dolmaWeb & web pages & 9,812 & 3,734 & 1,928 & 2,479 \\
{\renewcommand{\arraystretch}{1}\begin{tabular}[c]{@{}c@{}}GitHub\end{tabular}} & \dolmaCode & code & 1,043 & 210 & 260 & 411 \\
{\renewcommand{\arraystretch}{1}\begin{tabular}[c]{@{}c@{}}Reddit\end{tabular}} & \dolmaSocial & social media & 339 & 377 & 72 & 89 \\
{\renewcommand{\arraystretch}{1}\begin{tabular}[c]{@{}c@{}}Semantic Scholar\end{tabular}} & \dolmaPapers & papers & 268 & 38.8 & 50 & 70 \\
{\renewcommand{\arraystretch}{1}\begin{tabular}[c]{@{}c@{}}Project Gutenberg\end{tabular}} & \dolmaBooks & books & 20.4 & 0.056 & 4.0 & 6.0 \\
{\renewcommand{\arraystretch}{1}\begin{tabular}[c]{@{}c@{}}Wikipedia, Wikibooks\end{tabular}} & \dolmaRefs& encyclopedic & 16.2 & 6.2 & 3.7 & 4.3\\
\midrule
\multicolumn{3}{c}{\textbf{Total}} & \textbf{11,519} & \textbf{4,367} & \textbf{2,318} & \textbf{3,059} \\
\bottomrule
\end{tabular}
\vspace{1em}
\caption{
     The \dolma corpus at-a-glance. 
    It consists of three trillion tokens sampled from a diverse set of domains; sourced from approximately 200 TB of raw text before curation down to an 11 TB dataset.
    It has been extensively cleaned for language model pretraining use. Tokens calculated using the LLaMA tokenizer.
}
\label{tab:statistics}
\end{table*}

\extrafootertext{{\color{olmoOrange}$^\varheartsuit$}Core authors. See \autoref{sec:contrib} for list of contributions.}
\pagebreak
\footnotetext[1]{This manuscript was prepared for \textbf{\dolma \texttt{v.1.6}}. As our work on open data for language modeling continues, we will continue to improve \dolma. Updated versions can be found in the provided links.}

\section{Introduction}

Language models are now central to tackling myriad natural language processing tasks, including few-shot learning, summarization, question answering, and more.
Increasingly, the most powerful language models are built by a few organizations who withhold most model development details~\citep{claude-announce,OpenAI2023GPT4TR,Anil2023PaLM2T,Gemini-Team2023-gn}. 
In particular, the composition of language model pretraining data is often vaguely described, even in cases where the model itself is released for public use, such as Llama~2~\citep{Touvron2023Llama2O}.  
This hinders understanding of the effects of pretraining corpus composition on model capabilities and limitations, with impacts on scientific progress as well as on the public who interfaces with these models. 
Our aim is to increase participation in scientific research of language models through open corpora:
\begin{itemize}[leftmargin=1em]
    \item  Data transparency helps developers and users of \textbf{applications} that rely on language models to make more informed decisions~\citep{gebru2021datasheets}.  
    For example, models have shown to perform better on tasks that are more similar to their pretraining data~\citep{razeghi-etal-2022-impact,pmlr-v202-kandpal23a},
    or social biases in models' pretraining data may necessitate additional consideration when using them~\citep{feng-etal-2023-pretraining,Navigli2023Biases,seshadri2023bias}.
    \item Open pretraining data is necessary to \textbf{analyze} 
    how its composition influences model behavior, allowing those training models to interrogate and improve current data practices~\citep{Longpre2023APG,Gao2021EmpircalEQ,wimbd}. 
    Examples of this research include memorization~\citep{Carlini2022-mj,Chang2023SpeakMA}, deduplication~\citep{lee-etal-2022-deduplicating}, adversarial attacks~\citep{wallace-etal-2021-concealed}, benchmark contamination~\citep{magar-schwartz-2022-data}, and training data attribution~\citep{Hammoudeh2022TrainingDI,Grosse2023StudyingLL}.
\end{itemize}

To support broader participation and inquiry in these lines of research, we present \textbf{D}ata for \textbf{O}pen \textbf{L}anguage \textbf{M}odels' \textbf{A}ppetite (\dolma), an open corpus of three trillion tokens designed to support language model pretraining research. 
We source much of our data from sources similar to those present in past work, including a mix of web text from Common Crawl, scientific research from Semantic Scholar, code from GitHub, public domain books, social media posts from Reddit, and encyclopedic materials from Wikipedia.
Compared to other publicly-available pretraining corpora, \dolma offers a larger pool of tokens at comparable quality while maintaining diverse data composition. 
In summary, our contributions are two-fold:

\begin{itemize}[leftmargin=1em,topsep=1mm,itemsep=1mm]
    \item We release the \textbf{\dolma Corpus}, a diverse, \textbf{multi-source} collection of 3T tokens\footnote{We follow the definition of ``token'' as a subword obtained using a tokenizer (such as LLaMA’s or GPT-NeoX’s), which is distinct from ``word'', as in a unit of text as defined by the \href{https://unicode.org/reports/tr29/}{Unicode text segmentation standard}.} across over 4B documents acquired from 6 different data sources that are (\textit{i}) commonly seen in large-scale language model pretraining and (\textit{ii}) made accessible to the general public.  
    Table~\ref{tab:statistics} provides a high-level overview of the amount of data from each source.
    \item We open source the \textbf{\dolma Toolkit}, a high-performance, portable tool designed to efficiently curate large datasets for language model pretraining. 
    Through this toolkit, practitioners can not only reproduce our dataset, but also study and improve data curation practices. 
\end{itemize}

\section{Related Work}
\label{sec:related-work}

\paragraph{Closed data curation practices in language model pretraining research.} 
Pretraining data practices for language model research have grown increasingly closed, both with respect to \textbf{access} to data as well as \textbf{documentation} of key details about the data itself or its curation practices that would enable reproduction efforts or further scientific study.
Proprietary models (e.g., GPT-4, \citealp{OpenAI2023GPT4TR}; PaLM 2, \citealp{Anil2023PaLM2T}; Claude, \citealp{claude-announce}) disclose little to no information (not even corpus size, or data provenance), and do not share data artifacts.
Despite increasing access to powerful open models, few are released alongside their training data; exceptions include T5 on C4~\citep{raffel2020exploring}, BLOOM~\citep{leong-etal-2022-bloom} on ROOTS~\citep{piktus-etal-2023-roots}, GPT-J~\citep{gpt-j}, GPT-NeoX~\citep{Black2022GPTNeoX20BAO}, Pythia~\citep{Biderman2023PythiaAS} on Pile~\citep{Gao2020ThePA}, and INCITE~\citep{together2023incite} on RedPajama v1~\citep{together2023redpajama}.
The most powerful open models (e.g., Llama 2~\citep{Touvron2023Llama2O}, Mistral~\citep{mistral}, Yi~\citep{qwen}, Qwen~\citep{yi34b}) do not share their data nor provide sufficient details for reproduction.
Among large-scale language model pretraining efforts, the ones accompanied with transparent data curation documentation include LLaMA~\citep{Touvron2023LLaMAOA} (\emph{released model, unreleased data}), Gopher~\citep{Rae2021ScalingLM} (\emph{unreleased model and data}), and Falcon~\citep{falcon40b} (\emph{released model, released partial data}).
Appendix~\S\ref{sec:pretrain-data-llms} further illustrates the many unknowns of data curation practices of open and closed models, as well as recent trends away from open data practices that have motivated our work.

\paragraph{Open corpora for language model pretraining.} 
We recognize prior efforts to curate, document, and release open corpora to support language model pretraining research.
However, limitations in these prior corpora have motivated us to curate a new dataset:
\begin{itemize}[leftmargin=1em,topsep=1mm,itemsep=1mm]
    \item C4~\citep{raffel2020exploring} (175B tokens) and Pile~\citep{Gao2020ThePA} (387B tokens) are high-quality datasets with demonstrated use in training language models, but are unfortunately \textbf{limited in scale}. 
    ROOTS~\citep{piktus-etal-2023-roots} is large ($\approx$400B tokens) but given its multilingual focus, its English-only portion is only 30\% of the dataset and thus contributes too few tokens to train English-only models. We recognize that scale and English-only concentration do not imply a ``higher-quality'' dataset; rather, certain threads of research necessitate these foci, motivating our new corpus (see \autoref{sec:desiderata}).
    \item While Falcon~\citep{falcon40b} (580B tokens) and RedPajama v2~\citep{redpajama2} (30T tokens) meet our scale criterion, they are entirely derived from Common Crawl web pages, and thus \textbf{lack source diversity} commonly targeted when curating data for the largest language models (e.g., scientific papers, code). We also note that RedPajama v2 is only \textbf{lightly-curated}, distributing content output by CCNet~\citep{wenzek-etal-2020-ccnet} mostly as-is,
    thus placing the onus on model developers to decide their own filtering before training.
    \item RedPajama v1~\citep{together2023redpajama} ($\approx$1.2T tokens) is most similar to our effort and a source of inspiration when designing \dolma. While RedPajama v1 was a \textbf{specific} reproduction of the LLaMA~\citep{Touvron2023LLaMAOA} training data, we have a \textbf{broader} reproduction target which required diving into data sources that RedPajama v1 did not pursue, including larger collections of scientific papers and social media forums like Reddit (see \autoref{sec:desiderata}).
    Further, recent work has identified data quality issues suggesting significant additional cleanup of RedPajama v1 is recommended before costly language model training~\citep{cerebras2023slimpajama,wimbd}.
\end{itemize}

{\noindent While this manuscript was under review, several other open corpora for language modeling have been released, including FineWeb~\citep{fineweb}, Zyda~\citep{tokpanov2024zyda}, and the datasets used to train LLM360 Amber~\citep{liu2023llm360}, LLM360 K2~\citep{llm360K2}, and MAP-Neo~\citep{zhang2024mapneo} models.}

\section{Data Design Goals}
\label{sec:desiderata}

We present the design goals of \dolma and discuss how these goals guided our decision-making during data curation.
In sharing these, we hope to inform users of \dolma's strengths and limitations while also reinforcing practice around such disclosures in dataset curation research (see curation rationales in \citet{bender-friedman-2018-data} and motivation questions in \citet{gebru2021datasheets}).

\paragraph{Be consistent with prior language model pretraining recipes.}
By matching data sources and methods used to create other language modeling corpora, to the extent they are known, we enable the broader research community to use our artifacts to 
study (and scrutinize) language models being developed today, even those developed behind closed doors.
In this \textbf{reproduction} effort, we follow established practices 
to the extent they are known.
Notably, this also means scoping \dolma to \textbf{English-only} text to better leverage known curation practices and maximize generalizability of scientific work on \dolma to existing language models.\footnote{Recognizing that this focus reinforces the assumption of English as the ``default'' language, we hope to expand \dolma to more languages in the future. We release our data curation tools to support such efforts.
}

\paragraph{When in doubt, make evidence-backed decisions.} Still, there remain myriad data curation decisions for which there is no single clear recipe from prior work, both when best practice isn't known as well as when implementations differ in subtle ways.
In such cases, we prioritize decisions that \textbf{maximize performance} of language models trained on \dolma over a diverse suite of tasks and datasets (see \S\ref{sec:experimental-methodology}).

\paragraph{Large scale data to train large models.} 
\citet{Hoffmann2022TrainingCL}  suggested that one can train compute-optimal models by maintaining a fixed ratio between language model size (in parameters) and a minimum number of training tokens. Recent works that follow these ``scaling laws,'' such as Llama 2, show that there is still room for performance improvement by increasing the number of training tokens.
We aim for a sufficiently large corpus---\textbf{2--3T tokens}---to allow further study of the relationship between model and dataset size.

\paragraph{Make necessary adjustments to preserve openness.}
A core tenet of our work is openness, which we define to mean (\textit{i}) \textbf{sharing the data itself} and (\textit{ii}) \textbf{documenting the process to curate it}. 
This requirement means we occasionally must deviate from known recipes due to additional practical, legal or ethical considerations that arise when pursuing dataset research in the open.
For example, despite their use in training language models like LLaMA, we avoid sources like Books3~\citep{Gao2020ThePA} which are the center of ongoing legal cases around AI use of copyrighted materials~\citep{books3copyright}.
Similarly, despite the lack of discussion around the removal of personally identifiable information in prior recipes, we perform this filtering to mitigate risks associated with data release~\citep{subramani-etal-2023-detecting}.

\section{Data Curation Methodology}
\label{sec:creation}

\subsection{The \dolma Toolkit}
\label{sec:toolkit}
Pretraining data curation requires defining complex pipelines that transform raw data from multiple sources into a single collection of cleaned, plain text documents~\citep{wenzek-etal-2020-ccnet,falcon40b}.
To curate \dolma, we create and open-source a high-performance toolkit to facilitate efficient processing on hundreds of terabytes of text content.
Our toolkit unifies common dataset curation steps into ``filtering'' and ``mixing'' operations: 

\paragraph{\toolkitFilterText} We unify common data transformations like language, quality or content filters into a single implementation. 
Given a configuration---a text unit (e.g., document, paragraph,\footnote{
    We define a paragraph to be a span of text ending in a newline \utf character \texttt{``\textbackslash{}n''}.
} sentence, etc.), a scoring method (e.g., linear classifier, language model perplexity, regular expression matches), and a removal policy (e.g., delete, replace with string)---our toolkit parallelizes filtering operations by identifying and removing undesirable text at massive scale.
For \dolma, we use these to filter non-English, ``low quality'' or unnatural,\footnote{
    The term ``quality filter,'' while widely used in literature, does not appropriately describe the outcome of filtering a dataset. Quality might be perceived as a comment on the informativeness, comprehensiveness, or other characteristics valued by humans. However, the filters used in \dolma and other language models efforts select text according to criteria that are inherently ideological~\citep{gururangan-etal-2022-whose}.
}
toxicity,\footnote{
    Similar to ``quality'', there is no single definition for ``toxicity''.
    Rather, specific definitions vary depending on task~\citep{Vidgen2020-bh} and dataset curators' social identities~\citep{Santy2023-sh};
    annotators' beliefs also influence toxic language detection~\citep{Sap2021-ne}.
    Predicting toxicity remains challenging~\citep{welbl-etal-2021-challenges-detoxifying,markov-2023-holistic-approach-undesired-content-detection}, especially as existing methods have been shown to discriminate against minoritized groups~\citep{xu-etal-2021-detoxifying}.
}
and PII at the document and sub-document levels.
In internal tests to replicate C4 recipe, our toolkit performed filtering at a rate of 122 CPU hours per TB; for reference, processing the full ``raw'' \dolma files totaling 200 TB on a \texttt{c6a.48xlarge} instance with 192 vCPUs would take 5 days.

\paragraph{\toolkitDedupeText} We unify common cross-file operations, like up/down-sampling, deduplication and decontamination, into a single Rust module that ``mixes'' content across files into a smaller set of files.
For example, we can achieve up-sampling by repeatedly reading the same file paths when mixing.
We also implement a Bloom filter~\citep{Bloom1970} compatible with our mixer which enables linear-time probabilistic detection of duplicates.
We can repurpose this for test set decontamination by first seeding the Bloom filter with test examples, then flagging any detected duplicates when mixing the pretraining data.

\subsection{Data Ablations}
\label{sec:experimental-methodology}

To help us make informed decisions, we conduct \textbf{data ablations} 
in which we train language models on a dataset following a specific data curation decision, or \emph{intervention}, and evaluate the resulting model's performance on a range of test datasets against a \emph{baseline} dataset.
By comparing intervention and baseline results while controlling for model architecture and training, we can isolate the impact of specific dataset curation decisions on downstream models.

\paragraph{Model training.}
We conduct data ablations using a 1.2 billion parameter decoder-only model from the OLMo family of open language models~\citep{olmo20247b}.
This is in line with similar model sizes that have been used for ablations in prior work~\citep{le-scao-etal-2022-language}.
As training such models to completion is prohibitively expensive, especially when one must perform these experiments for each significant data curation decision, we only train these models up to 150 billion tokens before terminating them early.
Further details of our training setup in Appendix~\ref{sec:setup:abl}.

\paragraph{Tasks and test datasets.} 
To select our evaluation tasks and datasets, we prioritize those that \emph{(i)} have been used in prior language model pretraining evaluation, \emph{(ii)} capture a diverse range of language model knowledge and capabilities, and \emph{(iii)} for which we can avoid test set contamination~\citep{dodge-etal-2021-documenting,Yang2023RethinkingBA}.
We arrive at \textbf{8 datasets} in our evaluation suite (full details in Appendix~\S\ref{sec:setup}) that have been used in prior language modeling research (e.g., LLaMA, Llama 2, etc.) and capture a range of capabilities (e.g., question answering, commonsense reasoning, etc.).  
Full test set contamination analysis validating our dataset choices in Appendix~\S\ref{appendix:sec:decontam-dolma-perplexity-paloma}.

\paragraph{Evaluation.} We perform evaluation of our data ablation models using zero-shot in-context prompting, casting every task as (ranked) text classification, following in-context prompt truncation from \citet{min-etal-2022-metaicl}, prompts from PromptSource~\citep{bach2022promptsource}, and using an in-house evaluation harness similar to the Eleuther harness~\citep{eval-harness}.

\section{\dolmaWeb~Curating Dolma-Web}
\label{sec:common-crawl}

In this section, we describe the web subset of \dolma, which  consists of 2.28T tokens derived from \textbf{Common Crawl},\footnote{\href{https://commoncrawl.org}{\path{commoncrawl.org}}}
a collection of over 250 billion pages that were crawled since 2007.
Common Crawl is organized in snapshots, each corresponding to a full crawl over its seed URLs;
as of Feb 2024, there are 97 snapshots.
We used 25 snapshots between \texttt{2020-05} to \texttt{2023-06}.\footnote{To minimize storage and compute costs, we only acquired enough shards of Common Crawl to meet our target 2-3T token corpus size, assuming at least a 10x reduction from the sum of all data cleaning efforts, including CCNet (\S\ref{sec:desiderata}).}

\subsection{\toolkitWget~Acquisition \& \toolkitFilter~Language Filtering}
\label{sec:web:acquisition}

Our web pipeline leverages CCNet~\citep{wenzek-etal-2020-ccnet} to perform language filtering and initial content deduplication.
CCNet has been used to develop other language model datasets like that for LLaMA, RedPajama v1, RedPajama v2.
CCNet processes each web page with a FastText~\citep{joulin2016fasttext} language ID model\footnote{\href{https://fasttext.cc/docs/en/language-identification.html}{\path{fasttext.cc/docs/en/language-identification}}} to determine the primary language for each document; 
we keep all pages with English document score greater than or equal to 0.5 (removed 61.7\% of the data, by byte size). 
Further, CCNet identifies and removes very common paragraphs by grouping shards in each snapshot into small sets and removing duplicated paragraphs in each.
This step removed approximately 70\% of paragraphs, primarily consisting of headers and navigation elements.
Overall, CCNet pipeline filters out 84.2\% of the content in Common Crawl, from 175.1 TB to 27.7 TB. 
More details are provided in our Datasheet~\S\ref{sec:datasheet}.

\subsection{\toolkitFilter~Quality Filtering} 
\label{sub:web:quality}

Web crawled data requires significant cleanup before language model training; undesirable content ranges from artifacts introduced by HTML to plain text conversion (\textit{e.g.}, page headers, ill-formatted text) to pages lacking ``prose-like'' content (\textit{e.g.}, boilerplate text, short segments). 
Per arguments posed in \citet{Rae2021ScalingLM} and \citet{falcon40b} against model-based quality filters, we approach quality filtering by combining heuristics introduced by Gopher and C4.
Specifically, we keep all the Gopher rules (\texttt{Gopher All}) and keep a single heuristic from C4 designed to remove paragraphs that do not end in punctuation (\texttt{C4 NoPunc}), as opposed to adopting the full set of C4 rules (\texttt{C4 All}).
Implementation details of all filtering rules are provided in our Datasheet~\S\ref{sec:datasheet}.

\begin{figure}[h!]
    \centering
    \includegraphics[width=\linewidth]{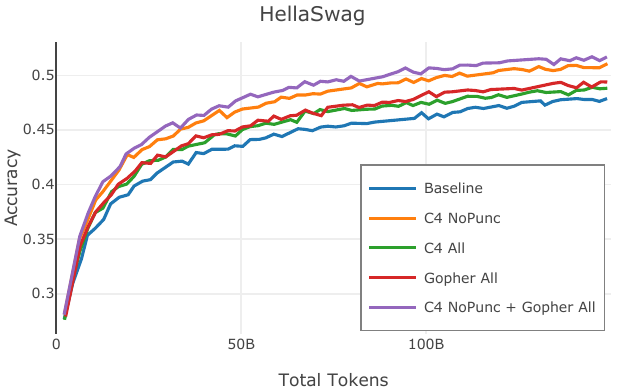}
    \caption{We find a positive effect of web data quality filters on 1.2B model performance, evaluated across training iterations, over a no-filtering baseline.
    We only show results on HellaSwag here; all figures for other evaluation datasets are in the Appendix~\S\ref{app:raw}.
    }
    \label{fig:abl_quality_only}
\end{figure}

Ablation results shown in \S\ref{fig:abl_quality_only} validate our filtering strategy: 
we find that \texttt{C4 NoPunc} on its own outperforms both \texttt{C4 All} as well as \texttt{Gopher All} on both perplexity and downstream tasks. 
Finally, combining \texttt{Gopher All} + \texttt{C4 NoPunc} offers the best performance.
In all, \texttt{Gopher All} tagged 15.23\% of \utf characters for removal, while \texttt{C4 NoPunc} tagged 22.73\% of characters for removal. 

\paragraph{Model and heuristic filters are orthogonal.} CCNet also provides quality scores using KenLM~\citep{heafield-2011-kenlm} perplexity that groups documents based on Wikipedia-likeness; these buckets are often interpreted as high (21.9\%), medium (28.5\%), or low (49.6\%) quality content, in which more Wikipedia-like is often associated with higher quality. To our surprise, we found our heuristic filtering rules did not affect these proportions, suggesting that such model-based quality filters may capture other signals orthogonal to heuristic filters.

\subsection{\toolkitFilter~Content Filtering} 
\label{sec:toxicity}
\label{sub:web:content}

\paragraph{Filtering Toxic Content}

Data sampled from the web often contains harmful or toxic content~\citep{Matic2020IdentifyingSU,luccioni-viviano-2021-whats,Birhane2023IntoTL,Birhane2023OnHS}.
Such content is often filtered to minimize the likelihood that downstream language models are prone to toxic content generation~\citep{Anil2023PaLM2T,Rae2021ScalingLM,Thoppilan2022-fn,Hoffmann2022TrainingCL,Longpre2023APG}.
To remove this content from \dolma, we train our own FastText classifiers on the Jigsaw Toxic Comments~\citep{jigsaw} dataset, producing two models that identify ``\texttt{hate}'' and ``\texttt{NSFW}'' content, respectively.
See Appendix~\S\ref{app:toxic} for implementation details.
We run these classifiers on Common Crawl sentences\footnote{Using BlingFire sentence splitter~\citep{blingfire}.} 
and remove any sentence scored \emph{above} a set threshold.

\begin{figure}[h!]
    \centering
    \includegraphics[width=\linewidth]{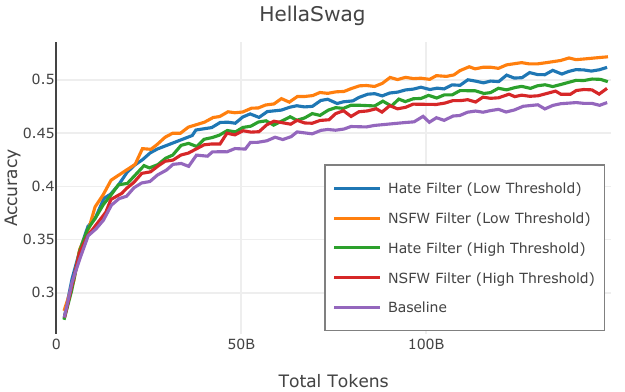}
    \caption{
    We find a positive effect of web data content filters on 1.2B model performance, evaluated across training iterations, over a no-filtering baseline.
    We only show results on HellaSwag here; all figures for other evaluation  datasets are in the Appendix~\S\ref{app:raw}.
    }
    \label{fig:abl_toxic}
\end{figure}

To understand filter thresholding effects on \dolma, we conduct a data ablation choosing two very different thresholds for these content filters (\S\ref{fig:abl_toxic}).
We find the ``High Threshold'' ($\tau=0.4$) removes \textit{less} content (5.5--7.3\%), but generally yields
lower downstream performance than the ``Low Threshold'' ($\tau=0.0004$) which removes \emph{more} content (29.1--34.9\%).\footnote{
    Manual inspection of the distribution of sentence scores revealed a bi-modal distribution with peaks near 0.0 and 1.0 (e.g., Figure~\ref{fig:reddit-location-toxicity}).
    As such, we chose ``Low'' to remove even slightly toxic data ($> 0.0$), and ``High'' to limit our max data removal amount to preserve our target dataset scale.
}

Weighing the tradeoff between dataset scale (``High'') and performance maximization (``Low''), we adopt the more permissive ``High'' threshold to ensure we meet our minimum token count requirement.
The cause of this was surprising: Our quality, content, and deduplication filters overlap very little in which texts they remove (Figure~\ref{fig:corr}), resulting in a compounded filtering effect when combining them.
In future versions of \dolma, we will start with more shards of Common Crawl and adopt stricter filter thresholds.

\paragraph{Filtering Personally Identifiable Information}

Data sampled from the web can also leak personally identifiable information (PII) of users~\citep{luccioni-viviano-2021-whats,subramani-etal-2023-detecting}.
Traces of PII are abundant in large-scale datasets~\citep{wimbd}, and 
language models have also been shown to reproduce PII at inference time~\citep{Carlini2022-mj,Chen2023-zm}. 
\dolma's size makes it impractical to use model-based PII detectors like Presidio~\citep{presidio};
instead, we rely on carefully-crafted regular expressions that sacrifice some accuracy for significant speed-up.
Following \citet{subramani-etal-2023-detecting}, we focus on three kinds of PII that are detectable with high precision: 
email addresses, IP addresses and phone numbers.
For documents with 5 or fewer PII spans, we replace the span with a special token (e.g., \texttt{|\negthickspace|\negthickspace|EMAIL\_ADDRESS|\negthickspace|\negthickspace|}); this affects 0.02\% of documents.
Otherwise, we remove entire documents with higher density of PII spans; this affects 0.001\% of documents.
In data ablation experiments, we find that execution details around PII (e.g., removal versus special token replacement) had no effect on model performance, which is expected given the tiny percentage of affected data.
See Appendix~\S\ref{app:pii-filter-details} for implementation details; all figures for results on evaluation suite are in the Appendix~\S\ref{app:raw}.

\subsection{\toolkitDedupe~Deduplication} 
\label{sub:web:dedup}

Deduplication of pretraining data has been shown to be effective for improving token efficiency during model training~\citep{lee-etal-2022-deduplicating,Abbas2023SemDeDupDL,Tirumala2023D4IL};
as such, it has become common practice among pretraining data recipes.
In \dolma, we perform three stages of deduplication:
\begin{enumerate}[label=(\roman*),leftmargin=2.0em,itemsep=1mm,topsep=1mm]
    \item\label{item:dup:url}\textbf{Exact \textsc{url} dedup}
    filters 53.2\% of documents. 
    \item\label{item:dup:doc}\textbf{Exact document dedup}  
    filters 14.9\% of \textsc{url}-deduped documents, including empty documents.
    \item\label{item:dup:par}\textbf{Exact paragraph dedup} 
    filters 18.7\% of paragraphs from the \textsc{url}-deduped documents, including empty paragraphs.
\end{enumerate}

This multi-stage approach is designed to increase efficiency: 
Stage~\ref{item:dup:url} is commonly used first thanks to its computational efficiency~\citep{urldedupe1,urldedupe2,Penedo2023TheRD}.
Stages~\ref{item:dup:url}~and~\ref{item:dup:doc} are designed to remove copies of the same item, such as re-crawls of the same URL and identical pages with multiple URLs (e.g., same news article in multiple online newspapers). 
Performing these early before any content or quality filtering greatly reduces the number of documents to process.
In contrast, Stage~\ref{item:dup:par} removes common boilerplate content (e.g., the byline under all articles by the same author);
as paragraph removal risks disrupting content analysis, we perform it last.
We perform all three stages using the Bloom filter in \S\ref{sec:toolkit}.

\subsection{{\small\toolkitWget\toolkitFilter\toolkitDedupe}\enskip Putting It All Together}

To summarize, the \dolma web pipeline transforms the output of CCNet through URL and document-level deduplication, then quality and content filtering, and finally paragraph-level deduplication.

\begin{figure}[h!]
    \centering
    \includegraphics[width=\linewidth]{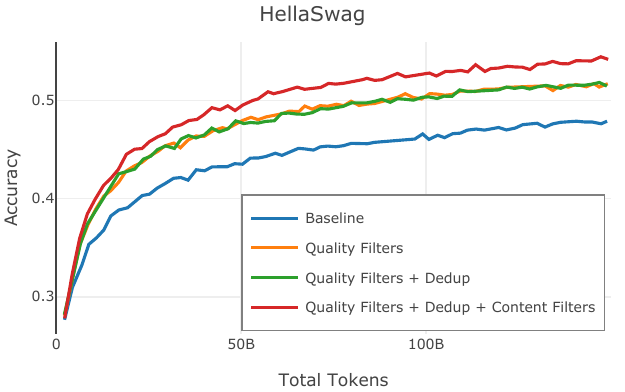}
    \caption{We find a positive compounding effect on 1.2B model performance, evaluated across training iterations, 
    when stacking quality filtering, content filtering and paragraph-level deduplication, over a no-filtering baseline.
    We show results on HellaSwag here; all figures for other evaluation datasets are in the Appendix~\S\ref{app:raw}.
    }
    \label{fig:abl_web_all}
\end{figure}

We show the positive compounding effect of all stages of our web pipeline on downstream model performance, as assessed through our data ablations \S\ref{sec:experimental-methodology}.
We present summary statistics in Appendix~\S\ref{sec:dolma-data-distribution-wimbd}.

\section{\dolmaCode~Curating Dolma-Code}
\label{sec:curating-dolma-code}

In this section, we describe the code subset of \dolma, which consists of 411B tokens derived from \textbf{GitHub}.

\subsection{\toolkitWget~Acquisition \& \toolkitFilter~Language Filtering}

Like prior work in code language models (e.g., StarCoder~\citep{Li2023StarCoderMT}), we also acquire code through the Stack~\citep{kocetkov2022stack}, a deduplicated but otherwise unfiltered collection of permissively-licensed GitHub repositories.
The raw version of this dataset was collected in March 2023.
We filter data-heavy 
files with extensions such as \texttt{JSON} and \texttt{CSV}.

\subsection{\toolkitFilter~Quality Filtering} 

We apply heuristics derived from the code subset of RedPajama v1 and StarCoder. 
RedPajama v1 uses rules to remove repetitive file preambles, such as license statements
and documents with excessively long lines or mostly numerical content. 
Overall, RedPajama v1 is removes files that are mostly data or generated through templates.
To select high-quality code snippets, we also use rules from the StarCoder pipeline;
these heuristics filter GitHub repositories with no to few stars, files with too few or too many comments, and HTML files with low code-to-text ratio.
Implementation details of all filtering rules are provided in our Datasheet~\S\ref{sec:datasheet}.

When conducting data ablations, we find that, compared to RedPajama v1 rules alone, RedPajama v1 and StarCoder rules combined lead to lower perplexity on code datasets (\textit{e.g.}, HumanEval;~\citealp{chen2021evaluating}) and
improved performance on datasets in our evaluation suite.\footnote{All figures for results on evaluation suite in Appendix~\S\ref{app:raw}.
}
Therefore, we chose to use this combination of the two filtering rules for this \dolma code subset.

\subsection{\toolkitFilter~Content Filtering} 

We apply the same heuristics to filter and mask PII used in the web subset (\S\ref{sec:common-crawl}). 
Additionally, we filter any documents containing code secrets and software-specific personal information by running the \texttt{detect-secrets} library~\citep{DetectSecrets} and removing any documents with a match.

\subsection{\toolkitDedupe~Deduplication} 

We started from the already-deduplicated version of the Stack, which used the pipeline first introduced by~\citet{Allal2023SantaCoder}, which uses MinHash~\citep{Broder2002-ra} and Locally Sensitive Hashing to find similar documents.

\section{\dolmaSocial~Curating Dolma-Social}
\label{sec:curating-dolma-social}

\label{sec:pushshift-reddit-forum}

In this section, we describe the social media subset of \dolma,  which consists of 80B tokens derived from \textbf{Reddit} data.

\subsection{\toolkitWget~Acquisition \& \toolkitFilter~Language Filtering}

We derive this subset from 378M posts from December 2005 until March 2023 obtained through  Pushshift \citep{pushshift}.
We include both \textit{submissions}---initial message in conversations on Reddit---and \textit{comments}---replies to messages.
The tree-like structure of Reddit threads allows for multiple possible data formats depending on how the various components of a thread are linearized for language model pretraining.
To better inform this transformation, we conduct a data ablation over several approaches:

\begin{enumerate}
    \item \textbf{Atomic Content}. Treats all comments and submissionas independent documents.
    \item \textbf{Partial Threads}. Comments from the same thread combined into a multi-round dialogue between users. Submissions as separate documents. 
    \item \textbf{Full Threads}. Combines submissions with all child comments into one document.
\end{enumerate}

\begin{figure}[h!]
    \centering
    \includegraphics[width=\linewidth]{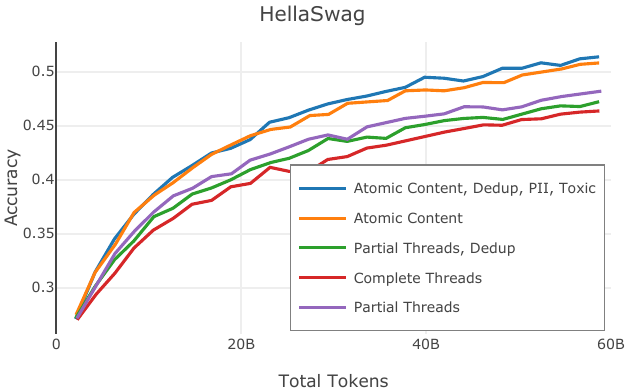}
    \caption{
    Experimenting with different Reddit thread linearization methods with 1.2B models, evaluated across training iterations.
    We only show results on HellaSwag here;  all figures for other evaluation  datasets are in the Appendix~\S\ref{app:raw}.
    }
    \label{fig:abl_reddit}
\end{figure}

See Appendix~\S\ref{sec:how-to-thread} for implementation details.
From results in Figure~\ref{fig:abl_reddit}, we see treating submissions and comments as independent documents (Atomic Content) leads to better performance on our evaluation suite.
We hypothesize that artificial formatting introduced when combining thread elements negatively impacts language model training; we leave further investigation to future work.
Finally, we filter non-English content using the approach from \S\ref{sec:web:acquisition}.

\subsection{\toolkitFilter~Quality Filtering} 
\label{sub:reddit:quality}

Like web crawled data, social media posts also require significant cleanup before language model training. 
We repurpose the pipeline introduced by~\citet{Henderson2019} to filter submissions and comments.
We remove comments shorter than 500 characters, and submissions shorter than 400 characters.\footnote{Qualitative inspection of the data suggested that submissions are of higher quality than comments; thus, we use a more permissive minimum length.}
We also remove documents over 40,000 characters. 

We remove comments with fewer than 3 votes\footnote{The total votes for each document are obtained by computing the difference between positive votes, also known as ``upvotes'', negative votes or ``downvotes''.}, as lower scores are more likely for comments that are deeply nested in a conversational thread~\citep{Weninger2013-kz} or content that is more likely to results in emotionally-charged discourse~\citep{Davis2021-rg}.
Votes have been used as a signal in constructing the WebText~\citep{Radford2019-vq} and OpenWebText~\citep{PetersonUnknown-dv} corpora.
We discard documents that have been deleted by their authors, removed by moderators, or labeled by their authors as \textit{``over 18''}.
We exclude any document originated from a set 26,123 banned or NSFW subreddits.\footnote{Available on GitHub as part of \dolma Toolkit (see \href{https://github.com/allenai/dolma/blob/a9eaf294b677af4e724dfc39014014fd3af0627f/sources/reddit/atomic_content_v5/subreddit_blocklist.txt}{\path{subreddit_blocklist.txt}})
. The list was curated by merging several sources that tracked banned subreddits. We also include any subreddit with over 10\% of posts tagged as NSFW.}

\subsection{\toolkitFilter~Content Filtering} 

We apply the same content filtering in \S\ref{sec:toxicity}, except due to the short length of many Reddit documents, instead of masking PII, we fully remove the document.

\subsection{\toolkitDedupe~Deduplication}

We employ the same strategy used in the web pipeline (\S\ref{sub:web:dedup}). 
Since submissions and comments are shorter than web documents, we only deduplicate at a document-level. 
This strategy is useful to reduce the incidence of ``\textit{copypasta}'' (identical text repeated across comments and subreddits for comedic effect) and other repetitive information.

\section{Assembling Other Data Sources}
\label{sec:other-data-sources}

In this section, we briefly summarize additional high-quality sources that were used to derive \dolma. 
More details on collection and processing in Datasheet~\S\ref{sec:datasheet}.

\paragraph{\dolmaWeb~C4 for Curated Web Content}

Similar to data recipes for LLaMA and Llama 2,  we supplement our web subset with C4~\citep{raffel2020exploring}.
We further refine this data by reprocessing it through our full web pipeline (excluding URL deduplication) (\S\ref{sec:common-crawl}) which removed additional content, including more low-quality and duplicated texts, and performed PII masking.

\paragraph{\dolmaPapers~Semantic Scholar for Academic Literature} 

The peS2o dataset~\citep{peS2o} is a collection of approximately 40 million open-access academic papers that have been cleaned, filtered, deduplicated, and formatted for pretraining language models. 
It is derived from the Semantic Scholar Open Research Corpus (S2ORC;~\citealp{lo-etal-2020-s2orc}).
As this dataset has been created for language modeling purposes, we use it as-is. 

\paragraph{\dolmaBooks~Project Gutenberg for Books} 
Project Gutenberg is a repository of over 70 thousand public domain books. 
We collected Project Gutenberg's archive in April 2023. 
We use English language books, which we filter using the same approach described in \S\ref{sec:web:acquisition}.
We deduplicate this dataset based on book title exact match.

\paragraph{\dolmaRefs~Wikipedia and Wikibooks for Encyclopedic Content} 

This dataset was derived by March 2023 Wikimedia dumps. 
We use the ``English'' and ``Simple'' editions of Wikipedia and Wikibooks as base for the Encyclopedic subset of \dolma. 
Sources were processed using WikiExtractor\citep{wikiextractor2023}.
We remove any document with 25 or fewer \utf-segmented words, as we found shorter pages to either be the result of short, templated pages (\textit{e.g.}, pages containing only a few words and an information box) or XML parsing errors.
By design, this dataset does not contain duplicated documents.

\section{Training a Language Model on \dolma}
\label{sec:using}

As a final validation step of the \dolma pipeline, we train, evaluate and release a decoder-only, autoregressive language model which we call \OlmoTiny.  
We present zero-shot experimental results of \OlmoTiny on a range of downstream tasks demonstrating comparable quality to other released language models of comparable size.

\subsection{Evaluating \OlmoTiny}
\label{sec:evaluating-1b}

\begin{table}[h!]
    \centering
    \small
    \renewcommand{\arraystretch}{1.1}
    \begin{tabular}{c@{}c@{}c@{}c@{}c}
    \toprule
        {
            \renewcommand{\arraystretch}{1}
            \begin{tabular}[c]{@{}c@{}}
            \textbf{Task} 
            \vspace{0em}
            \end{tabular}
        }  & 
        {
            \renewcommand{\arraystretch}{1}
            \rotatebox{70}{\begin{tabular}[c]{@{}c@{}}
            {\footnotesize \hyperlink{cite.stablelm}{StableLM$_\mathbf{2}$}} \\
            {\footnotesize (1.6B) }
            \vspace{0em}
            \end{tabular}}
        }  & 
        {
            \renewcommand{\arraystretch}{1}
            \rotatebox{70}{\begin{tabular}[c]{@{}c@{}}
            {\footnotesize \hyperlink{cite.Biderman2023PythiaAS}{Pythia}} \\
            {\footnotesize (1.1B) }
            \vspace{0em}
            \end{tabular}}
        }  & 
        {
            \renewcommand{\arraystretch}{1}
            \rotatebox{70}{\begin{tabular}[c]{@{}c@{}}
            {\footnotesize \hyperlink{cite.tinyllama}{TinyLlama}} \\
            {\footnotesize (1.1B) }
            \vspace{0em}
            \end{tabular}}
        }  & 
        {
            \renewcommand{\arraystretch}{1}
            \rotatebox{70}{\begin{tabular}[c]{@{}c@{}}
            {\footnotesize \textbf{\color{VarnishBravoThree}\OlmoTiny} } \\
            {\footnotesize \color{VarnishBravoThree} (1.2B) }
            \vspace{0em}
            \end{tabular}}
        }   \\
    \midrule
    {\textit{\hyperlink{cite.arc}{ARC-E}}} & 63.7 & 50.2 & 53.2 & {\color{VarnishBravoThree}58.1} \\
    {\textit{\hyperlink{cite.arc}{ARC-C}}} & 43.8 & 33.1 & 34.8 & {\color{VarnishBravoThree}34.5} \\
    {\textit{\hyperlink{cite.clark2019boolq}{BoolQ}}} & 76.6 & 61.8 & 64.6 & {\color{VarnishBravoThree}60.7} \\
    {\textit{\hyperlink{cite.zellers2019hellaswag}{HellaSwag}}} & 68.2 & 44.7 & 58.7 & {\color{VarnishBravoThree}62.5} \\
    {\textit{\hyperlink{cite.openbookqa}{OpenBookQA}}} & 45.8 & 37.8 & 43.6 & {\color{VarnishBravoThree}46.4} \\
    {\textit{\hyperlink{cite.piqa}{PIQA}}} & 74.0 & 69.1 & 71.1 & {\color{VarnishBravoThree}73.7} \\
    {\textit{\hyperlink{cite.sciq}{SciQ}}} & 94.7 & 86.0 & 90.5 & {\color{VarnishBravoThree}88.1} \\
    {\textit{\hyperlink{cite.winogrande}{WinoGrande}}} & 64.9 & 53.3 & 58.9 & {\color{VarnishBravoThree}58.9} \\
    \midrule
    \textbf{Average} & \textbf{\textit{66.5}} & \textbf{54.5} & \textbf{59.4} & {\color{VarnishBravoThree}\textbf{60.3}} \\
    \bottomrule
    \end{tabular}
    \vspace{1em}
    \caption{
        Comparison of \OlmoTiny and other similarly-sized language models on our evaluation suite. 
    }
    \label{table:1b-eval}
\end{table}

In \autoref{table:1b-eval} we compare \OlmoTiny with other 1B models. 
We note that, while all models share a roughly comparable number of parameters, only TinyLlama was trained on roughly the same number of tokens as \OlmoTiny.
Pythia was trained on nearly 10 times fewer tokens and 
$\text{StableLM}_2$ was trained on 2 trillion tokens for two epochs (data composition not shared).
Nevertheless, we find that \OlmoTiny performs better on average than the most comparable model, TinyLlama, outperforming it in 4 out of 8 tasks from our evaluation suite \S\ref{sec:experimental-methodology}. 
Though zero-shot evaluations of such tasks are often challenging for smaller 1B models, we see that performance across all tasks and models is above naive random performance.

\subsection{Measuring Domain Fit}
\label{sec:ppl_eval}

In \S\ref{sec:desiderata}, we motivated our decision in curating \dolma to cover a diverse set of sources.
In this section, we use \OlmoTiny to assess \dolma's distribution of documents leads to pretrained language models that fit well to diverse textual domains, compared to training on other open corpora.
To represent diverse domains, we use Paloma~\citep{paloma}, a stratified collection of hundreds of fine-grained textual sources; thus, training on more diverse datasets should result in models with lower overall perplexity on Paloma. 
We repeat our data ablation methodology, training 1.2B models on 150B token samples from C4, mC4 (English-only)~\citep{mc4}, RedPajama v1, RefinedWeb~\citep{falcon40b}, Pile, and \dolma.

\begin{figure}[h!]
    \centering
    \includegraphics[width=.9\linewidth]{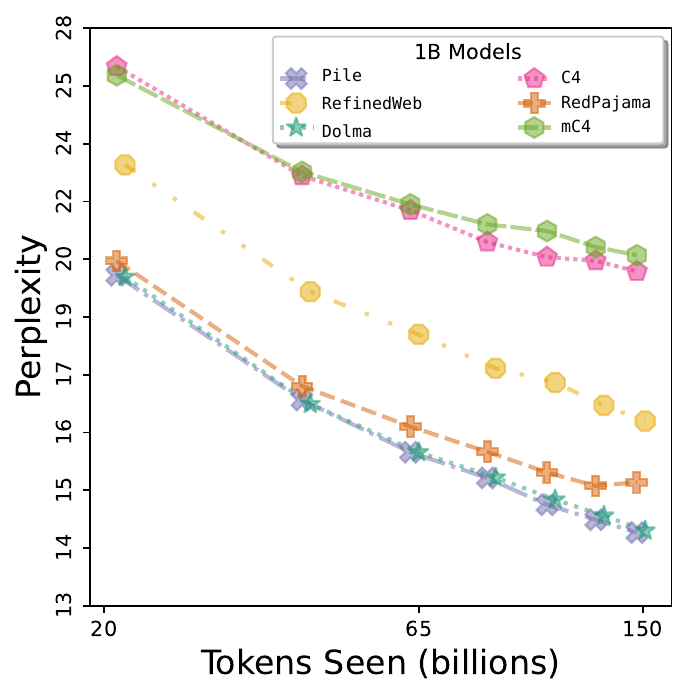}
    \caption{
    1.2B parameter language models trained on 150B tokens from \dolma and other open corpora, evaluated across training iterations on perplexity over diverse domains in Paloma~\citep{paloma}.
    }
    \label{fig:models_by_ppl_over_all_tasks}
\end{figure}

From the results in Figure~\ref{fig:models_by_ppl_over_all_tasks}, we observe the following: (1) The model trained on Pile performs well as it is comprised of many diverse sources, despite its overall smaller scale. (2) Larger multi-source datasets like \dolma and, to a lesser extent, RedPajama v1 yield models with similar coverage of diverse domains to Pile.
(3) Finally, training on single-source corpora like C4, mC4 (English-only), and RefinedWeb leads to models with poor fit to diverse domains as indicated by higher average perplexity.

Our controlled perplexity analysis reveals the importance of including non-web data from diverse curated sources. 
The metric that we use from Paloma surfaces how models fit more heterogeneous data, because it samples marked domains from each source equally rather than by their unequal proportions in the source. 
Intuitively, the model trained on the Pile is well-fit to such data as that pretraining corpus is mostly sourced from similar smaller, hand-picked sources. 
But as we wish to scale the total number of tokens in a corpus, the challenge becomes how to integrate more available web data without losing sample efficiency on diverse evaluations such as Paloma.
In this case, we see that \OlmoTiny nearly matches the perplexity curve of the Pile model despite a much larger fraction of web data included.

\section*{Conclusion}
\label{sec:conclusion}

In this manuscript, we introduce \dolma, a three trillion token English corpus for language model pretraining. 
The \dolma corpus is comprised of a diverse set of content, including web documents, scientific papers, code, public-domain books, social media, and encyclopedic materials.
Building off a list of explicit desiderata, we document our data curation pipelines, providing experimental results that support our decisions. 
We freely release \dolma and open-source all tools we used to curate this dataset as part of the OLMo project~\citep{olmo20247b}. 
Since the time of writing, we have made improvements to \dolma and have continued to make releases; for example, our follow-on release of \textbf{\dolma~\texttt{v.1.7}} yields significant performance improvement on downstream tasks, holding the model constant.\footnote{
\href{https://blog.allenai.org/olmo-1-7-7b-a-24-point-improvement-on-mmlu-92b43f7d269d}{\path{medium.com/p/92b43f7d269d}}}
We hope this line of work can promote transparency, reproducibility, and further research in the field of language modeling, 
as well as address the current gap in the availability of pretraining data of commercial and open language models.
We release \dolma under ODC-By
and our toolkit under Apache 2.0.

\section*{Limitations}
\label{sec:limitations}

\paragraph{English-only corpus.} 
\dolma was curated to contain English data. 
As tools for language identification may have false negatives, \dolma might contain a small percentage of non-English data.
Traces of non-English data are unlikely to lead to any meaningful downstream performance on non-English tasks for any model trained on \dolma. 
Thus, \dolma reinforces the expectation of English being the ``default'' language for NLP. 

\paragraph{Representativeness of sources in \dolma.} 
As mentioned in \S\ref{sec:desiderata}, it is impossible to curate a corpus that is representative of all language model data curation practices. 
Further, many open and close language models are trained on content that cannot be acquired or redistributed, and thus could not be included in \dolma. 

\paragraph{Single model configuration for ablations.} 
The experimental setup we use to validate our data curation pipeline only covers a subset of model types used to create language models. 
For example, while many language models are in the 7 billion to 70 billion parameters range, we train 1 billion parameter models;
further, we did not investigate the use of any alternative architectures to dense auto-regressive transformer models.
This choice was dictated by the need to efficiently iterate over many possible configurations, but it might result in design decisions that are not relevant at larger model sizes.
We expect downstream model developers to scrutinize \dolma before using it to train their language models, similar to the process we sketch in~\S\ref{sec:using}.

\paragraph{Limited tasks in evaluation suite.}
As detailed in \S\ref{sec:experimental-methodology}, we select tasks that have been used to evaluate previous base language models, and that are not present in our training data (i.e., \dolma is not contaminated against them). 
As such, we can only assess a subset of tasks language models are routinely used for. 
For example, the effect of adding code to pretraining data cannot be fully measured until models are able to generate executable code; such capability is typically observed only after models are finetuned to follow instructions~\citep{muennighoff2023octopack,zhuo2024astraios}.

\paragraph{Manual inspection and evaluation of \dolma is infeasible.}
Given the corpus size, it is impossible to fully inspect \dolma to assess its content. 
While tools like WIMBD~\citep{wimbd} and Data Portraits~\citep{Marone2023DataPR} aid programmatic inspection of subsets of data, they cannot provide an assessment of all documents in a corpus. 
As such, we cannot fully describe the properties of \dolma in terms of data distribution, content quality, and potential harms due to the inclusion or exclusion of particular content.

\section*{Ethical Considerations}
\label{sec:ethics}

\paragraph{Minimize risk of harm to individuals during data curation.}

Curating a pretraining corpus may introduce risk to individuals, either by facilitating access to information that is present in the corpus, or by enabling training of harmful models
that disclose personal information~\citep{Carlini2020-jx} or produce toxic content~\citep{gehman-etal-2020-realtoxicityprompts,Ngo2021-fn}.
To minimize these risks while meeting our stated goals, we engaged with legal and ethics experts 
early in the project and evaluated data design decisions based on their feedback on a case-by-case basis.
Broadly, we follow accepted practices when available (\textit{e.g.}, masking of certain personal identifiable information), and take a measured approach when diverging opinions exist in the literature (\textit{e.g.}, most effective approach to identify and remove toxic content). 
Further, we will provide tools to request data removal\footnote{\label{foot:removal}Available at {\href{https://docs.google.com/forms/d/e/1FAIpQLSfL6KzFR7xNJj6MPyV1uikIpj-VmrftC9mjty2nXzSClU2rnw/viewform}{\path{forms.gle/FzpUXLJhE57JLJ3f8}}}}
We believe in compromising on desired research artifact properties like model reproducibility, performance, and extensibility in cases of significant harm to individuals.

Besides a risk-based approach, alternative frameworks for considering the ethical implications of language model data have also been proposed.
Data stewardship~\citep{Jernite2022DataGI} seeks to create a framework to collect and reflect explicit interests of data owners. 
Data trusts~\citep{datatrust} or data licensing~\citep{licensing} can also enable explicit consent in sharing data for AI training. 
As no current state-of-the-art model is trained on data collected through these frameworks, these approaches would limit the representativeness goal stated in \S\ref{sec:desiderata}.
As these principles are adopted, we will consider them for future versions of \dolma.

\paragraph{Copyright and fair use considerations.} 
At the time of writing, the landscape governing applicability of copyright law and fair use doctrine (also known as ``fair dealing'') and language models is largely undetermined~\citep{Cooper2023GenerativeAILaw,Lee2023Talking}. 
In the United States, legal scholars and practitioners have suggested that training models on copyright content might constitute fair use~\citep{iSchoolUIUC2023,UCBerkeley2023,Henderson2023FoundationMA}, while also recognizing limitations of existing doctrine in this application~\citep{uspto-copyright-ai2}.
Further, legal assessments regarding the use of copyrighted data in language models vary widely depending on jurisdiction: 
in early 2024, Israel~\citep{Israel2022} and Japan~\citep{Technomancersai2023-rx} allow copyrighted content to be used for AI training data, although the latter is currently re-considering this framework.
While most datasets we used were curated with copyright and licensing in mind (e.g., open access papers in peS2o~\citep{peS2o}, open source repositories in the Stack~\citep{kocetkov2022stack}) or were already permissively licensed (e.g., Wikipedia is released under a Creative Commons license), we recognize that large web crawls may also contain copyrighted material. 
Yet, given current tools, it's not possible to reliably or scalably detect copyrighted materials in a corpus of this size.
Our decision to curate and distribute \dolma factors in several considerations, including that all our data sources were publicly available and already being used in large-scale language model pretraining (both open and closed). 
We recognize that the legal landscape of AI is changing rapidly, especially as it pertains to use of copyrighted materials for training models.

\bibliography{references,anthology}

\appendix

\section{Acknowledgements}
\label{sec:ack}

\dolma would not have been possible without the support of many individuals and institutions. 
The experimental components of this work were made possible through a partnership with AMD and CSC, enabling use of the LUMI supercomputer.
We thank Jonathan Frankle, Cody Blakeney, Matthew Leavitt and Daniel King and the rest of the MosaicML team for sharing findings from experiments on preliminary versions of our data.
We thank Vitaliy Chiley for messaging us on Twitter with a \href{https://twitter.com/vitaliychiley/status/1675594766799769600}{suggestion} for resolving a random number generator bug that was affecting our data shuffling.
We thank Erfan Al-Hossami, Shayne Longpre, and Gregory Yauney for sharing findings from their own large-scale pretraining data experiments.
We thank Ce Zhang and Maurice Weber of Together AI for thoughtful discussion on open datasets and data distribution format.
We thank Stella Biderman and Aviya Skowron for discussions around data licensing and data processing framework.
We thank our teammates at AI2 Nicole DeCario, Matt Latzke, Darrell Plessas, Kelsey MacMillan, Carissa Schoenick, Sam Skjonsberg, and Michael Schmitz for their help with the website, design, internal and external communications, budgeting, and other activities that supported smooth progress on this project.
Finally, we also express gratitude for the helpful discussions and feedback from our teammates at AI2 and close collaborators, including Prithviraj (Raj) Ammanabrolu, Maria Antoniak, Chris Callison-Burch, Peter Clark, Pradeep Dasigi, Nicole DeCario, Doug Downey, Ali Farhadi, Suchin Gururangan, Sydney Levine, Maarten Sap, Ludwig Schmidt, Will Smith, Yulia Tsvetkov, and Daniel S. Weld.

\section{Author Contributions}
\label{sec:contrib}

Dolma would not be possible without the help of our many teammates and collaborators. Weekly project meetings, messaging apps and documentation were accessible for anyone at AI2. Major decisions about Dolma were often made in these channels, with exception for certain topics (e.g., legal, funding). While many were involved in the Dolma effort (see Acknowledgements~\S\ref{sec:ack}), the authors of this paper were those who owned and delivered a critical piece of the puzzle. We detail their contributions below (authors in alphabetical order):

Contributors to \textbf{data acquisition and source-specific data processing} include Akshita Bhagia, Dirk Groeneveld, Rodney Kinney, Kyle Lo, Dustin Schwenk, and Luca Soldaini. Everyone contributed to literature review on available sources and best practices and decisions around sources to pursue. Akshita Bhagia, Rodney Kinney, Dustin Schwenk, and Luca Soldaini handled the bulk of data acquisition and processing and ablation experiments with 1B models for source-specific design decisions. Kyle Lo and Luca Soldaini handled discussions with legal to inform our choice of sources.

Contributors to \textbf{infrastructure and tooling} include Russell Authur, Dirk Groeneveld, Rodney Kinney, Kyle Lo, and Luca Soldaini. Rodney Kinney, Kyle Lo, and Luca Soldaini designed and implemented the shared toolkit used for processing our corpus at scale. Dirk Groeneveld wrote the Bloom filter for deduplication and decontamination. Russell Authur wrote a toolkit for acquisition and storage of Common Crawl data.

Contributors to \textbf{source-agnostic data processing} include Khyathi Chandu, Yanai Elazar, Rodney Kinney, Kyle Lo, Xinxi Lyu, Ian Magnusson, Aakanksha Naik, Abhilasha Ravichander, Zejiang Shen, and Luca Soldaini. Khyathi Chandu, and Aakanksha Naik developed the toxic text filter. Kyle Lo, and Xinxi Lyu helped evaluate it. Luca Soldaini developed the language filtering approach. Rodney Kinney, Zejiang Shen, and Luca Soldaini developed the ``quality'' filter. Yanai Elazar identified repeating $n$-gram sequences. Abhilasha Ravichander, Kyle Lo, and Luca Soldaini developed the PII filter. Jesse Dodge and Ian Magnusson developed the evaluation set decontamination approach. 

Contributors to \textbf{ablation experiments} include Iz Beltagy, Akshita Bhagia, Jesse Dodge, Dirk Groeneveld, Rodney Kinney, Kyle Lo, Ian Magnusson, Matthew Peters, Kyle Richardson, Dustin Schwenk, Luca Soldaini, Nishant Subramani, Oyvind Tafjord, and Pete Walsh. This work included designing and prioritizing experiments given compute constraints, implementing and running the 1B model experiments, and interpreting results. In particular, Oyvind Tafjord's work on the evaluation toolkit and Pete Walsh's work on the model implementation were critical.

Contributors to \textbf{posthoc experiments and analysis} on the final Dolma artifacts. Ben Bogin led the probing experiments on 1B model weights to assess impact of differing code mixtures with support from Kyle Lo and Niklas Muennighoff. Yanai Elazar ran the data analysis tool to summarize and document Dolma's composition. Valentin Hofmann led the tokenization fertility analysis with support from Kyle Lo. 
Ananya Harsh Jha and Ian Magnusson performed experiments training and evaluating baseline 1B models on other open datasets with support from Luca Soldaini.
Sachin Kumar and Jacob Morrison performed analysis of systematic issues in our choice of language identification and toxicity classifiers with support from Kyle Lo.
Niklas Muennighoff led analysis of correlation between different filters employed on Common Crawl data with support from Kyle Lo and Luca Soldaini.

Contributors to \textbf{licensing and release policy} include David Atkinson, Jesse Dodge, Jennifer Dumas, Nathan Lambert, Kyle Lo, Crystal Nam, and Luca Soldaini. 
David Atkinson, Jesse Dodge, Jennifer Dumas, and Crystal Nam led the bulk of this, including research into data licenses, risk-level determination for pretraining data, and defining the release policy. 
Kyle Lo and Luca Soldaini provided feedback throughout this process and handled technical details needed for the release.
Nathan Lambert provided feedback on release process and handled the actual release strategy, particularly around external communication.

All of the contributors above helped with \textbf{documentation and writing} of their respective components. In particular, Li Lucy provided an extensive literature review of language models, open corpora and pretraining corpus creation practices. Emma Strubell gave valuable feedback on our manuscript.
Nathan Lambert helped with feedback on the blog post and other forms of external-facing communication about Dolma.

Hannaneh Hajishirzi, Noah Smith, and Luke Zettlemoyer \textbf{advised} on the project, including broad strategy, writing, recruiting and providing resources. As OLMo project leads, Iz Beltagy, Jesse Dodge, and Dirk Groeneveld helped with \textbf{visibility and coordination} with other critical OLMo project workstreams. Notably, we credit Noah Smith for coming up with the name Dolma.

Finally, Kyle Lo and Luca Soldaini \textbf{led} the overall Dolma project and were involved in all aspects, including project management, planning and design, discussions with legal and ethics committees, data and compute partnerships, infrastructure, tooling, implementation, experiments, writing/documentation, etc.

\section{(Lack of) details about pretraining data curation for both open and closed language models}
\label{sec:pretrain-data-llms}

We provide a high-level overview of the pretraining data curation practices (or lack of reporting therof) of the largest, most performant language models (in no particular order) to illustrate the need for clear documentation and transparency around dataset curation.

\subsection{PaLM 2~\citep{Anil2023PaLM2T}}\label{appendix:llm:palm2} 

\citet{Anil2023PaLM2T} provides limited information on pretraining data used for PaLM 2; we summarize what we could from gather from their manuscript's Sections 3 and D1:

\begin{enumerate}
    \item \textbf{Corpus size}. Unreported other than it's larger than what was used to train PaLM~\citep{Chowdhery2022PaLMSL}
    \item \textbf{Data provenance}. Unreported other than they use web documents, books, code, mathematics, and conversational data. 
    \item \textbf{PII}. Reported as performed filtering, but without further details.
    \item \textbf{Toxicity}. Toxic text identified using \href{https://perspectiveapi.com/}{Perspective API} but lacking details needed for reproduction (i.e., text unit, threshold). No details on removal. They did report tackling toxicity through the use of control tokens, but do not provide enough details on this method.
    \item \textbf{Language ID}. Reports the most frequent languages included as well as their frequencies. Lacking details needed for reproduction (i.e., text unit, tools used, threshold).
    \item \textbf{Quality}.  Reported as performed filtering, but without further details.
    \item \textbf{Deduplication}. Reported as performed filtering, but without further details.
    \item \textbf{Decontamination}. N/A.
    \item \textbf{Other}. \citet{Anil2023PaLM2T} report aggregated statistics of how often certain demographic identities are represented (or not) in the data. Such statistics include identities (e.g., American) or English pronouns. These were identified using tools such as \href{https://knowyourdata.withgoogle.com/}{KnowYourData} or those available on \href{https://cloud.google.com/natural-language/docs/classifying-text}{GoogleCloud}, but the manuscript lacks specifics necessary for reproduction.
\end{enumerate}

\subsection{GPT-4~\citep{OpenAI2023GPT4TR}}\label{appendix:llm:gpt4} 

\citet{OpenAI2023GPT4TR} provides limited information on pretraining data used for GPT-4; we summarize what we could from gather from their manuscript's Section 2, Appendix C and D, footnotes 5, 6, 10 and 27, and Sections 1.1 and 3.1 in the System Card:

\begin{enumerate}
    \item \textbf{Corpus size}. N/A
    \item \textbf{Data provenance}. N/A aside from reporting that (1) data was sourced from both the Internet as well as third-party providers, (2) data was sourced mainly before September 2021 with trace amounts of more recent data, and (3) they included GSM-8K~\citep{Cobbe2021TrainingVT} as a tiny fraction of the total pretraining mix.
    \item \textbf{PII}. N/A.
    \item \textbf{Toxicity}. Removed documents that violate their usage policies from pretraining, including ``erotic content,'' using a combination of lexicon-based heuristics and bespoke classifiers following \citet{markov-2023-holistic-approach-undesired-content-detection}.
    \item \textbf{Language ID}. N/A aside from reporting that the majority of pretraining data is in English.
    \item \textbf{Quality}. N/A.
    \item \textbf{Deduplication}. N/A.
    \item \textbf{Decontamination}. No discussion of decontamination procedures, but instead reported post-hoc statistics measuring extent of contamination on professional and academic exams, as well as several academic benchmarks. Method for identifying contamination based on exact substring match (after removing whitespaces) of a test example against a pretraining data example. They reported some contamination with BIG-Bench~\citep{srivastava2023beyond}.
    \item \textbf{Other}. There are myraid works performing ``data archeology'' on GPT-4 that is, attempting to glean information about the pretraining data used in GPT-4 through probes for memorization. For example, \citet{Chang2023SpeakMA} show GPT-4 can generate sequences from copyrighted books. We do not attempt to survey all of these investigative works.
\end{enumerate}

\subsection{Claude~\citep{claude-announce}}\label{appendix:llm:claude} 

Unfortunately, we know next to nothing about the pretraining data used for Claude.

\subsection{Llama 2~\citep{Touvron2023Llama2O}}\label{appendix:llm:llama2} 

\citet{Touvron2023Llama2O} provides limited information on pretraining data used for Llama 2; we summarize what we could from gather from their manuscript's Sections 2.1, 4.1, and A.6:

\begin{enumerate}
    \item \textbf{Corpus size}. 2T tokens.
    \item \textbf{Data provenance}. N/A aside from they avoided using Meta user data.
    \item \textbf{PII}. Reported as excluded data from certain websites known to contain high volumes of PII, though what these sites are was not disclosed.
    \item \textbf{Toxicity}. Not explicitly discussed, but appears to not have performed toxicity filtering, opting instead to handle toxic text generation in a later training stage. They do report results from a post hoc analysis in which they used a HateBERT~\citep{caselli-etal-2021-hatebert} classifier finetuned on ToxiGen~\citep{hartvigsen-etal-2022-toxigen} to score each document line (and averaged to produce a document-level score).
    \item \textbf{Language ID}. Not stated as used in pretraining data curation, but they provide a post hoc analysis of the pretraining dataset using FastText Language ID with a 0.5 threshold for detected language. We assume this is likely the same protocol they used for pretraining data curation as it is also seen in the CCNet library~\citep{wenzek-etal-2020-ccnet}, which was used for Llama~\citep{Touvron2023LLaMAOA}.
    \item \textbf{Quality}. N/A.
    \item \textbf{Deduplication}. N/A.
    \item \textbf{Decontamination}. They provide extensive reporting on their deduplication method, which relies on a modified version of the ngram deduplication tool from \citet{lee-etal-2022-deduplicating}.
    \item \textbf{Other}. Reported upsampling certain sources, but without further details. They also report a similar analysis as in PaLM 2~\citep{Anil2023PaLM2T} on aggregate statistics about demographic identities and English pronouns.
\end{enumerate}

\subsection{LLaMA~\citep{Touvron2023LLaMAOA}}\label{appendix:llm:llama} 

\citet{Touvron2023LLaMAOA} provides some information on pretraining data used for training LLaMA; we summarize what we could gather from their manuscript's Section 2.1.

\begin{enumerate}
    \item \textbf{Corpus size}. 1.4T tokens. 
    \item \textbf{Data provenance}. LLaMA used data with known provenance, including five shards of CommonCrawl between 2017 and 2020, C4~\citep{raffel2020exploring}, GitHub code from \href{https://cloud.google.com/bigquery/public-data/}{Google BigQuery public datasets} (restricted to Apache, BSD and MIT licenses), Wikipedia dumps from June to August 2022, Project Gutenberg books, Books3 from The Pile~\citep{Gao2020ThePA}, LaTeX files from arXiv, and StackExchange pages. 
    \item \textbf{PII}. N/A.
    \item \textbf{Toxicity}. N/A. Reports evaluation on the RealToxicityPrompts~\citep{gehman-etal-2020-realtoxicityprompts} benchmark.
    \item \textbf{Language ID}. Reports use of the CCNet library~\citep{wenzek-etal-2020-ccnet}, which employs FastText~\citep{joulin2016fasttext} classifiers to remove non-English text (below a 0.5 threshold). No additional language ID reported for C4, GitHub, Books, arXiv, and StackExchange sets. For Wikipedia, reported restriction of pages to those using Latin or Cyrillic scripts: bg, ca, cs, da, de, en, es, fr, hr, hu, it, nl, pl, pt, ro, ru, sl, sr, sv, uk.
    \item \textbf{Quality}. Reports use of the CCNet library~\citep{wenzek-etal-2020-ccnet} to remove low-quality content from CommonCrawl; CCNet uses KenLM~\citep{heafield-2011-kenlm}, an $n$-gram language model to score perplexity of text as a measure of similarity to Wikipedia text. They do not report their chosen threshold for filtering. They also report use of a linear model trained to classify pages as Wikipedia Reference-like or not. They also report light heuristic filtering of boilerplate content for GitHub and Wikipedia subsets.
    \item \textbf{Deduplication}. Reports use of the CCNet library~\citep{wenzek-etal-2020-ccnet} to identify duplicated lines for Common Crawl texts, file-level exact match deduplication for GitHub code, and deduplicating books with over 90\% for Gutenberg and Books3 subsets. 
    \item \textbf{Decontamination}. N/A.
    \item \textbf{Mixture}. The manuscript reports a mixture of 67\% CommonCrawl, 15\% C4, 4.5\% GitHub, 4.5\% Wikipedia, 4.5\% Books, 2.5\% arXiv, and 2.0\% StackExchange. Model training was a single epoch over this mixture except for an upsampling of Wikipedia and Books (2 epochs).
\end{enumerate}

\subsection{OPT~\citep{zhang-2022-language}}\label{appendix:llm:opt} 

From \citet{zhang-2022-language}'s manuscript and provided datasheet~\citep{gebru2021datasheets}, we summarize the following:

The OPT model was trained on \textbf{180B tokens} from data sources with known \textbf{provenance}: the datasets used for RoBERTa~\citep{Liu2019RoBERTaAR}, a subset of the Pile~\citep{Gao2020ThePA}, and the Pushshift Reddit Dataset~\citep{pushshift} as processed by \citep{roller-etal-2021-recipes}. They made several notable changes to these sources:

\begin{enumerate}
    \item \emph{RoBERTa}.  Reports updated the CC-News collection up to September 2021.
    \item \emph{Pile}. Reports restricted to the following collections: CommonCrawl, DM Mathematics, Project Gutenberg, HackerNews, OpenSubtitles, OpenWebText2, USPTO and Wikipedia. \citep{zhang-2022-language} report omission of other Pile subsets due to gradient norm spikes at the 1B model scale.
    \item \emph{Pushshift Reddit}. 
    Reports restricted to only the longest chain of comments in each thread; an operation that reportedly reduced the dataset by 66\%.
\end{enumerate}

Also describes: (1) \textbf{deduplication} using MinHashLSH~\citep{rajaraman-ullman-minhash-lsh} with a Jaccard similarity threshold of 0.95, and (2) \textbf{language ID} filtering to English-only text, though they do not describe the method used. 

They do not discuss whether they do (or do not) perform any processing for \textbf{PII}, \textbf{toxicity}, \textbf{quality}, or \textbf{decontamination}.

\section{Experimental Setup}
\label{sec:setup}

\subsection{Ablation Setup}
\label{sec:setup:abl}

For all data ablations described in this section, we train a 1B parameter model on up to 150B tokens. 
We follow model architecture and training from OLMo~\citep{olmo20247b}; we summarize key details here, but direct the reader to the manuscript for further details.
Each model is an decoder-only transformer model with 16 layers, 16 attention heads, and 2048 dimensionality. 
We use ALiBi positional embeddings~\citep{Press2021Alibi}, SwiGLU activation~\citep{Shazeer2020swiglu}, and mixed precision; model context size is set to $2048$ tokens. 
We use EleutherAI's GPT NeoX tokenizer~\citep{Black2022GPTNeoX20BAO}.
The model is trained using the LionW optimizer~\citep{Chen2023Lion} with $1\text{e-}4$ peak learning rate, warm-up of $2000$ steps, cosine decay, and $1\text{e-}2$ weight decay. 
Batch size was set to $1024$.
While we set our max number of steps to 95k (which is approximately 200B tokens), we conclude our experiments at 150B tokens.

We use 64 \href{https://www.amd.com/en/products/server-accelerators/instinct-mi250x}{AMD Instinct MI250X accelerators}. Each MI250X accelerator contains two logical nodes; therefore, from the point of view of our training code, our experiments ran on 128 compute units grouped in 16 nodes. Per each logical unit, we use a micro-batch size of 8. 
We implement our experiments using the \texttt{anonymized} codebase.

\subsection{Perplexity Evaluation Suite}
\label{sec:setup:ppl}

For data ablations, we keep track of language model perplexity using Paloma~\citep{paloma}. Datasets included:

\begin{itemize}[leftmargin=1em]
    \item \textbf{C4}~\citep{raffel2020exploring,dodge-etal-2021-documenting}: Standard contemporary LM pretraining corpus automatically filtered from the April 2019 Common Crawl scrape.
    \item \textbf{mC4}~\citep{mc4}; \textit{English subset}: the English language portion of a pretraining corpus automatically filtered from 71 Common Crawl scrapes.
    \item \textbf{Pile}~\citep{Gao2020ThePA}, \textit{validation set}: widely-used language modeling pretraining corpus; 
    contains documents curated from multiple sources including several non-web sources.
    \item \textbf{WikiText 103}~\citep{merity2016pointer}: a standard collection of verified “Good” and “Featured” articles on Wikipedia.
    \item \textbf{Penn Tree Bank}~\citep{marcus-etal-1994-penn}: widely-used NLP corpus derived from Wall Street Journal articles.
    
    \item \textbf{M2D2}~\citep{reid-etal-2022-m2d2}, \textit{S2ORC subset}: papers from Semantic Scholar~\citep{lo-etal-2020-s2orc} grouped by hierarchical academic field categories.
    \item \textbf{M2D2}~\citep{reid-etal-2022-m2d2}, \textit{Wiki subset}: Wikipedia articles grouped by hierarchical categories in the Wikipedia ontology
    \item \textbf{C4 100 domains}~\citep{chronopoulou-etal-2022-efficient}: balanced samples of the top 100 domains in C4.
    \item \textbf{Gab}~\citep{zannettou2018gab}: data from 2016-2018 from an alt-right, free-speech-oriented social media platform that has been shown to contain more hate speech than mainstream platforms. 
    
    \item \textbf{ICE}~\citep{greenbaum1991ice}: English from around the world curated by local experts, with subsets for Canada, East Africa, Hong Kong, India, Ireland, Jamaica, Philippines, Singapore, and the USA.

    \item \textbf{Twitter AAE}~\citep{blodgett-etal-2016-demographic}: balanced sets of tweets labeled as African American or white-aligned English.
    \item \textbf{Manosphere}~\citep{ribeiroevolution2021}: sample of 9 forums where a set of related masculinist ideologies developed over the past decade.
    \item \textbf{4chan}~\citep{papasavva2020raiders}: data from 2016-2019 politics subsection of an anonymity-focused forum found shown to contain high rates of toxic content.
    
\end{itemize}

We also curated held-out sets from other open language model corpora to augment Paloma:
\begin{itemize}[leftmargin=1em]
    \item \textbf{\dolma}~(this work), \textit{uniform sample}: A sample 8,358 documents from the \dolma corpus across all of its subsets (13 from books, 1,642 from Common Crawl web pages, 4,545 Reddit submissions, 450 scientific articles, 1,708 Wikipedia and Wikibooks entries).
    \item \textbf{RedPajama v1}~\citep{redpajama2}: 1 trillion tokens replication of the LLaMA 1~\citep{Touvron2023LLaMAOA} pretraining corpus.
    \item \textbf{Falcon RefinedWeb}~\citep{Penedo2023TheRD}: A corpus of English sampled from all Common Crawl scrapes until June 2023, more aggressively filtered and deduplicated than C4 and mC4-en.
    \item \textbf{Dolma 100 Subreddits}~(this work): Balanced samples of the top 100 subreddits by number of posts, sourced from the \dolma Reddit subset.
    \item \textbf{Dolma 100 Programming Languages}~(this work): Balanced samples of the top 100 programming languages by number of tokens, sourced from the \dolma Stack subset.
\end{itemize}

\subsection{Downstream Evaluation Suite}
\label{sec:setup:downstream}

We primarily base our data ablation decisions on the performance of models on this evaluation suite:

\begin{itemize}[leftmargin=1em]
    \item \textbf{AI2 Reasoning Challenge}~\citep{arc}: A science question-answering dataset broken into \emph{easy} and \emph{challenge} subsets. Only the easy subset was used in online evaluations. The challenge subset was, however,  included in offline evaluations.
    \item \textbf{BoolQ}~\citep{clark2019boolq}: A reading comprehension dataset consisting of naturally occurring yes/no boolean questions and background contexts. 
    \item \textbf{HellaSwag}~\citep{zellers2019hellaswag}: A multiple-choice question-answering dataset that tests situational understanding and commonsense. 
    \item \textbf{OpenBookQA}~\citep{openbookQA}: A multiple-choice question-answering dataset modeled on open-book science exams.  
    \item \textbf{Physical Interaction: Question Answering (PIQA)}~\citep{piqa}: A multiple-choice question-answering dataset that focuses on physical commonsense and naive physics.  
    \item \textbf{SciQ}~\citep{sciq}: A crowdsourced multiple-choice question-answering dataset consisting of everyday questions about physics, chemistry and biology, among other areas of science.   
    \item \textbf{WinoGrande}~\citep{winogrande}: A dataset of pronoun resolution problems involving various forms of commonsense. Modeled after the Winograd challenge from \citet{wsc}.  
\end{itemize}

\subsection{Training Setup for \OlmoTiny}
\label{appendix:training-olmo-1b}

For \OlmoTiny, we follow the experimental setup outlined for dataset ablation experiments in \autoref{sec:setup}, with the following differences:

\begin{itemize}[leftmargin=1em]
    \item We set the max number of steps to 739,328 (which is roughly 3.1T tokens).
    \item We double the batch size to $2048$ and do so by scaling up to 256 compute units (double what we used for data ablations).
    \item Due to instabilities we found in the LionW optimizer, we switched to using AdamW.
\end{itemize}

\section{Construction of Conversational Threads in Forums Data} 
\label{sec:how-to-thread}

Content comes from Reddit's data API in two separate but linked forms: \textit{submissions} and \textit{comments}. \textit{Submissions} are either "link posts" to external content (e.g. news articles, blogs, or even multimedia content) or "self posts" (submissions written by the poster meant to initiate a discussion thread on a topic). \textit{Comments} are user replies to either the initiating post (top level comments) or to another user's comment. Posts, top-level comments, and replies to comments form a nested conversational thread with a submission post at it's root and comments branching out into multiple possible dialogue trees.

The tree-like structure of Reddit threads allows for multiple possible data formats depending on how the various components of a thread are combined. 
We investigate three formats for their potential as LM pretraining data:
\begin{itemize}[leftmargin=1em]
    \item \textbf{Atomic content}. This simple format treats all comments and submissions as independent documents without any structure or connection to the thread they appear in.
    \item \textbf{Partial threads}. This format assembles comments from the same thread into a structured, multi-round dialogue between users. Submissions are left as separate documents. Assembled dialogues are limited to a maximum parent depth, and the resulting documents are only snippets of a their originating thread (which are spread across several documents).
    \item \textbf{Full threads}. This complex format combines a given submission and all of its child comments into a single document encompassing an entire thread. Code-like indentation is used to indicate the depth of a comment in the thread's hierarchy.
\end{itemize}

We experimentally evaluated these strategies for assembling documents in \autoref{fig:abl_reddit}. We found that, for language modeling purposes, treating comments and submissions as atomic units leads to better downstream performance compared to partial and full threads. 
We hypothesize that the more complex formatting required to handle dialogues might introduce undesirable content for language modeling, such as short and repeated comments. 
We leave the study of better formatting for forum content for language modeling to future work.

\section{Tokenization Analysis}
\label{app:tokenization}

\begin{figure*}[t!]
        \centering      
        \begin{subfigure}[b]{0.3\textwidth}  
            \includegraphics[width=\textwidth]{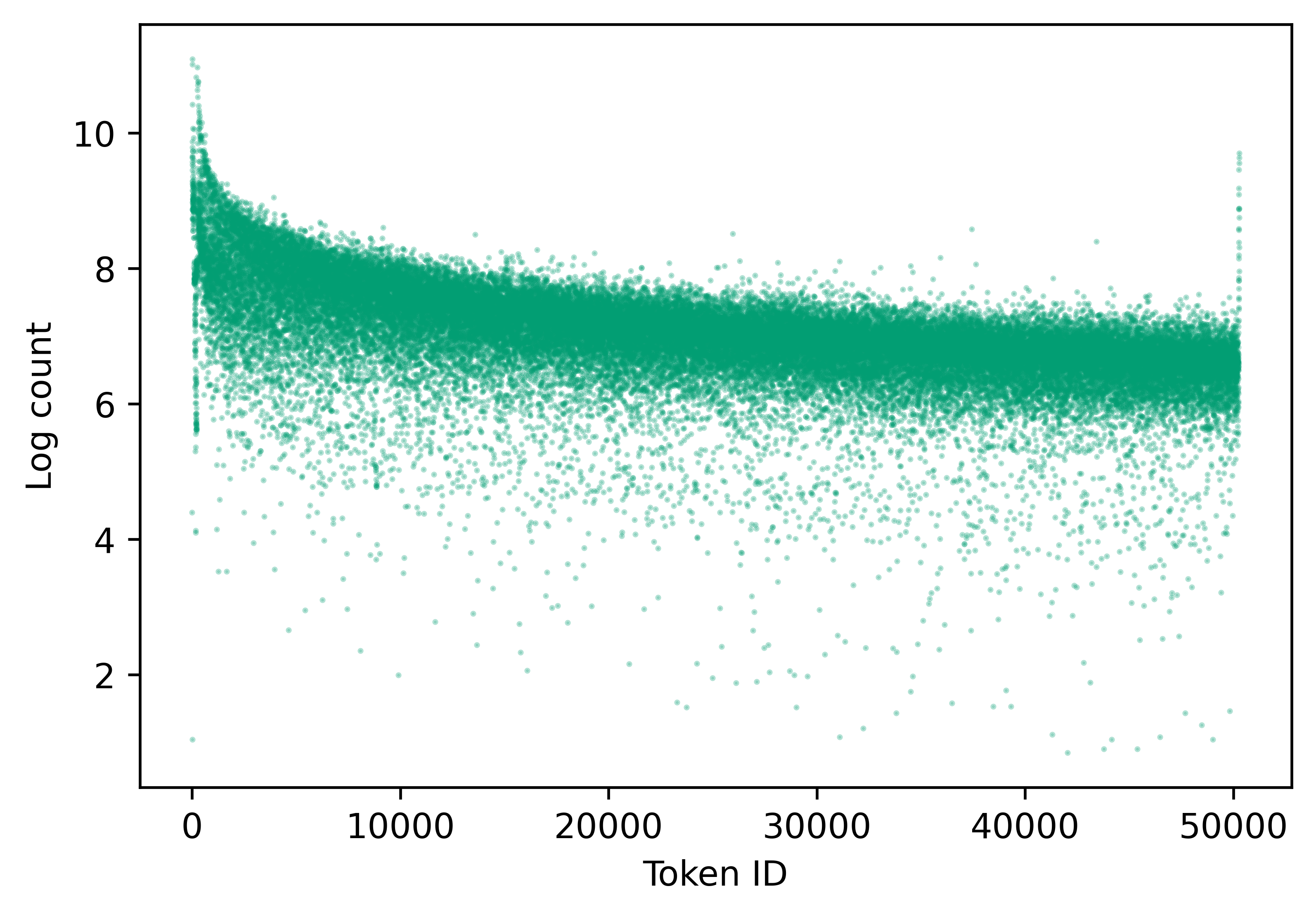}
            \caption[]%
            {{\small Count analysis\vspace{0.25cm}}}    
            \label{fig:tokenization-frequency-analysis}
        \end{subfigure} 
        \begin{subfigure}[b]{0.3\textwidth}   
            \includegraphics[width=\textwidth]{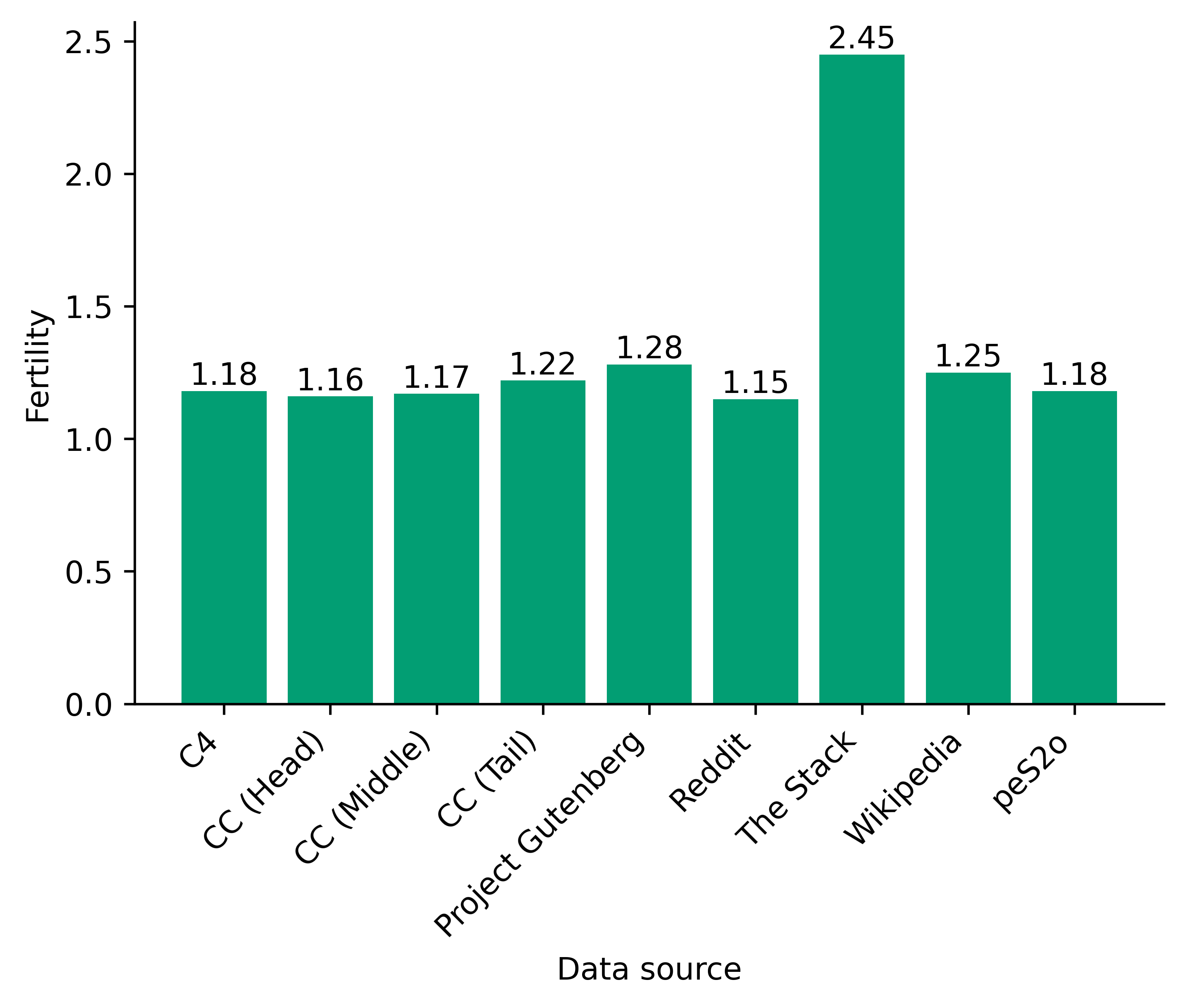}
            \caption[]%
            {{\small Fertility analysis}}   
            \label{fig:tokenization-fertility-analysis}
        \end{subfigure}
        \begin{subfigure}[b]{0.3\textwidth}   
            \includegraphics[width=\textwidth]{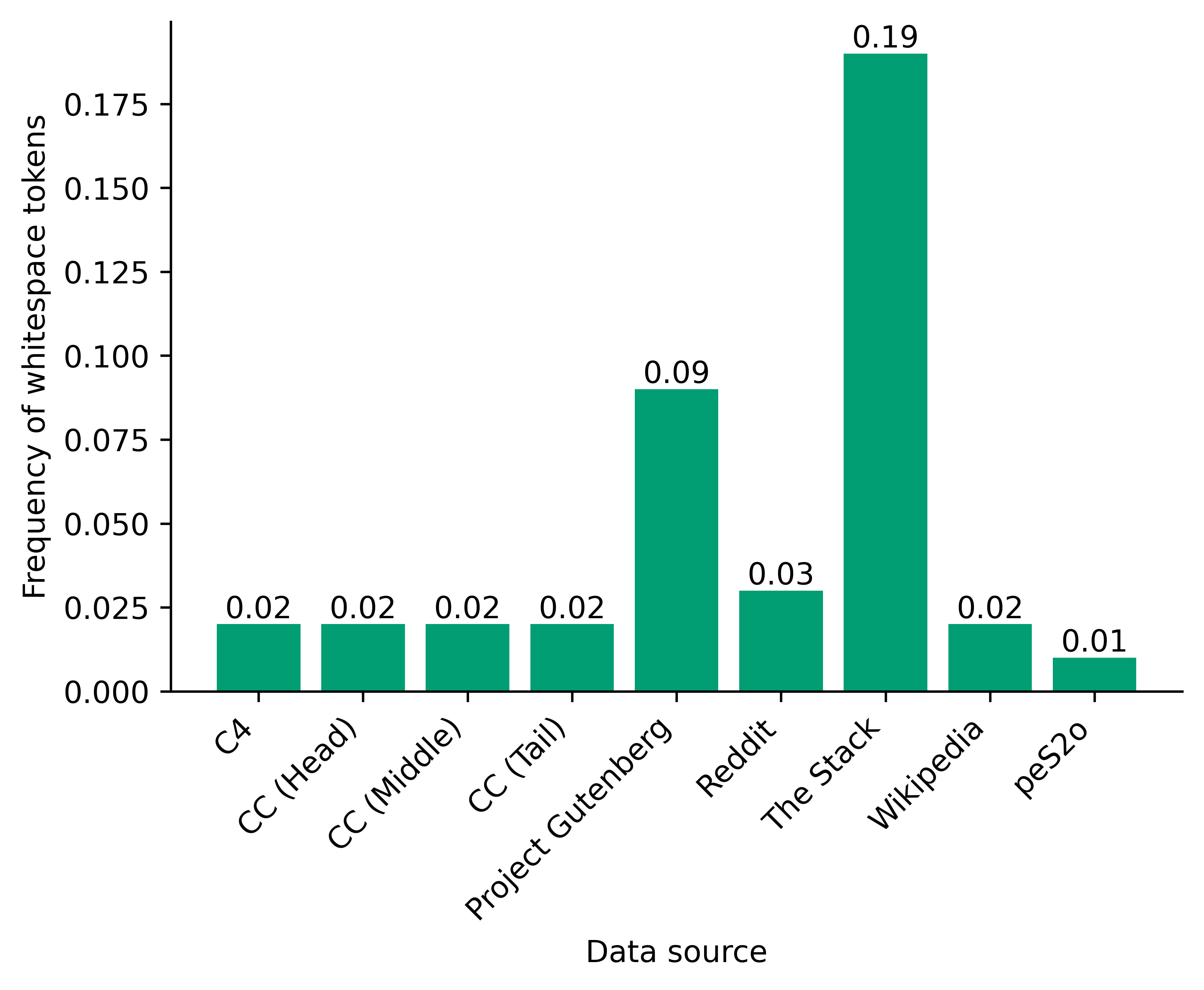}
            \caption[]%
            {{\small Whitespace analysis}}   
            \label{fig:tokenization-whitespace-analysis}
        \end{subfigure}
        \caption[]{Tokenization analysis. Tokens with small IDs, which have a high count in the tokenizer training data, also tend to have a high count in \dolma (a). The Stack has a substantially higher fertility compared to the other data sources (b), which can be explained by the higher relative frequency of whitespace characters such ``\textbackslash n'' and ``\textbackslash t'' (c). See text for more details.}
        \label{fig:tokenization-analysis}
\end{figure*}

The first step of processing text with LMs is \emph{tokenization}, i.e., mapping the text to a sequence of tokens with corresponding input embeddings \citep{sennrich-etal-2016-neural, kudo-2018-subword, kudo-richardson-2018-sentencepiece}. Recently, there has been a growing interest in the question of how well LM tokenizers fit different data sources \citep[e.g., data in different languages;][]{ahia2023languages, petrov2023language}
Inspired by this emerging line of work, we conduct an explorative analysis of the GPTNeoX tokenizer~\citep{Black2022GPTNeoX20BAO} applied to \dolma, which provides a first picture of how challenging the different data sources comprised by \dolma are for current LM tokenizers. 

We start by taking a global look at the tokenizer's fit to \dolma. Out of the 50,280 tokens in the tokenizer vocabulary, 50,057 are present in the tokenized text of \dolma. In other words, 223 tokens are never used, amounting to roughly 0.4\% of the tokenizer vocabulary. The 223 tokens mostly consist of combinations of whitespace characters (e.g., ``\textbackslash n\textbackslash n {} '', two newline characters followed by two blank space characters). Note that when training an LM with the examined tokenizer on \dolma, the input embeddings corresponding to these tokens would not be updated. 
In terms of the count distribution of tokens, we find that tokens with smaller IDs tend to have higher counts in \dolma (see Figure \ref{fig:tokenization-frequency-analysis}), which is also reflected by a strong Spearman's correlation between (i) the ranking of tokens based on their counts in \dolma and (ii) the token IDs ($r =$ 0.638, $p <$ 0.001). Given how the tokenizer was trained \citep{sennrich-etal-2016-neural, Black2022GPTNeoX20BAO}, smaller IDs correspond to byte pairs merged earlier and hence tokens occurring more frequently in the tokenizer training data 
Overall, these results suggest a good fit of the GPTNeoX tokenizer to \dolma.

Does the tokenizer fit all data sources included in \dolma equally well? To examine this question, we analyze fertility, which is defined as the average number of tokens per word generated by a tokenizer \citep{acs2019,Scao2022BLOOMA1}, in our case measured on a specific data source.
We find that fertility is similar for most data sources, ranging between 1.15 (conversational forum subset) and 1.28 (books subset), with the exception of the code subset, which has a substantially higher fertility of 2.45 (see Figure \ref{fig:tokenization-fertility-analysis}). This means that the costs of processing the code subset --- be they computational or financial in nature \citep{petrov2023language} --- are more than twice as high compared to the other data sources. 

What causes this discrepancy? We find that in the code subset (which mostly contains code), words are often preceded by whitespace characters \emph{other than} 
a blank space (e.g., newline, tab, return). Crucially, while a blank space before a word is tokenized as part of that word (e.g., \textit{I love you} $\rightarrow$ ``I'', `` love'', `` you''), other whitespace characters yield separate tokens (e.g., \textit{I \quad love \quad you} $\rightarrow$ ``I'', ``\textbackslash t'', ``love'', ``\textbackslash t'', ``you''). This can also be seen by plotting the relative frequency of tokens representing whitespace characters by data source, which is one order of magnitude higher for The Stack compared to most other data sources (see Figure \ref{fig:tokenization-whitespace-analysis}). When training LMs on The Stack (or code more generally), it thus might be advisable to add special tokens to the tokenizer \citep[e.g., ``\textbackslash nif'';][]{hong-etal-2021-avocado}. It is important to notice that this observation applies to most tokenizers in use today (e.g., the tokenizer used by GPT-4), which tend to lack tokens such as ``\textbackslash nif''.

\section{Auditing our Language Filter}
\label{app:langid}

To analyze the impact of the FastText language identification classifier, we ran an external audit on the International Corpus of English (ICE)~\citep{Kirk2018TheIC}, a dataset containing spoken and written English from nine countries around the world. We ran our language ID tool on all documents in the ICE dataset to estimate how many documents from each region would have been erroneously filtered. 
The ground truth in this analysis is that every document is in English, and should be classified as such. Interestingly, we found that at our fairly permissive threshold (keeping documents with at least a 0.5 score for English)  correctly identified all English-language documents in ICE each as English, no matter the region it was from.

\begin{figure}[h]
    \centering
    \includegraphics[width=.8\linewidth]{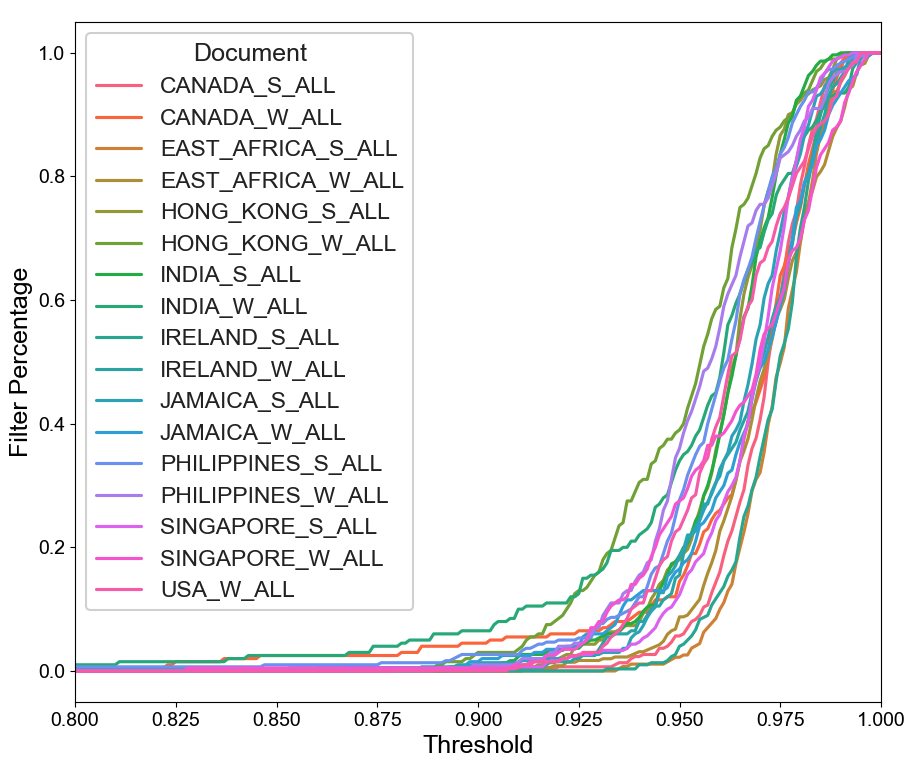}   
    \caption{Percentage of English-language documents in the International Corpus of English (ICE)~\citep{Kirk2018TheIC} that would be misidentified as non-English as a result of thresholding the FastText classifier's predicted English score. We find a majority of English documents in ICE remain identified as English even with a threshold of 0.90.}
    \label{fig:lang-id-ice}
\end{figure}

\section{Details on Toxicity Filters} 
\label{app:toxic}

\paragraph{Implementation.} To remove toxic content from \dolma, 
we used the Jigsaw Toxic Comments dataset~\citep{jigsaw}, 
which contains forum comments tagged with (multilabel) categories ``\texttt{toxic}'', ``\texttt{severe toxic}'', ``\texttt{threat}'', ``\texttt{insult}'', ``\texttt{obscene}'', and/or ``\texttt{identity hate}'' alongside unlabeled comments, to train two FastText classifiers---a binary ``\texttt{hate}'' detector and a binary ``\texttt{NSFW}'' detector:

\begin{enumerate}[leftmargin=1em]
    \item For our ``\texttt{hate}'' detector, we group all unlabeled comments and ``\texttt{obscene}''-only comments as negatives and leave remaining comments as positives.
    \item For our ``\texttt{NSFW}'' detector, we take all comments tagged as ``\texttt{obscene}'' as positives and leave other remaining comments as negatives. It is important to note this detector only filters \emph{toxic content} that mentions sexual or obscene topics, not sexual content in general. 
\end{enumerate}

\begin{figure}[h!]
    \centering
    \includegraphics[width=.8\linewidth]{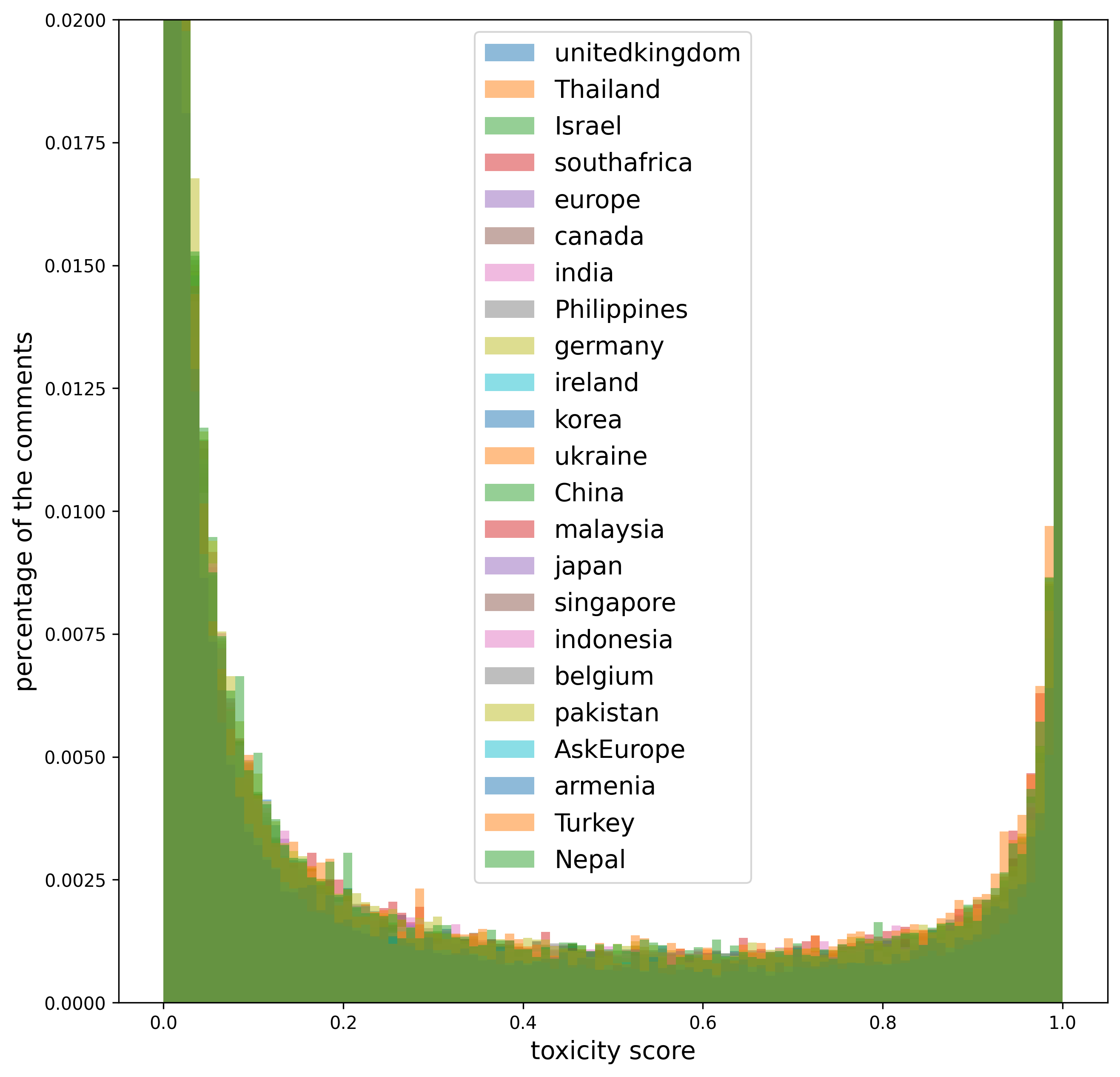}   
    \caption{Distribution of Reddit comments labeled as toxic by English variation.}
    \label{fig:reddit-location-toxicity}
\end{figure}

\paragraph{Analysis of resulting classifier.} To measure dialectal biases in the FastText toxicity classifier, we analyze its proclivity to predict English variations spoken in different countries as toxic. %
Starting with the unfiltered Reddit corpus, we create a dataset of comments from location-based subreddits,\footnote{\href{https://www.reddit.com/r/LocationReddits/wiki/index/}{\path{reddit.com/r/LocationReddits/wiki/index}}} filtering for country-specific subreddits with more than 50K comments. This dataset serves as a crude proxy for different dialects of English, assuming most commenters live in the respective locations and speak the variation. We further assume the fraction of actually toxic comments in each of these subreddits to be roughly the same. We compute the toxicity score for each comment in this dataset using the FastText classifier and 
report the percentage of comments marked as toxic against different classifier thresholds in \autoref{fig:reddit-location-toxicity}.  For all thresholds, for any two locations, we find <5\% difference in the fraction of comments marked as toxic suggesting little to no bias. 
Further, we plot the distribution of toxicity scores for comments in each subreddit and find that scores assigned to the comments often fall at the extremes (close to 0 or close to 1), suggesting that any reasonable threshold (lying between 0.1 to 0.9) to predict toxicity will lead to similar outcomes.%

\section{Details on PII Filters}
\label{app:pii-filter-details}

\paragraph{Filter implementation.} The Common Crawl, C4, Reddit, and GitHub subsets used the same regular expressions for identifying PII.
We refer the reader to our \href{https://anonymous.4open.science/r/dolma-55C2/}{GitHub} for exact implementations of our regular expressions for each of the PII types --- email address, phone number, and IP address.
Once spans are tagged, we employ different processing strategies based on the their density on each document:
\begin{itemize}[leftmargin=1em,itemsep=1mm,topsep=1mm]
    \item \textit{5 or fewer PII spans detected}: 
    we replace all spans on a page with special tokens \texttt{|\negthickspace|\negthickspace|EMAIL\_ADDRESS|\negthickspace|\negthickspace|}, \texttt{|\negthickspace|\negthickspace|PHONE\_NUMBER|\negthickspace|\negthickspace|}, and \texttt{|\negthickspace|\negthickspace|IP\_ADDRESS|\negthickspace|\negthickspace|} for email addresses, phone numbers, and IP addresses respectively.\footnote{When training models on \dolma, we add these special tokens to the tokenizer vocabulary.}
    In total, we find that 0.02\% of documents in the 25 Common Crawl snapshots match this filter. 
    \item \textit{6 or more PII spans detected}: 
    we remove any document that contains 6 or more matching PII spans. 
    We use this approach because pages containing abundant phone numbers and email addresses are likely to pose a greater risk of disclosing other PII classes. 
    0.001\% of documents in the 25 Common Crawl snapshots match this filter.
\end{itemize}

\section{Do quality and content filters have similar effects?}
\label{sec:quality-content-correlation}

\begin{figure*}[htbp]
    \centering
    \subfloat[\centering High]{{\includegraphics[width=0.25\textwidth]{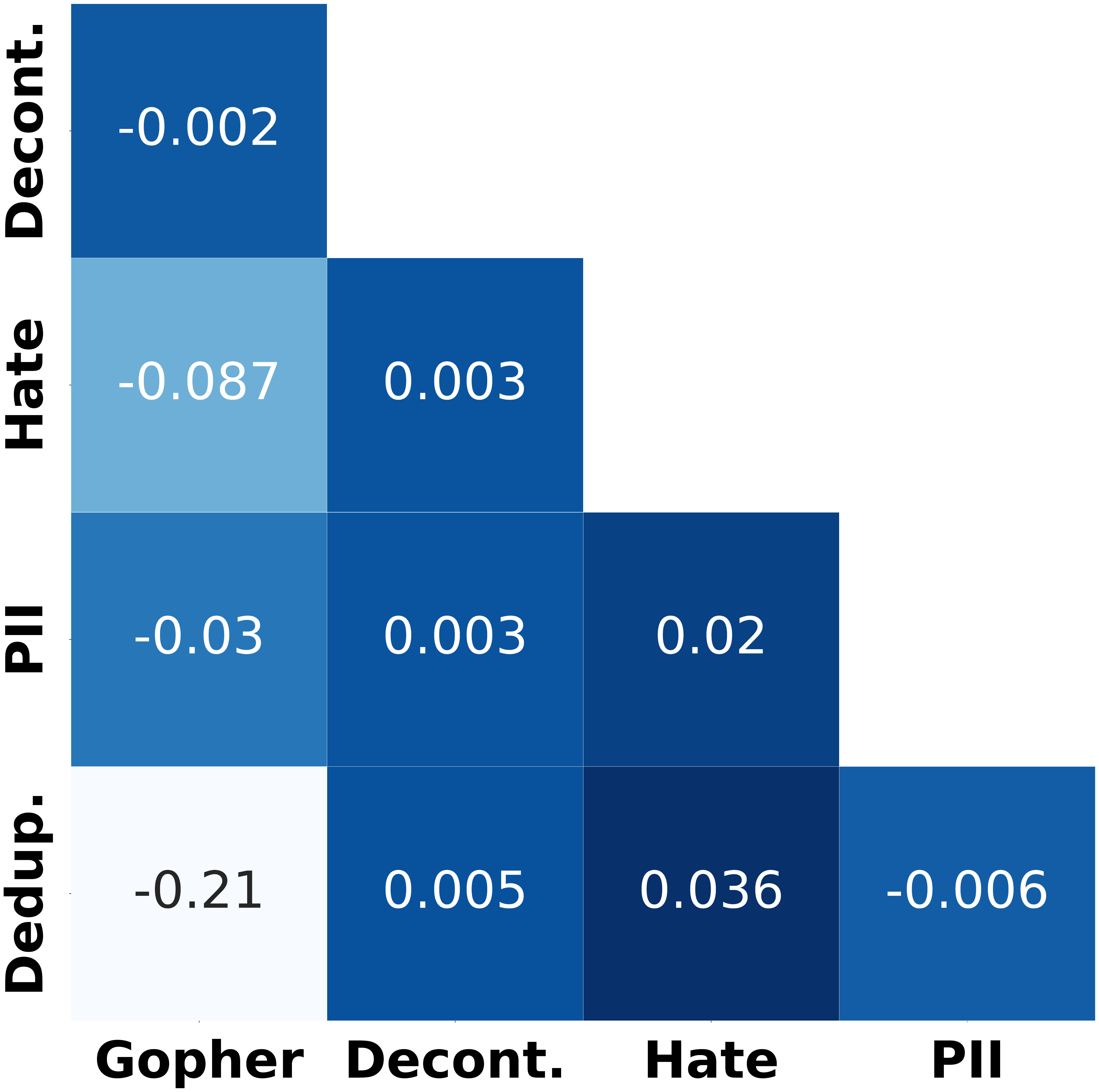}}}
    \qquad
    {\subfloat[\centering Medium]{{\includegraphics[width=0.25\textwidth]{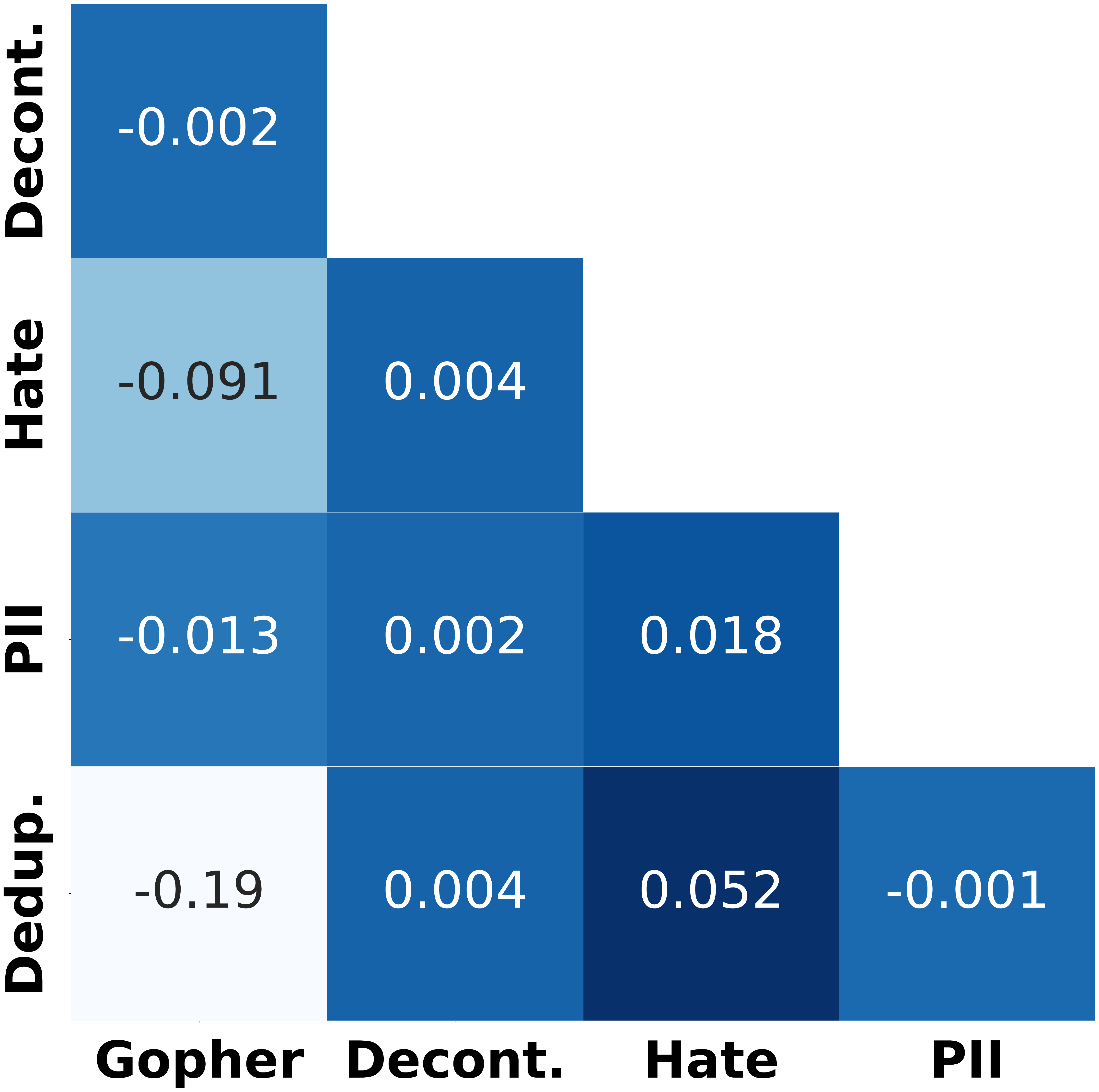}}}}
    \qquad
    {\subfloat[\centering Low]{{\includegraphics[width=0.25\textwidth]{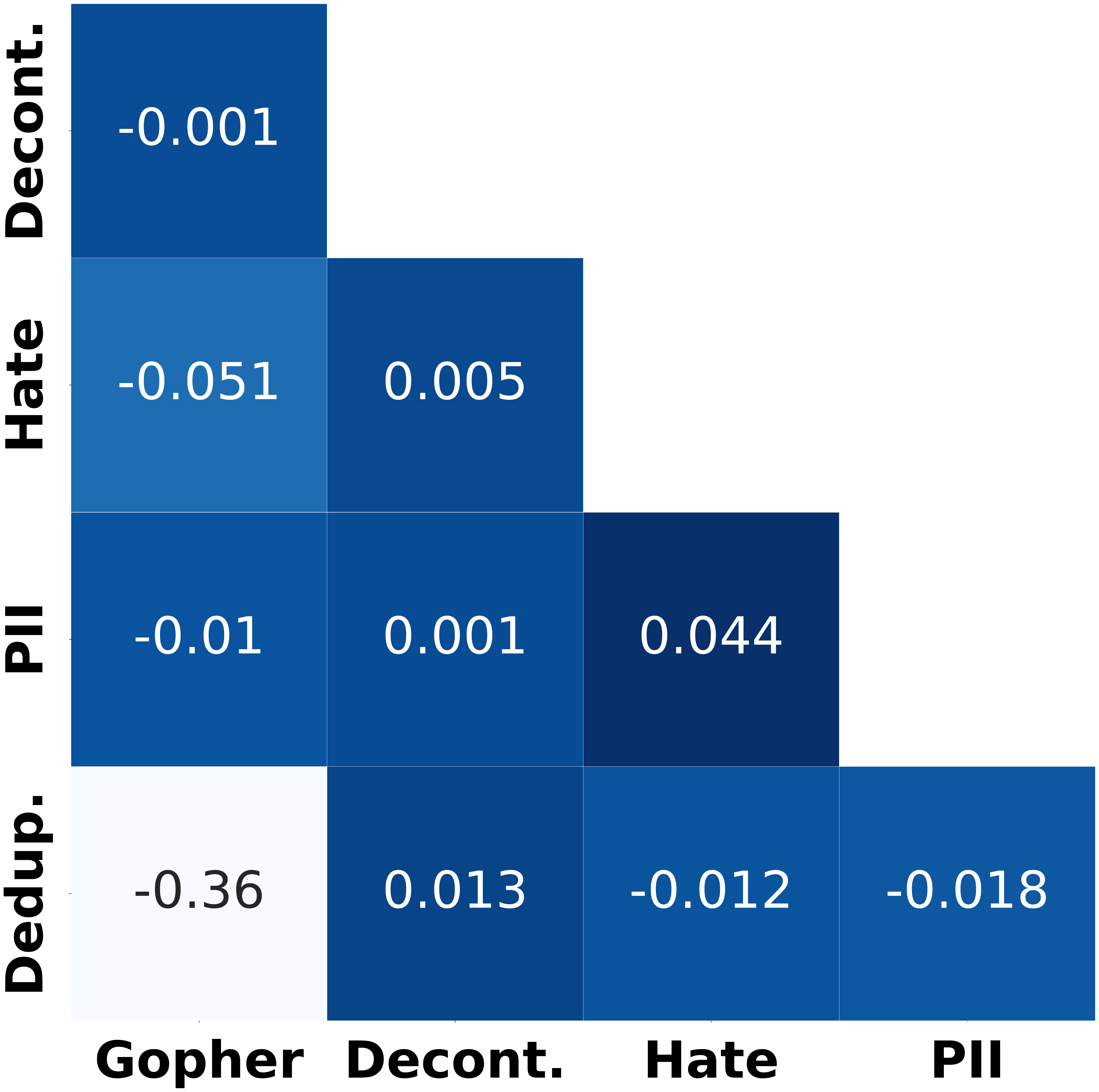}}}}    
    \caption{Pearson Correlation of various \dolma filters on the High, Medium, and Low buckets of our Common Crawl data, computed over 24M, 20M, and 43M documents, respectively. The filters are Gopher=Gopher rules from \citet{Rae2021ScalingLM}, Dedup.=Deduplication, PII=Personally Identifiable Information, Hate=Toxicity and Decont.=Decontamination.
    Calculated at the document-level: two filters contribute to positive correlation when any span in a document is tagged by both filters.
    We find our various filters remove different documents and are not redundant.
    }
    \label{fig:corr}
\end{figure*}

In order to further understand how filters described in \autoref{sub:web:quality}, \autoref{sub:web:content}, and \autoref{sub:web:dedup} interact with each other, we perform a correlation analysis on a subset of documents sampled from our pipeline. 
The correlation among the documents flagged for removal by our Common Crawl filters is depicted in Figure~\ref{fig:corr}.
Overall, we find that correlations are generally low, thus our filters select fairly different documents and are not redundant.

There is some positive correlation between our PII (Personal Identifiable Information) filters and filters removing hate speech. 
This is likely because hate speech is often directed at people. 
The Gopher filtering rules correlate negatively with our deduplication, especially for the high-perplexity tail part of our data. 
This is due to the Gopher rules removing many high-perplexity documents such as random strings, which are not caught by deduplication due to their randomness. 
As these random strings likely do not contribute to a better understanding of language, it is important to filter them out and thus rely on filters beyond deduplication.

\section{\dolma data distribution figures using WIMBD}
\label{sec:dolma-data-distribution-wimbd}

\begin{figure*}[h!]
    \subfloat[Web (URL) domains]{{\includegraphics[width=.33\linewidth]{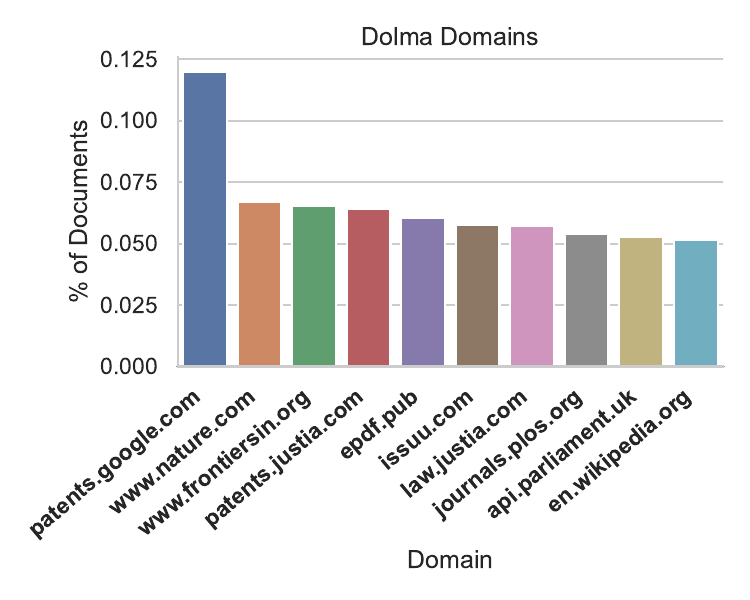} }}\label{fig:wimbd-domains}%
    \subfloat[Dates of documents]{{\includegraphics[width=.33\linewidth]{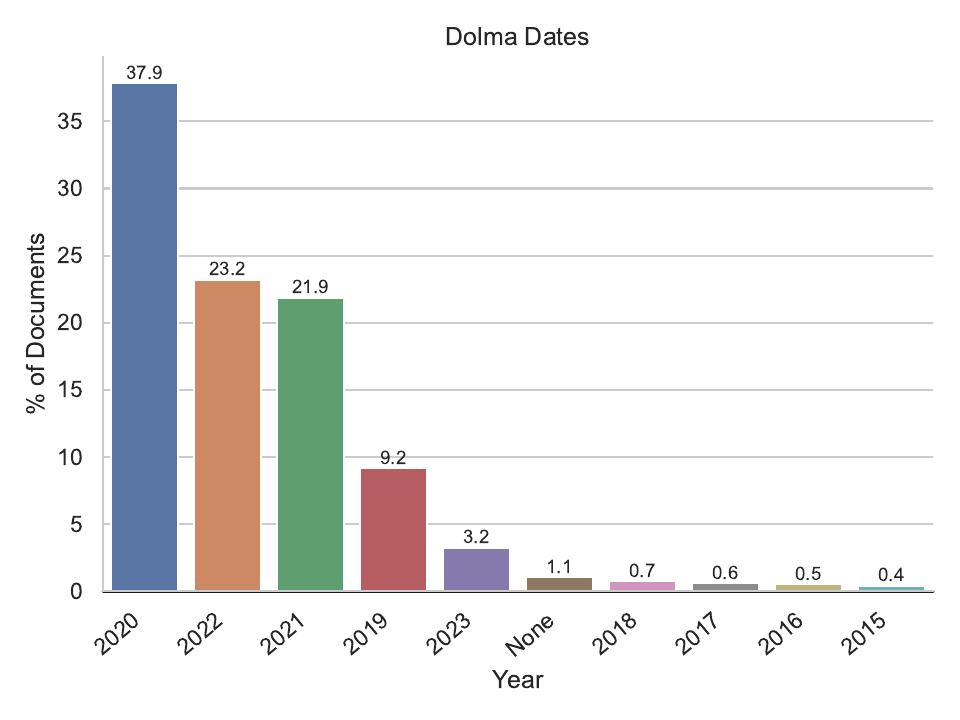} }}\label{fig:wimbd-dates}%
    \subfloat[Non-English languages]{{\includegraphics[width=.33\linewidth]{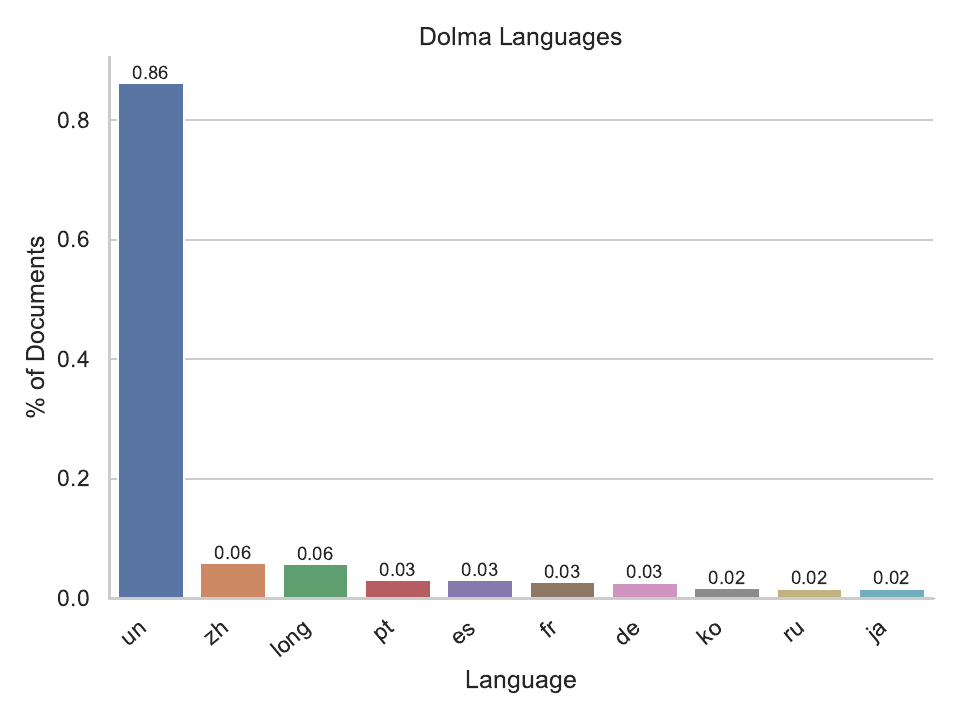} }}\label{fig:wimbd-lang}%
    \caption{
        Frequencies over different document metadata as computed using the WIMBD tool from \citet{wimbd}. 
        In subfigure (c), \texttt{un} denotes documents whose language could not be identified;
        \texttt{long} indicates documents that are too long to be processed with the tool's language ID module. 
    }%
    \label{fig:wimbd-domains-dates-lang-dist}%
\end{figure*}

We use the tool from \citet{wimbd} to inspect the final data composition in Figure~\ref{fig:wimbd-domains-dates-lang-dist}.
In particular, we analyze web domain, year, and language distributions.

We note that \dolma contains documents from a broad set of internet domains, mostly from 2020, 2022, and 2021.
The most common internet domains in \dolma, per token, are \url{patents.google.com}, followed by \url{www.nature.com} and \url{www.frontiersin.org}.
In fact, similar to other corpora reported in \citet{wimbd}, 63.6\% of Dolma's web documents are from `.com' sites (followed then by `.org' and `.co.uk' sites).
Finally, as all language identification tools are imperfect, we summarize what languages are remaining post English-only filtering: We find the most common language after English is not well identified (`un') with 0.86\% of the documents, followed by 0.06\% of the documents identified as Chinese.

\section{Test Set Contamination in \dolma}
\label{appendix:sec:decontam-dolma-perplexity-paloma}

\paragraph{Decontamination for perplexity evaluation.}
Using the paragraph deduplication tools described in \S\ref{sub:web:dedup}, 
we mark any paragraph in \dolma as contaminated if (\textit{i}) it is longer than 13 Unicode-segmented tokens\footnote{Like in \citet{wimbd}, we only consider paragraphs of sufficient length to avoid false positive matches.} and (\textit{ii}) it appears in any of the documents in Paloma.

To train \OlmoTiny, we remove any document with at least one paragraph marked as contaminated. 
This approach, while prone to false positives, has a negligible  impact on the final removal rate ($\leq 0.001\%$ characters in \dolma contaminated, $\leq 0.02\%$ of documents removed.), and reduces likelihood of false negatives.

\begin{figure*}[h]
    \centering
\subfloat{{\includegraphics[width=.9\linewidth]{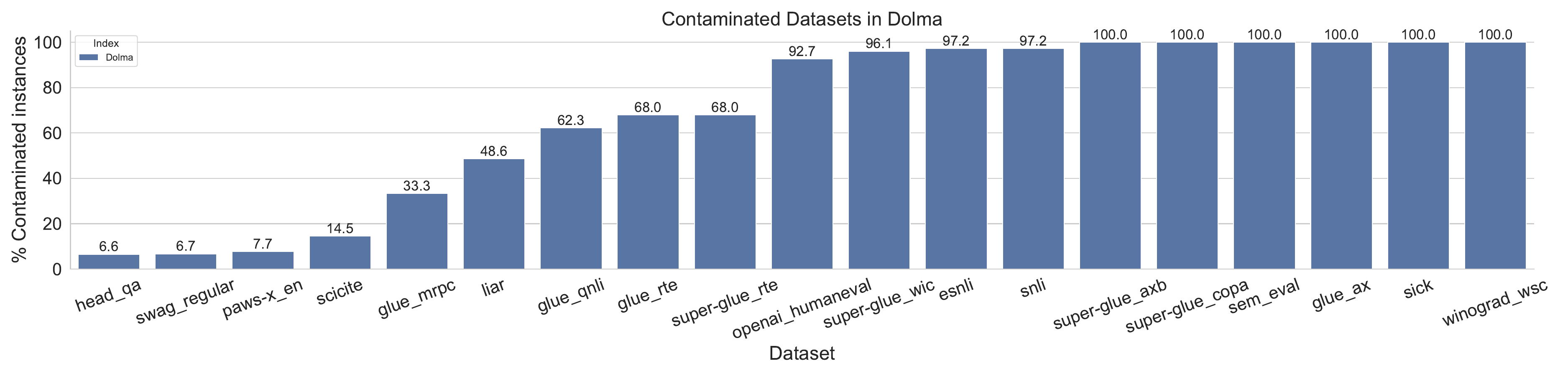} }}%
    \caption{Contamination percentages of datasets from PromptSource~\citep{bach2022promptsource}.}%
    \label{fig:wimbd-contamination}%
\end{figure*}

\paragraph{Decontamination of downstream tasks.} 
Using WIMBD~\cite{wimbd}, we analyze test set contamination in \dolma.
We find contamination of entire datasets from popular benchmarks like 
GLUE~\citep{wang-etal-2018-glue} and SuperGLUE~\citep{wang2019superglue}, and evaluation datasets like SNLI~\citep{bowman-etal-2015-large} and the Winograd Schema Challenge~\citep{wsc}.
Further analysis reveals that many of these sets are contaminated in our code subset, as public repositories in GitHub often contains copies of these datasets.
We report the top contaminated datasets in Figure \ref{fig:wimbd-contamination}.

Results indicate that portion of datasets in Promptsource appear in \dolma. 
Six datasets are completely contaminated (100\%): the Winograd Schema Challenge \citep{wsc}, 
Sick \citep{marelli-etal-2014-sick}, 
AX from GLUE \citep{wang-etal-2018-glue}, 
SemEval (specifically, Task 1 from 2014), 
COPA from SuperGLUE \citep{roemmele2011choice}, 
and AX$_b$ (the diagnostic task) from SuperGLUE \citep{wang2019superglue}.
In addition, other datasets are mostly contaminated, with over 90\% of their test sets appearing in Dolma documents: OpenAI HumanEval \citep{chen2021evaluating}, WIC from SuperGLUE \citep{pilehvar2019wic}, ESNLI \citep{esnli}, and SNLI \citep{snli}.
We note that the contaminated datasets have been excluded from the downstream tasks we use for model evaluation (\textit{c.r.f.} \autoref{sec:setup}).

\section{Strategies for Subsets Mixing and Upsampling with \dolma}
\label{sec:mixtures}

Like the pretraining corpora of nearly every large-scale language model, \dolma is a multi-source dataset. 
Training on \dolma thus requires a mixing strategy that determines how much data from each source to include, and potentially which sources to upsample. 
Like other multi-source corpora (e.g., ROOTS~\citep{Laurenccon2023TheBR}, the Pile~\citep{Gao2020ThePA}, RedPajama v1~\citep{together2023redpajama}),\footnote{RedPajama v1 was a reproduction of the multi-source corpus used in LLaMA~\citep{Touvron2023LLaMAOA}. RedPajama v2~\citep{redpajama2} focuses solely on Common Crawl and is thus single-source.} \dolma does not prescribe a single mixing strategy. 
We refer the reader to \citet{Rae2021ScalingLM} for an example of how one might programmatically search over mixing configurations to maximize performance. 
Here, we perform mixing experiments as an opportunity to answer some research questions about how different data sources interact. 
We use the same ablation setup described in \autoref{sec:creation}.

\paragraph{How much code is important for pretraining?} It is common practice for language models to be pretrained on some amount of code, even if code generation is not the intended task.
Some research has suggested that mixing code into training over plain text documents improves performance on reasoning tasks~\citep{madaan-etal-2022-language}.
We investigate whether this observation holds for models trained on \dolma, and if so, how much code is needed?

\begin{table*}[h]
    \centering
   
    \small
    \begin{tabular}{llll}
        \toprule
        \textbf{Dataset} & \textbf{0\% Code} & \textbf{5\% Code} & \textbf{15\% Code} \\
        \midrule
	bAbI (ICL) & 0.0~~~\textpm~0.0 & 8.8~~~\textpm~0.9 & 10.1 \textpm~2.8 \\
	WebNLG (ICL) & 16.8 \textpm~1.1 & 19.3 \textpm~1.1 & 22.0 \textpm~1.3 \\
	GSM8K (FT) & 0.0~~~\textpm~0.0 & 0.0~~~\textpm~0.0 & 0.0~~~\textpm~0.0 \\
	GSM8K+PAL (FT) & 11.8 \textpm~0.8 & 14.2 \textpm~1.3 & 14.7 \textpm~0.9 \\
    \bottomrule
    
    \end{tabular}
    \caption{
    Performance of three models pre-trained with increasing amounts of code on three datasets, across 5 random seeds. 
    We measure exact match for bAbI and GSM8K, and Rouge-2 for WebNLG.
    }
    \label{tab:code-reasoning}
\end{table*}

We create three mixtures from the C4 and Stack subsets containing 0\%, 5\% and 15\% of code data.
On each, we train a 1B model. 
We evaluate these models on three different reasoning tasks: bAbI~\citep{Weston2015TowardsAQ}, WebNLG \cite{gardent-etal-2017-creating} and GSM8k \cite{Cobbe2021TrainingVT}. For the first two tasks, we follow the experimental setup of \citet{muennighoff2023scaling} and evaluate each model in an ICL setup with a changing number of demonstrations (0-5) across 5 random seeds. \citet{muennighoff2023scaling} show that adding code to pre-training data improves ICL performance on bAbI and WebNLG and they suggest that code improves long-range state-tracking capabilities. Our experiments, as shown in Table~\ref{tab:code-reasoning}, corroborate these findings: while the C4-only model fails on all bAbI tasks, adding code improves performance, with a similar trend for WebNLG. 

On the more difficult GSM8k benchmark, all models failed to get any correct answer in an ICL setup, and even when fine-tuning the models on the entire training set. However, we find that by fine-tuning on program-aided output, where questions are solved by writing Python snippets as described in \cite{gao2022pal}, code models outperform the C4-only model. These results show that models pre-trained on code can leverage code generation to answer challenging reasoning tasks even when the original task does not directly involve code.

\paragraph{Evaluating mixing strategies for pretraining on \dolma}
While \dolma does not prescribe a specific source mixture, we analyze some commonly used strategies\footnote{
    We did not include any social data in these mixes as it was not ready at the time of this experiment.
} and compare their effect using the Paloma evaluation suite~\citep{paloma}.
Specifically, we present and evaluate four possible data mixtures in \autoref{table:mix}.

\begin{table*}[h]
\centering
\small
\renewcommand{\arraystretch}{1.2}
\begin{tabular}{lm{19em}cc}
    \toprule
    \textbf{Mix Name} &
    \textbf{Description} &
    \textbf{Sampling} &
    \textbf{Proportion} \\
    \midrule
    \textbf{
        Na\"ive
    } & {
        Sample each source in \autoref{tab:statistics} equally.
    }  &  {
        \renewcommand{\arraystretch}{1}
        \scriptsize
        \begin{tabular}{@{}lr@{}}
        {\tiny\dolmaWeb}~Web & 100\% \\
        {\tiny\dolmaCode}~Code & 100\% \\
        {\tiny\dolmaRefs~\dolmaPapers}~Ref. & 100\% \\
        {\tiny\dolmaBooks}~Books & 100\% \\
        \end{tabular}
    }  & {
        \renewcommand{\arraystretch}{1}
        \scriptsize
        \begin{tabular}{@{}lr@{}}
        {\tiny\dolmaWeb}~Web & 83.5\% \\
        {\tiny\dolmaCode}~Code & 13.8\% \\
        {\tiny\dolmaRefs~\dolmaPapers}~Ref. & 2.5\% \\
        {\tiny\dolmaBooks}~Books & 0.2\% \\
        \end{tabular}
        \vspace{1em}
    } \\
    \textbf{Web Only} &
    {
        Similar to~\citet{ayoola-etal-2022-refined}, we test a mixture that only uses web data.

    }  &  {
        \renewcommand{\arraystretch}{1}
        \scriptsize
        \begin{tabular}{@{}lr@{}}
        {\tiny\dolmaWeb}~Web & 100\% \\
        {\tiny\dolmaCode}~Code & 0\% \\
        {\tiny\dolmaRefs~\dolmaPapers}~Ref. & 0\% \\
        {\tiny\dolmaBooks}~Books & 0\% \\
        \end{tabular}
    }  & {
        \renewcommand{\arraystretch}{1}
        \scriptsize
        \begin{tabular}{@{}lr@{}}
        {\tiny\dolmaWeb}~Web & 100\% \\
        {\tiny\dolmaCode}~Code & 0\% \\
        {\tiny\dolmaRefs~\dolmaPapers}~Ref. & 0\% \\
        {\tiny\dolmaBooks}~Books & 0\% \\
        \end{tabular}
    }  \\
    \textbf{
        Reference+
    } & {
        It is common practice to upsamole knowledge-intensive documents when composing training mixture. 
        In our case, we upsample the PeS2o papers, Wikipedia, Wikibooks, and Gutenberg books subsets by 2x.
    }&  {
        \renewcommand{\arraystretch}{1}
        \scriptsize
        \begin{tabular}{@{}lr@{}}
        {\tiny\dolmaWeb}~Web & 100\% \\
        {\tiny\dolmaCode}~Code & 100\% \\
        {\tiny\dolmaRefs~\dolmaPapers}~Ref. & 200\% \\
        {\tiny\dolmaBooks}~Books & 200\% \\
        \end{tabular}
    }  & {
        \renewcommand{\arraystretch}{1}
        \scriptsize
        \begin{tabular}{@{}lr@{}}
        {\tiny\dolmaWeb}~Web & 81.2\% \\
        {\tiny\dolmaCode}~Code & 13.5\% \\
        {\tiny\dolmaRefs~\dolmaPapers}~Ref. & 4.9\% \\
        {\tiny\dolmaBooks}~Books & 0.4\% \\
        \end{tabular}
        \vspace{1em}
    }  \\
    \textbf{
        Gopher-like
    } & {
        Following~\citet{Rae2021ScalingLM}, we create a mix that is heavily biased towards reference material. 
        As we do not have access to the same sources, an exact replication of their mix is not possible.
    }&  {
        \renewcommand{\arraystretch}{1}
        \scriptsize
        \begin{tabular}{@{}lr@{}}
        {\tiny\dolmaWeb}~Web & 17\% \\
        {\tiny\dolmaCode}~Code & 8\% \\
        {\tiny\dolmaRefs~\dolmaPapers}~Ref. & 200\% \\
        {\tiny\dolmaBooks}~Books & 200\% \\
        \end{tabular}
    }  & {
        \renewcommand{\arraystretch}{1}
        \scriptsize
        \begin{tabular}{@{}lr@{}}
        {\tiny\dolmaWeb}~Web & 68.4\% \\
        {\tiny\dolmaCode}~Code & 5.4\% \\
        {\tiny\dolmaRefs~\dolmaPapers}~Ref. & 24.2\% \\
        {\tiny\dolmaBooks}~Books & 2.0\% \\
        \end{tabular}
    }  \\
    \bottomrule
\end{tabular}
\vspace{1em}
\caption{
    Overview of the mixtures and their composition. 
}
\label{table:mix}
\end{table*}

\begin{figure*}[h]
    \centering
    \begin{subfigure}{.31\textwidth}
    \includegraphics[width=\linewidth]{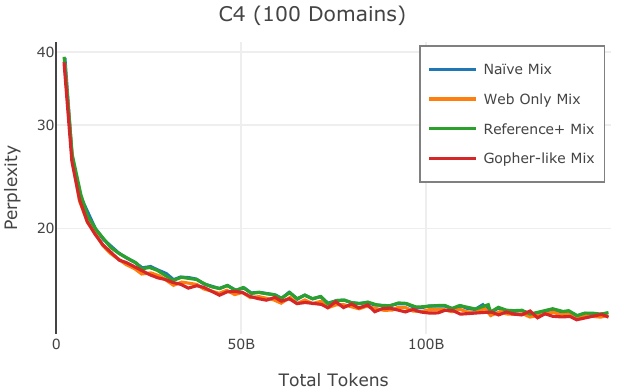}
    
    \label{fig:dolma_mix:c4_100}
    \end{subfigure}
    \quad
    \begin{subfigure}{.31\textwidth}
    \includegraphics[width=\linewidth]{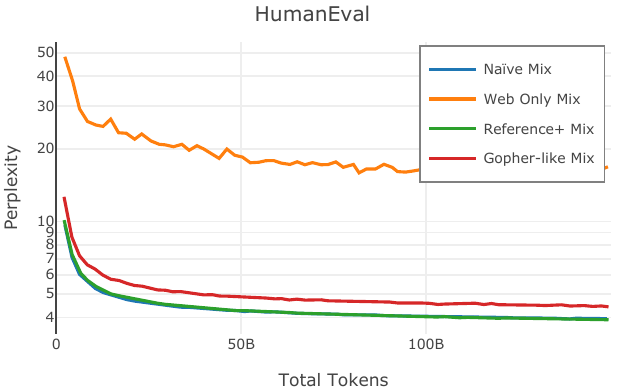}
    \end{subfigure}
    \quad
    \begin{subfigure}{.31\textwidth}
    \includegraphics[width=\linewidth]{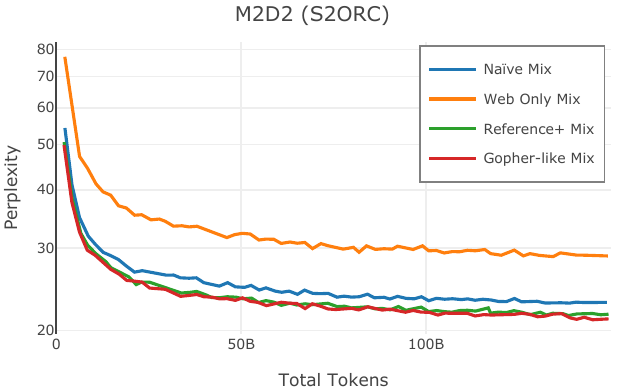}
    \end{subfigure}
    \caption{
    1B model ablations for different proportions of \dolma data. All mixture perform similarly on web data (left), while excluding code increases perplexity on code datasets (center). 
    Finally, increasing reference material by upsampling papers and Wikipedia yields lower perplexity on S2ORC (right). 
    Overall, source distribution is linked to downstream capabilities; thus, \dolma users should sample subsets according to their needs.
}%
    \label{fig:dolma_mix}
\end{figure*}

We show results of mixtures in \autoref{fig:dolma_mix}. 
Overall, we observe that the different mixtures have an effect on the ability of resulting models to capture specific subdomains. 
All mixtures show similar perplexity scores on pages sampled from 100 domains from C4 (\autoref{fig:dolma_mix}, left), indicating their general effectiveness at modeling web documents.  
On the other hand, we note how models struggle to model specialized domains unless they are exposed to them. 
As an example, a model trained on the \textit{Web-only} mix struggles to represent data in the code domain (\autoref{fig:dolma_mix}, center, HumanEval).
Finally, we use results on the S2ORC subset of M2D2, which consists of academic papers, to illustrate how different data mixtures affect perplexity. 
As is it the case with code, \textit{Web-only} model exhibits higer perplexity due to domain mismatch. 
On the other hand, models trained on \textit{Reference+} and \textit{Gopher-like} mixes achieve lower perplexity than the model trained on the \textit{Na\"ive} mix, due to more in-domain content. 
However, we note that, despite significant differences in the amount of academic papers between \textit{Reference+} and \textit{Gopher-like} (4.9\% vs 24.2\%), they achieve nearly identical results, suggesting that even a relatively small percentage of in-domain data is sufficient to achieve good domain fit.

\section{Datasheet}
\label{sec:datasheet}

Following the template by \citet{gebru2021datasheets}, we provide a Datasheet for \dolma.

\subsection{Motivation for Dataset Creation}

\paragraph{Why was the dataset created?}~ 

\dolma was created with the primary purpose of training \olmo autoregressive language model. It is a mixture of documents from multiple data sources. Documents have been transformed using a combination of rule-based and statistical tools to extract textual content, remove layout information, and filter for English content. 

\dolma contains data sourced from different domains. In particular, it contains a mixture of text obtained from a web scrape, scientific content extracted from academic PDFs and its associated metadata, code over a variety of programming languages, reference material from Wikipedia and Wikibooks, as well as public domain books from Project Gutenberg.

\paragraph{What (other) tasks could the dataset be used for?}~

We expect this dataset to be useful to train other language models, either in its current form or through further filtering and combining it with other datasets. 

Beside language model training, this dataset could be used to study interaction between pretraining corpora and models trained on them. For example, one could study provenance of generations from the model, or perform further corpus analysis. 

Specific subset of \dolma could be used to train domain specific models. For example, the code subset could be used to train an AI programming assistant. 

\paragraph{Are there obvious tasks for which it should not be used?}~

Due to the myriad transformations applied to the original source materials to derive our dataset, we believe it is ill-suited as a replacement for users seeking to directly consume the original content. 
We refer users of our dataset to our license and terms on the Hugging Face Hub \href{https://huggingface.co/datasets/allenai/dolma}{\path{huggingface.co/datasets/allenai/dolma}} which detail any use restrictions.

\paragraph{Has the dataset been used for any tasks already?}~

The \olmo~\citep{olmo20247b} model family is trained on this dataset. 

\paragraph{If so, where are the results so others can compare?}~

Experimental results are detailed in this paper and in the \olmo~\citep{olmo20247b} manuscript.

\paragraph{Who funded the creation of the dataset?}~

All individuals who are responsible for this dataset are employed by the Allen Institute for AI. Similarly, computing resources are provided by AI2. 

\paragraph{If there is an associated grant, provide the grant number.}~

Compute for the \olmo project is provided by AMD and CSC, using GPUs on the LUMI supercomputer.

\subsection{Dataset Composition}

\paragraph{What are the instances? Are there multiple types of instances?}~

Instances are plain-text spans on English text or computer code.
Each instance was obtained by processing web pages (which might include news, documents, forums, etc), academic articles, computer code from GitHub, encyclopedic content from Wikipedia, or Project Gutenberg books. 

\paragraph{Are relationships between instances made explicit in the data?}~

Metadata for subsets of \dolma could be used to reconstruct relationships between items:

\begin{itemize}[leftmargin=1em]
    \item \textbf{Common Crawl}. Each document uses the URL of the web page from which it was extracted as its identifier; therefore, it can be used to identify relationships between documents.
    \item \textbf{C4}. The URL of each web page from which documents were extracted is included as metadata; therefore, it can be used to identify relationships between documents.
    \item \textbf{Reddit}. The originating subreddits and thread ids of documents are included in the metadata. 
    \item \textbf{Semantic Scholar}. The id of each document is the Semantic Scholar Corpus ID of its corresponding manuscript. Metadata for each manuscript can be obtained using the Semantic Scholar APIs~\citep{kinney2023semantic}.
    \item \textbf{GitHub}. The name of the GitHub repository each document belongs to is included as metadata. 
    \item \textbf{Project Gutenberg}. The title of each book is included as the first line of each document.
    \item \textbf{Wikipedia}, \textbf{Wikibooks}. For both, metadata includes the URL corresponding to the page content was extracted from. Structure and connections between documents can be recovered through the URL.
\end{itemize}

\paragraph{How many instances of each type are there?}~

Summary statistics are reported in Table 1.

\paragraph{What data does each instance consist of? ``Raw'' data (e.g., unprocessed text or images)? Features/attributes?}~

For each source, raw data is not available directly but could be recovered using source-specific methods:

\begin{itemize}[leftmargin=1em]
    \item \textbf{Common Crawl}. We obtain data from common crawl snapshots from 2020-05 to 2023-06. WARC files from Common Crawl can be intersected with \dolma ids to recover original HTML files.
    \item \textbf{C4}. We obtained this corpus from the Hugging Face Hub\footnote{\label{foot:c4}\href{https://huggingface.co/datasets/allenai/c4}{\path{hf.co/datasets/allenai/c4}}}. In turn, documents in C4 have been derived from a Common Crawl snapshot for 04/2019. URLs in C4 can be used to recover HTML files.
    \item \textbf{Reddit}. The complete set of monthly data dumps used in this work are no longer distributed by Pushshift, however they can still be obtained through torrents and some public web archives.   
    \item \textbf{Semantic Scholar}. peS2o is derived from S2ORC~\citep{lo-etal-2020-s2orc}. Original parsed documents can be obtained from extracting documents in S2ORC that share the same ID with peS2o. Further, metadata in S2ORC can be used to obtain original PDF.
    \item \textbf{GitHub}. The filename and repository name, both available in metadata, can be used to recover original file contents.
    \item \textbf{Project Gutenberg}. The title of each book is the first line of each document.
    \item \textbf{Wikipedia}, \textbf{Wikibooks}. For both, metadata includes the URL corresponding to the page content was extracted from. Structure and connections between documents can be recovered through the URL.
\end{itemize}

\paragraph{Is there a label/target associated with instances? If the instances are related to people, are subpopulations identified (e.g., by age, gender, etc.) and what is their distribution?}~

There are no labels associated with instances. 
Many text instances were likely created by people or groups of people, but in the vast majority of cases authorship information is unavailable let alone subpopulation metadata. we leave aggregation and reporting of these statistics to future work.

\paragraph{Is everything included or does the data rely on external resources? (e.g., websites, tweets, datasets) If external resources, a) are there guarantees that they will exist, and remain constant, over time; b) is there an official archival version. Are there licenses, fees or rights associated with any of the data?}~

The data are derived from the web and the original resources may not persist over time. However, each source represents an archival snapshot of that data that should remain fixed and available: 

\begin{itemize}[leftmargin=1em]
    \item \textbf{Common Crawl}. The Common Crawl data is available on Amazon S3 as part of the Amazon Web Services’ Open Data Sponsorship program and can be freely downloaded\footnote{\href{https://commoncrawl.org/the-data/get-started/}{\path{commoncrawl.org/the-data/get-started}}}. 
    We followed Common Crawl terms of use\footnote{\label{foot:cc-start}\href{https://commoncrawl.org/terms-of-use/}{\path{commoncrawl.org/terms-of-use}}}.
    \item \textbf{C4}. This corpus can be obtained from from the Hugging Face Hub\fnref{foot:c4} and is released under ODC-By 1.0~\citep{odc-by}. 
    \item \textbf{Reddit}. Pushshift no longer distributes this dataset due to changes to the Reddit API's terms. Unofficial copies of the data might be be available through torrents and some public web archives. Pushshift data dumps inherit\footnote{\href{https://www.reddit.com/r/pushshift/comments/d6luj5/comment/f0ugpqp}{\path{reddit.com/r/pushshift/comments/d6luj5/comment/f0ugpqp}}} the Terms of use of the Reddit API at the time of their collection (March 2023).
    \item \textbf{Semantic Scholar}. peS2o is derived from S2ORC~\citep{lo-etal-2020-s2orc}. S2ORC is released through the Semantic Scholar Public API\footnote{\href{https://www.semanticscholar.org/product/api}{\path{semanticscholar.org/product/api}}} under ODC-By 1.0~\citep{odc-by}.
    \item \textbf{GitHub}. The corpus is available on the Hugging Face Hub\footnote{\label{foot:stack}\href{https://huggingface.co/datasets/bigcode/the-stack-dedup}{\path{hf.co/datasets/bigcode/the-stack-dedup}}} and consists of code released under a variety of permissive licenses. More details including terms of use for hosting or sharing the corpus are provided in the datacard at the link above.
    \item \textbf{Project Gutenberg}. Project Gutenberg consists of books that are not protected under U.S. copyright law. The corpus is available at \href{https://www.gutenberg.org/}{\path{gutenberg.org}}.
    \item \textbf{Wikipedia}, \textbf{Wikibooks}. Wikimedia data dumps are freely available\footnote{\label{foot:wiki}\href{https://dumps.wikimedia.org}{\path{dumps.wikimedia.org}}} and released under CC BY-SA 4.0 license~\citep{cc-by-sa-4}.
\end{itemize}

\paragraph{Are there recommended data splits or evaluation measures? (e.g., training, development, testing; accuracy/AUC)}~

No. 
See current manuscript Section \S\ref{sec:experimental-methodology}.

\paragraph{What experiments were initially run on this dataset? Have a summary of those results and, if available, provide the link to a paper with more information here.}~

See current manuscript Section \S\ref{sec:experimental-methodology} for description of data ablation methodology, and remainder of paper for full set of experiments. 
Every experimental result is available through links provided in the manuscript.

\subsection{Data Collection Process}
\label{datasheet:collection}

\paragraph{How was the data collected? (e.g., hardware apparatus/sensor, manual human curation, software program, software interface/API; how were these constructs/measures/methods validated?)}~

Data acquisition for each subset was performed as follows:

\begin{itemize}[leftmargin=1em]
    \item \textbf{Common Crawl}. snapshots were downloaded from Common Crawl's official S3 bucket\footnote{\texttt{s3://commoncrawl/}} using the \texttt{cc\_net} pipeline~\citep{wenzek-etal-2020-ccnet}. Data was obtained between March 17$^\textrm{th}$ and March 27$^\textrm{th}$, 2023.
    \item \textbf{C4}. We clone C4 from the Hugging Face Hub\fnref{foot:c4} using Git with the Git-LFS extension. Repository cloned on May 24$^\textrm{th}$, 2023.
    \item \textbf{Reddit}. Reddit was acquired in the form of monthly data dumps of  comments and submissions collected and distributed by the Pushshift project\footnote{\href{https://files.pushshift.io/reddit/submissions/}{\path{files.pushshift.io/reddit/submissions}} and \href{https://files.pushshift.io/reddit/comments/}{\path{files.pushshift.io/reddit/comments}}}. We used the complete set of 422 publicly available dumps (208 comments, 214 submissions) spanning a period from 06/2005--03/2023. The majority of Dumps were acquired in March, 2023 with the last dumps downloaded in May of 2023.
    \item \textbf{Semantic Scholar}. We clone peS2o from the Hugging Face Hub\footnote{\href{https://huggingface.co/datasets/allenai/peS2o}{\path{hf.co/datasets/allenai/peS2o}}} using Git with the Git-LFS extension. We use pes2o V2. Repository cloned on June 30$^\textrm{th}$, 2023.
    \item \textbf{GitHub}. We clone The Stack (deduplicated) from the Hugging Face Hub\fnref{foot:stack} using Git with the Git-LFS extension. Repository cloned on May 28$^\textrm{th}$, 2023.
    \item \textbf{Project Gutenberg}. Data was downloaded directly from \url{gutenberg.org}. We used \texttt{GutenbergPy}~\citep{GutenbergPy} to extract books. Website accessed on April 3$^\textrm{rd}$, 2023.
    \item \textbf{Wikipedia}, \textbf{Wikibooks}. Dumps were downloaded from Wikimedia's website\fnref{foot:wiki}. We use the dump from March 20$^\textrm{th}$, 2023.
\end{itemize}

\paragraph{Who was involved in the data collection process? (e.g., students, crowdworkers) How were they compensated? (e.g., how much were crowdworkers paid?)}~

Data was collected and postprocessed by full-time employees at the Allen Institute for AI. No instances in this dataset are manually annotated.

\paragraph{Over what time-frame was the data collected? Does the collection time-frame match the creation time-frame?}~

Please see list above. 

\paragraph{How was the data associated with each instance acquired? Was the data directly observable (e.g., raw text, movie ratings), reported by subjects (e.g., survey responses), or indirectly inferred/derived from other data (e.g., part of speech tags; model-based guesses for age or language)? If the latter two, were they validated/verified and if so how?}~

Any metadata associated with each instance was obtained directly from each source.

\paragraph{Does the dataset contain all possible instances? Or is it, for instance, a sample (not necessarily random) from a larger set of instances? 
If the dataset is a sample, then what is the population? What was the sampling strategy (e.g., deterministic, probabilistic with specific sampling probabilities)? Is the sample representative of the larger set (e.g., geographic coverage)? If not, why not (e.g., to cover a more diverse range of instances)? How does this affect possible uses?}~

Sampling for each subset was performed as follows:

\begin{itemize}[leftmargin=1em]
    \item \textbf{Common Crawl}. Common Crawl is not a representative sample of the web. Summary statistics about Common Crawl are reported through the \texttt{cc-crawl-statistics}~\citep{cc-crawl-statistics} project, available at \href{https://commoncrawl.github.io/cc-crawl-statistics/}{\texttt{commoncrawl.github.io/cc-crawl-statistics}}. \dolma uses  Common Crawl snapshots from \texttt{2020-05} to \texttt{2023-06}\footnote{Common Crawl snapshots follow naming convention \texttt{xxxx-yy}, where \texttt{xxxx} is the year the snapshot was finalized, and \texttt{yy} is the week, ranging from \texttt{01} to \texttt{52}.}.
    \item \textbf{C4}. We use C4 in its entirety. 
    \item \textbf{Reddit}. We use all available Reddit content from from 06/2005--03/2023.
    \item \textbf{GitHub}. We use The Stack (deduplicated) in its entirety. 
    \item \textbf{Semantic Scholar}. We use pes2o V2 in its entirety. 
    \item \textbf{Project Gutenberg}. We process all Gutenberg books. 
    \item \textbf{Wikipedia}, \textbf{Wikibooks}. We use the \textit{English} and \textit{Simple} subset of Wikipedia and Wikibooks in their entirety.  
\end{itemize}

\paragraph{Is there information missing from the dataset and why? (this does not include intentionally dropped instances; it might include, e.g., redacted text, withheld documents) Is this data missing because it was unavailable?}~

Common Crawl is the only source we did not use in its entirety. 
We use only about a quarter of all snapshots available. 
This amount was deemed sufficient for the goal of the \dolma project.
We decided to use the 24 most recent Common Crawl snapshots at the time.

\paragraph{Are there any known errors, sources of noise, or redundancies in the data?}~

Not that we are aware of, although a negligible portion of Common Crawl data could have been lost due to network issues with S3 storage. 
When accessing Common Crawl, we implemented retry mechanisms, but copy could have failed due to exceeding the retry limits. 

\subsection{Data Preprocessing}
\label{datasheet:processing}

\paragraph{What preprocessing/cleaning was done? (e.g., discretization or bucketing, tokenization, part-of-speech tagging, SIFT feature extraction, removal of instances, processing of missing values, etc.)}~

All data sources are filtered using FastText language identification models~\citep{joulin2016fasttext,joulin2016bag} with an English threshold of 0.5.

For the \textbf{Common Crawl} and \textbf{C4} subsets, we use the following filters that substantially modify the original data. 
Note that data might be tagged for removal by one or more filter.

\begin{itemize}[leftmargin=1em]
    \item \textbf{\textit{Only \underline{Common Crawl}, as part of their distribution pipeline}}: Linearize all HTML into plain text files (\texttt{WET} files generation\fnref{foot:cc-start});
    \item \textbf{\textit{Only \underline{Common Crawl}, as part of CCNet pipeline}}: We remove frequently occurring paragraph in Common Crawl by identifying repeated paragraphs on small subsets of each snapshots. This step gets rid of headers that are shared across many pages, such as navigational headers. 
    Removal is operationalized as follows: given $1\ldots,n,\ldots,N$ shards each snapshot is comprised to, group shards in sets $S=\{n-k, n\}$; then, remove exact duplicates of paragraphs in $S$. Paragraphs are defined as newline-separated slices of documents, and compared using their SHA1. We choose $k$ such that each set is at most 20GB\footnote{This is a slight modification of the original CCNet pipeline, where $k$ is chose so that each set is 2\% of snapshot. We chose to use a fixed shard size, rather an a percentage of the corpus, because fixed size is more predictable in terms of resource usage, leading to less-error prone code. Conceptually it’s equivalent to putting a threshold on the absolute probability of a paragraph occurring}. (\textit{approximately 70\% of paragraph removed});
    \item \textbf{\textit{Only \underline{Common Crawl}, deduplication by URL}}: We deduplicate pages by URL (\textit{53\% of duplicates removed});
    \item \textbf{Language identification}: remove all documents with an English score lower than 0.5, as determined by FastText language identification models~\citep{joulin2016fasttext,joulin2016bag} (\textit{removed 61.69\% of web pages by size});
    \item \textbf{Quality filter\footnote{\label{foot:quality}The term ``quality filter'', while widely used in literature, does not appropriately describe the outcome of filtering a dataset. Quality might be perceived as a comment on the informativeness, comprehensiveness, or other characteristics valued by humans. However, the filters used in \dolma and other language models efforts select text according to criteria that are inherently ideological~\citep{gururangan-etal-2022-whose}.}}: Remove documents with more than half of their line not ending in ``\texttt{.}'', ``\texttt{?}'', ``\texttt{!}'', or ``\texttt{"}''. (\textit{22.73\% of characters tagged for removal});
    \item \textbf{Quality filter}\fnref{foot:quality}: Remove any document that does not pass any of the Gopher rules~\citep{Rae2021ScalingLM} (\textit{15.23\% of characters tagged for removal}); 
    \begin{itemize}
        \item Fraction of characters in most common ngram greater than a threshold\footnote{For bigrams, threshold of 0.20. For trigrams, 0.18. For 4-grams, 0.16.}
        \item Fraction of characters in duplicate ngrams greater than a threshold\footnote{For 5-grams, 0.15. For 6-grams, 0.14. For 7-grams, 0.13. For 8-grams, 0.12. For 9-grams, 0.11. For 10-grams, 0.10.}
        \item Contains fewer than 50 or more than 100K words
        \item Median word length is less than 3 or greater than 10
        \item Symbol to word ratio greater than 0.10
        \item Fraction of words with alpha character less than 0.80
        \item Contains fewer than 2 of a set of required words\footnote{``the'', ``be'', ``to'', ``of'', ``and'', ``that'', ``have'', ``with''}
        \item Fraction of lines in document starting with bullet point greater than 0.90
        \item Fraction of lines in document ending with ellipsis greater than 0.30
        \item Fraction of lines in document that are duplicated greater than 0.30
        \item Fraction of characters in duplicated lines greater than 0.30
    \end{itemize}
    \item \textbf{Quality filter}\fnref{foot:quality}: Remove any document that contains a token or sequence of tokens repeating over 100 times\footnote{\label{foot:reps}We use \href{https://huggingface.co/allenai/gpt-neox-olmo-dolma-v1_5}{\path{allenai/gpt-neox-olmo-dolma-v1\_5}} to obtain tokens.}
    (\textit{0.003\% of characters tagged for removal}); 
    \item \textbf{Content filter}: Remove sentences that get ranked as toxic by a FastText classifier (score above $0.4$). We train a bigram classifier on the Jigsaw dataset~\citep{jigsaw} (\textit{1.01\% of data tagged for removal});
    \item  \textbf{Content filter}: Mask Personal Identifiable Information (PII) using regular expressions that identify emails, phone numbers, 
    and IP addresses; pages containing 6 or more PIIs are completely removed from the corpus (\textit{0.05\% tagged for masking, 0.11\% tagged for removal}); 
    \item \textbf{Exact document deduplication}: duplicate documents the same text. No punctuation or whitespace is removed. Empty documents count as duplicates (\textit{14.9\% of documents tagged for removal}).
    \item \textbf{\textit{Only \underline{Common Crawl}, deduplication by paragraph}}: We deduplicate the web subset at a paragraph level using a Bloom filter (\textit{19.1\% of UTF-8 characters tagged for removal}).
\end{itemize}

For the \textbf{Reddit} subset, we use the following filters that substantially reduce the original data. 

\begin{itemize}[leftmargin=1em]
    \item \textbf{Language identification}: remove all documents with an English score lower than 0.5, as determined by a FastText language identification model.
    \item \textbf{Quality filter\fnref{foot:quality}}: Remove comments and submissions shorter than 500 characters in length.
    \item \textbf{Quality filter\fnref{foot:quality}}: Remove user comments with fewer than three upvotes (Reddit users vote on the quality of submissions and comments).
    \item \textbf{Content filter\fnref{foot:quality}}: Remove comments and submissions from banned,  toxic, or NSFW subreddits.
    \item \textbf{Content filter\fnref{foot:quality}}: Remove sentences that get ranked as toxic or as hatespeech by a FastText classifier (score above $0.4$).
    \item  \textbf{Content filter}: Mask Personal Identifiable Information (PII) using regular expressions that identify emails, phone numbers, 
    and IP addresses
    \item \textbf{Deduplication}: We deduplicate comments and submissions (jointly) at a paragraph level using a Bloom filter.
\end{itemize}

For the code subset derived from The Stack (deduplicated), we use the following filters:

\begin{itemize}[leftmargin=1em]
    \item \textbf{Language filter}: Removed files associated with the following programming languages:
    \begin{itemize}
        \item Data or numerical content: \textit{csv, json, json5, jsonld, jsoniq, svg}
        \item Assembly code: \textit{assembly}
    \end{itemize}
    \item \textbf{Quality filter\fnref{foot:quality}}: Removed copyright statements in code files from document preamble\footnote{Code license and provenance is still tracked in metadata.};
    \item \textbf{Quality filter\fnref{foot:quality}}: Removed documents matching any of the RedPajama v1~\citep{together2023redpajama} code filters (\textit{41.49\% of data tagged for removal}):
    \begin{itemize}
    \item Maximum line length > 1000 characters.
    \item Average line length > 100 characters.
    \item Proportion of alpha-numeric characters < 0.25.
    \item Ratio of alphabetical characters to number of tokens < 1.5\footnote{Tokens counted using whitespace tokenizer}.
    \end{itemize}
    \item \textbf{Quality filter\fnref{foot:quality}}: Removed documents matching any of the following Starcoder filters~\citep{Li2023StarCoderMT}:
    \begin{itemize}
        \item Contains XML template code.
        \item HTML code-to-text ratio <= 0.2.
        \item Java, Javascript, Python code-to-comment ratio <= 0.01 or > 0.8.
    \end{itemize}
    \item  \textbf{Content filter}: Mask Personal Identifiable Information (PII) using regular expressions that identify emails, phone numbers, 
    and IP addresses; pages containing 6 or more PIIs are completely removed from the corpus. 
\end{itemize}

The \textbf{Common Crawl}, \textbf{C4}, \textbf{Reddit}, and \textbf{Code} subsets used the same regular expressions for identifying PII:

\begin{itemize}
    \item \textbf{Email addresses:}\\
    \texttt{[.\textbackslash{}s@,?!;:)(]*([\textbackslash{}\textasciicircum{}\textbackslash{}s@]+@[\textbackslash{}\textasciicircum{}\textbackslash{}s@,?!;:)}
    \newline\texttt{(]+?)[.\textbackslash{}s@,?!;:)}\texttt{(]?[\textbackslash{}s\textbackslash{}n\textbackslash{}r]
}
    \item \textbf{IP addresses:} \\
    \texttt{\textbackslash{}s+\textbackslash{}(?(\textbackslash{}d\{3\})\textbackslash{})?[-\textbackslash{}.}
    \newline\texttt{]*(\textbackslash{}d\{3\})[-. ]?(\textbackslash{}d\{4\})}
    \item \textbf{Phone numbers:} \\
    \texttt{(?:(?:25[0-5]|2[0-4][0-9]|[01]?[0-9]}
    \newline\texttt{\{1,2\})\textbackslash.)\{3\}}\texttt{(?:25[0-5]|2[0-4][0-9]|}
    \newline\texttt{[01]?[0-9]\{1,2\})}
\end{itemize}

For the \textbf{Wikipedia and Wikibooks} subsets, we remove pages that contain fewer than 25 UTF-8 words. 

For the \textbf{Gutenberg} subset:
\begin{itemize}[leftmargin=1em]
    \item \textbf{Language identification}: for each paragraph (defined as newline-separated spans of text), we use FastText to perform language identification. 
    Then, we compute the average language score by averaging the score for all passages. 
    If a document has a language score lower than $0.5$, it is discarded;
    \item \textbf{Quality filter\fnref{foot:quality}}: we remove pages that contain fewer than 25 UTF-8 words;
    \item \textbf{Quality filter}\fnref{foot:quality}: Remove any document that contains a token or sequence of tokens repeating over 100 times\fnref{foot:reps}.
\end{itemize}

For the \textbf{Semantic Scholar} subset, we remove any document that contains a token or sequence of tokens repeating over 100 times\fnref{foot:reps} .

For \dolma versions \texttt{1.0} and \texttt{1.5}, we perform decontamination for all subsets of \dolma. In particular, we remove paragraphs that are shared with documents in the Paloma evaluation suite~\citep{paloma}.
Overall, only 0.003\% of our dataset is removed due to contamination with this evaluation set. 
\dolma version \texttt{1.6} is not decontaminated.

\paragraph{Was the ``raw'' data saved in addition to the preprocessed/cleaned data? (e.g., to support unanticipated future uses)}~

Raw data is available for all subsets except Common Crawl. 
Due to space constrains, we only keep linearized version of Common Crawl snapshots, filtered by Language ID as described above. 

Raw data is not available for download outside the Allen Institute for AI. 
Interested individuals may contact authors of this manuscript if they require access to raw data.

\paragraph{Is the preprocessing software available?}~

Yes, all preprocessing software is available on GitHub at \href{https://github.com/allenai/dolma}{\texttt{github.com/allenai/dolma}} and on PyPI\footnote{\href{https://pypi.org/project/dolma/}{\path{pypi.org/project/dolma}}}.

\paragraph{Does this dataset collection/processing procedure achieve the motivation for creating the dataset stated in the first section of this datasheet?}~

Yes, it does. 

\subsection{Dataset Distribution}

\paragraph{How is the dataset distributed? (e.g., website, API, etc.; does the data have a DOI; is it archived redundantly?)}~

\dolma is distributed via the Hugging Face Hub, which offers access via the \texttt{datasets}~\citep{Lhoest_Datasets_A_Community_2021} Python package, direct download, and Git using the Git-LFS extension.
Additionally, a copy is stored on the cloud storage of 
the Allen Institute for AI.

\paragraph{When will the dataset be released/first distributed? (Is there a canonical paper/reference for this dataset?)}~

The dataset is available now. 
This manuscript serves as a reference for the dataset.

\paragraph{What license (if any) is it distributed under? Are there any copyrights on the data?}~

Information about the license associated with \dolma are available on its release page on the Hugging Face Hub: \href{https://huggingface.co/datasets/allenai/dolma}{\path{huggingface.co/datasets/allenai/dolma}}.

\paragraph{Are there any fees or access/export restrictions?}~

The dataset is distributed for free. 
Users should verify any restrictions on its release page on the Hugging Face Hub: \href{https://huggingface.co/datasets/allenai/dolma}{\path{huggingface.co/datasets/allenai/dolma}}.

\subsection{Dataset Maintenance}

\paragraph{Who is supporting/hosting/maintaining the dataset? How does one contact the owner/curator/manager of the dataset (e.g. email address, or other contact info)?}~

The Allen Institute for AI maintains the dataset. 
For support questions, users are invited to open an issue on GitHub\footnote{\href{https://github.com/allenai/dolma/issues}{\path{github.com/allenai/dolma/issues}}} or on the community tab of dataset page\footnote{\href{https://huggingface.co/datasets/allenai/dolma/discussions}{\path{hf.co/datasets/allenai/dolma/discussions}}} (the former being preferred over the latter).
Any other inquiry should be sent to \texttt{ai2-info@allenai.org}.

\paragraph{Will the dataset be updated? How often and by whom? How will updates/revisions be documented and communicated (e.g., mailing list, GitHub)? Is there an erratum?}~

Dataset will be uploaded on a need-to basis by maintainers at the Allen Institute for AI.
Newer version of the dataset will be labeled accordingly. 
The latest version of the dataset, as well as a changelog, will be made available starting from the first revision.

\paragraph{If the dataset becomes obsolete how will this be communicated? Is there a repository to link to any/all papers/systems that use this dataset?}~

Users should keep track of the version of the dataset in use. 
Information about latest version of \dolma  are available on its release page on the Hugging Face Hub: \href{https://huggingface.co/datasets/allenai/dolma}{\path{huggingface.co/datasets/allenai/dolma}}.
\dolma users should cite this manuscript when using this data.

\paragraph{If others want to extend/augment/build on this dataset, is there a mechanism for them to do so? If so, is there a process for tracking/assessing the quality of those contributions. What is the process for communicating/distributing these contributions to users?}~

Creation and distribution of derivatives is described above. 
In case contributors want to flow their improvement back to future \dolma releases, they should contact corresponding authors of this manuscript.

\subsection{Legal \& Ethical Considerations}
\paragraph{If the dataset relates to people (e.g., their attributes) or was generated by people, were they informed about the data collection? (e.g., datasets that collect writing, photos, interactions, transactions, etc.)}~

Subsets of \dolma derived from web data are likely created by people or groups of people, however authorship information is often unavailable. 

Authors were not directly informed about the data collection. 
For encyclopedic and web content, logs of web servers will contain records of spiders ran by Common Crawl.
For academic content, the pes2o subset~\citep{peS2o} is derived from manuscripts that are licensed for permissive distribution by their authors.
Reddit content was acquired through a public API adherent to terms of service; individual authors of Reddit posts were not contacted directly.
Finally, the Allen Institute for AI did not contact Project Gutenberg.
 
\paragraph{If it relates to other ethically protected subjects, have appropriate obligations been met? (e.g., medical data might include information collected from animals)}~

Due to the nature of and size of \dolma, it is impossible to determine which obligations, if any, are appropriate.

\paragraph{If it relates to people, were there any ethical review applications/reviews/approvals? (e.g. Institutional Review Board applications) If it relates to people, were they told what the dataset would be used for and did they consent? What community norms exist for data collected from human communications? If consent was obtained, how? Were the people provided with any mechanism to revoke their consent in the future or for certain uses?}~

The \dolma project includes Ethics committee comprised of 
internal and external members to the Allen Institute for AI.
Plans for the creation of \dolma were reviewed with the committee, and we incorporated their recommendations. 

Following practices established in similar efforts, no consent was collected from individuals who might be represented in the dataset.
We make available a form\footnote{\label{foot:data-removal}\href{https://forms.gle/q4BNUUxUxKwKkfdT6}{\path{forms.gle/q4BNUUxUxKwKkfdT6}}} for individuals who wish to be removed from the dataset. 

\paragraph{If it relates to people, could this dataset expose people to harm or legal action? (e.g., financial social or otherwise) What was done to mitigate or reduce the potential for harm?}~

\dolma contains text instances that have been derived from web pages Common Crawl crawled from the web. 
Content might contain sensitive information including personal information, or financial information users of the web chose to put publicly online. 
This data is taken only from public places, so the same data is or has been accessible via browsing the web. 
We have measured a variety of types of personal information, and built tools specifically to remove some types of sensitive information, and through our license we restrict what users can do with this data. 

We recommend individuals to submit a request using through our form\fnref{foot:data-removal} if they wish their information to be removed. 

\paragraph{If it relates to people, does it unfairly advantage or disadvantage a particular social group? In what ways? How was this mitigated?}~

\dolma is not a representative sample of none of its sources. 
It might underrepresent or overrepresent some communities on the internet;
further, papers in the peS2o subset are skewed towards STEM disciplines;
books in the Gutenberg library are mostly from the public domain (at the time of publication, books published before 1927);
finally, the English and Simple subset of Wikipedia and Wikibooks might be biased towards events and people from the global north.

We did not attempt to alter distribution of social groups in \dolma. 
Large-scale interventions to correct societal biases in large datasets remain challenging, and are left to future work.

\paragraph{If it relates to people, were they provided with privacy guarantees? If so, what guarantees and how are these ensured?}~

This datasets contains text that was derived from web paged scraped by Common Crawl from the web. 
For much of that data it’s not possible identify the authors. 
In many instances, creators purposely choose to post anonymously online, so aiming to infer authorship can be ethically fraught. We provide access to our data, and encourage any creators that would likely to have data from or about them removed to reach out.

\paragraph{Does the dataset comply with the EU General Data Protection Regulation (GDPR)? Does it comply with any other standards, such as the US Equal Employment Opportunity Act?}~

We created this dataset in aggregate, not separately identifying any individual’s content or information. 
We took reasonable steps to remove types of personal information that were possible to reliably detect. 
We restrict who has access to the data, and we release this under a license that prohibits uses that might be deemed discriminatory. 
We also provide an avenue for any person to contact us to have text from or about them removed from our corpus\fnref{foot:data-removal}.

\paragraph{Does the dataset contain information that might be considered sensitive or confidential? (e.g., personally identifying information) Does the dataset contain information that might be considered inappropriate or offensive?}~

This datasets contains text that was derived from web paged scraped by Common Crawl from the web. 
Therefore, it can contain text posted on public websites by creators on the internet. 
If an author publicly posted personal information or offensive content, it could be included in this dataset. 
We took reasonable steps to remove types of personal information that were possible to reliably detect. 
We also removed documents that contained sentences that were classified as being toxic.

\section{All Raw Ablation Results}
\label{app:raw}

\subsection{Comparing \dolma With Other Corpora}
\label{sec:150b_runs}

\label{sec:150b_runs:ppl}

\begin{figure}[h!]
	\centering
	\begin{subfigure}{0.31\textwidth}
		\includegraphics[width=\linewidth]{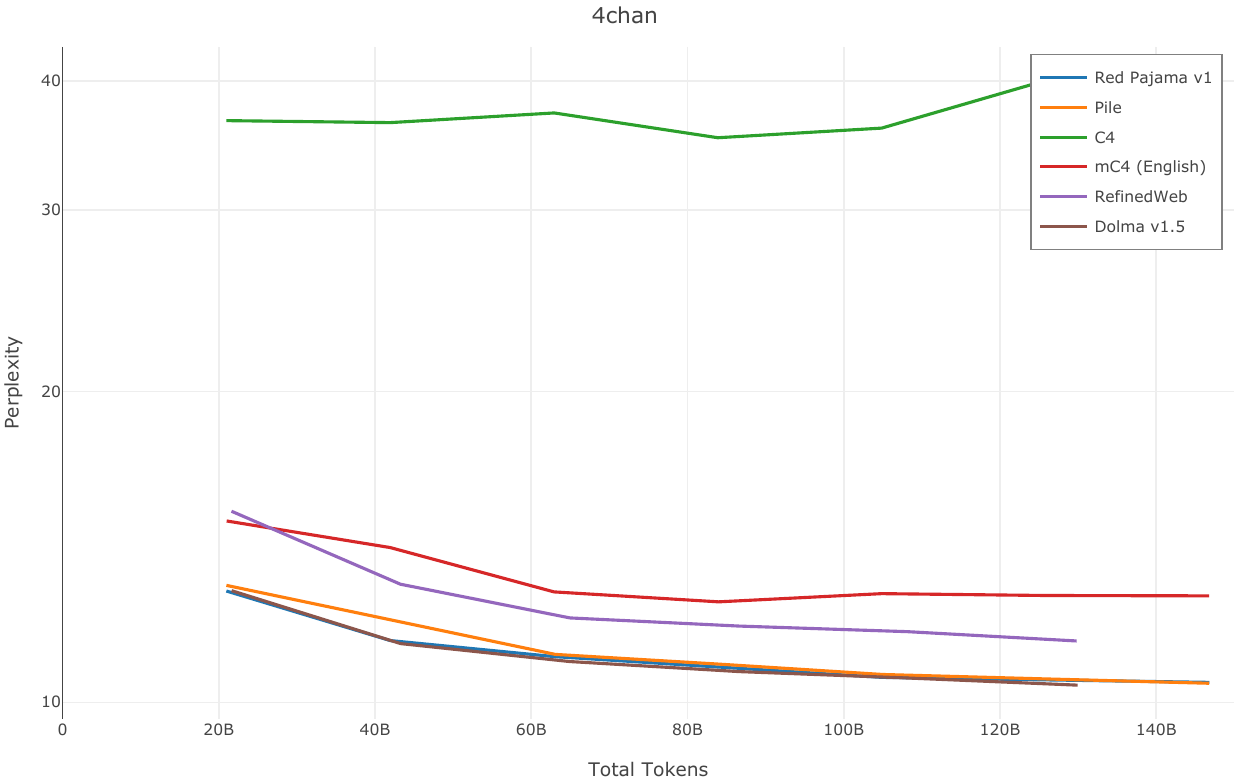}
	\end{subfigure}
	\quad
	\begin{subfigure}{0.31\textwidth}
		\includegraphics[width=\linewidth]{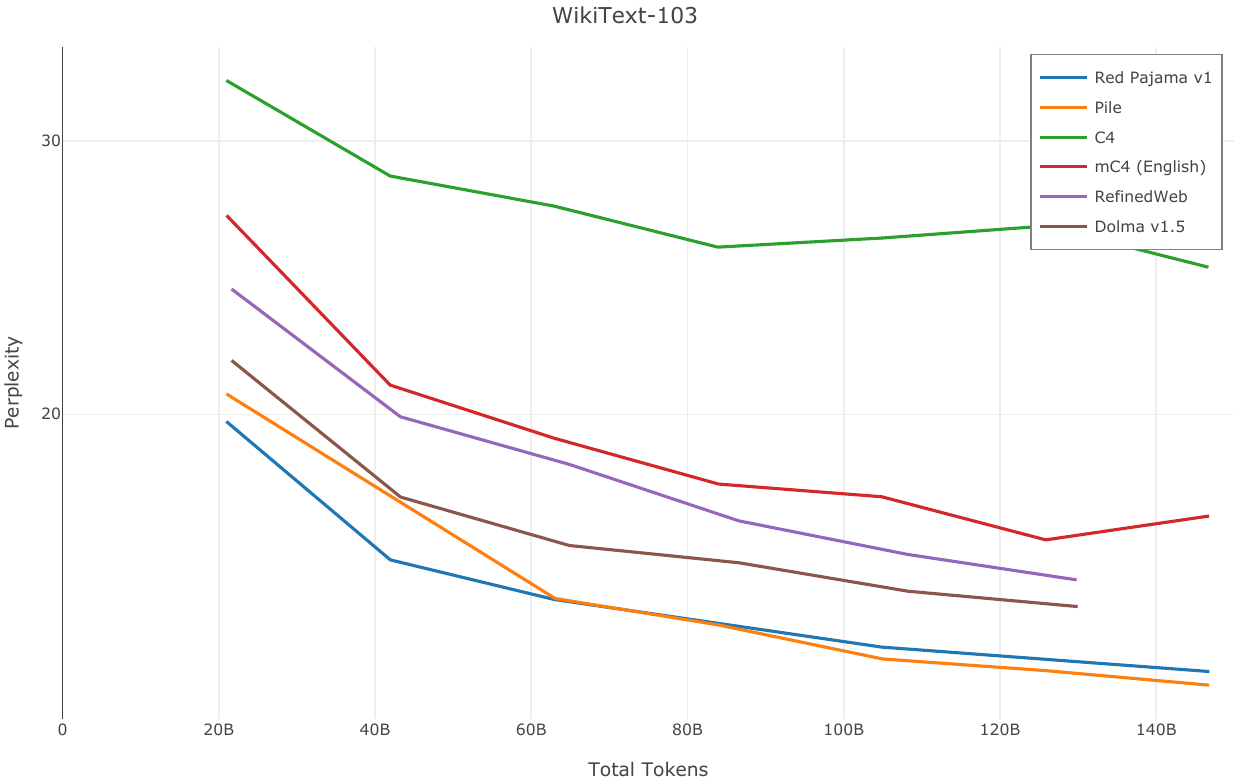}
	\end{subfigure}
	\quad
	\begin{subfigure}{0.31\textwidth}
		\includegraphics[width=\linewidth]{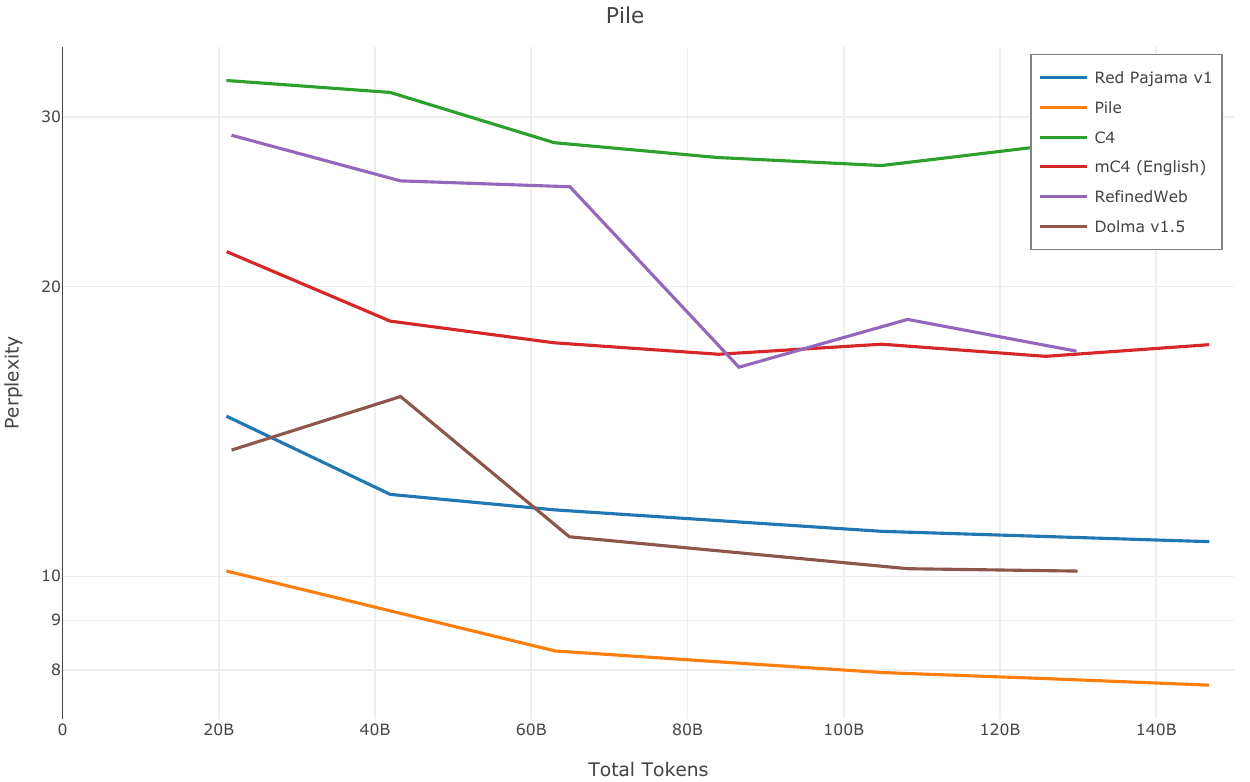}
	\end{subfigure}
	\caption{Perplexity results on Paloma~\citep{paloma}; subsets 4chan~\citep{papasavva2020raiders}, WikiText 103~\citep{merity2016pointer}, and Pile~\citep{Gao2020ThePA} (Val)}
\end{figure}

\begin{figure}[h!]
	\centering
	\begin{subfigure}{0.31\textwidth}
		\includegraphics[width=\linewidth]{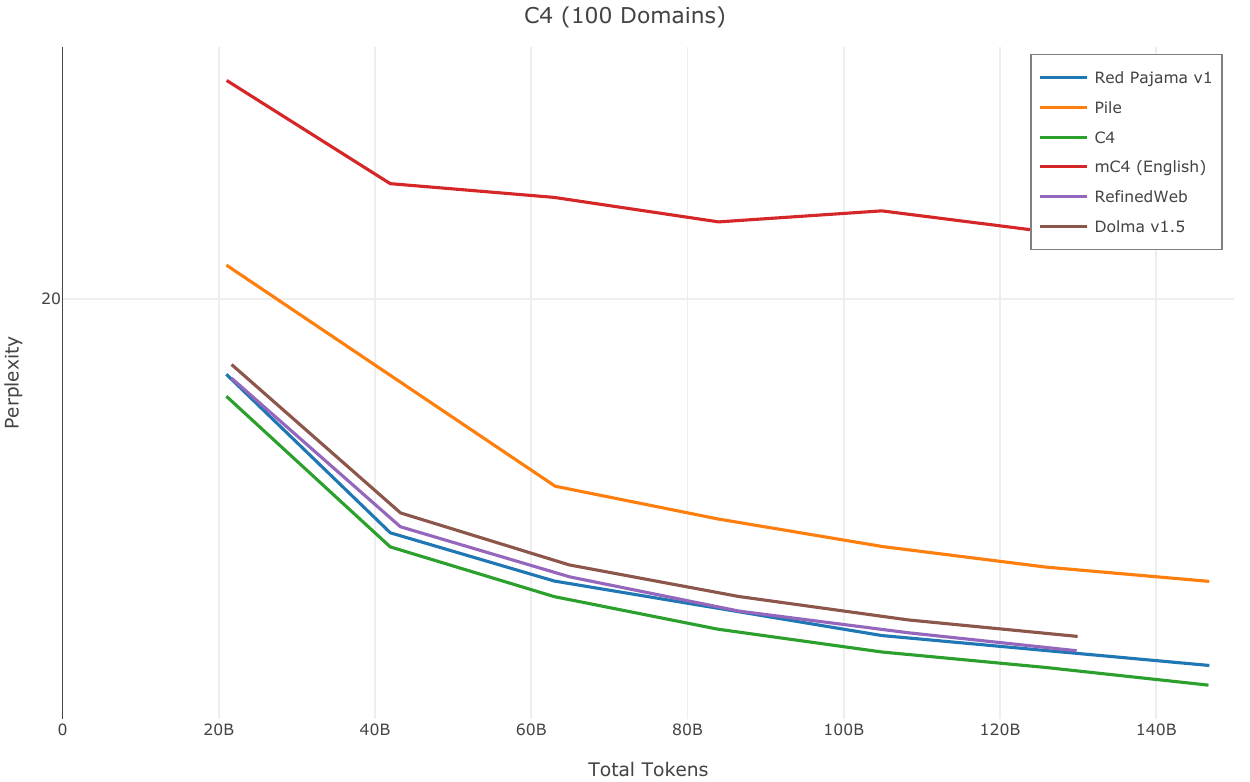}
	\end{subfigure}
	\quad
	\begin{subfigure}{0.31\textwidth}
		\includegraphics[width=\linewidth]{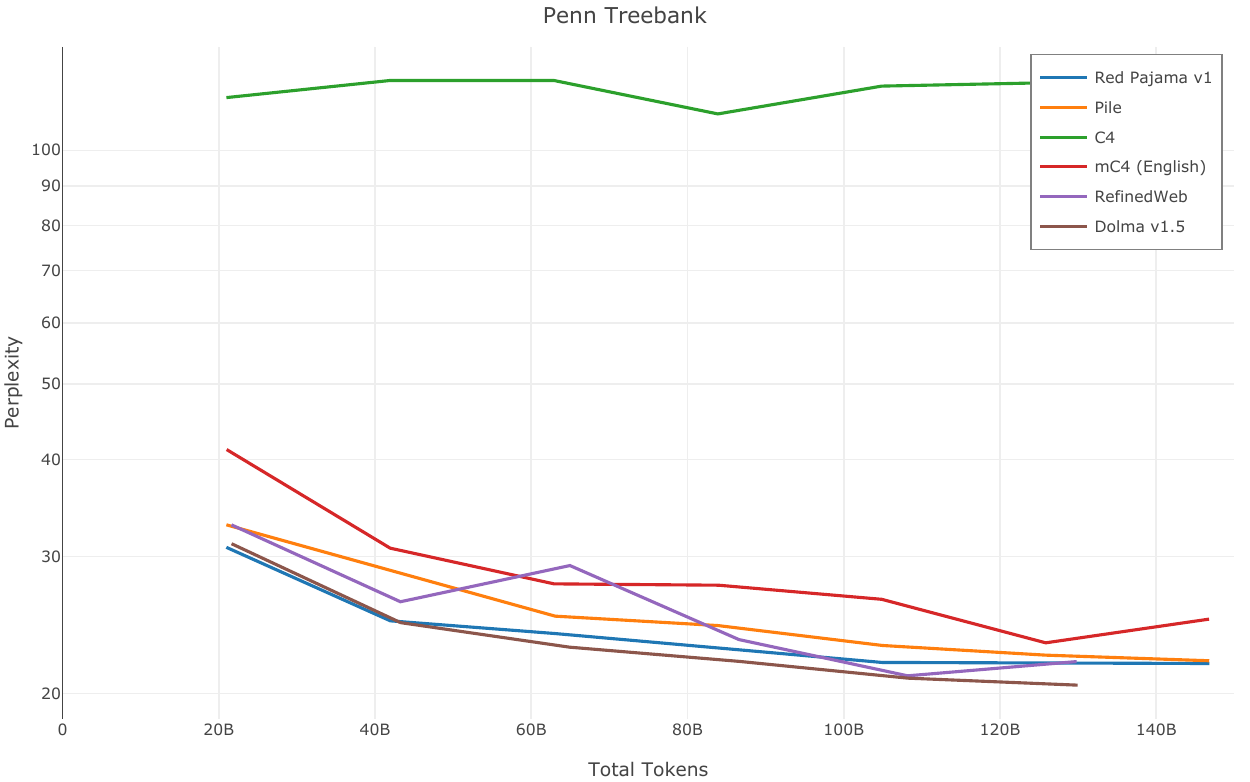}
	\end{subfigure}
	\quad
	\begin{subfigure}{0.31\textwidth}
		\includegraphics[width=\linewidth]{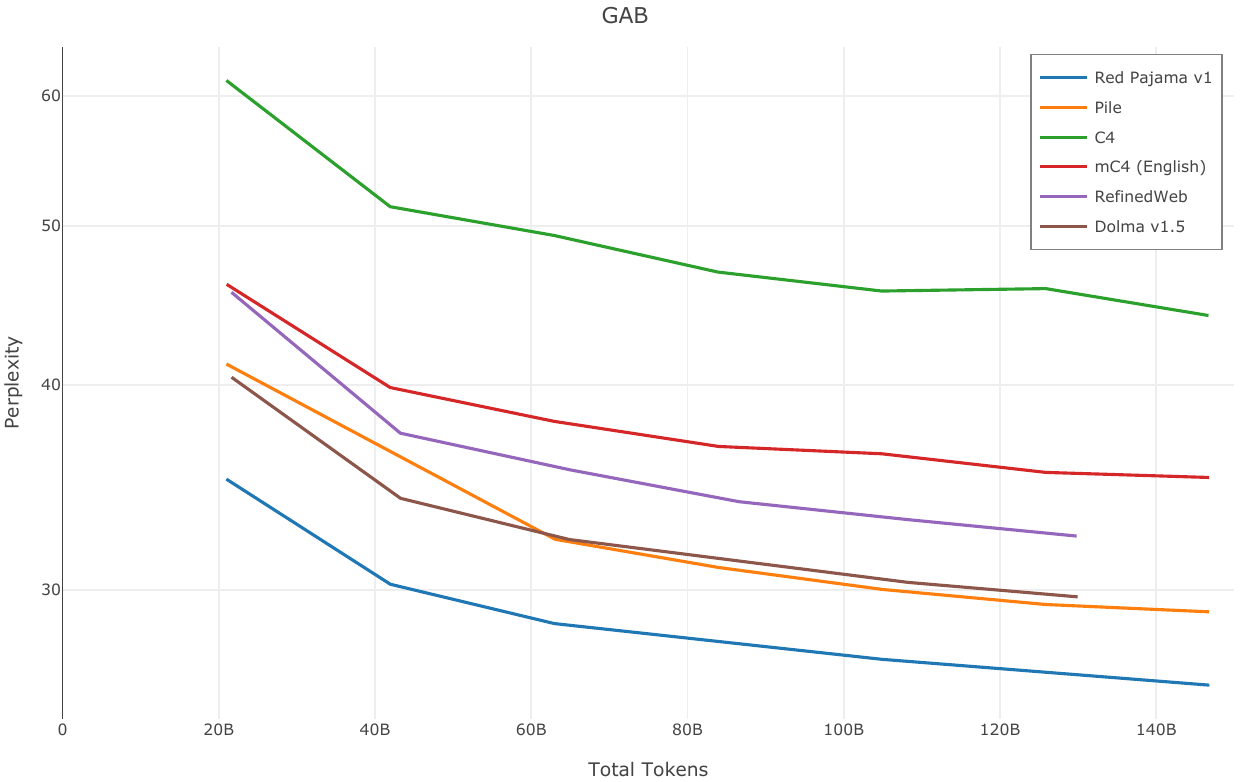}
	\end{subfigure}
	\caption{Perplexity results on Paloma~\citep{paloma}; subsets C4 100 dom~\citep{chronopoulou-etal-2022-efficient}, Penn Tree Bank~\citep{marcus-etal-1994-penn}, and Gab~\citep{zannettou2018gab}}
\end{figure}

\begin{figure}[h!]
	\centering
	\begin{subfigure}{0.31\textwidth}
		\includegraphics[width=\linewidth]{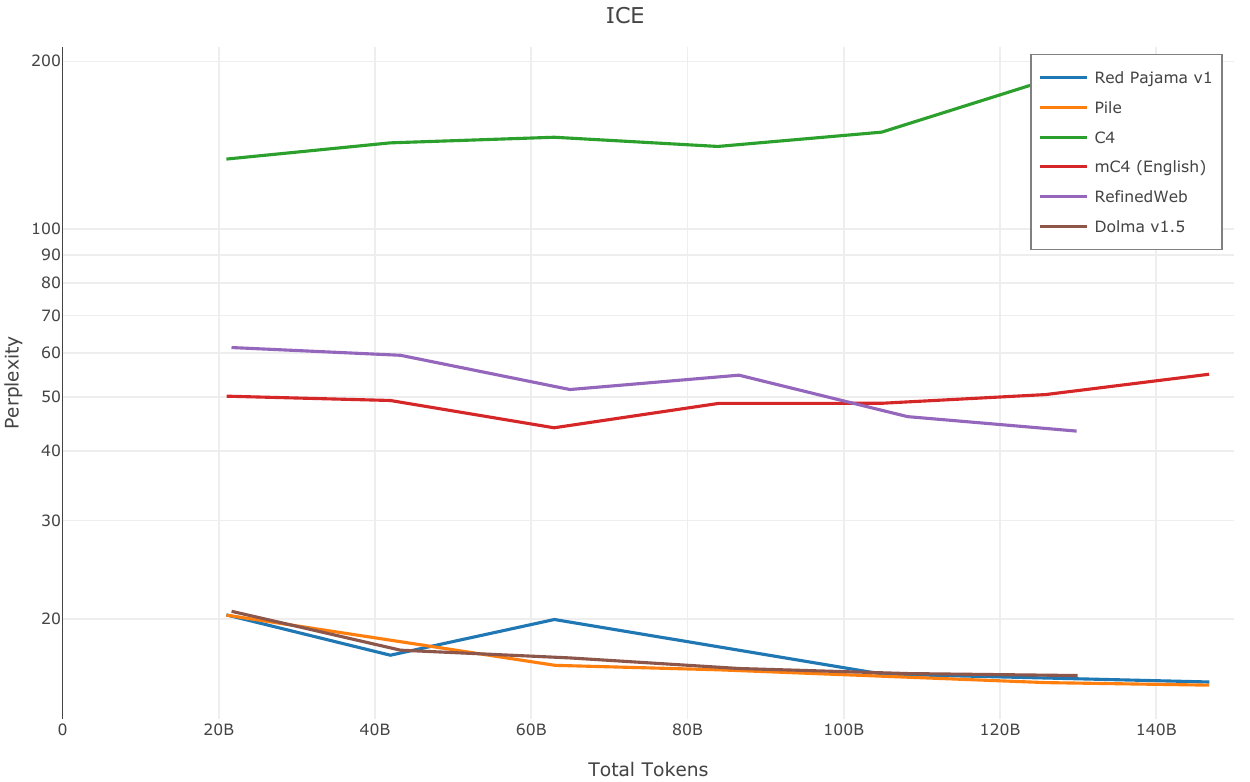}
	\end{subfigure}
	\quad
	\begin{subfigure}{0.31\textwidth}
		\includegraphics[width=\linewidth]{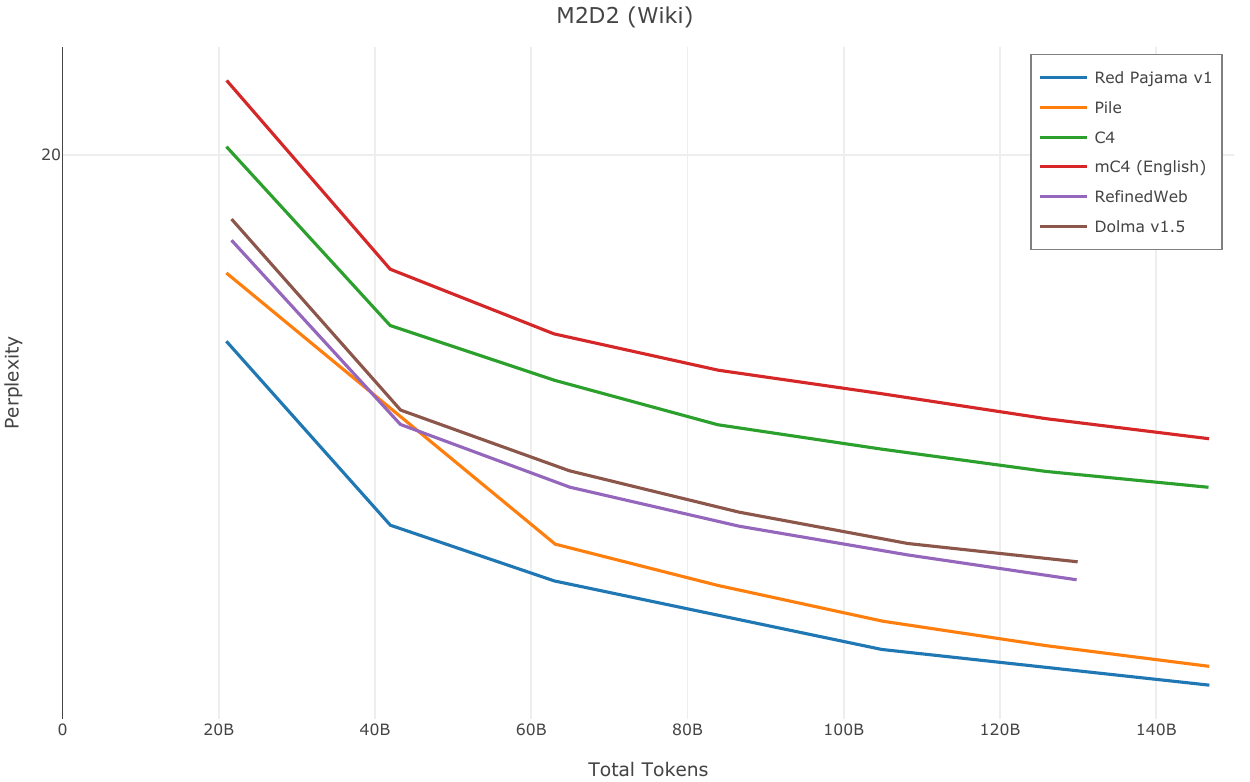}
	\end{subfigure}
	\quad
	\begin{subfigure}{0.31\textwidth}
		\includegraphics[width=\linewidth]{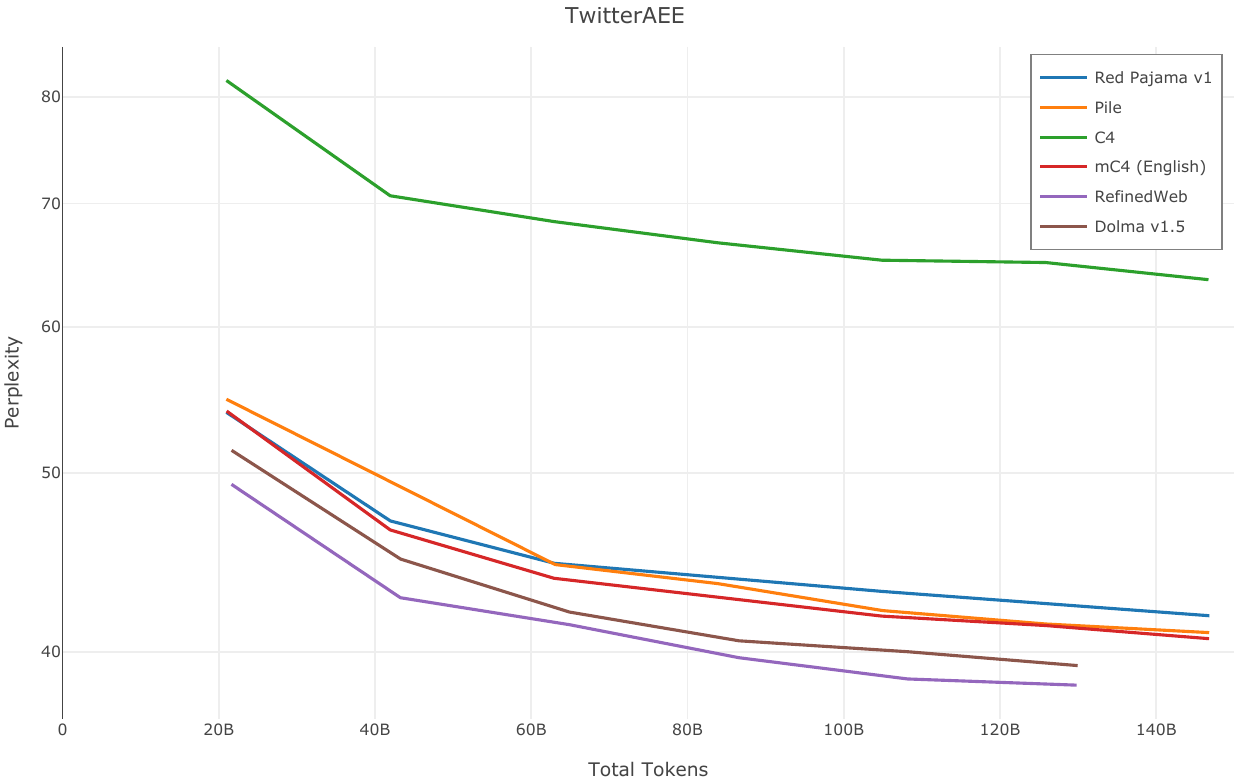}
	\end{subfigure}
	\caption{Perplexity results on Paloma~\citep{paloma}; subsets ICE~\citep{greenbaum1991ice}, M2D2~\citep{reid-etal-2022-m2d2} (Wiki), and Twitter AAE~\citep{blodgett-etal-2016-demographic}}
\end{figure}

\begin{figure}[h!]
	\centering
	\begin{subfigure}{0.31\textwidth}
		\includegraphics[width=\linewidth]{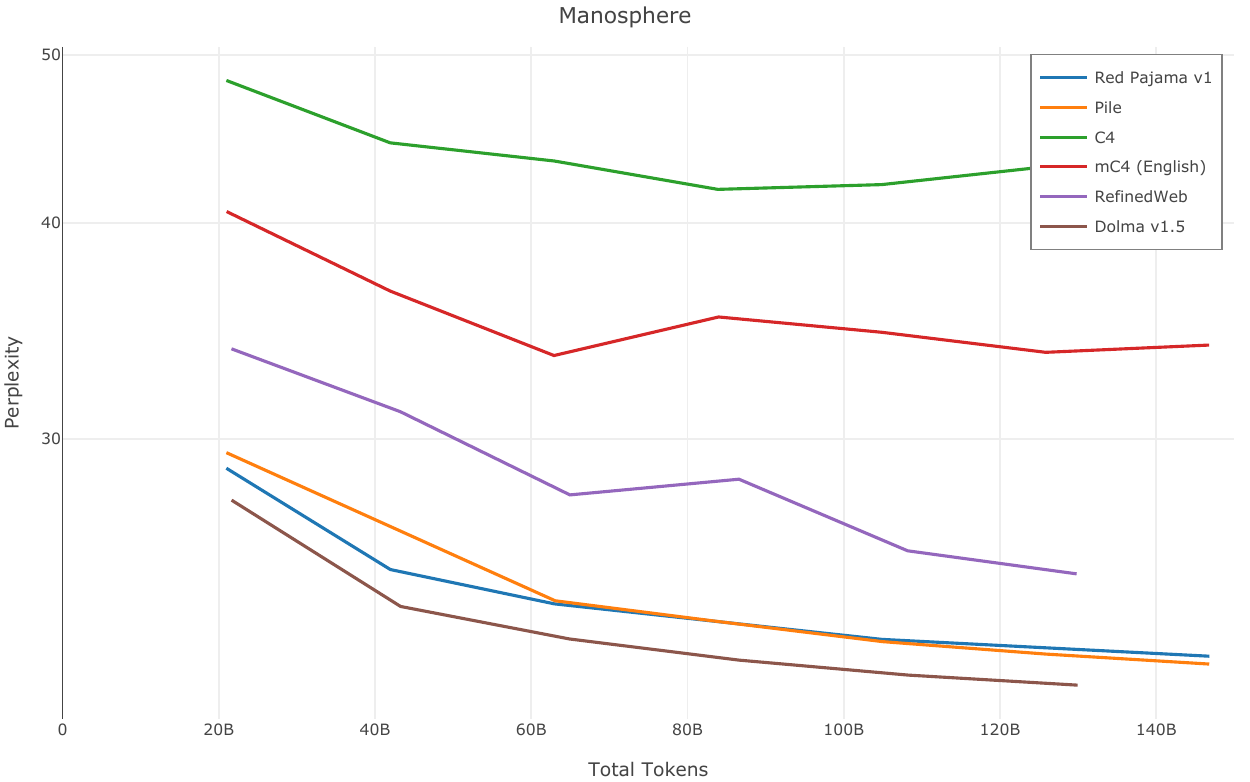}
	\end{subfigure}
	\caption{Perplexity results on Paloma~\citep{paloma}; subsets Manosphere~\citep{ribeiroevolution2021}}
\end{figure}

\begin{figure}[h!]
	\centering
	\begin{subfigure}{0.31\textwidth}
		\includegraphics[width=\linewidth]{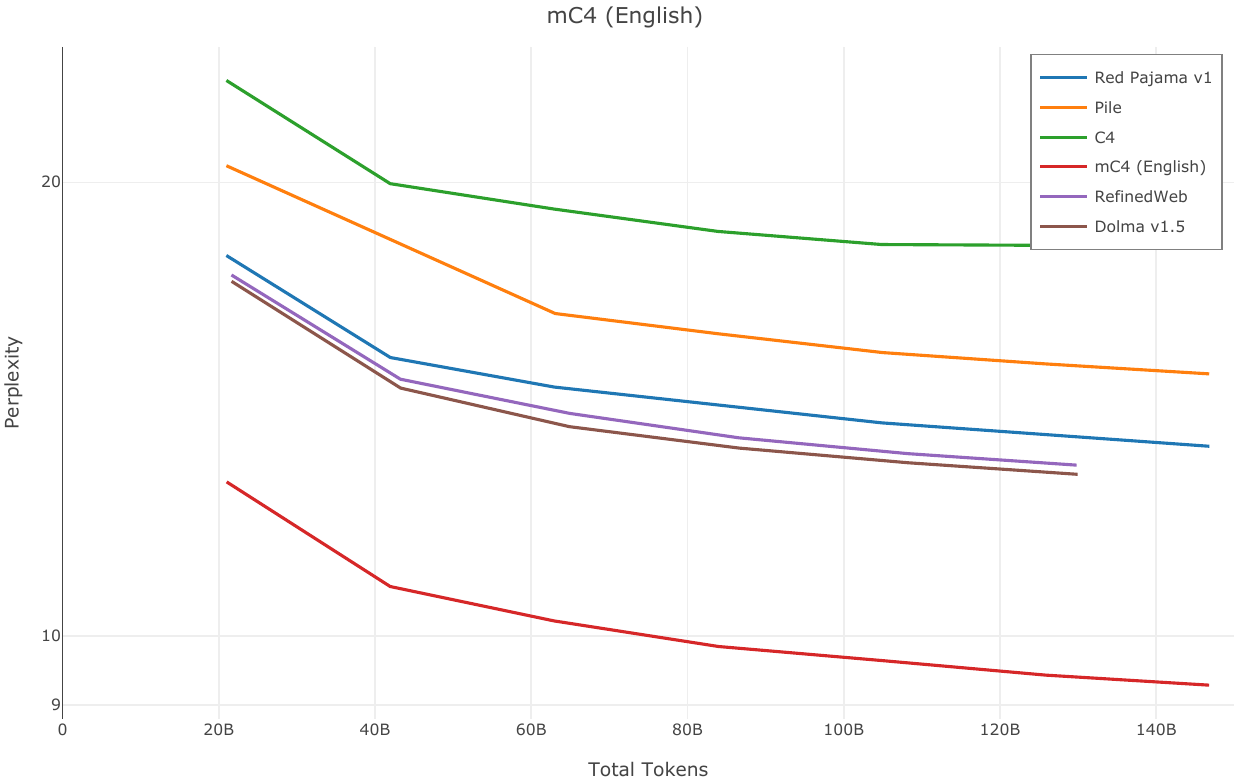}
	\end{subfigure}
	\quad
	\begin{subfigure}{0.31\textwidth}
		\includegraphics[width=\linewidth]{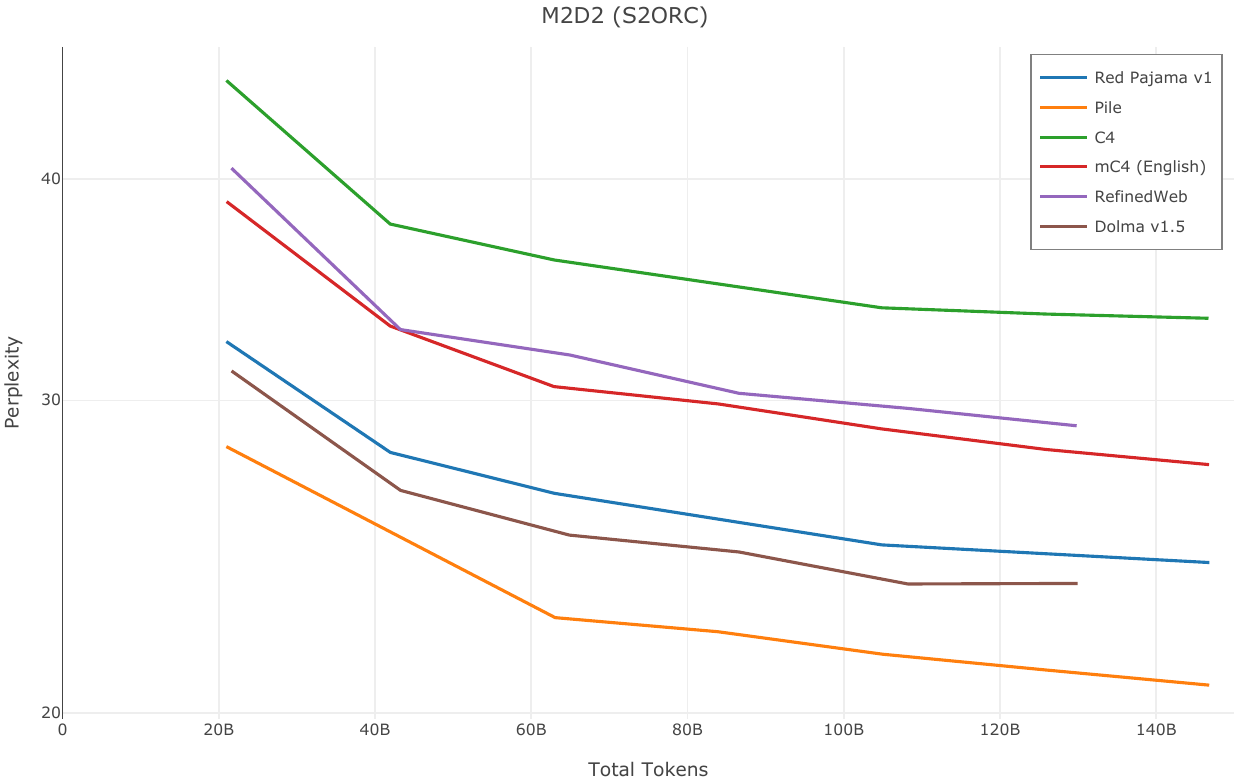}
	\end{subfigure}
	\quad
	\begin{subfigure}{0.31\textwidth}
		\includegraphics[width=\linewidth]{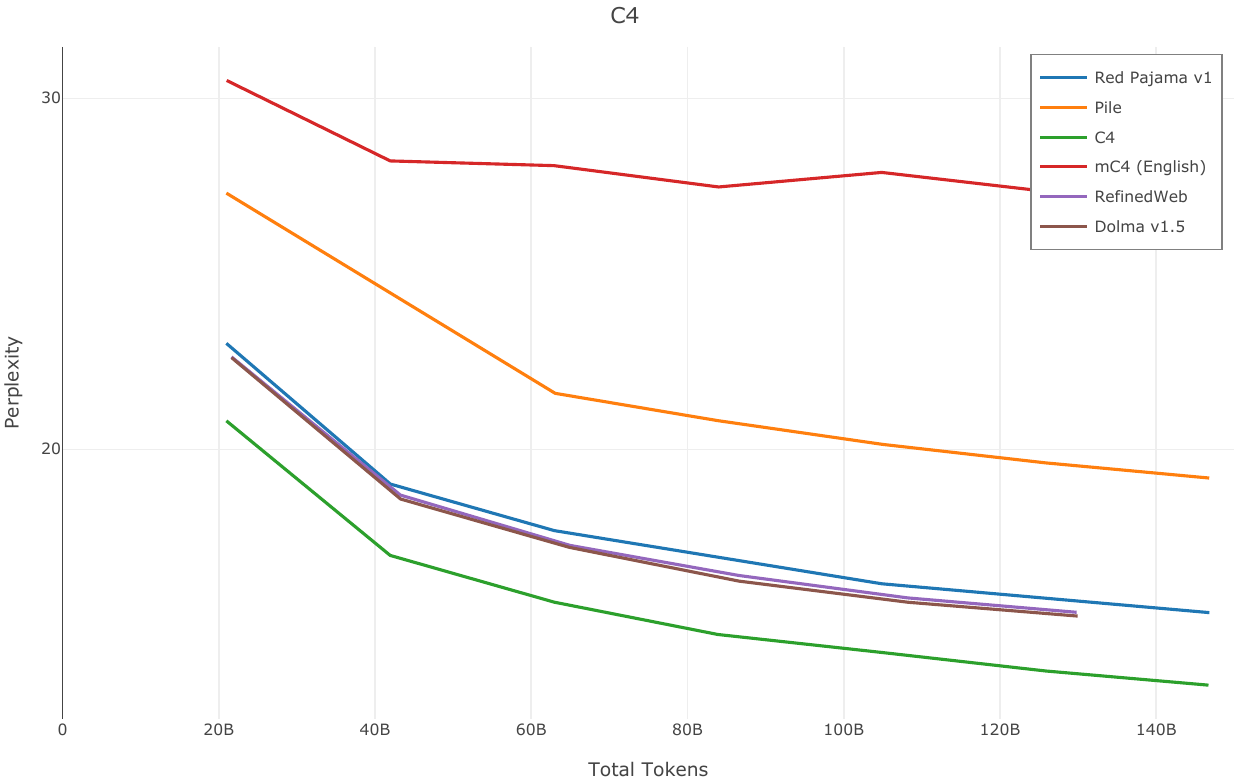}
	\end{subfigure}
	\caption{Perplexity results on Paloma~\citep{paloma}; subsets mC4~\citep{mc4} (English), M2D2~\citep{reid-etal-2022-m2d2} (S2ORC), and C4~\citep{raffel2020exploring,dodge-etal-2021-documenting}}
\end{figure}

\label{sec:150b_runs:downstream}

\begin{figure}[h!]
	\centering
	\begin{subfigure}{0.31\textwidth}
		\includegraphics[width=\linewidth]{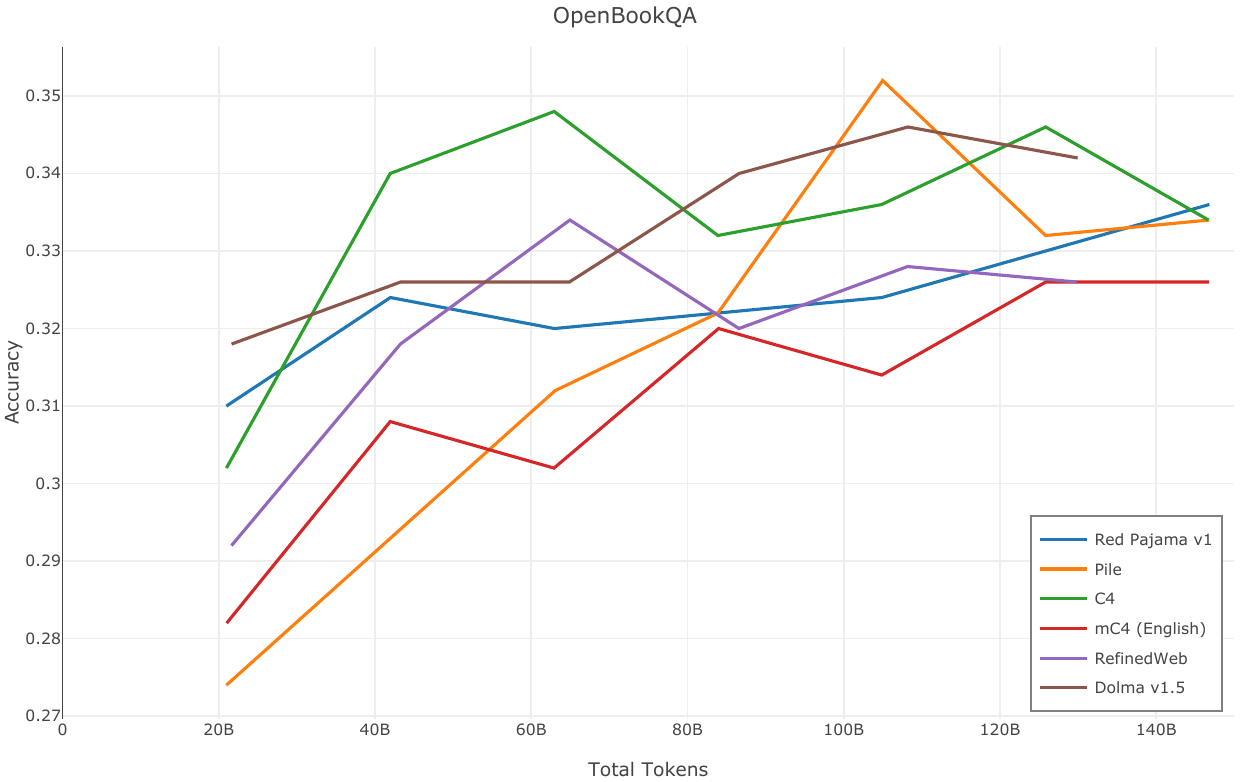}
	\end{subfigure}
	\quad
	\begin{subfigure}{0.31\textwidth}
		\includegraphics[width=\linewidth]{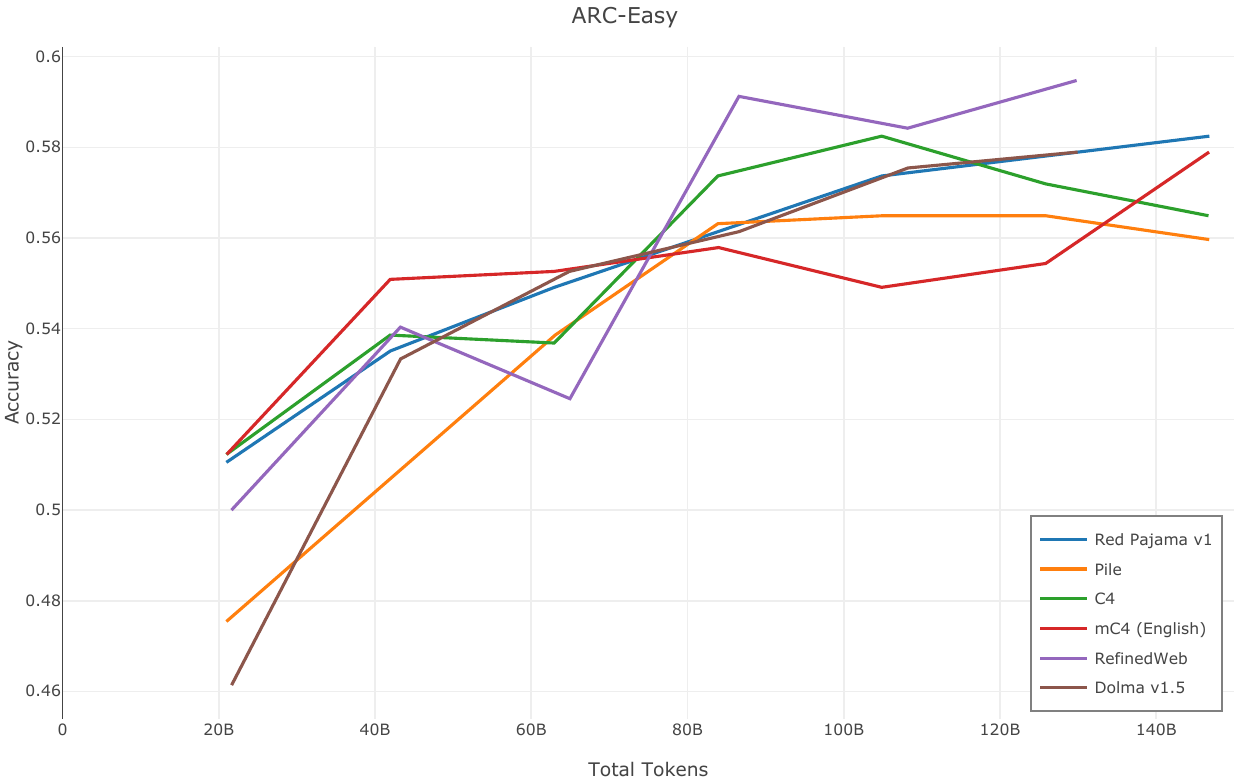}
	\end{subfigure}
	\quad
	\begin{subfigure}{0.31\textwidth}
		\includegraphics[width=\linewidth]{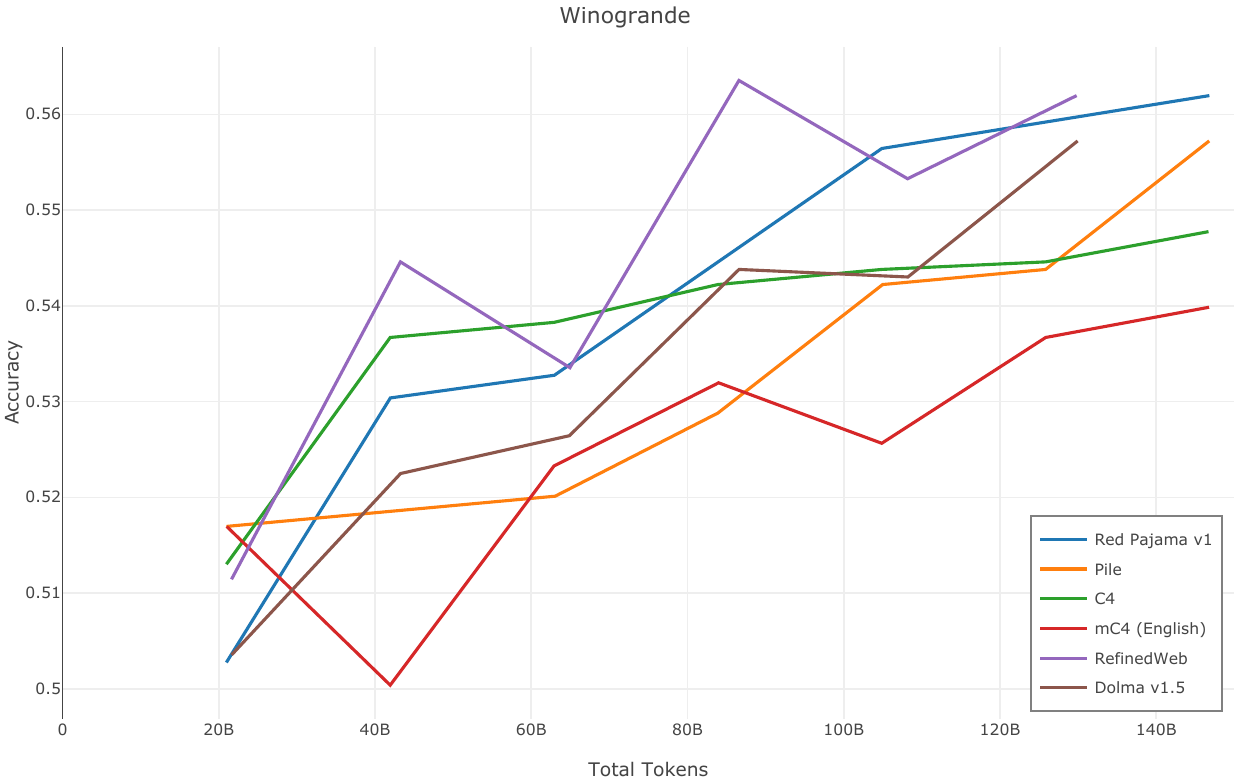}
	\end{subfigure}
	\caption{Results downstream tasks OpenBookQA~\citep{openbookQA}, ARC-E~\citep{arc}, and WinoGrande~\citep{winogrande}}
\end{figure}

\begin{figure}[h!]
	\centering
	\begin{subfigure}{0.31\textwidth}
		\includegraphics[width=\linewidth]{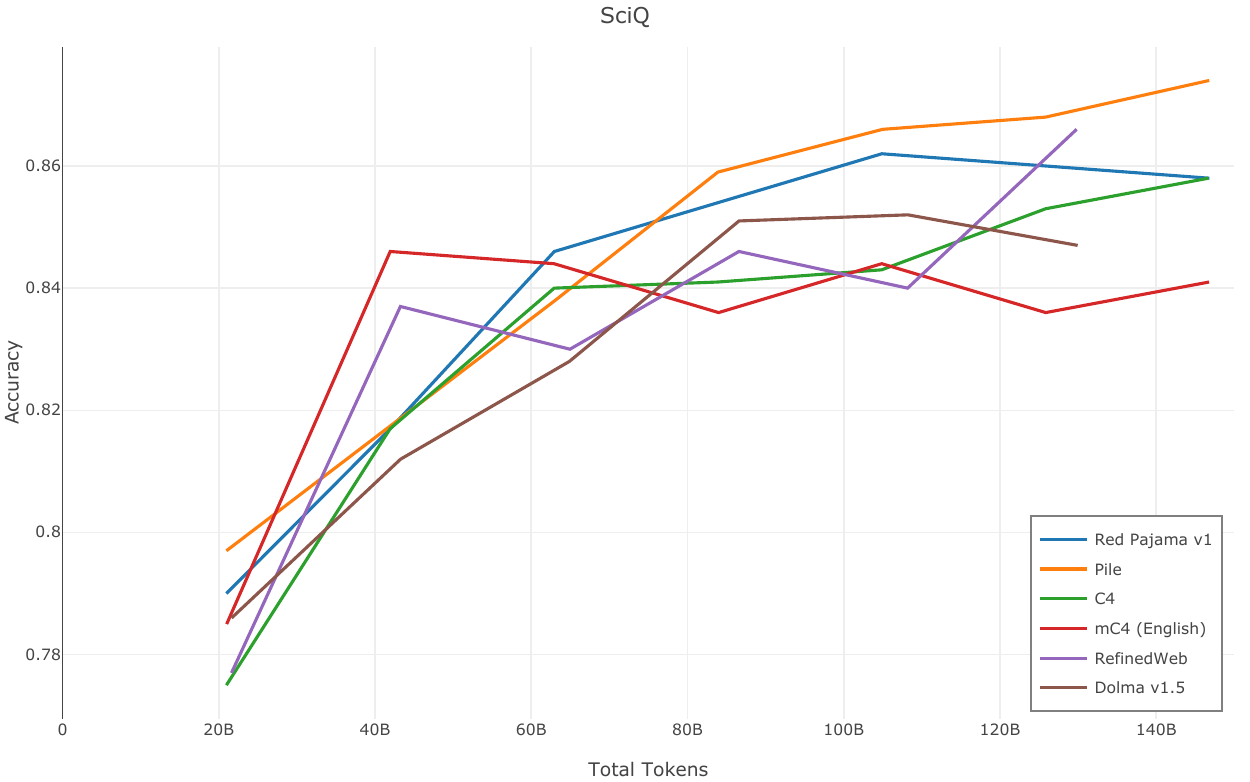}
	\end{subfigure}
	\quad
	\begin{subfigure}{0.31\textwidth}
		\includegraphics[width=\linewidth]{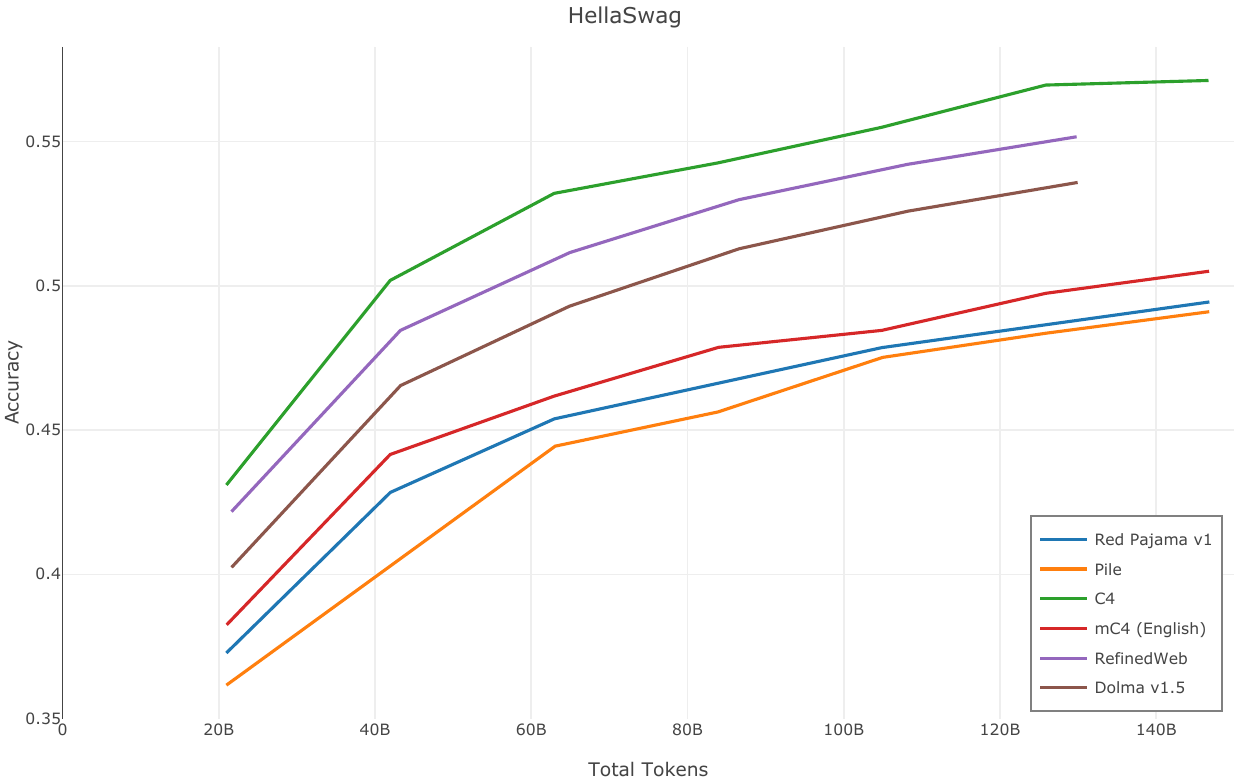}
	\end{subfigure}
	\quad
	\begin{subfigure}{0.31\textwidth}
		\includegraphics[width=\linewidth]{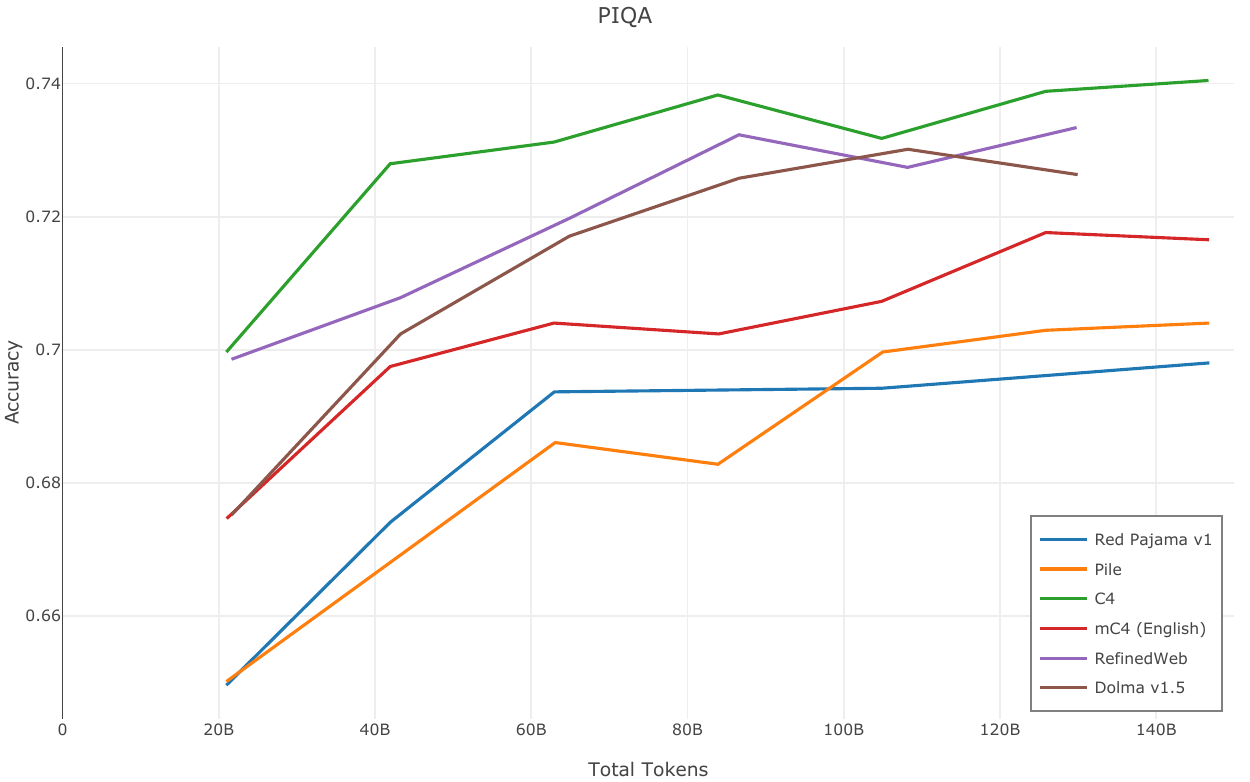}
	\end{subfigure}
	\caption{Results downstream tasks SciQ~\citep{sciq}, HellaSwag~\citep{zellers2019hellaswag}, and PIQA~\citep{piqa}}
\end{figure}

\label{sec:150b_runs:train}

\begin{figure}[h!]
	\centering
	\begin{subfigure}{0.31\textwidth}
		\includegraphics[width=\linewidth]{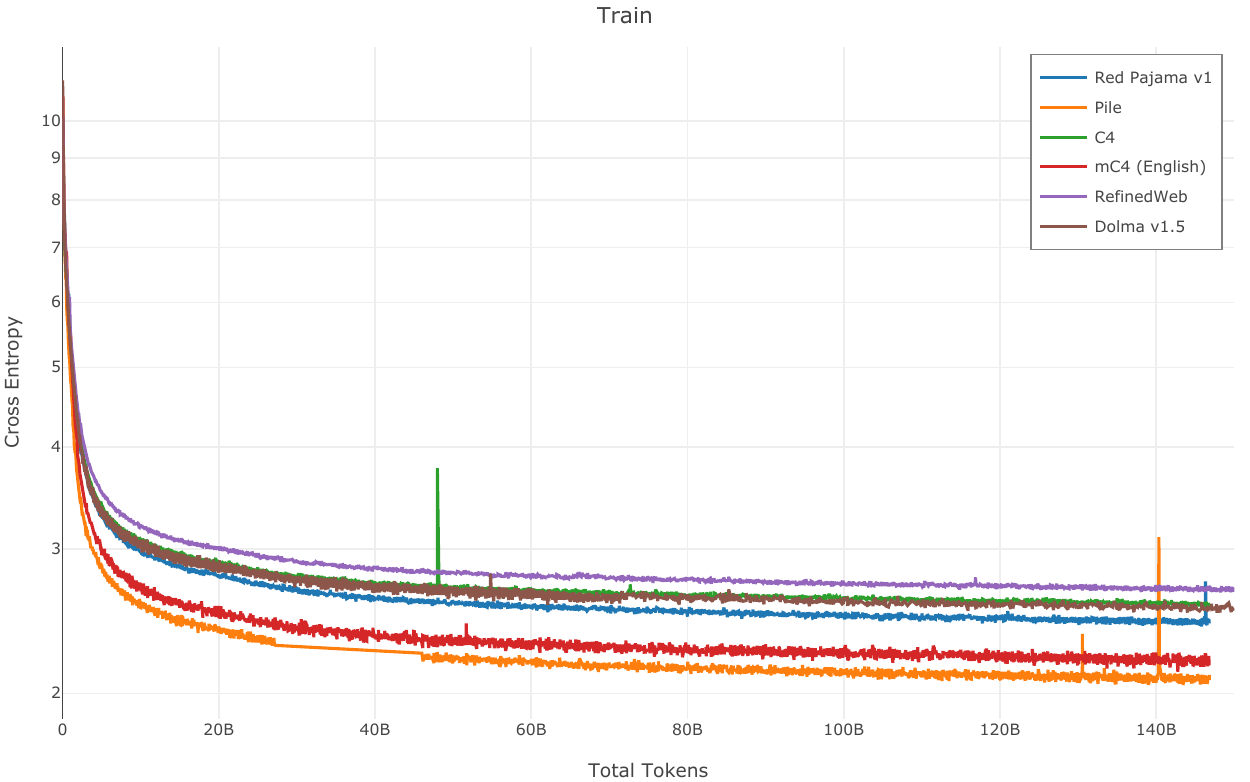}
	\end{subfigure}
	\caption{Training Cross Entropy}
\end{figure}

\clearpage

\subsection{Deduping Strategy}
\label{sec:ablations_cc_dedupe}

\label{sec:ablations_cc_dedupe:ppl}

\begin{figure}[h!]
	\centering
	\begin{subfigure}{0.31\textwidth}
		\includegraphics[width=\linewidth]{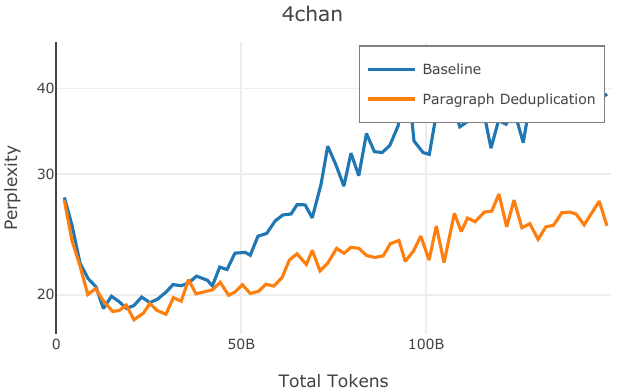}
	\end{subfigure}
	\quad
	\begin{subfigure}{0.31\textwidth}
		\includegraphics[width=\linewidth]{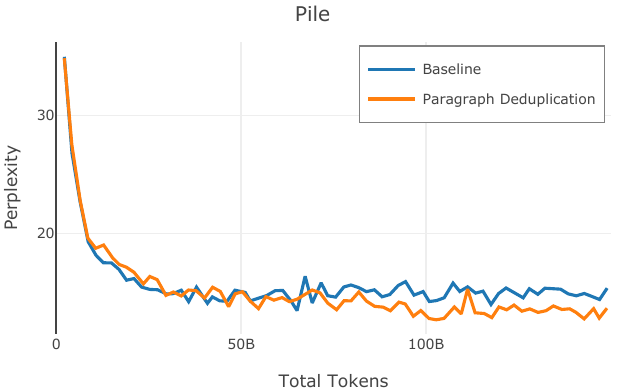}
	\end{subfigure}
	\quad
	\begin{subfigure}{0.31\textwidth}
		\includegraphics[width=\linewidth]{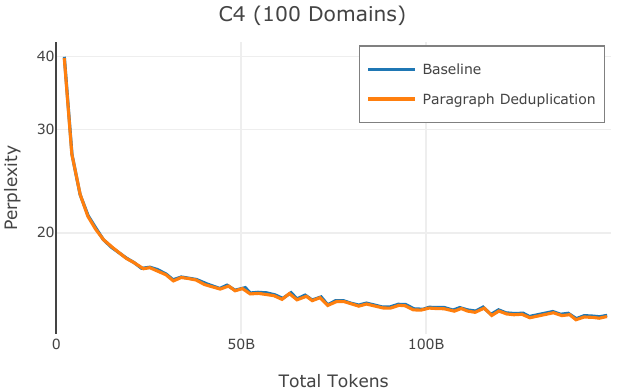}
	\end{subfigure}
	\caption{Perplexity results on Paloma~\citep{paloma}; subsets 4chan~\citep{papasavva2020raiders}, Pile~\citep{Gao2020ThePA} (Val), and C4 100 dom~\citep{chronopoulou-etal-2022-efficient}}
\end{figure}

\begin{figure}[h!]
	\centering
	\begin{subfigure}{0.31\textwidth}
		\includegraphics[width=\linewidth]{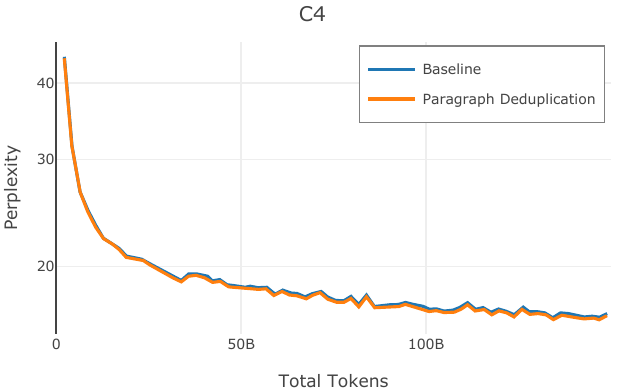}
	\end{subfigure}
	\quad
	\begin{subfigure}{0.31\textwidth}
		\includegraphics[width=\linewidth]{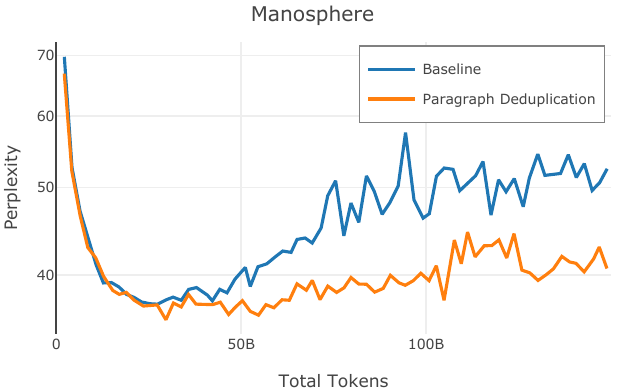}
	\end{subfigure}
	\caption{Perplexity results on Paloma~\citep{paloma}; subsets C4~\citep{raffel2020exploring,dodge-etal-2021-documenting} and Manosphere~\citep{ribeiroevolution2021}}
\end{figure}

\begin{figure}[h!]
	\centering
	\begin{subfigure}{0.31\textwidth}
		\includegraphics[width=\linewidth]{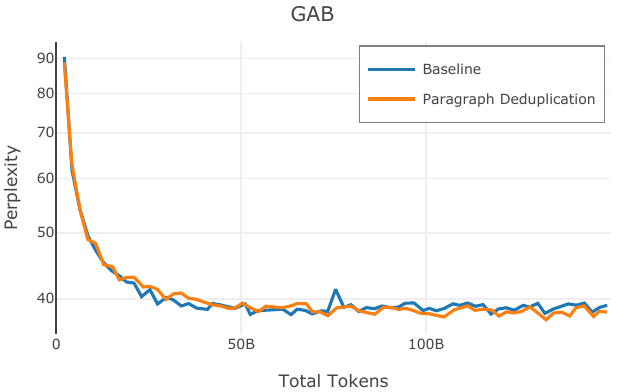}
	\end{subfigure}
	\quad
	\begin{subfigure}{0.31\textwidth}
		\includegraphics[width=\linewidth]{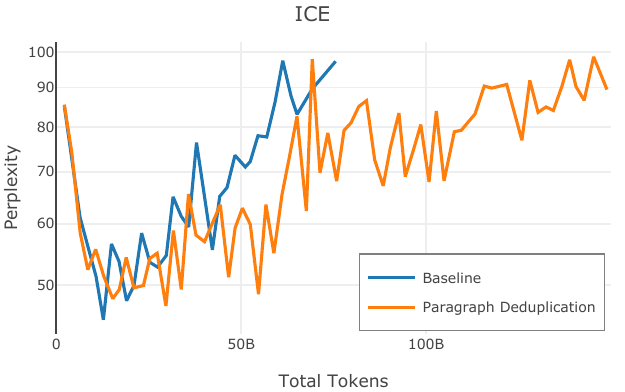}
	\end{subfigure}
	\quad
	\begin{subfigure}{0.31\textwidth}
		\includegraphics[width=\linewidth]{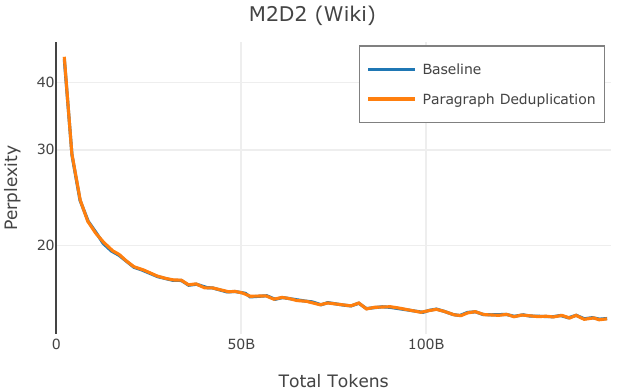}
	\end{subfigure}
	\caption{Perplexity results on Paloma~\citep{paloma}; subsets Gab~\citep{zannettou2018gab}, ICE~\citep{greenbaum1991ice}, and M2D2~\citep{reid-etal-2022-m2d2} (Wiki)}
\end{figure}

\begin{figure}[h!]
	\centering
	\begin{subfigure}{0.31\textwidth}
		\includegraphics[width=\linewidth]{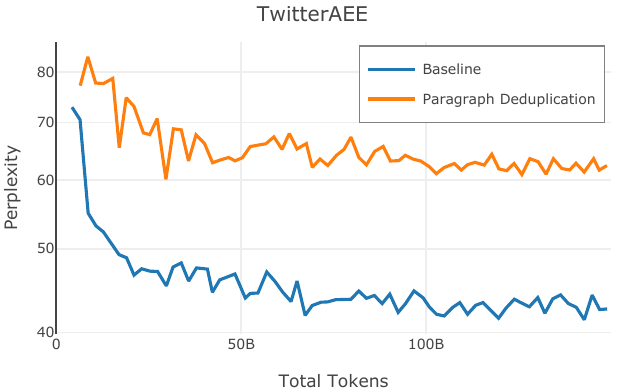}
	\end{subfigure}
	\quad
	\begin{subfigure}{0.31\textwidth}
		\includegraphics[width=\linewidth]{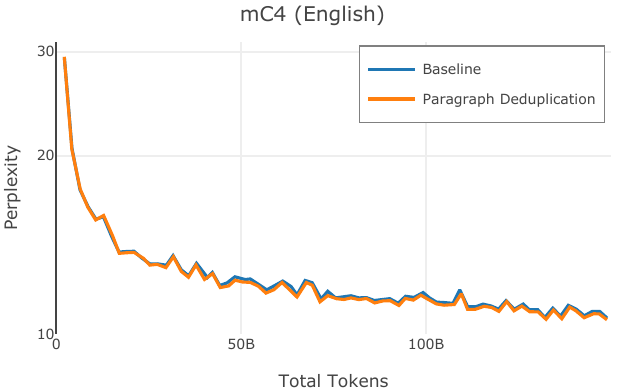}
	\end{subfigure}
	\quad
	\begin{subfigure}{0.31\textwidth}
		\includegraphics[width=\linewidth]{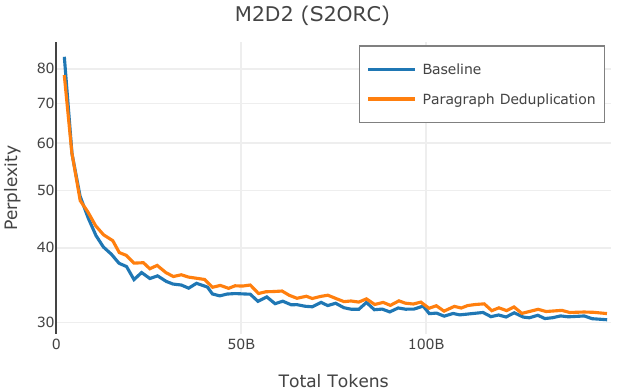}
	\end{subfigure}
	\caption{Perplexity results on Paloma~\citep{paloma}; subsets Twitter AAE~\citep{blodgett-etal-2016-demographic}, mC4~\citep{mc4} (English), and M2D2~\citep{reid-etal-2022-m2d2} (S2ORC)}
\end{figure}

\label{sec:ablations_cc_dedupe:downstream}

\begin{figure}[h!]
	\centering
	\begin{subfigure}{0.31\textwidth}
		\includegraphics[width=\linewidth]{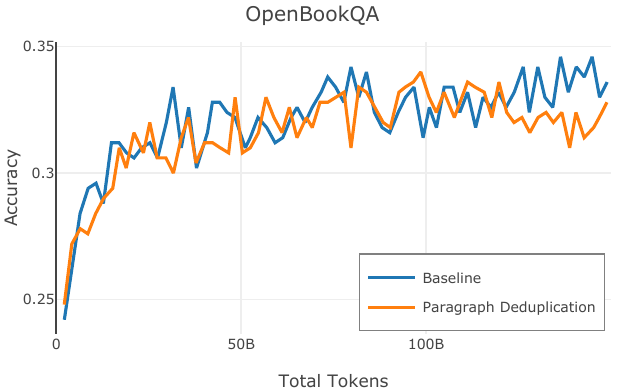}
	\end{subfigure}
	\quad
	\begin{subfigure}{0.31\textwidth}
		\includegraphics[width=\linewidth]{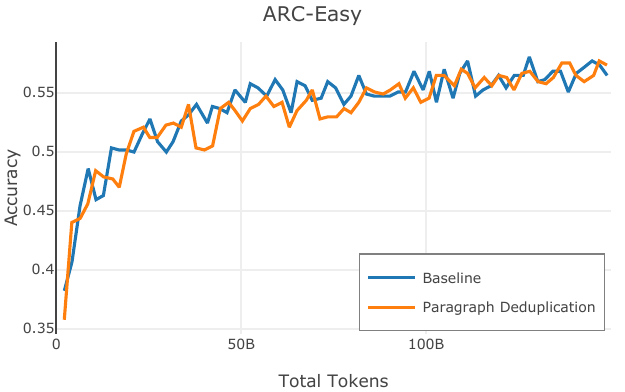}
	\end{subfigure}
	\quad
	\begin{subfigure}{0.31\textwidth}
		\includegraphics[width=\linewidth]{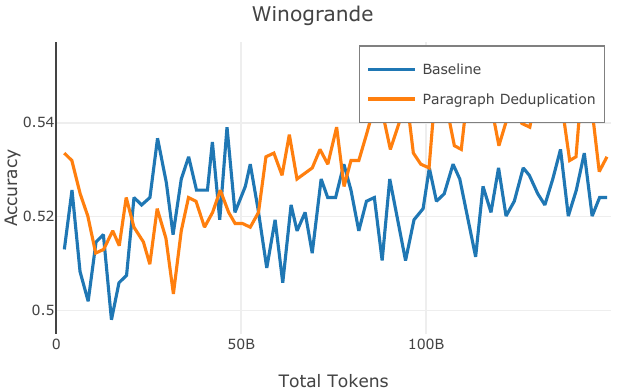}
	\end{subfigure}
	\caption{Results downstream tasks OpenBookQA~\citep{openbookQA}, ARC-E~\citep{arc}, and WinoGrande~\citep{winogrande}}
\end{figure}

\begin{figure}[h!]
	\centering
	\begin{subfigure}{0.31\textwidth}
		\includegraphics[width=\linewidth]{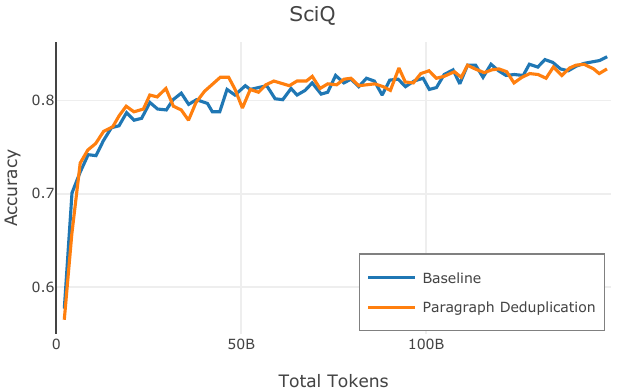}
	\end{subfigure}
	\quad
	\begin{subfigure}{0.31\textwidth}
		\includegraphics[width=\linewidth]{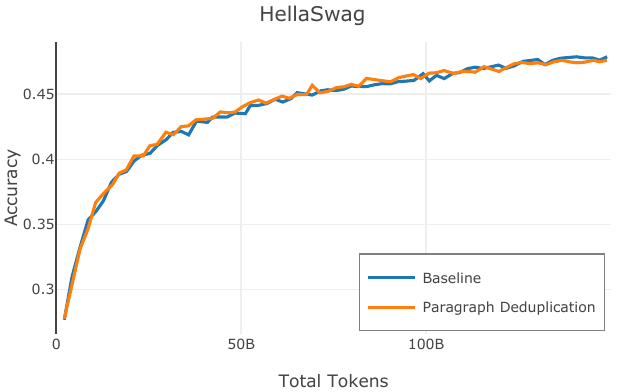}
	\end{subfigure}
	\quad
	\begin{subfigure}{0.31\textwidth}
		\includegraphics[width=\linewidth]{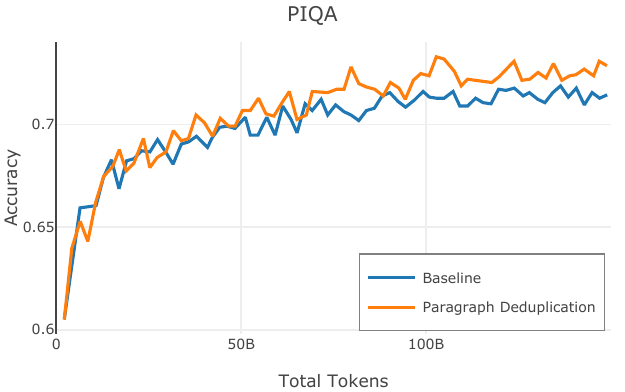}
	\end{subfigure}
	\caption{Results downstream tasks SciQ~\citep{sciq}, HellaSwag~\citep{zellers2019hellaswag}, and PIQA~\citep{piqa}}
\end{figure}

\label{sec:ablations_cc_dedupe:train}

\begin{figure}[h!]
	\centering
	\begin{subfigure}{0.31\textwidth}
		\includegraphics[width=\linewidth]{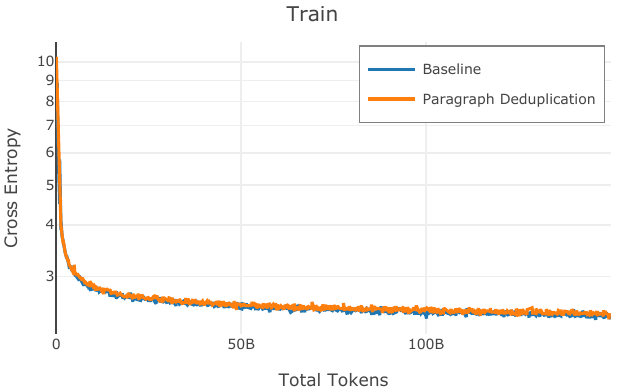}
	\end{subfigure}
	\caption{Training Cross Entropy}
\end{figure}

\clearpage

\subsection{Filtering of Personal Identifiable Information}
\label{sec:ablations_cc_pii_filtering}

\label{sec:ablations_cc_pii_filtering:ppl}

\begin{figure}[h!]
	\centering
	\begin{subfigure}{0.31\textwidth}
		\includegraphics[width=\linewidth]{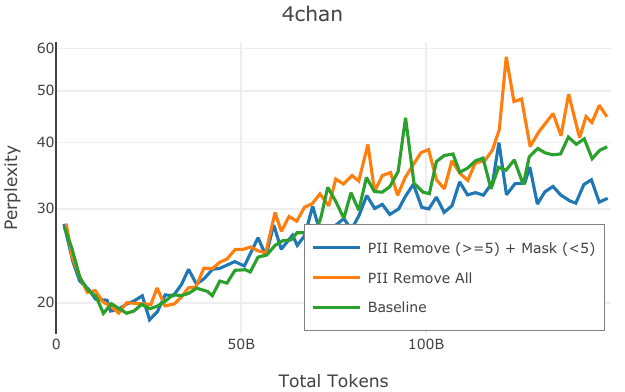}
	\end{subfigure}
	\quad
	\begin{subfigure}{0.31\textwidth}
		\includegraphics[width=\linewidth]{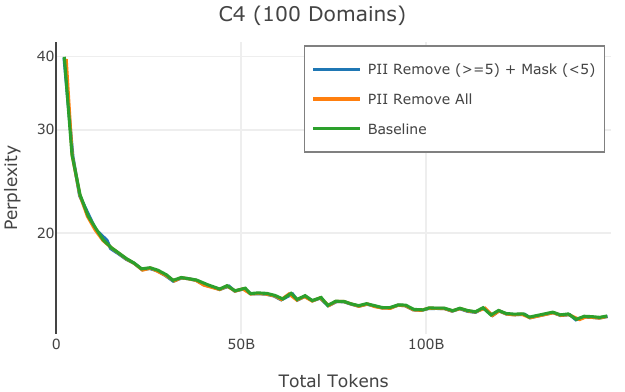}
	\end{subfigure}
	\quad
	\begin{subfigure}{0.31\textwidth}
		\includegraphics[width=\linewidth]{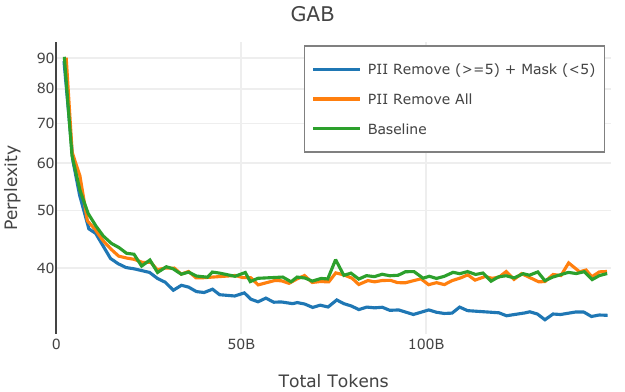}
	\end{subfigure}
	\caption{Perplexity results on Paloma~\citep{paloma}; subsets 4chan~\citep{papasavva2020raiders}, C4 100 dom~\citep{chronopoulou-etal-2022-efficient}, and Gab~\citep{zannettou2018gab}}
\end{figure}

\begin{figure}[h!]
	\centering
	\begin{subfigure}{0.31\textwidth}
		\includegraphics[width=\linewidth]{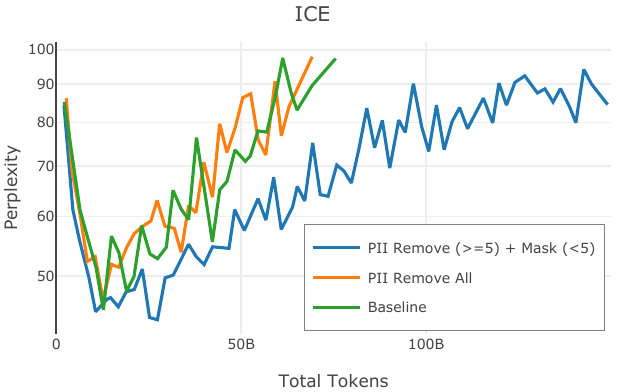}
	\end{subfigure}
	\quad
	\begin{subfigure}{0.31\textwidth}
		\includegraphics[width=\linewidth]{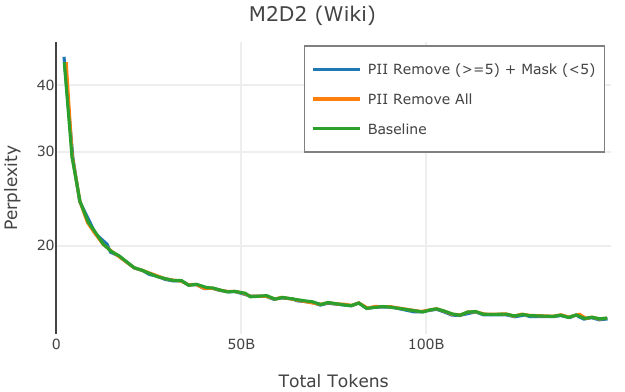}
	\end{subfigure}
	\quad
	\begin{subfigure}{0.31\textwidth}
		\includegraphics[width=\linewidth]{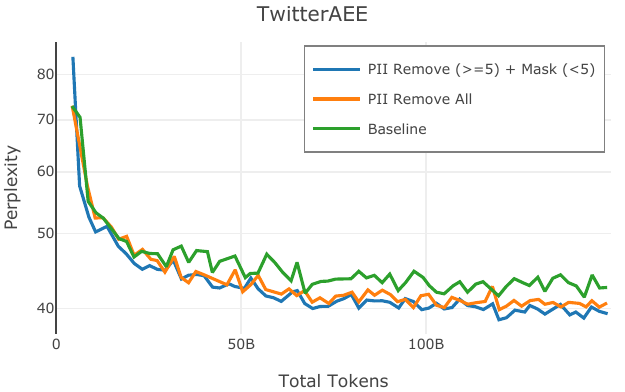}
	\end{subfigure}
	\caption{Perplexity results on Paloma~\citep{paloma}; subsets ICE~\citep{greenbaum1991ice}, M2D2~\citep{reid-etal-2022-m2d2} (Wiki), and Twitter AAE~\citep{blodgett-etal-2016-demographic}}
\end{figure}

\begin{figure}[h!]
	\centering
	\begin{subfigure}{0.31\textwidth}
		\includegraphics[width=\linewidth]{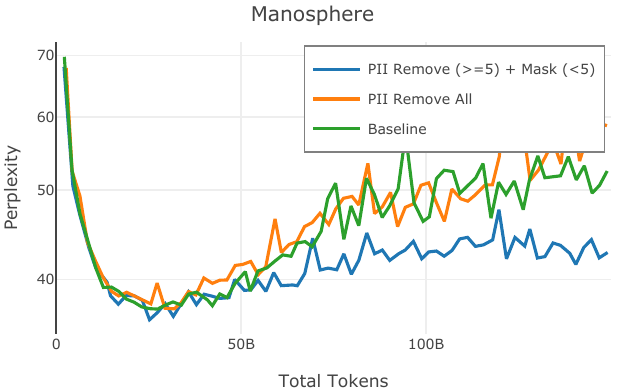}
	\end{subfigure}
	\caption{Perplexity results on Paloma~\citep{paloma}; subsets Manosphere~\citep{ribeiroevolution2021}}
\end{figure}

\begin{figure}[h!]
	\centering
	\begin{subfigure}{0.31\textwidth}
		\includegraphics[width=\linewidth]{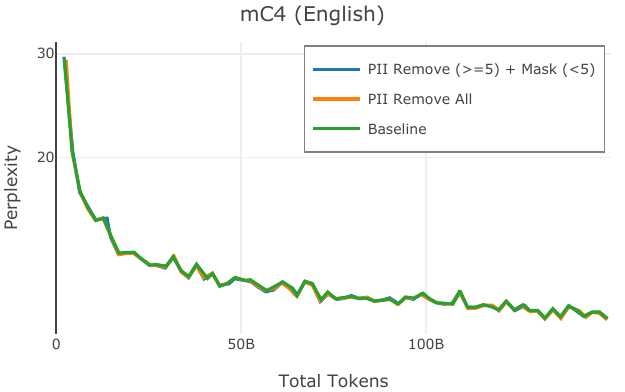}
	\end{subfigure}
	\quad
	\begin{subfigure}{0.31\textwidth}
		\includegraphics[width=\linewidth]{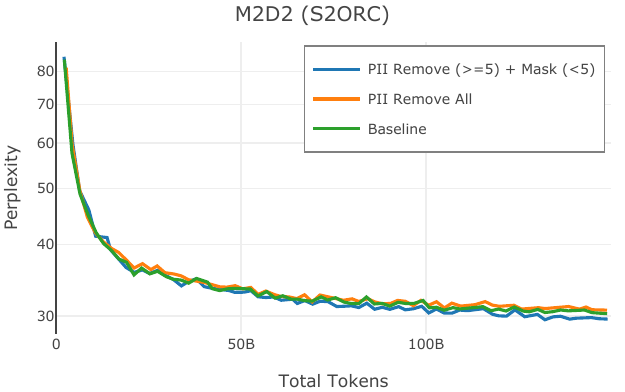}
	\end{subfigure}
	\quad
	\begin{subfigure}{0.31\textwidth}
		\includegraphics[width=\linewidth]{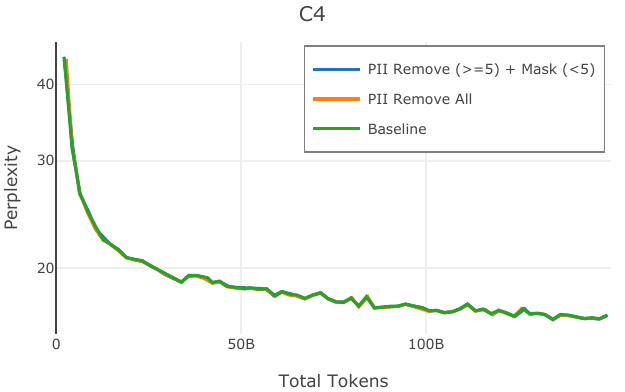}
	\end{subfigure}
	\caption{Perplexity results on Paloma~\citep{paloma}; subsets mC4~\citep{mc4} (English), M2D2~\citep{reid-etal-2022-m2d2} (S2ORC), and C4~\citep{raffel2020exploring,dodge-etal-2021-documenting}}
\end{figure}

\label{sec:ablations_cc_pii_filtering:downstream}

\begin{figure}[h!]
	\centering
	\begin{subfigure}{0.31\textwidth}
		\includegraphics[width=\linewidth]{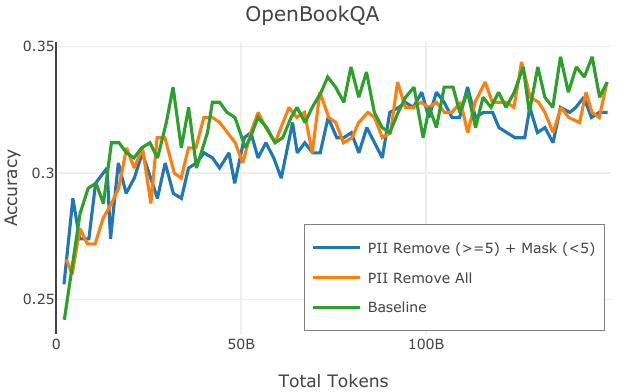}
	\end{subfigure}
	\quad
	\begin{subfigure}{0.31\textwidth}
		\includegraphics[width=\linewidth]{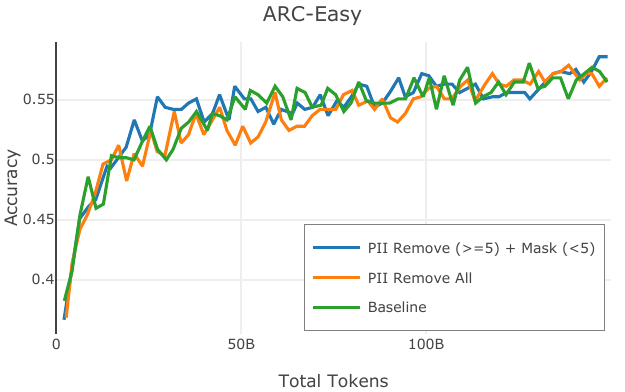}
	\end{subfigure}
	\quad
	\begin{subfigure}{0.31\textwidth}
		\includegraphics[width=\linewidth]{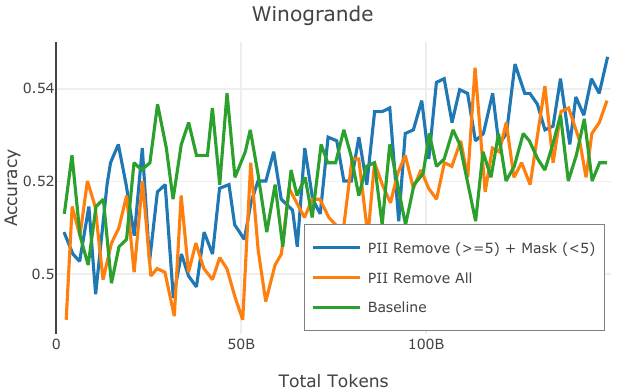}
	\end{subfigure}
	\caption{Results downstream tasks OpenBookQA~\citep{openbookQA}, ARC-E~\citep{arc}, and WinoGrande~\citep{winogrande}}
\end{figure}

\begin{figure}[h!]
	\centering
	\begin{subfigure}{0.31\textwidth}
		\includegraphics[width=\linewidth]{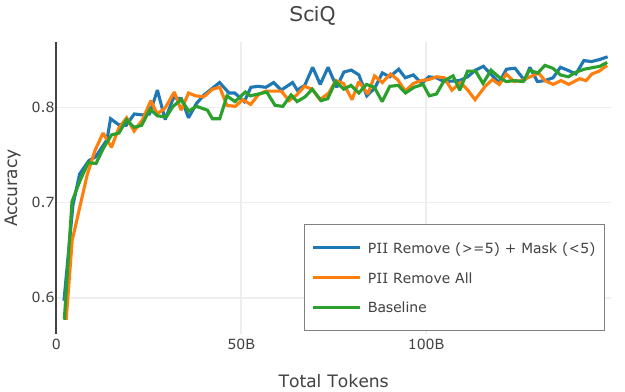}
	\end{subfigure}
	\quad
	\begin{subfigure}{0.31\textwidth}
		\includegraphics[width=\linewidth]{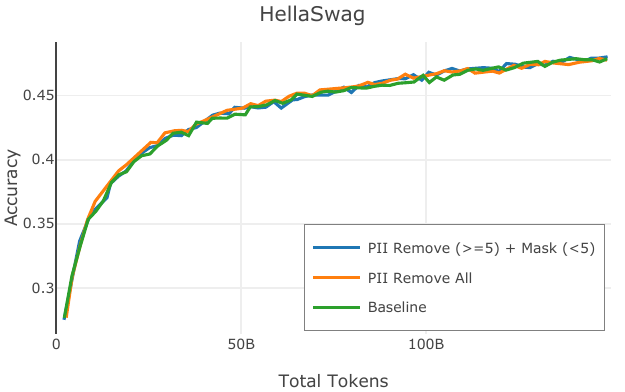}
	\end{subfigure}
	\quad
	\begin{subfigure}{0.31\textwidth}
		\includegraphics[width=\linewidth]{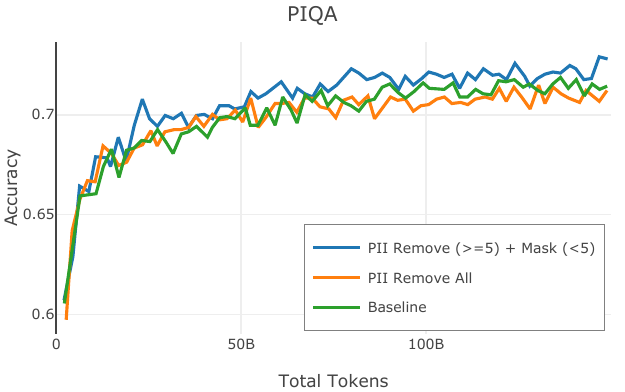}
	\end{subfigure}
	\caption{Results downstream tasks SciQ~\citep{sciq}, HellaSwag~\citep{zellers2019hellaswag}, and PIQA~\citep{piqa}}
\end{figure}

\label{sec:ablations_cc_pii_filtering:train}

\begin{figure}[h!]
	\centering
	\begin{subfigure}{0.31\textwidth}
		\includegraphics[width=\linewidth]{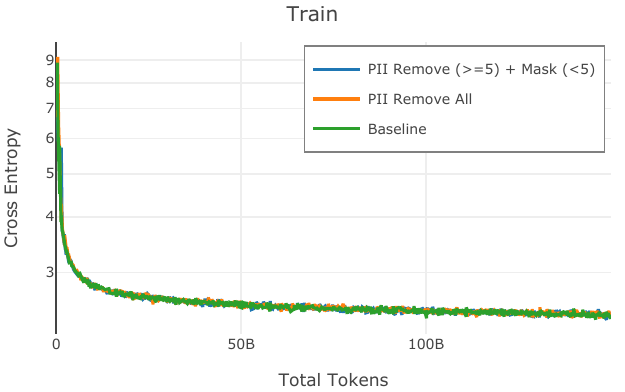}
	\end{subfigure}
	\caption{Training Cross Entropy}
\end{figure}

\FloatBarrier

\subsection{Comparing Quality Filters for Web Pipeline}
\label{sec:ablations_cc_quality_only}

\label{sec:ablations_cc_quality_only:ppl}

\begin{figure}[h!]
	\centering
	\begin{subfigure}{0.31\textwidth}
		\includegraphics[width=\linewidth]{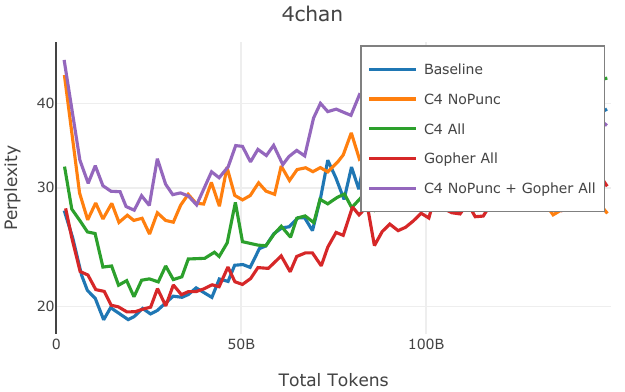}
	\end{subfigure}
	\quad
	\begin{subfigure}{0.31\textwidth}
		\includegraphics[width=\linewidth]{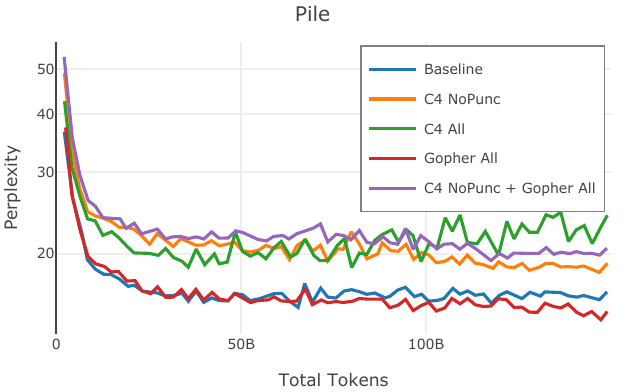}
	\end{subfigure}
	\quad
	\begin{subfigure}{0.31\textwidth}
		\includegraphics[width=\linewidth]{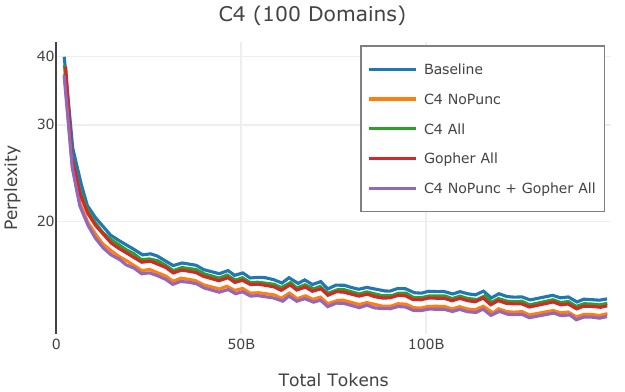}
	\end{subfigure}
	\caption{Perplexity results on Paloma~\citep{paloma}; subsets 4chan~\citep{papasavva2020raiders}, Pile~\citep{Gao2020ThePA} (Val), and C4 100 dom~\citep{chronopoulou-etal-2022-efficient}}
\end{figure}

\begin{figure}[h!]
	\centering
	\begin{subfigure}{0.31\textwidth}
		\includegraphics[width=\linewidth]{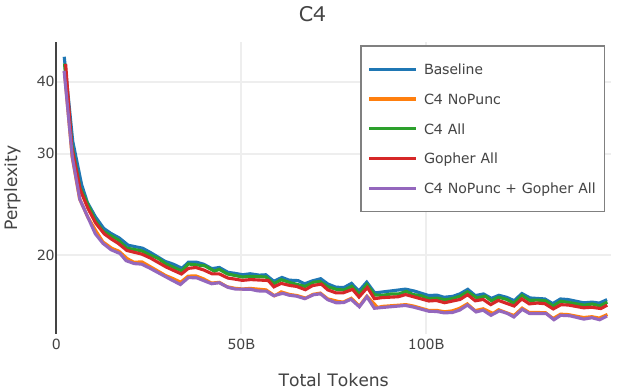}
	\end{subfigure}
	\quad
	\begin{subfigure}{0.31\textwidth}
		\includegraphics[width=\linewidth]{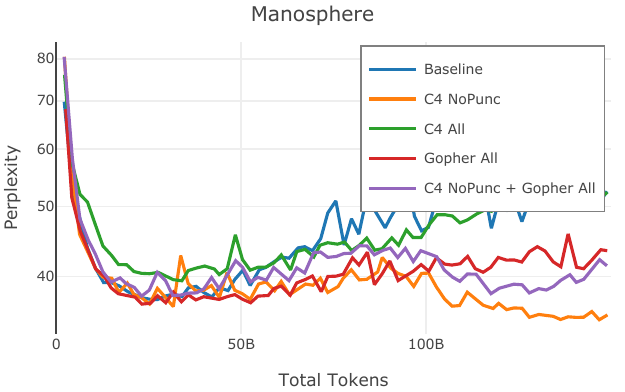}
	\end{subfigure}
	\caption{Perplexity results on Paloma~\citep{paloma}; subsets C4~\citep{raffel2020exploring,dodge-etal-2021-documenting} and Manosphere~\citep{ribeiroevolution2021}}
\end{figure}

\begin{figure}[h!]
	\centering
	\begin{subfigure}{0.31\textwidth}
		\includegraphics[width=\linewidth]{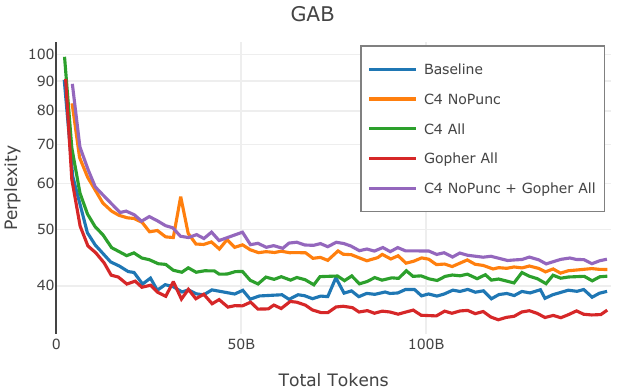}
	\end{subfigure}
	\quad
	\begin{subfigure}{0.31\textwidth}
		\includegraphics[width=\linewidth]{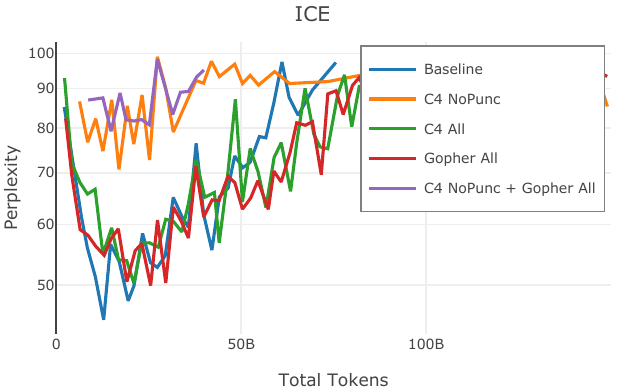}
	\end{subfigure}
	\quad
	\begin{subfigure}{0.31\textwidth}
		\includegraphics[width=\linewidth]{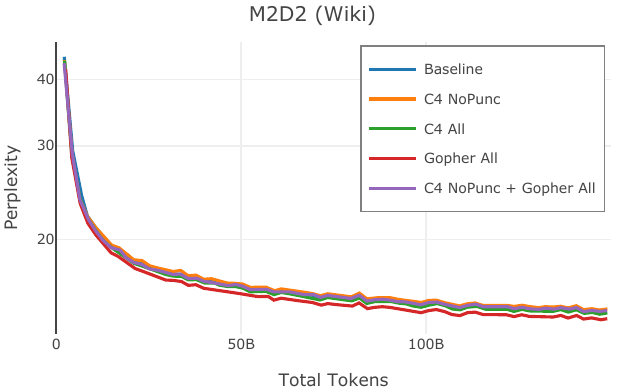}
	\end{subfigure}
	\caption{Perplexity results on Paloma~\citep{paloma}; subsets Gab~\citep{zannettou2018gab}, ICE~\citep{greenbaum1991ice}, and M2D2~\citep{reid-etal-2022-m2d2} (Wiki)}
\end{figure}

\begin{figure}[h!]
	\centering
	\begin{subfigure}{0.31\textwidth}
		\includegraphics[width=\linewidth]{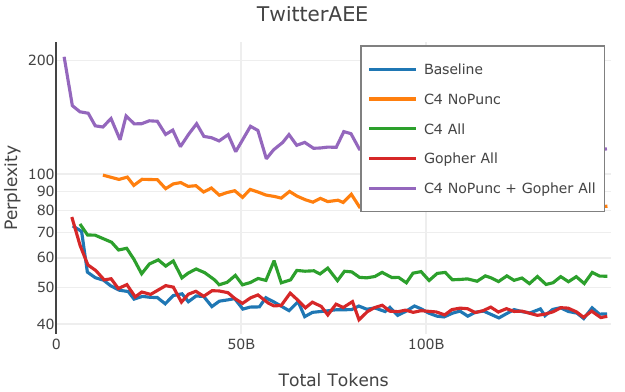}
	\end{subfigure}
	\quad
	\begin{subfigure}{0.31\textwidth}
		\includegraphics[width=\linewidth]{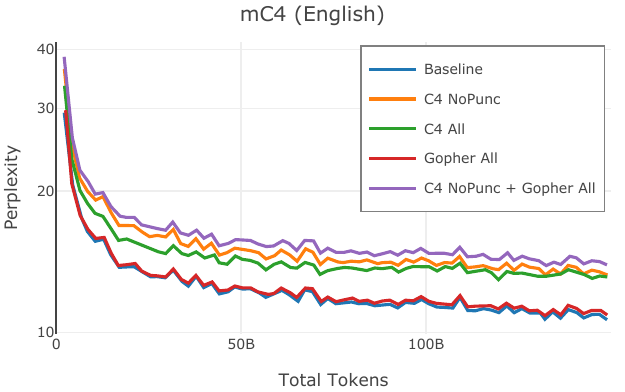}
	\end{subfigure}
	\quad
	\begin{subfigure}{0.31\textwidth}
		\includegraphics[width=\linewidth]{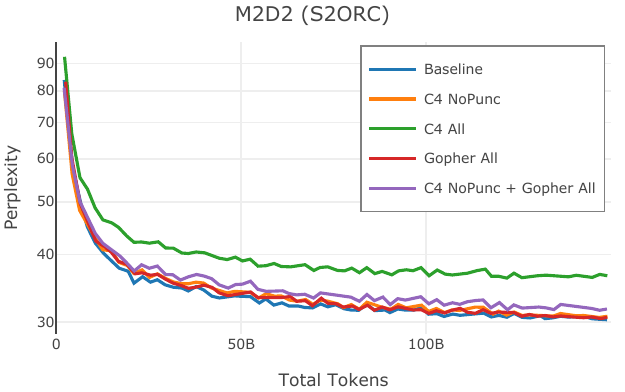}
	\end{subfigure}
	\caption{Perplexity results on Paloma~\citep{paloma}; subsets Twitter AAE~\citep{blodgett-etal-2016-demographic}, mC4~\citep{mc4} (English), and M2D2~\citep{reid-etal-2022-m2d2} (S2ORC)}
\end{figure}

\label{sec:ablations_cc_quality_only:downstream}

\begin{figure}[h!]
	\centering
	\begin{subfigure}{0.31\textwidth}
		\includegraphics[width=\linewidth]{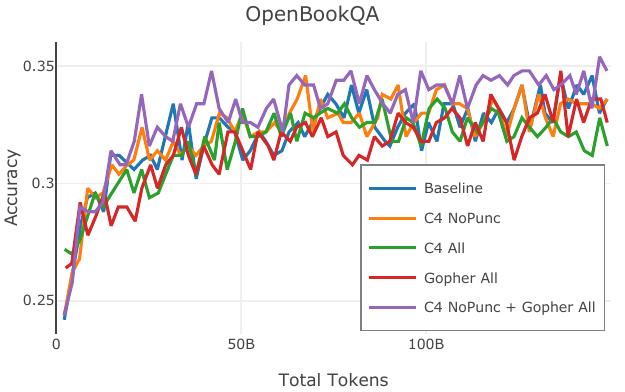}
	\end{subfigure}
	\quad
	\begin{subfigure}{0.31\textwidth}
		\includegraphics[width=\linewidth]{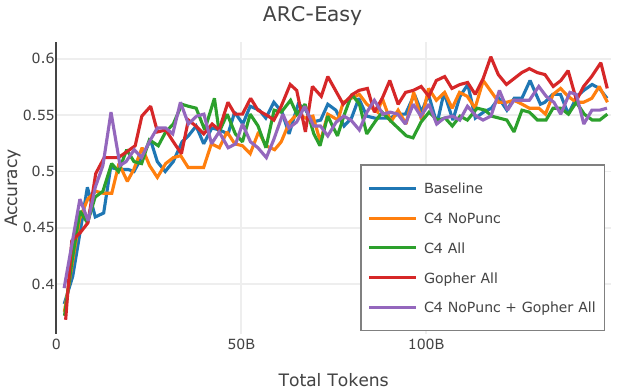}
	\end{subfigure}
	\quad
	\begin{subfigure}{0.31\textwidth}
		\includegraphics[width=\linewidth]{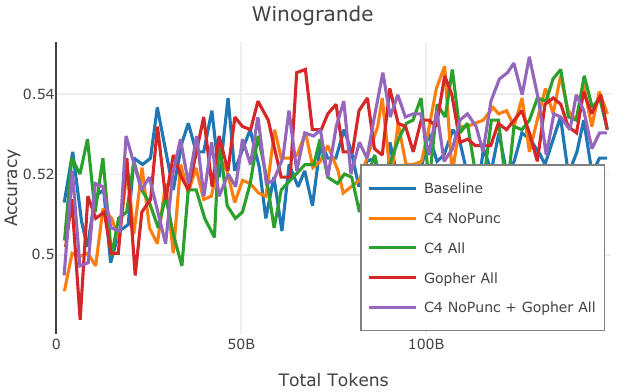}
	\end{subfigure}
	\caption{Results downstream tasks OpenBookQA~\citep{openbookQA}, ARC-E~\citep{arc}, and WinoGrande~\citep{winogrande}}
\end{figure}

\begin{figure}[h!]
	\centering
	\begin{subfigure}{0.31\textwidth}
		\includegraphics[width=\linewidth]{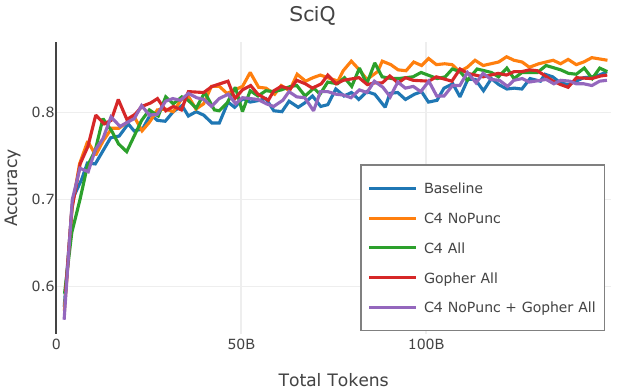}
	\end{subfigure}
	\quad
	\begin{subfigure}{0.31\textwidth}
		\includegraphics[width=\linewidth]{experiments/ablations_cc_quality_only/downstream/hellaswag.pdf}
	\end{subfigure}
	\quad
	\begin{subfigure}{0.31\textwidth}
		\includegraphics[width=\linewidth]{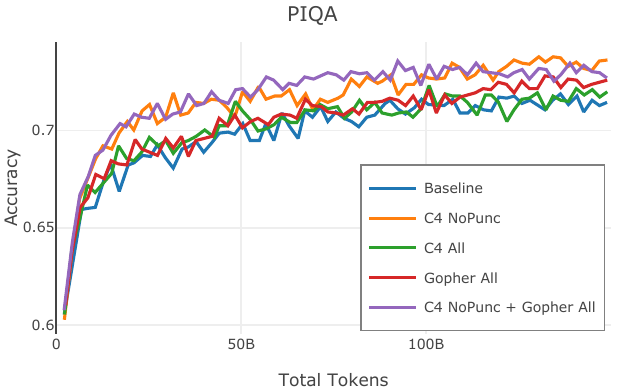}
	\end{subfigure}
	\caption{Results downstream tasks SciQ~\citep{sciq}, HellaSwag~\citep{zellers2019hellaswag}, and PIQA~\citep{piqa}}
\end{figure}

\label{sec:ablations_cc_quality_only:train}

\begin{figure}[h!]
	\centering
	\begin{subfigure}{0.31\textwidth}
		\includegraphics[width=\linewidth]{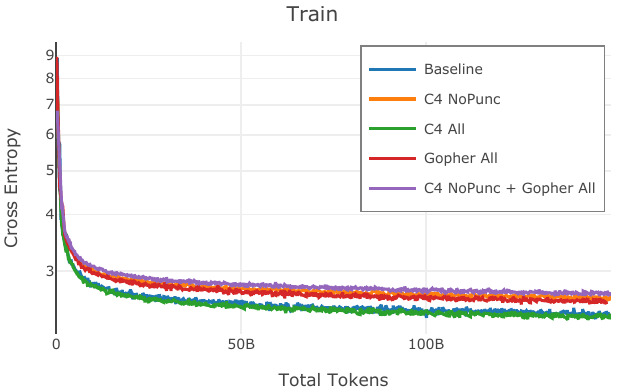}
	\end{subfigure}
	\caption{Training Cross Entropy}
\end{figure}

\FloatBarrier

\subsection{Full Comparison of Web Pipeline}
\label{sec:ablations_cc_to_quality_plus_content}

\label{sec:ablations_cc_to_quality_plus_content:ppl}

\begin{figure}[h!]
	\centering
	\begin{subfigure}{0.31\textwidth}
		\includegraphics[width=\linewidth]{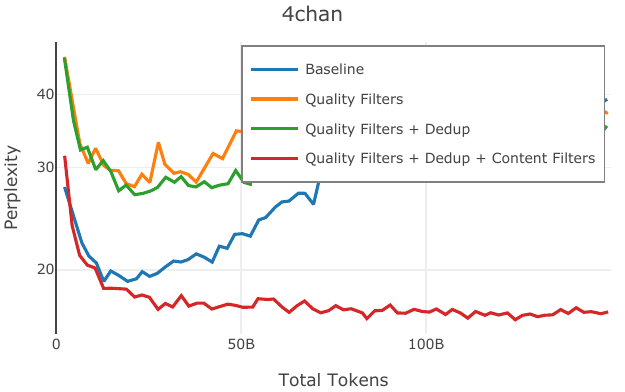}
	\end{subfigure}
	\quad
	\begin{subfigure}{0.31\textwidth}
		\includegraphics[width=\linewidth]{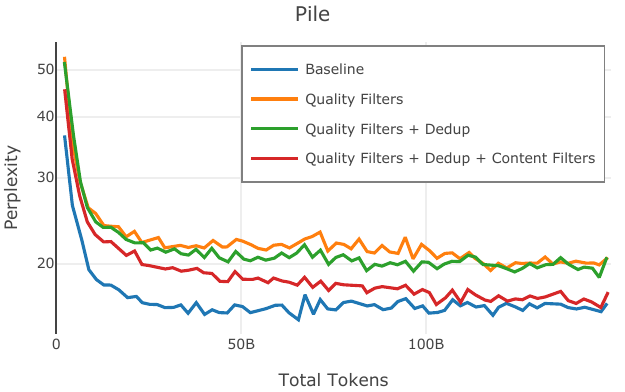}
	\end{subfigure}
	\quad
	\begin{subfigure}{0.31\textwidth}
		\includegraphics[width=\linewidth]{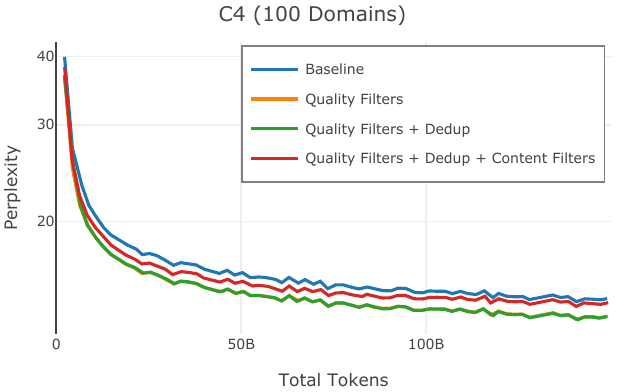}
	\end{subfigure}
	\caption{Perplexity results on Paloma~\citep{paloma}; subsets 4chan~\citep{papasavva2020raiders}, Pile~\citep{Gao2020ThePA} (Val), and C4 100 dom~\citep{chronopoulou-etal-2022-efficient}}
\end{figure}

\begin{figure}[h!]
	\centering
	\begin{subfigure}{0.31\textwidth}
		\includegraphics[width=\linewidth]{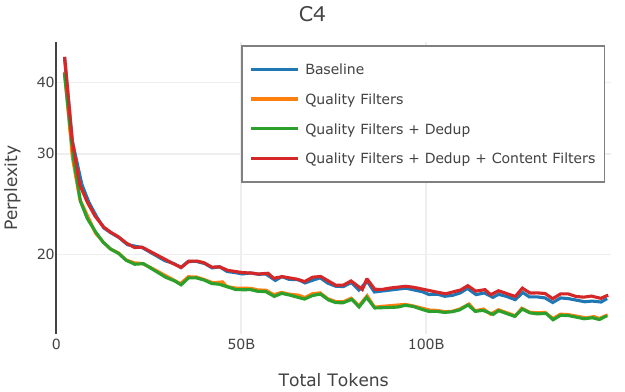}
	\end{subfigure}
	\quad
	\begin{subfigure}{0.31\textwidth}
		\includegraphics[width=\linewidth]{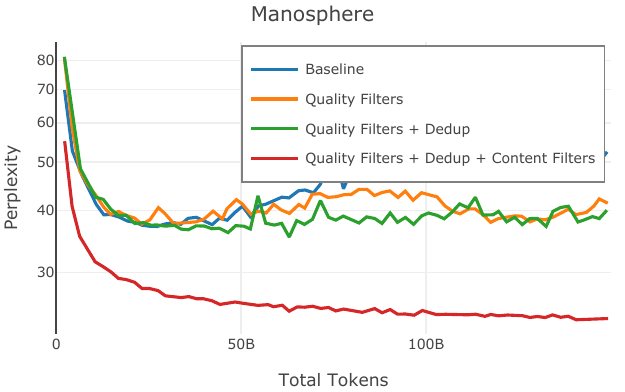}
	\end{subfigure}
	\caption{Perplexity results on Paloma~\citep{paloma}; subsets C4~\citep{raffel2020exploring,dodge-etal-2021-documenting} and Manosphere~\citep{ribeiroevolution2021}}
\end{figure}

\begin{figure}[h!]
	\centering
	\begin{subfigure}{0.31\textwidth}
		\includegraphics[width=\linewidth]{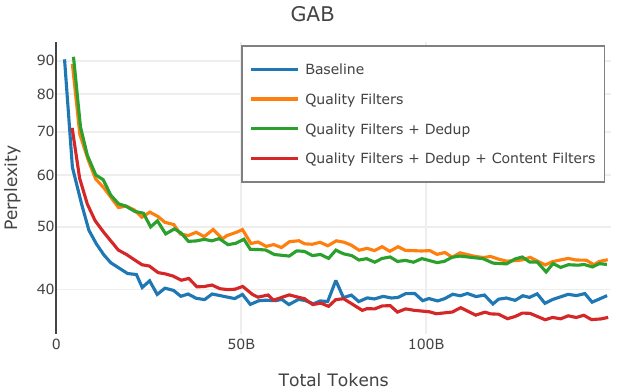}
	\end{subfigure}
	\quad
	\begin{subfigure}{0.31\textwidth}
		\includegraphics[width=\linewidth]{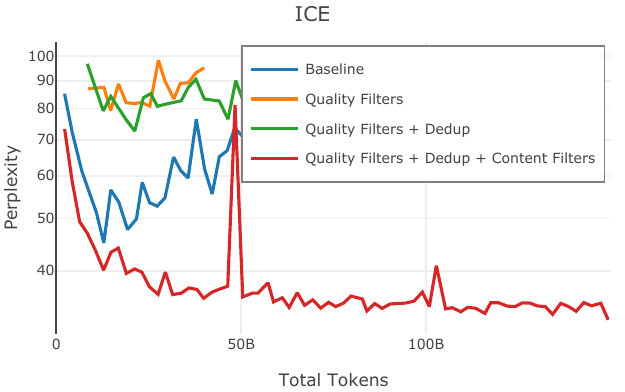}
	\end{subfigure}
	\quad
	\begin{subfigure}{0.31\textwidth}
		\includegraphics[width=\linewidth]{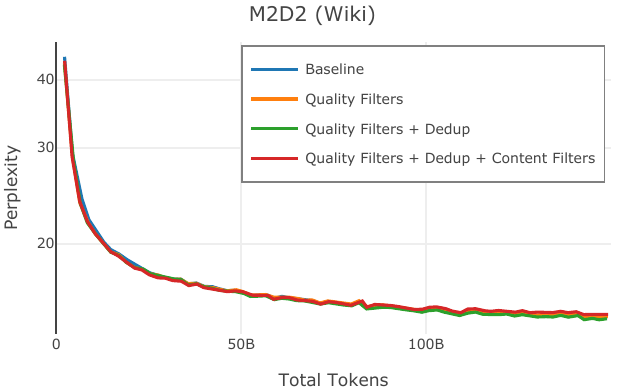}
	\end{subfigure}
	\caption{Perplexity results on Paloma~\citep{paloma}; subsets Gab~\citep{zannettou2018gab}, ICE~\citep{greenbaum1991ice}, and M2D2~\citep{reid-etal-2022-m2d2} (Wiki)}
\end{figure}

\begin{figure}[h!]
	\centering
	\begin{subfigure}{0.31\textwidth}
		\includegraphics[width=\linewidth]{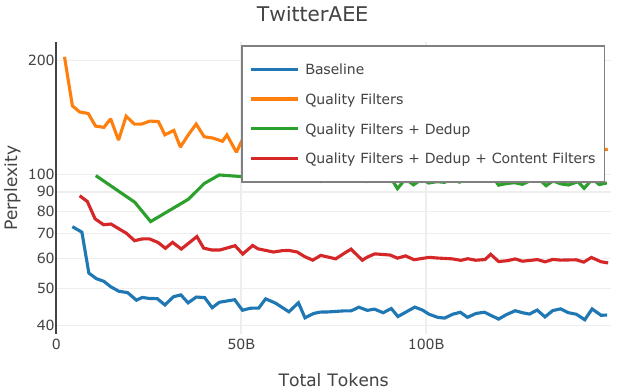}
	\end{subfigure}
	\quad
	\begin{subfigure}{0.31\textwidth}
		\includegraphics[width=\linewidth]{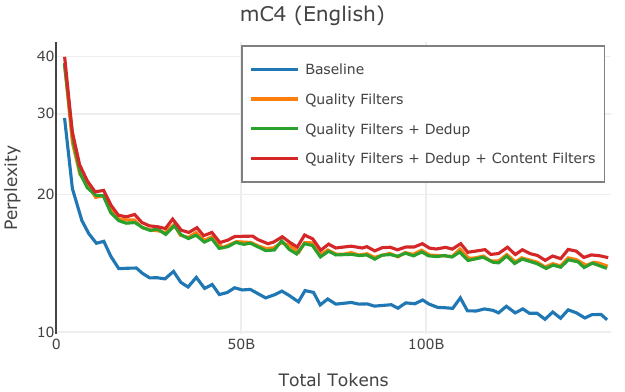}
	\end{subfigure}
	\quad
	\begin{subfigure}{0.31\textwidth}
		\includegraphics[width=\linewidth]{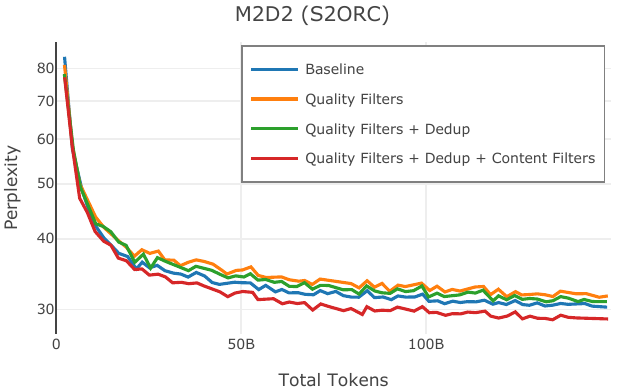}
	\end{subfigure}
	\caption{Perplexity results on Paloma~\citep{paloma}; subsets Twitter AAE~\citep{blodgett-etal-2016-demographic}, mC4~\citep{mc4} (English), and M2D2~\citep{reid-etal-2022-m2d2} (S2ORC)}
\end{figure}

\label{sec:ablations_cc_to_quality_plus_content:downstream}

\begin{figure}[h!]
	\centering
	\begin{subfigure}{0.31\textwidth}
		\includegraphics[width=\linewidth]{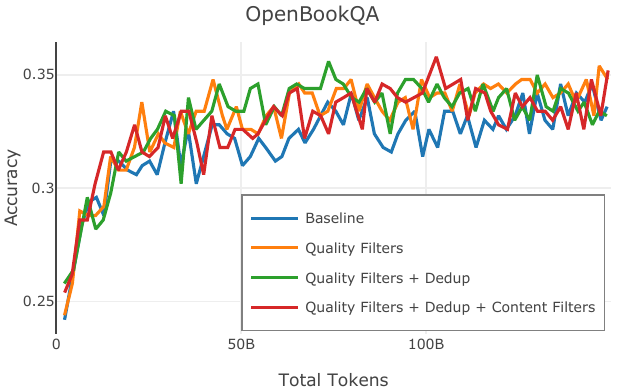}
	\end{subfigure}
	\quad
	\begin{subfigure}{0.31\textwidth}
		\includegraphics[width=\linewidth]{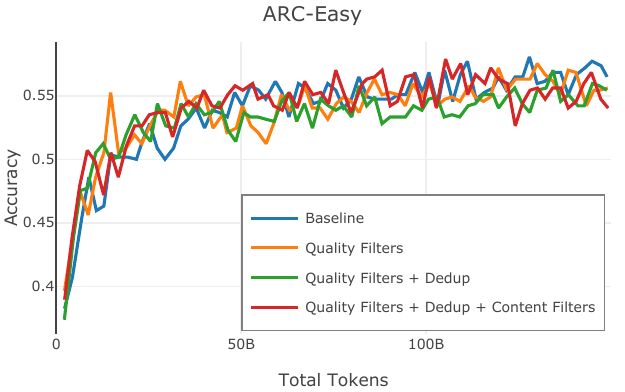}
	\end{subfigure}
	\quad
	\begin{subfigure}{0.31\textwidth}
		\includegraphics[width=\linewidth]{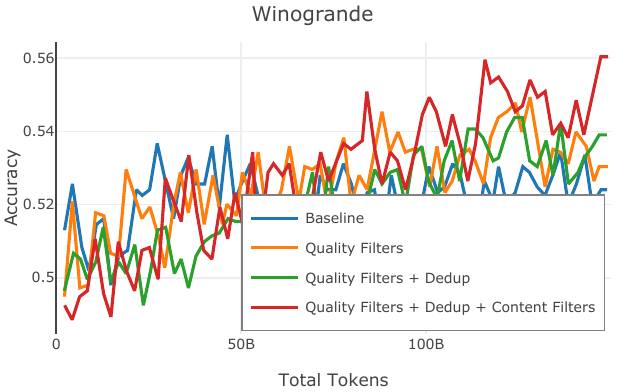}
	\end{subfigure}
	\caption{Results downstream tasks OpenBookQA~\citep{openbookQA}, ARC-E~\citep{arc}, and WinoGrande~\citep{winogrande}}
\end{figure}

\begin{figure}[h!]
	\centering
	\begin{subfigure}{0.31\textwidth}
		\includegraphics[width=\linewidth]{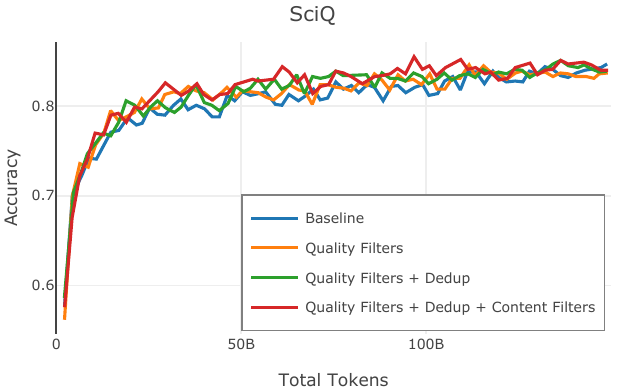}
	\end{subfigure}
	\quad
	\begin{subfigure}{0.31\textwidth}
		\includegraphics[width=\linewidth]{experiments/ablations_cc_to_quality_plus_content/downstream/hellaswag.pdf}
	\end{subfigure}
	\quad
	\begin{subfigure}{0.31\textwidth}
		\includegraphics[width=\linewidth]{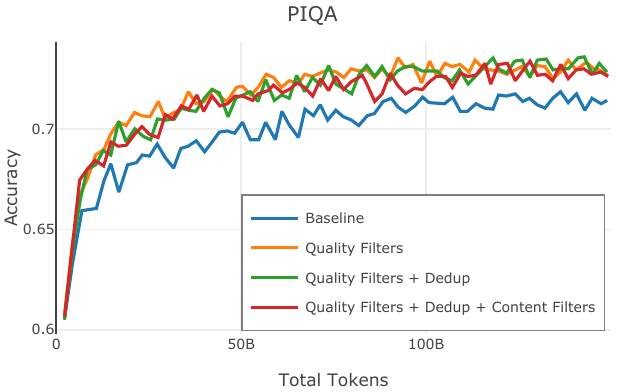}
	\end{subfigure}
	\caption{Results downstream tasks SciQ~\citep{sciq}, HellaSwag~\citep{zellers2019hellaswag}, and PIQA~\citep{piqa}}
\end{figure}

\label{sec:ablations_cc_to_quality_plus_content:train}

\begin{figure}[h!]
	\centering
	\begin{subfigure}{0.31\textwidth}
		\includegraphics[width=\linewidth]{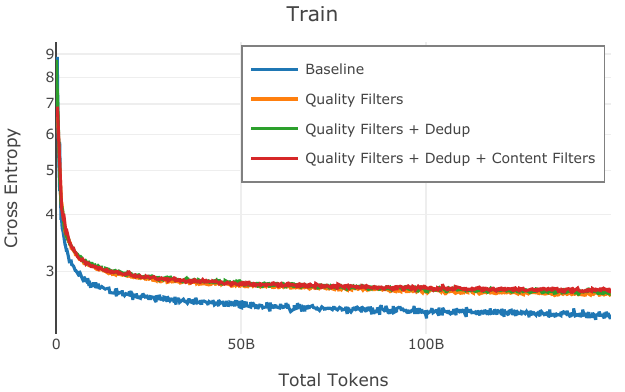}
	\end{subfigure}
	\caption{Training Cross Entropy}
\end{figure}

\FloatBarrier

\subsection{Toxicity Filtering in Web Pipeline}
\label{sec:ablations_cc_toxic_filtering}

\label{sec:ablations_cc_toxic_filtering:ppl}

\begin{figure}[h!]
	\centering
	\begin{subfigure}{0.31\textwidth}
		\includegraphics[width=\linewidth]{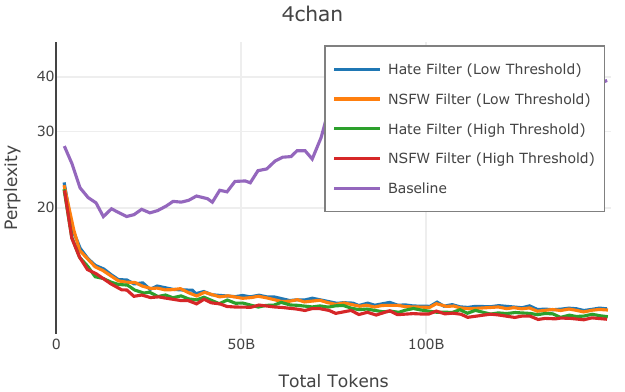}
	\end{subfigure}
	\quad
	\begin{subfigure}{0.31\textwidth}
		\includegraphics[width=\linewidth]{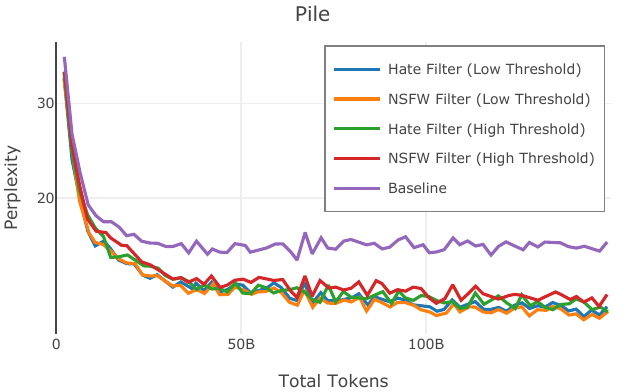}
	\end{subfigure}
	\quad
	\begin{subfigure}{0.31\textwidth}
		\includegraphics[width=\linewidth]{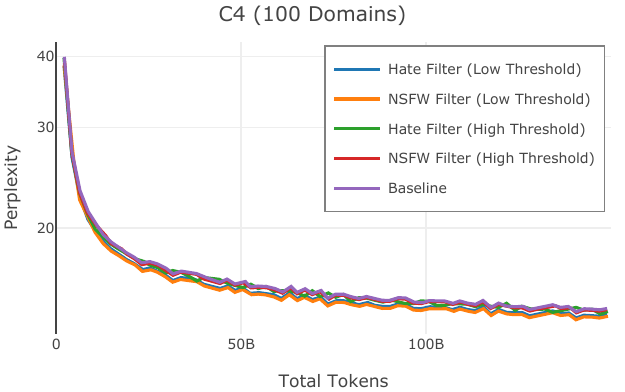}
	\end{subfigure}
	\caption{Perplexity results on Paloma~\citep{paloma}; subsets 4chan~\citep{papasavva2020raiders}, Pile~\citep{Gao2020ThePA} (Val), and C4 100 dom~\citep{chronopoulou-etal-2022-efficient}}
\end{figure}

\begin{figure}[h!]
	\centering
	\begin{subfigure}{0.31\textwidth}
		\includegraphics[width=\linewidth]{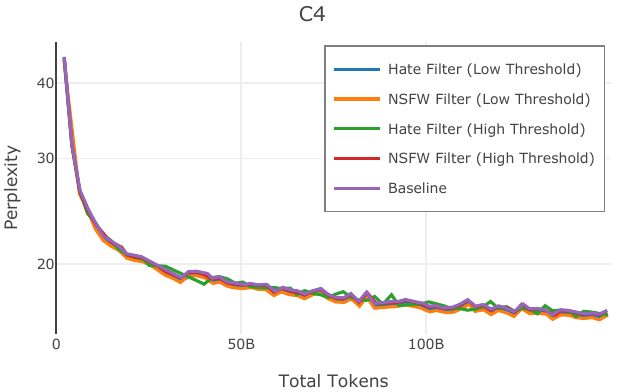}
	\end{subfigure}
	\quad
	\begin{subfigure}{0.31\textwidth}
		\includegraphics[width=\linewidth]{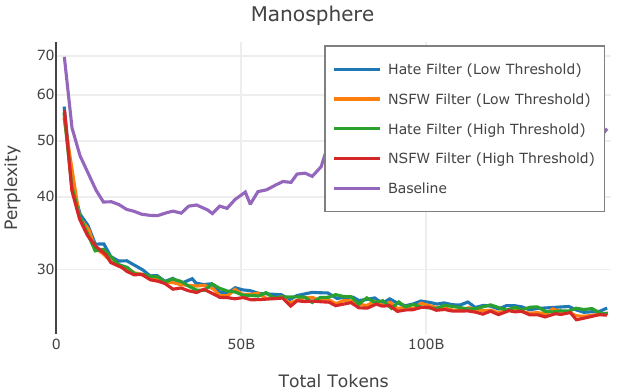}
	\end{subfigure}
	\caption{Perplexity results on Paloma~\citep{paloma}; subsets C4~\citep{raffel2020exploring,dodge-etal-2021-documenting} and Manosphere~\citep{ribeiroevolution2021}}
\end{figure}

\begin{figure}[h!]
	\centering
	\begin{subfigure}{0.31\textwidth}
		\includegraphics[width=\linewidth]{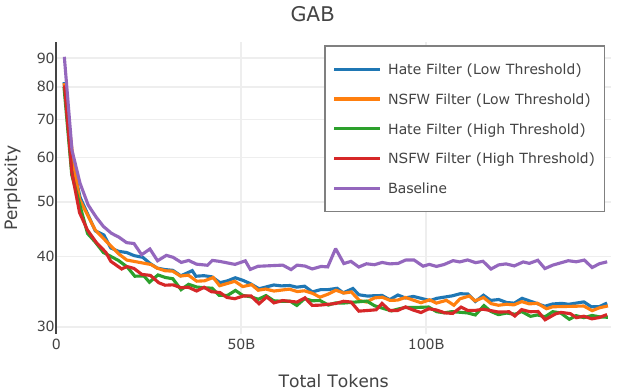}
	\end{subfigure}
	\quad
	\begin{subfigure}{0.31\textwidth}
		\includegraphics[width=\linewidth]{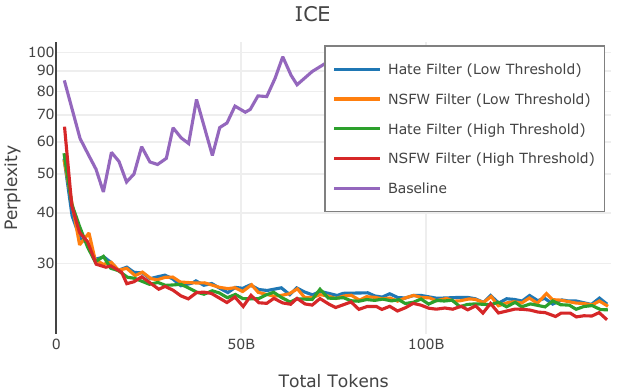}
	\end{subfigure}
	\quad
	\begin{subfigure}{0.31\textwidth}
		\includegraphics[width=\linewidth]{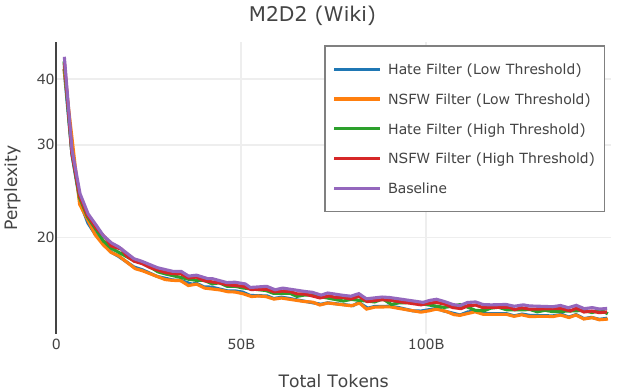}
	\end{subfigure}
	\caption{Perplexity results on Paloma~\citep{paloma}; subsets Gab~\citep{zannettou2018gab}, ICE~\citep{greenbaum1991ice}, and M2D2~\citep{reid-etal-2022-m2d2} (Wiki)}
\end{figure}

\begin{figure}[h!]
	\centering
	\begin{subfigure}{0.31\textwidth}
		\includegraphics[width=\linewidth]{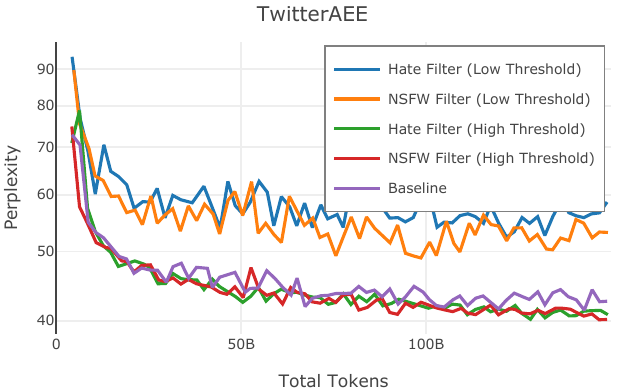}
	\end{subfigure}
	\quad
	\begin{subfigure}{0.31\textwidth}
		\includegraphics[width=\linewidth]{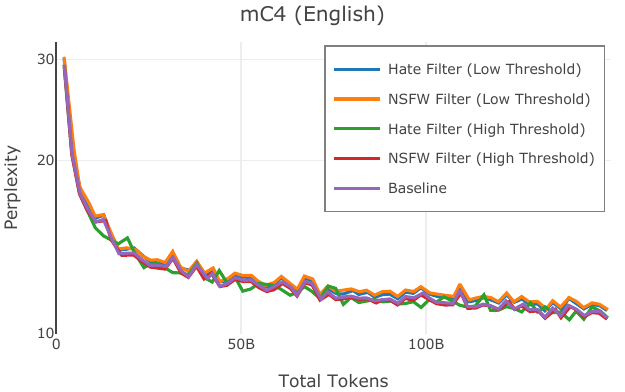}
	\end{subfigure}
	\quad
	\begin{subfigure}{0.31\textwidth}
		\includegraphics[width=\linewidth]{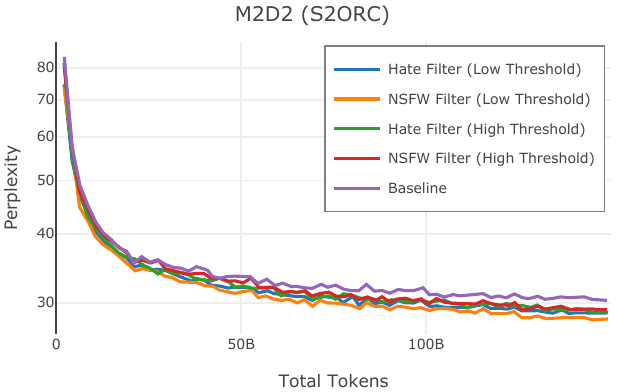}
	\end{subfigure}
	\caption{Perplexity results on Paloma~\citep{paloma}; subsets Twitter AAE~\citep{blodgett-etal-2016-demographic}, mC4~\citep{mc4} (English), and M2D2~\citep{reid-etal-2022-m2d2} (S2ORC)}
\end{figure}

\label{sec:ablations_cc_toxic_filtering:downstream}

\begin{figure}[h!]
	\centering
	\begin{subfigure}{0.31\textwidth}
		\includegraphics[width=\linewidth]{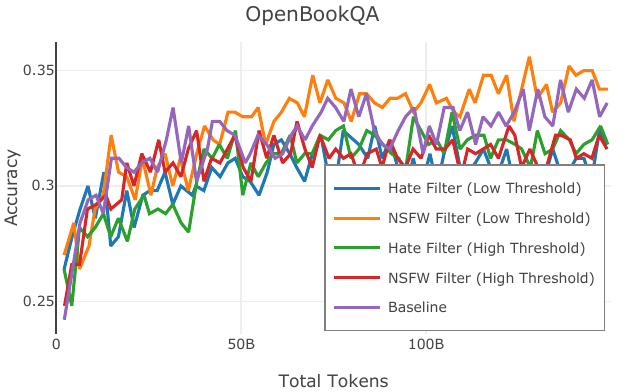}
	\end{subfigure}
	\quad
	\begin{subfigure}{0.31\textwidth}
		\includegraphics[width=\linewidth]{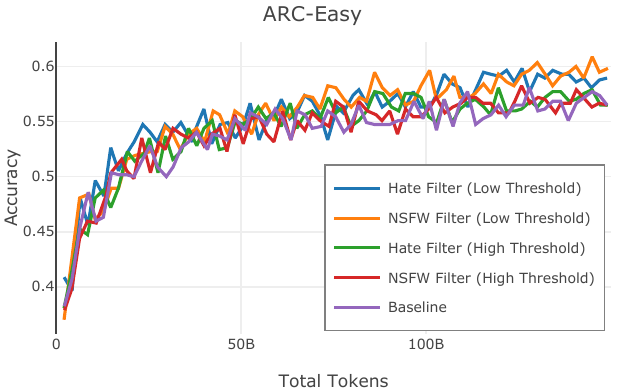}
	\end{subfigure}
	\quad
	\begin{subfigure}{0.31\textwidth}
		\includegraphics[width=\linewidth]{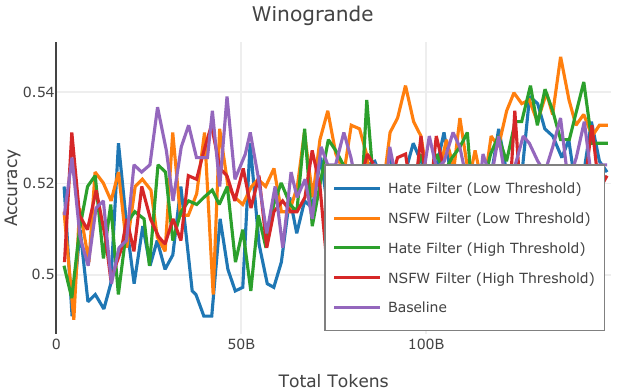}
	\end{subfigure}
	\caption{Results downstream tasks OpenBookQA~\citep{openbookQA}, ARC-E~\citep{arc}, and WinoGrande~\citep{winogrande}}
\end{figure}

\begin{figure}[h!]
	\centering
	\begin{subfigure}{0.31\textwidth}
		\includegraphics[width=\linewidth]{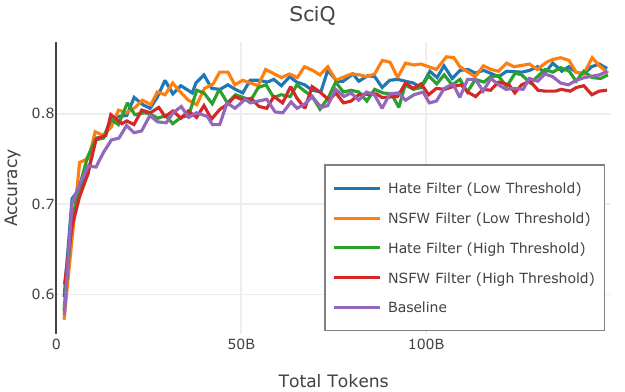}
	\end{subfigure}
	\quad
	\begin{subfigure}{0.31\textwidth}
		\includegraphics[width=\linewidth]{experiments/ablations_cc_toxic_filtering/downstream/hellaswag.pdf}
	\end{subfigure}
	\quad
	\begin{subfigure}{0.31\textwidth}
		\includegraphics[width=\linewidth]{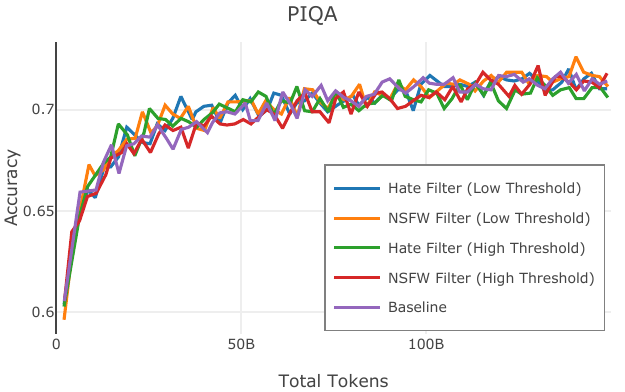}
	\end{subfigure}
	\caption{Results downstream tasks SciQ~\citep{sciq}, HellaSwag~\citep{zellers2019hellaswag}, and PIQA~\citep{piqa}}
\end{figure}

\label{sec:ablations_cc_toxic_filtering:train}

\begin{figure}[h!]
	\centering
	\begin{subfigure}{0.31\textwidth}
		\includegraphics[width=\linewidth]{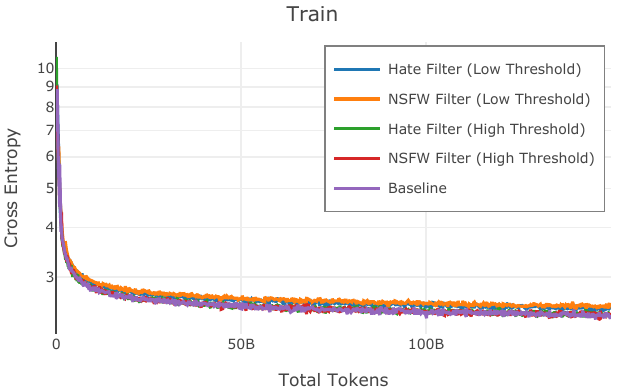}
	\end{subfigure}
	\caption{Training Cross Entropy}
\end{figure}

\FloatBarrier

\subsection{Comparing Code Processing Pipeline}
\label{sec:ablations_code_stack_v2_vs_v4}

\label{sec:ablations_code_stack_v2_vs_v4:ppl}

\begin{figure}[h!]
	\centering
	\begin{subfigure}{0.31\textwidth}
		\includegraphics[width=\linewidth]{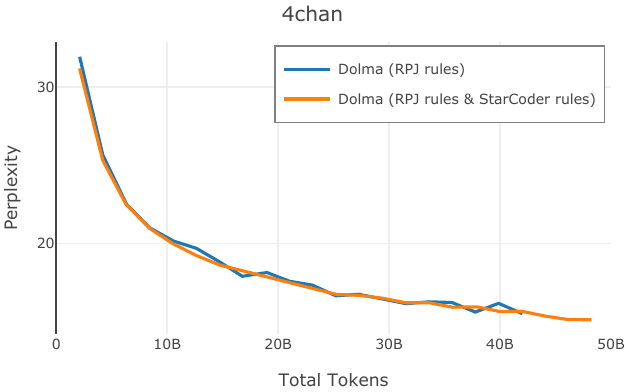}
	\end{subfigure}
	\quad
	\begin{subfigure}{0.31\textwidth}
		\includegraphics[width=\linewidth]{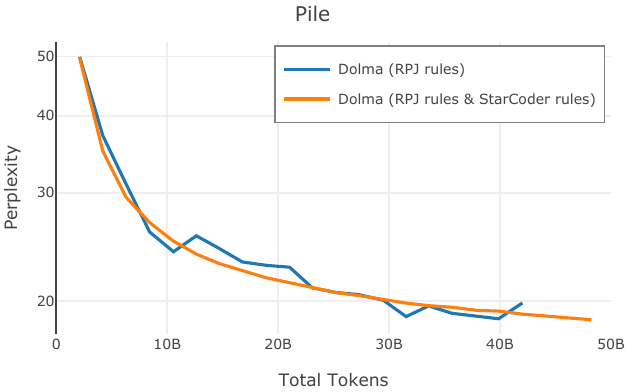}
	\end{subfigure}
	\quad
	\begin{subfigure}{0.31\textwidth}
		\includegraphics[width=\linewidth]{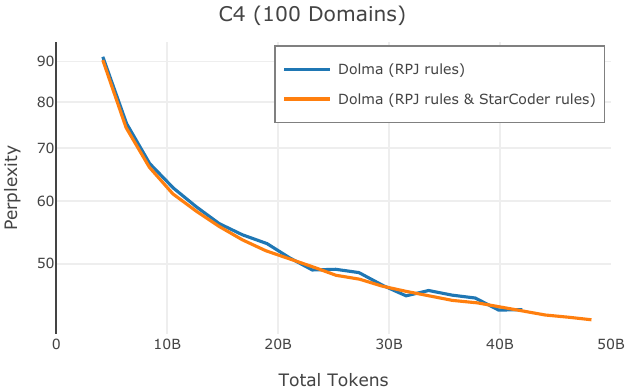}
	\end{subfigure}
	\caption{Perplexity results on Paloma~\citep{paloma}; subsets 4chan~\citep{papasavva2020raiders}, Pile~\citep{Gao2020ThePA} (Val), and C4 100 dom~\citep{chronopoulou-etal-2022-efficient}}
\end{figure}

\begin{figure}[h!]
	\centering
	\begin{subfigure}{0.31\textwidth}
		\includegraphics[width=\linewidth]{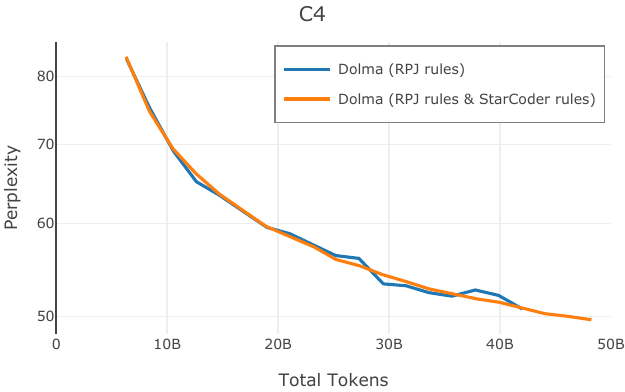}
	\end{subfigure}
	\quad
	\begin{subfigure}{0.31\textwidth}
		\includegraphics[width=\linewidth]{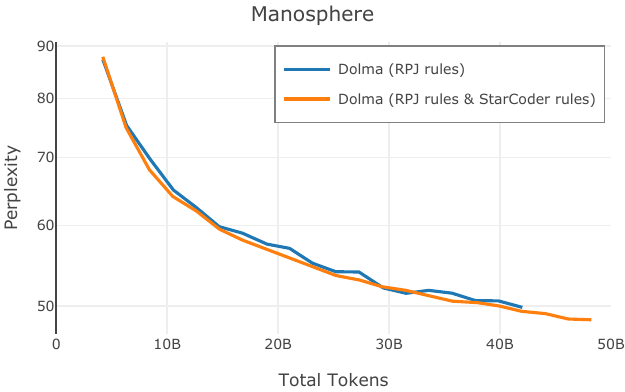}
	\end{subfigure}
	\caption{Perplexity results on Paloma~\citep{paloma}; subsets C4~\citep{raffel2020exploring,dodge-etal-2021-documenting} and Manosphere~\citep{ribeiroevolution2021}}
\end{figure}

\begin{figure}[h!]
	\centering
	\begin{subfigure}{0.31\textwidth}
		\includegraphics[width=\linewidth]{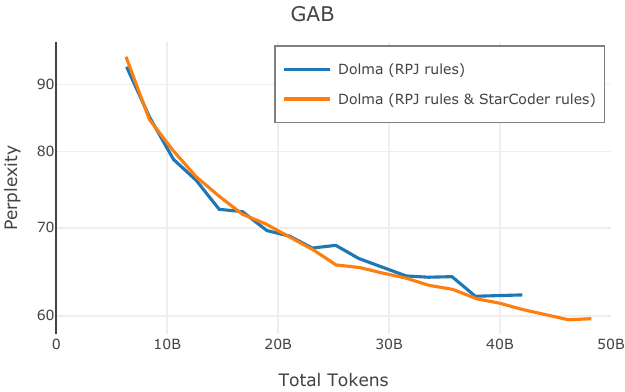}
	\end{subfigure}
	\quad
	\begin{subfigure}{0.31\textwidth}
		\includegraphics[width=\linewidth]{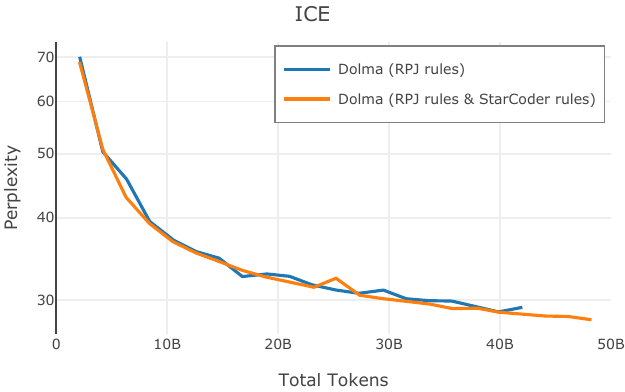}
	\end{subfigure}
	\quad
	\begin{subfigure}{0.31\textwidth}
		\includegraphics[width=\linewidth]{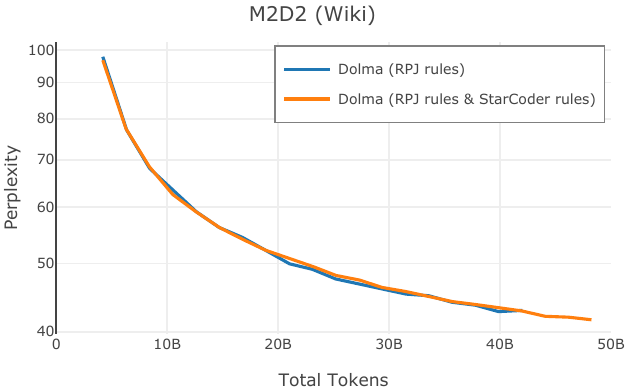}
	\end{subfigure}
	\caption{Perplexity results on Paloma~\citep{paloma}; subsets Gab~\citep{zannettou2018gab}, ICE~\citep{greenbaum1991ice}, and M2D2~\citep{reid-etal-2022-m2d2} (Wiki)}
\end{figure}

\begin{figure}[h!]
	\centering
	\begin{subfigure}{0.31\textwidth}
		\includegraphics[width=\linewidth]{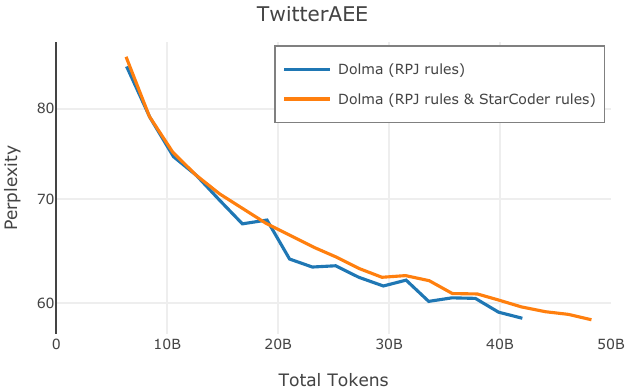}
	\end{subfigure}
	\quad
	\begin{subfigure}{0.31\textwidth}
		\includegraphics[width=\linewidth]{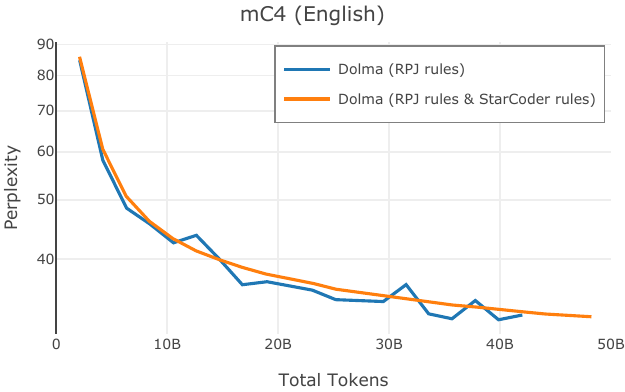}
	\end{subfigure}
	\quad
	\begin{subfigure}{0.31\textwidth}
		\includegraphics[width=\linewidth]{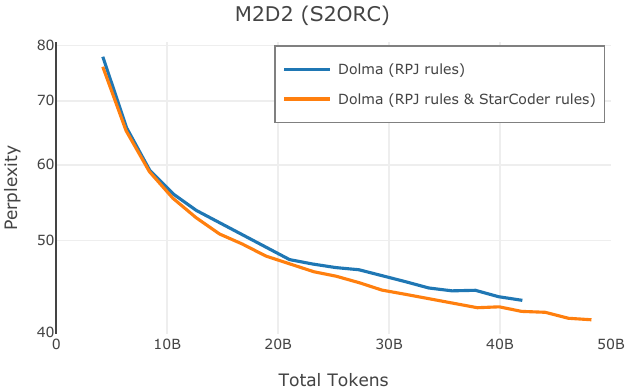}
	\end{subfigure}
	\caption{Perplexity results on Paloma~\citep{paloma}; subsets Twitter AAE~\citep{blodgett-etal-2016-demographic}, mC4~\citep{mc4} (English), and M2D2~\citep{reid-etal-2022-m2d2} (S2ORC)}
\end{figure}

\label{sec:ablations_code_stack_v2_vs_v4:downstream}

\begin{figure}[h!]
	\centering
	\begin{subfigure}{0.31\textwidth}
		\includegraphics[width=\linewidth]{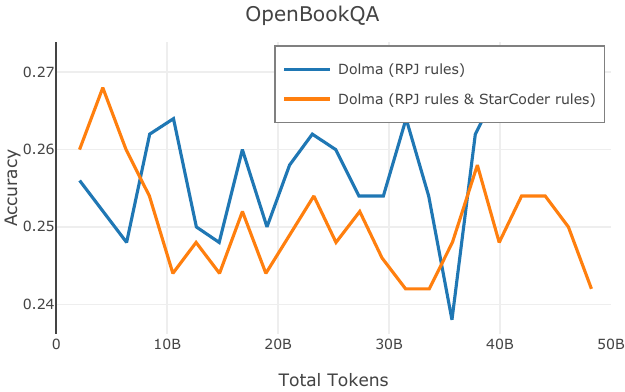}
	\end{subfigure}
	\quad
	\begin{subfigure}{0.31\textwidth}
		\includegraphics[width=\linewidth]{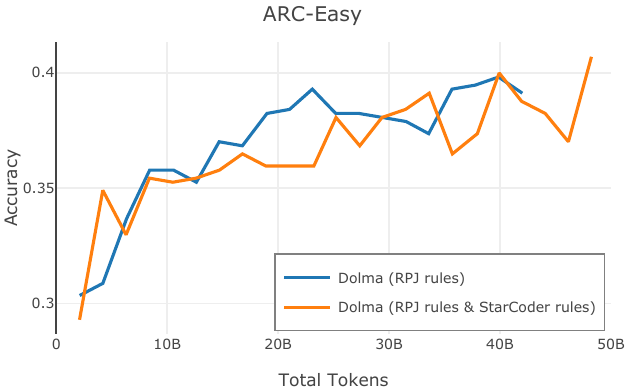}
	\end{subfigure}
	\quad
	\begin{subfigure}{0.31\textwidth}
		\includegraphics[width=\linewidth]{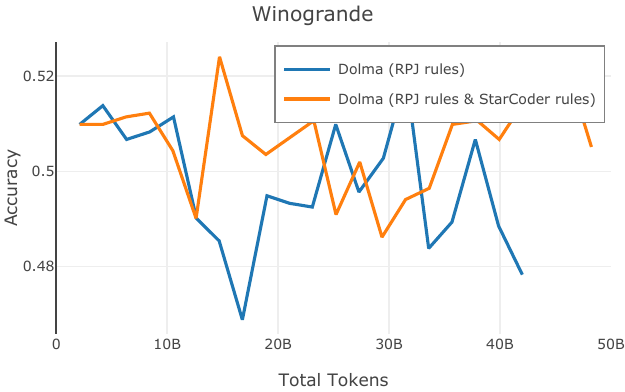}
	\end{subfigure}
	\caption{Results downstream tasks OpenBookQA~\citep{openbookQA}, ARC-E~\citep{arc}, and WinoGrande~\citep{winogrande}}
\end{figure}

\begin{figure}[h!]
	\centering
	\begin{subfigure}{0.31\textwidth}
		\includegraphics[width=\linewidth]{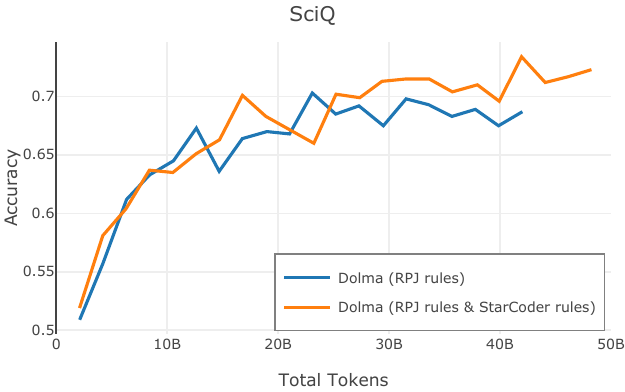}
	\end{subfigure}
	\quad
	\begin{subfigure}{0.31\textwidth}
		\includegraphics[width=\linewidth]{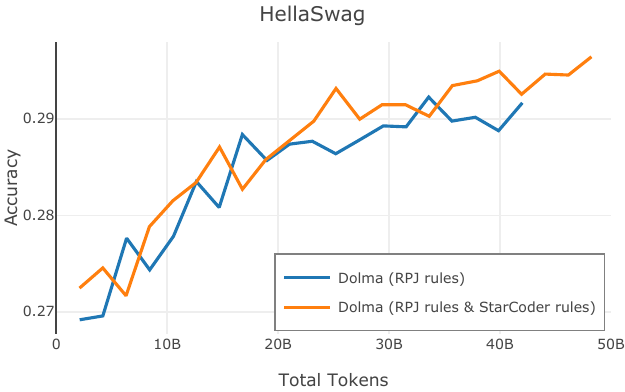}
	\end{subfigure}
	\quad
	\begin{subfigure}{0.31\textwidth}
		\includegraphics[width=\linewidth]{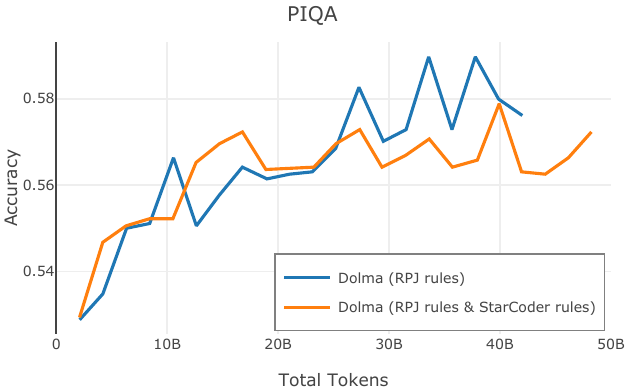}
	\end{subfigure}
	\caption{Results downstream tasks SciQ~\citep{sciq}, HellaSwag~\citep{zellers2019hellaswag}, and PIQA~\citep{piqa}}
\end{figure}

\label{sec:ablations_code_stack_v2_vs_v4:train}

\begin{figure}[h!]
	\centering
	\begin{subfigure}{0.31\textwidth}
		\includegraphics[width=\linewidth]{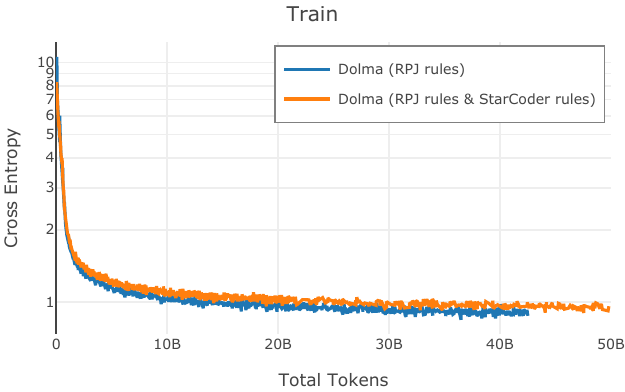}
	\end{subfigure}
	\caption{Training Cross Entropy}
\end{figure}

\FloatBarrier

\subsection{Studying \dolma Mixture}
\label{sec:ablations_dolma_mix}

\label{sec:ablations_dolma_mix:ppl}

\begin{figure}[h!]
	\centering
	\begin{subfigure}{0.31\textwidth}
		\includegraphics[width=\linewidth]{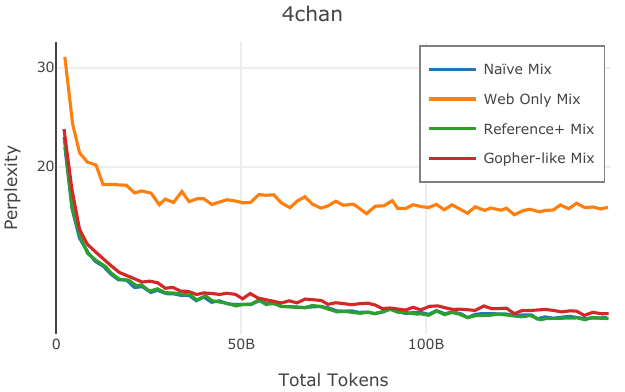}
	\end{subfigure}
	\quad
	\begin{subfigure}{0.31\textwidth}
		\includegraphics[width=\linewidth]{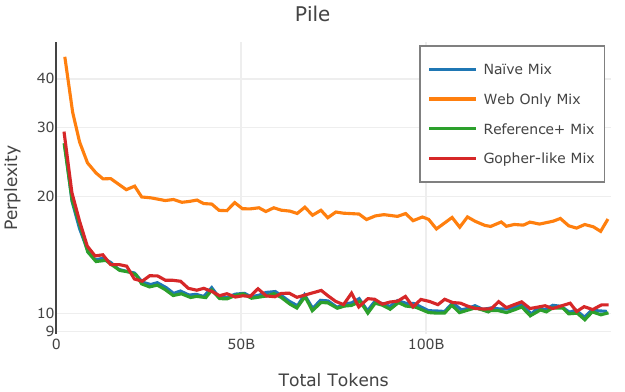}
	\end{subfigure}
	\quad
	\begin{subfigure}{0.31\textwidth}
		\includegraphics[width=\linewidth]{experiments/ablations_dolma_mix/ppl/c4_100_domains.pdf}
	\end{subfigure}
	\caption{Perplexity results on Paloma~\citep{paloma}; subsets 4chan~\citep{papasavva2020raiders}, Pile~\citep{Gao2020ThePA} (Val), and C4 100 dom~\citep{chronopoulou-etal-2022-efficient}}
\end{figure}

\begin{figure}[h!]
	\centering
	\begin{subfigure}{0.31\textwidth}
		\includegraphics[width=\linewidth]{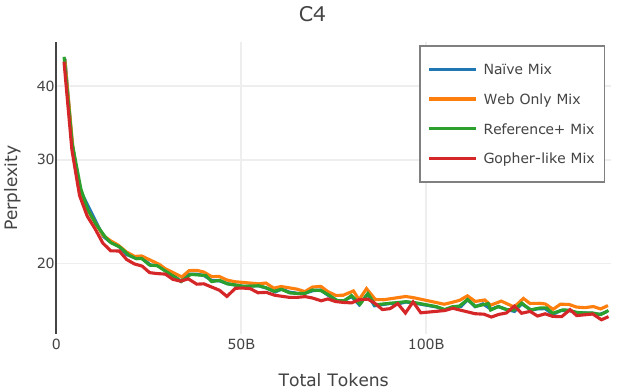}
	\end{subfigure}
	\quad
	\begin{subfigure}{0.31\textwidth}
		\includegraphics[width=\linewidth]{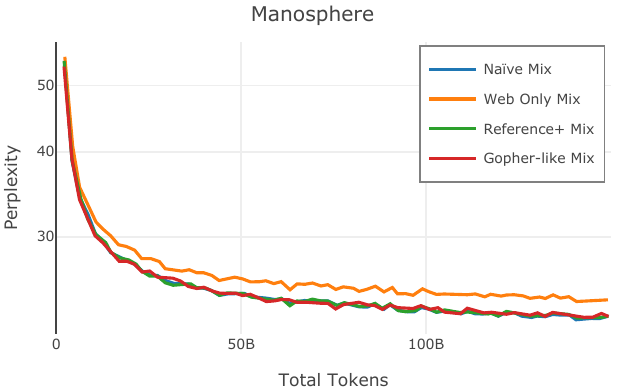}
	\end{subfigure}
	\caption{Perplexity results on Paloma~\citep{paloma}; subsets C4~\citep{raffel2020exploring,dodge-etal-2021-documenting} and Manosphere~\citep{ribeiroevolution2021}}
\end{figure}

\begin{figure}[h!]
	\centering
	\begin{subfigure}{0.31\textwidth}
		\includegraphics[width=\linewidth]{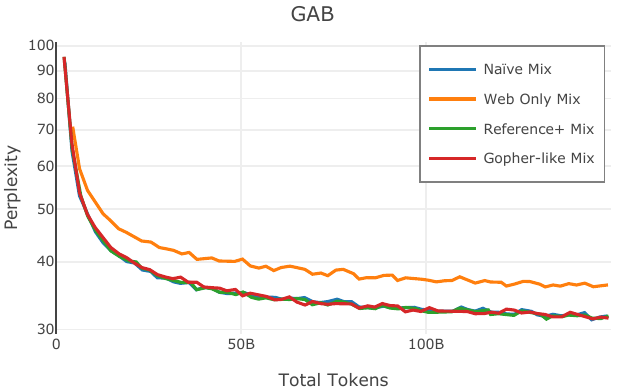}
	\end{subfigure}
	\quad
	\begin{subfigure}{0.31\textwidth}
		\includegraphics[width=\linewidth]{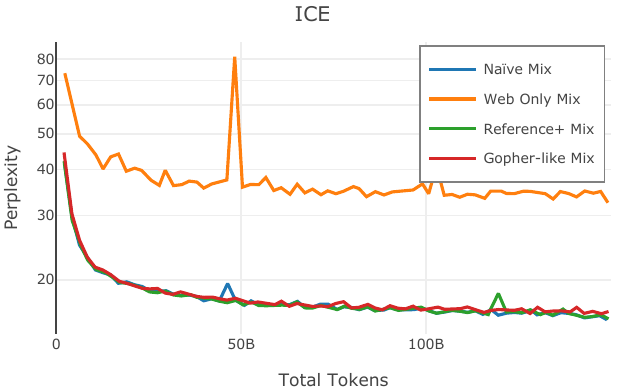}
	\end{subfigure}
	\quad
	\begin{subfigure}{0.31\textwidth}
		\includegraphics[width=\linewidth]{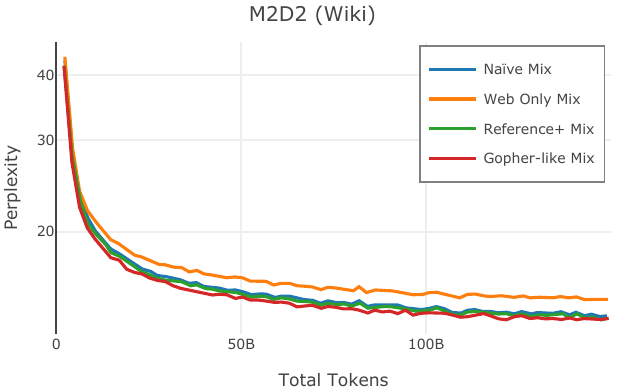}
	\end{subfigure}
	\caption{Perplexity results on Paloma~\citep{paloma}; subsets Gab~\citep{zannettou2018gab}, ICE~\citep{greenbaum1991ice}, and M2D2~\citep{reid-etal-2022-m2d2} (Wiki)}
\end{figure}

\begin{figure}[h!]
	\centering
	\begin{subfigure}{0.31\textwidth}
		\includegraphics[width=\linewidth]{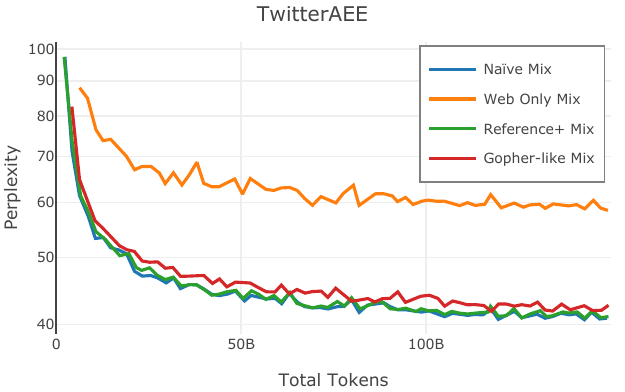}
	\end{subfigure}
	\quad
	\begin{subfigure}{0.31\textwidth}
		\includegraphics[width=\linewidth]{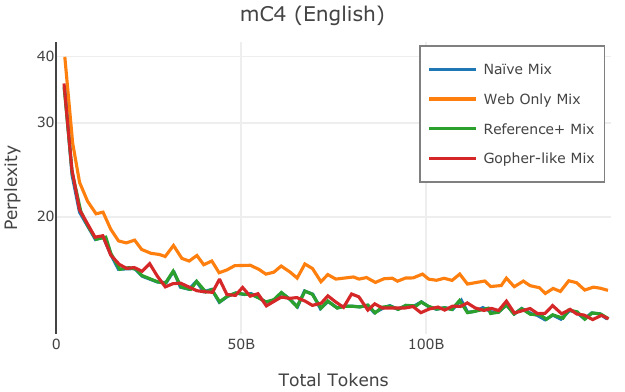}
	\end{subfigure}
	\quad
	\begin{subfigure}{0.31\textwidth}
		\includegraphics[width=\linewidth]{experiments/ablations_dolma_mix/ppl/m2d2_s2orc.pdf}
	\end{subfigure}
	\caption{Perplexity results on Paloma~\citep{paloma}; subsets Twitter AAE~\citep{blodgett-etal-2016-demographic}, mC4~\citep{mc4} (English), and M2D2~\citep{reid-etal-2022-m2d2} (S2ORC)}
\end{figure}

\label{sec:ablations_dolma_mix:downstream}

\begin{figure}[h!]
	\centering
	\begin{subfigure}{0.31\textwidth}
		\includegraphics[width=\linewidth]{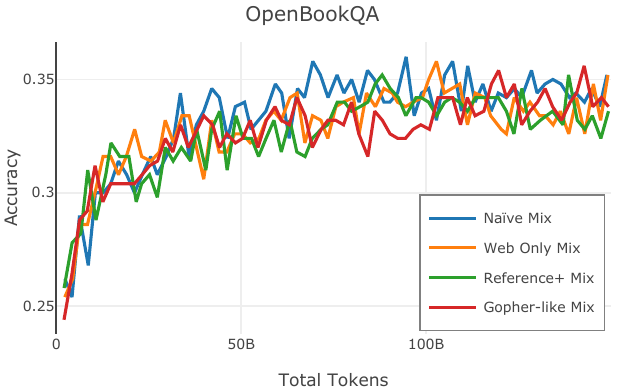}
	\end{subfigure}
	\quad
	\begin{subfigure}{0.31\textwidth}
		\includegraphics[width=\linewidth]{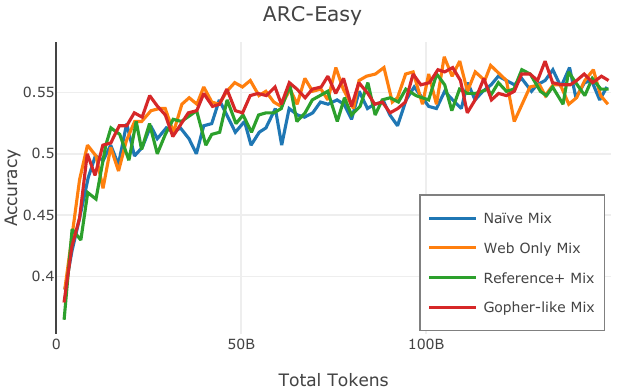}
	\end{subfigure}
	\quad
	\begin{subfigure}{0.31\textwidth}
		\includegraphics[width=\linewidth]{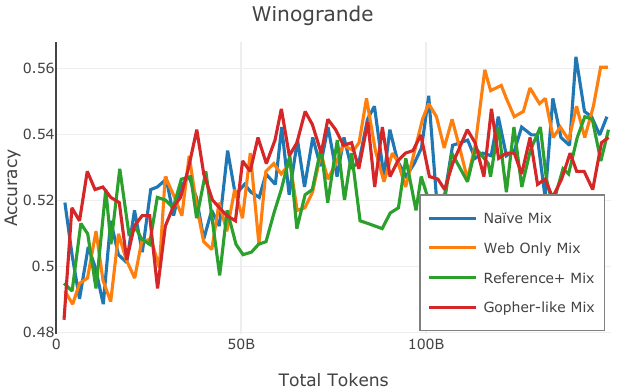}
	\end{subfigure}
	\caption{Results downstream tasks OpenBookQA~\citep{openbookQA}, ARC-E~\citep{arc}, and WinoGrande~\citep{winogrande}}
\end{figure}

\begin{figure}[h!]
	\centering
	\begin{subfigure}{0.31\textwidth}
		\includegraphics[width=\linewidth]{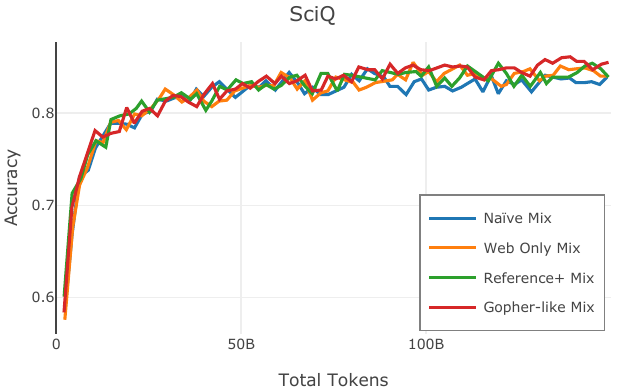}
	\end{subfigure}
	\quad
	\begin{subfigure}{0.31\textwidth}
		\includegraphics[width=\linewidth]{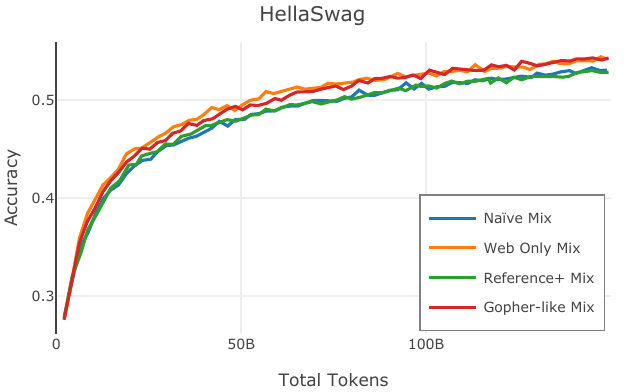}
	\end{subfigure}
	\quad
	\begin{subfigure}{0.31\textwidth}
		\includegraphics[width=\linewidth]{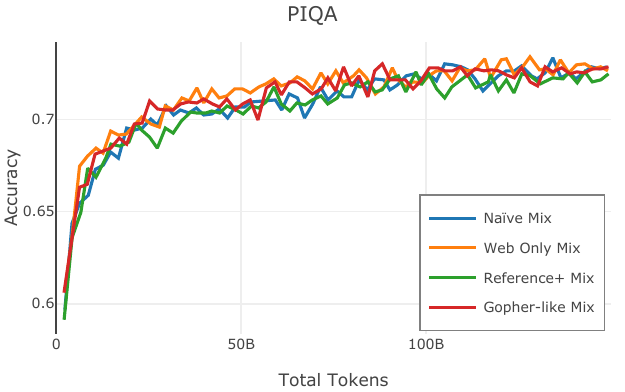}
	\end{subfigure}
	\caption{Results downstream tasks SciQ~\citep{sciq}, HellaSwag~\citep{zellers2019hellaswag}, and PIQA~\citep{piqa}}
\end{figure}

\clearpage

\subsection{Strategies to Format Conversational Forums Pipeline}
\label{sec:ablations_reddit_selection}

\label{sec:ablations_reddit_selection:ppl}

\begin{figure}[h!]
	\centering
	\begin{subfigure}{0.31\textwidth}
		\includegraphics[width=\linewidth]{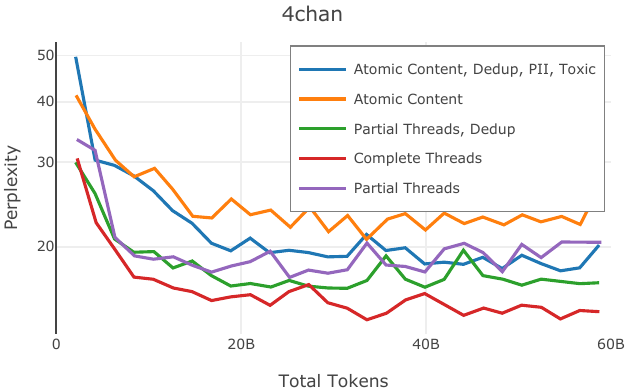}
	\end{subfigure}
	\quad
	\begin{subfigure}{0.31\textwidth}
		\includegraphics[width=\linewidth]{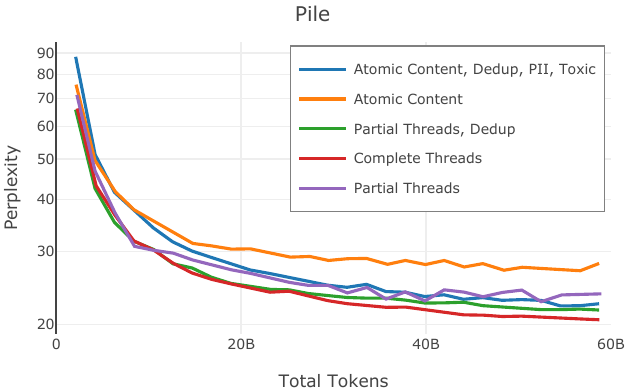}
	\end{subfigure}
	\quad
	\begin{subfigure}{0.31\textwidth}
		\includegraphics[width=\linewidth]{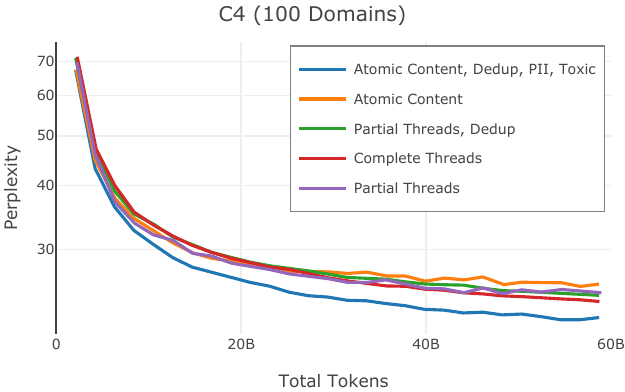}
	\end{subfigure}
	\caption{Perplexity results on Paloma~\citep{paloma}; subsets 4chan~\citep{papasavva2020raiders}, Pile~\citep{Gao2020ThePA} (Val), and C4 100 dom~\citep{chronopoulou-etal-2022-efficient}}
\end{figure}

\begin{figure}[h!]
	\centering
	\begin{subfigure}{0.31\textwidth}
		\includegraphics[width=\linewidth]{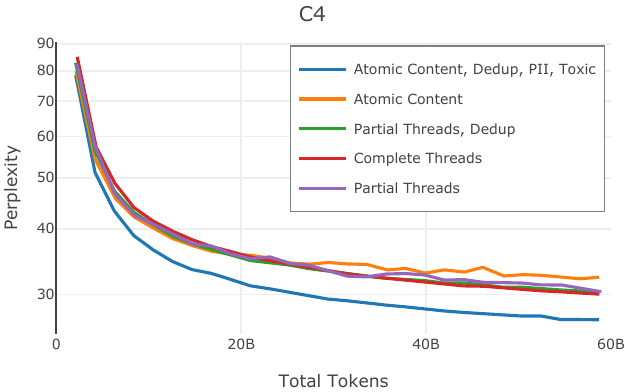}
	\end{subfigure}
	\quad
	\begin{subfigure}{0.31\textwidth}
		\includegraphics[width=\linewidth]{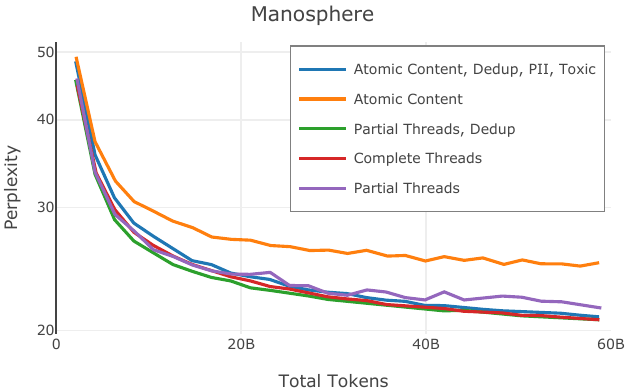}
	\end{subfigure}
	\caption{Perplexity results on Paloma~\citep{paloma}; subsets C4~\citep{raffel2020exploring,dodge-etal-2021-documenting} and Manosphere~\citep{ribeiroevolution2021}}
\end{figure}

\begin{figure}[h!]
	\centering
	\begin{subfigure}{0.31\textwidth}
		\includegraphics[width=\linewidth]{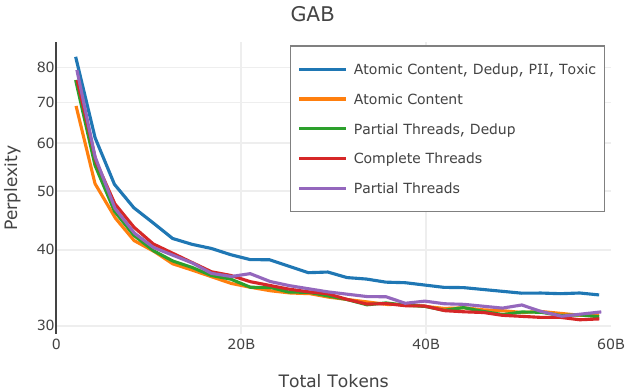}
	\end{subfigure}
	\quad
	\begin{subfigure}{0.31\textwidth}
		\includegraphics[width=\linewidth]{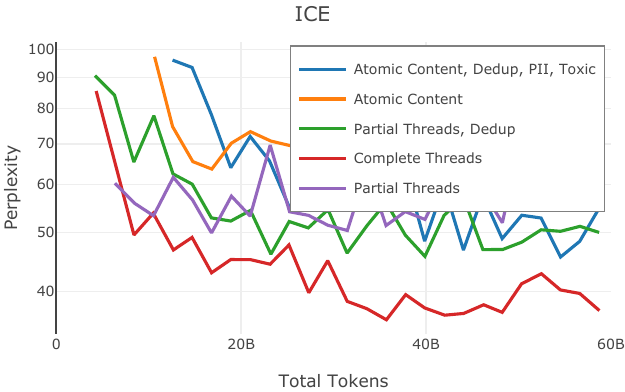}
	\end{subfigure}
	\quad
	\begin{subfigure}{0.31\textwidth}
		\includegraphics[width=\linewidth]{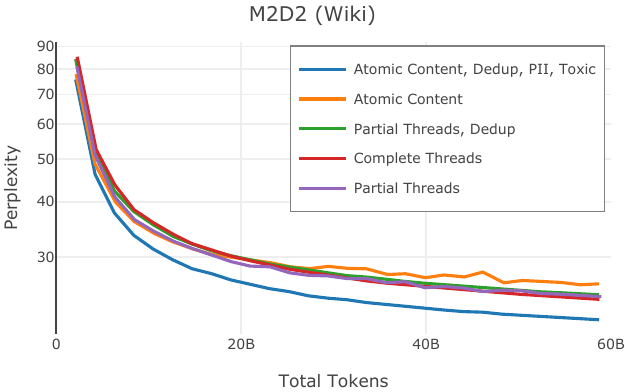}
	\end{subfigure}
	\caption{Perplexity results on Paloma~\citep{paloma}; subsets Gab~\citep{zannettou2018gab}, ICE~\citep{greenbaum1991ice}, and M2D2~\citep{reid-etal-2022-m2d2} (Wiki)}
\end{figure}

\begin{figure}[h!]
	\centering
	\begin{subfigure}{0.31\textwidth}
		\includegraphics[width=\linewidth]{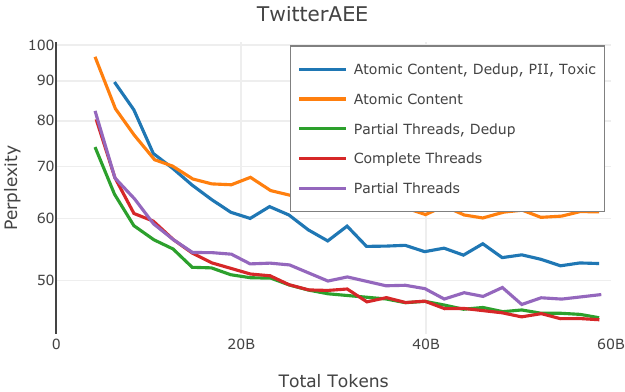}
	\end{subfigure}
	\quad
	\begin{subfigure}{0.31\textwidth}
		\includegraphics[width=\linewidth]{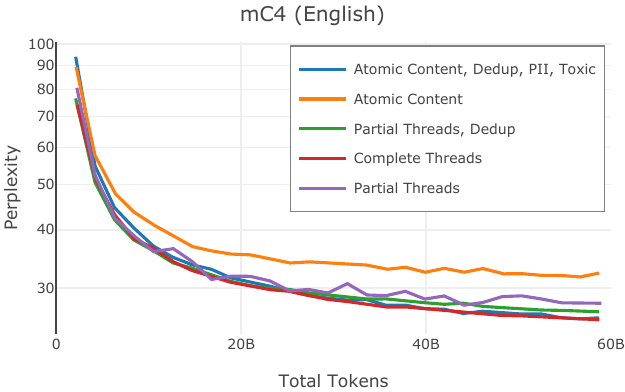}
	\end{subfigure}
	\quad
	\begin{subfigure}{0.31\textwidth}
		\includegraphics[width=\linewidth]{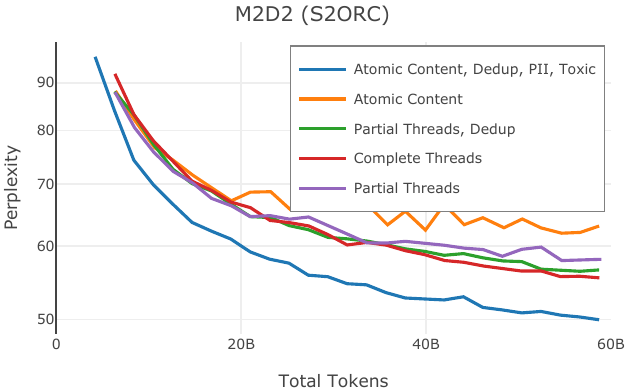}
	\end{subfigure}
	\caption{Perplexity results on Paloma~\citep{paloma}; subsets Twitter AAE~\citep{blodgett-etal-2016-demographic}, mC4~\citep{mc4} (English), and M2D2~\citep{reid-etal-2022-m2d2} (S2ORC)}
\end{figure}

\label{sec:ablations_reddit_selection:downstream}

\begin{figure}[h!]
	\centering
	\begin{subfigure}{0.31\textwidth}
		\includegraphics[width=\linewidth]{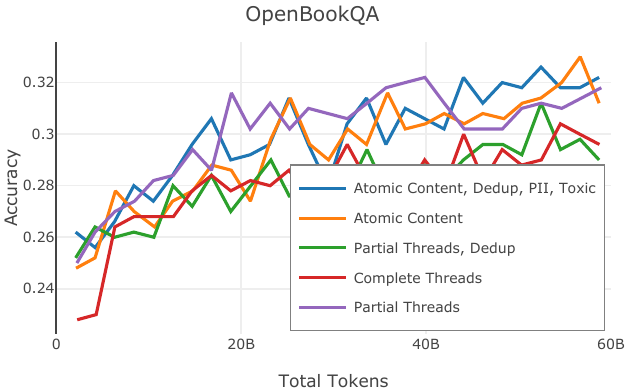}
	\end{subfigure}
	\quad
	\begin{subfigure}{0.31\textwidth}
		\includegraphics[width=\linewidth]{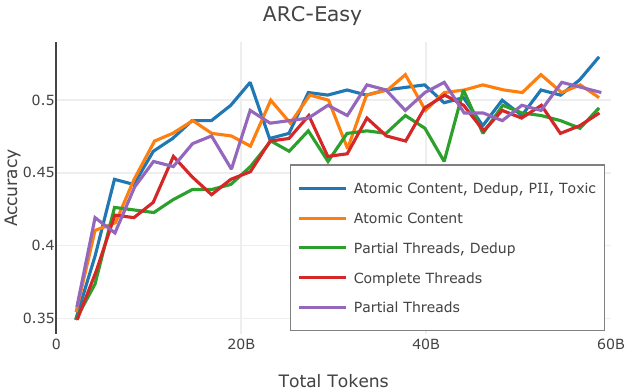}
	\end{subfigure}
	\quad
	\begin{subfigure}{0.31\textwidth}
		\includegraphics[width=\linewidth]{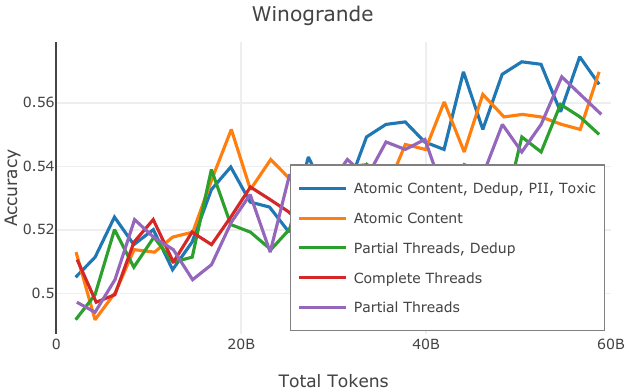}
	\end{subfigure}
	\caption{Results downstream tasks OpenBookQA~\citep{openbookQA}, ARC-E~\citep{arc}, and WinoGrande~\citep{winogrande}}
\end{figure}

\begin{figure}[h!]
	\centering
	\begin{subfigure}{0.31\textwidth}
		\includegraphics[width=\linewidth]{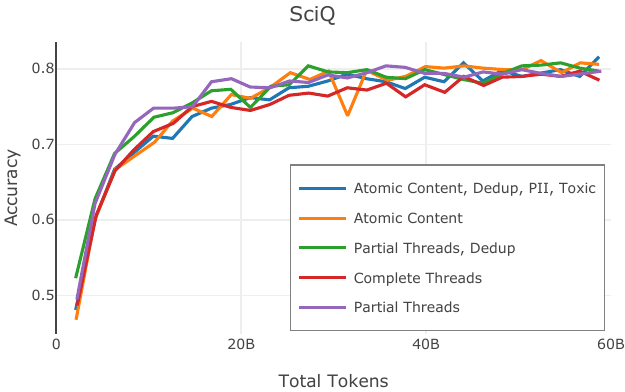}
	\end{subfigure}
	\quad
	\begin{subfigure}{0.31\textwidth}
		\includegraphics[width=\linewidth]{experiments/ablations_reddit_selection/downstream/hellaswag.pdf}
	\end{subfigure}
	\quad
	\begin{subfigure}{0.31\textwidth}
		\includegraphics[width=\linewidth]{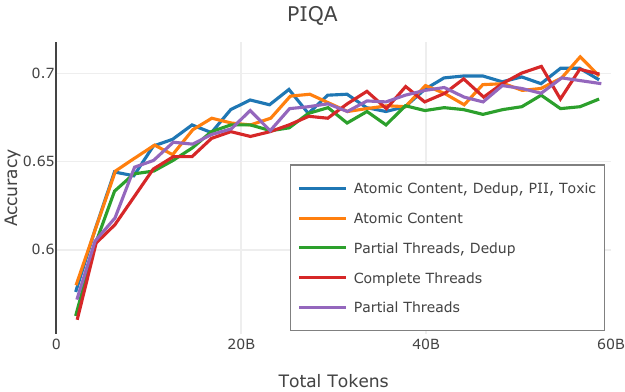}
	\end{subfigure}
	\caption{Results downstream tasks SciQ~\citep{sciq}, HellaSwag~\citep{zellers2019hellaswag}, and PIQA~\citep{piqa}}
\end{figure}

\label{sec:ablations_reddit_selection:train}

\begin{figure}[h!]
	\centering
	\begin{subfigure}{0.31\textwidth}
		\includegraphics[width=\linewidth]{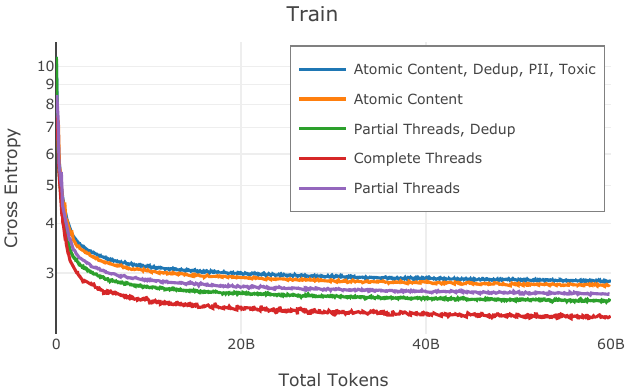}
	\end{subfigure}
	\caption{Training Cross Entropy}
\end{figure}

\FloatBarrier

\subsection{Evaluating Toxicity Filtering in Conversational Forums Pipeline}
\label{sec:ablations_reddit_toxic_filtering}

\label{sec:ablations_reddit_toxic_filtering:ppl}

\begin{figure}[h!]
	\centering
	\begin{subfigure}{0.31\textwidth}
		\includegraphics[width=\linewidth]{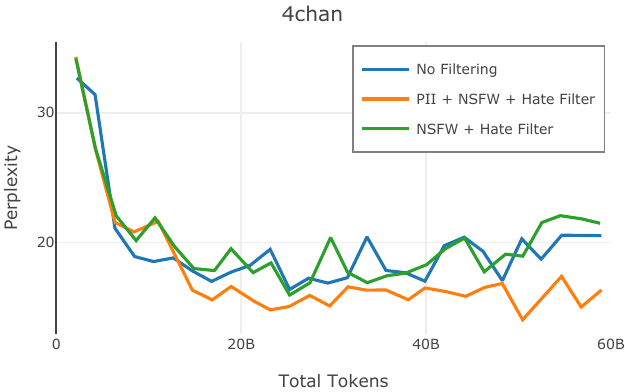}
	\end{subfigure}
	\quad
	\begin{subfigure}{0.31\textwidth}
		\includegraphics[width=\linewidth]{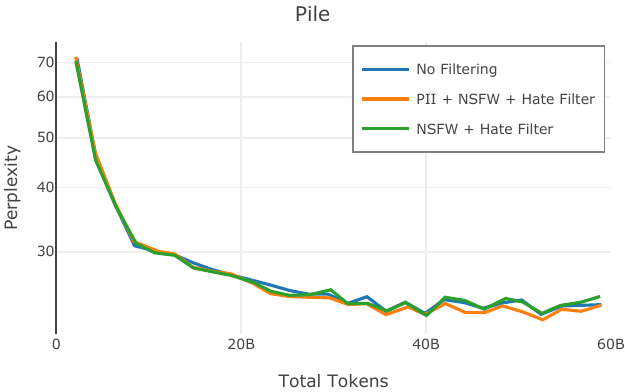}
	\end{subfigure}
	\quad
	\begin{subfigure}{0.31\textwidth}
		\includegraphics[width=\linewidth]{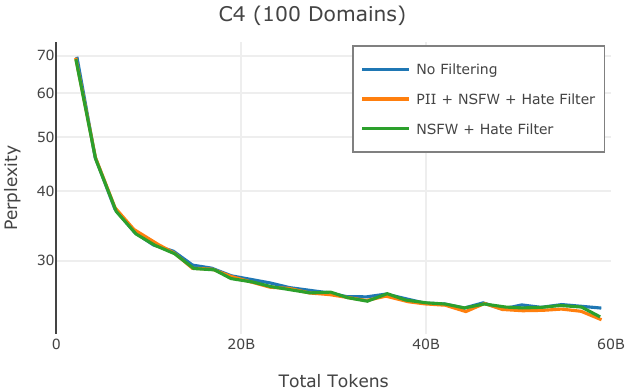}
	\end{subfigure}
	\caption{Perplexity results on Paloma~\citep{paloma}; subsets 4chan~\citep{papasavva2020raiders}, Pile~\citep{Gao2020ThePA} (Val), and C4 100 dom~\citep{chronopoulou-etal-2022-efficient}}
\end{figure}

\begin{figure}[h!]
	\centering
	\begin{subfigure}{0.31\textwidth}
		\includegraphics[width=\linewidth]{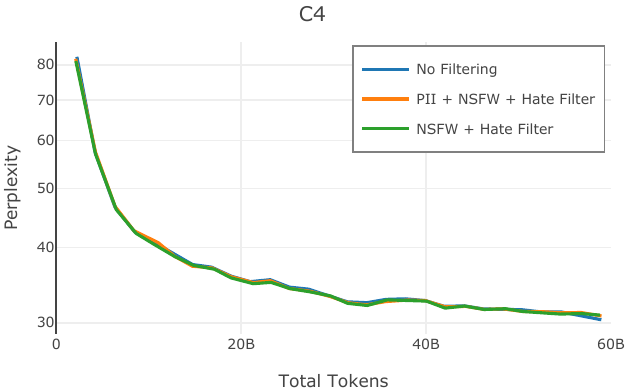}
	\end{subfigure}
	\quad
	\begin{subfigure}{0.31\textwidth}
		\includegraphics[width=\linewidth]{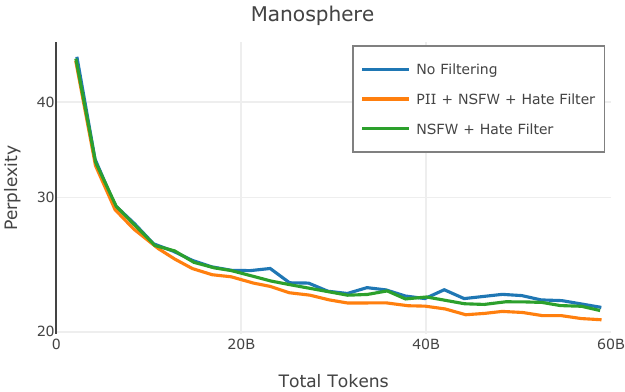}
	\end{subfigure}
	\caption{Perplexity results on Paloma~\citep{paloma}; subsets C4~\citep{raffel2020exploring,dodge-etal-2021-documenting} and Manosphere~\citep{ribeiroevolution2021}}
\end{figure}

\begin{figure}[h!]
	\centering
	\begin{subfigure}{0.31\textwidth}
		\includegraphics[width=\linewidth]{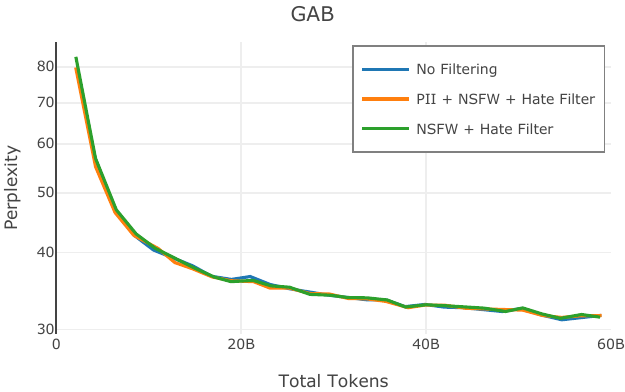}
	\end{subfigure}
	\quad
	\begin{subfigure}{0.31\textwidth}
		\includegraphics[width=\linewidth]{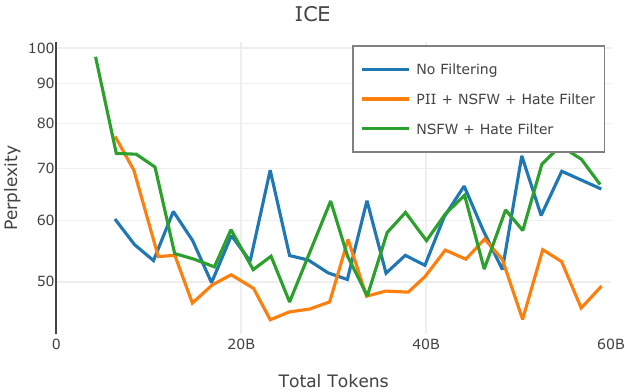}
	\end{subfigure}
	\quad
	\begin{subfigure}{0.31\textwidth}
		\includegraphics[width=\linewidth]{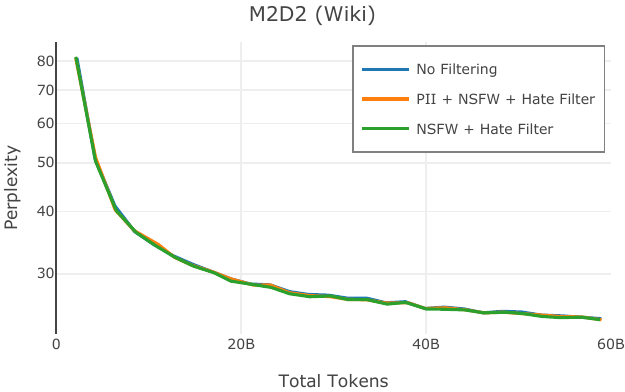}
	\end{subfigure}
	\caption{Perplexity results on Paloma~\citep{paloma}; subsets Gab~\citep{zannettou2018gab}, ICE~\citep{greenbaum1991ice}, and M2D2~\citep{reid-etal-2022-m2d2} (Wiki)}
\end{figure}

\begin{figure}[h!]
	\centering
	\begin{subfigure}{0.31\textwidth}
		\includegraphics[width=\linewidth]{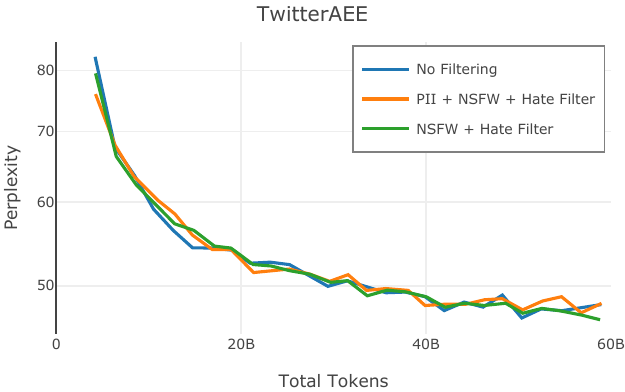}
	\end{subfigure}
	\quad
	\begin{subfigure}{0.31\textwidth}
		\includegraphics[width=\linewidth]{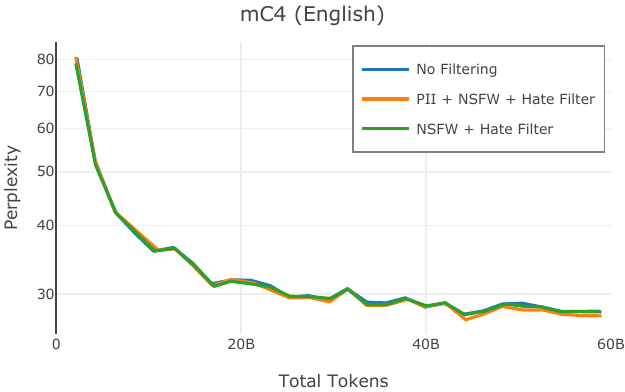}
	\end{subfigure}
	\quad
	\begin{subfigure}{0.31\textwidth}
		\includegraphics[width=\linewidth]{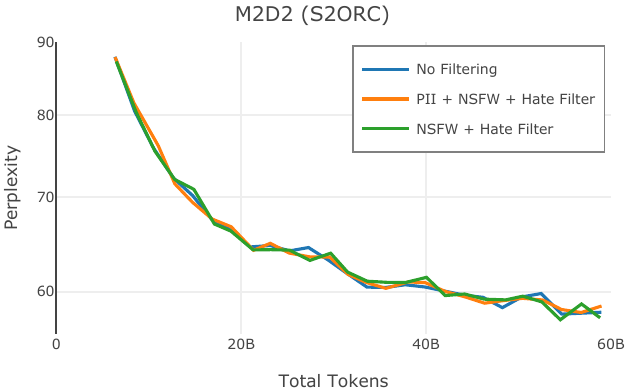}
	\end{subfigure}
	\caption{Perplexity results on Paloma~\citep{paloma}; subsets Twitter AAE~\citep{blodgett-etal-2016-demographic}, mC4~\citep{mc4} (English), and M2D2~\citep{reid-etal-2022-m2d2} (S2ORC)}
\end{figure}

\label{sec:ablations_reddit_toxic_filtering:downstream}

\begin{figure}[h!]
	\centering
	\begin{subfigure}{0.31\textwidth}
		\includegraphics[width=\linewidth]{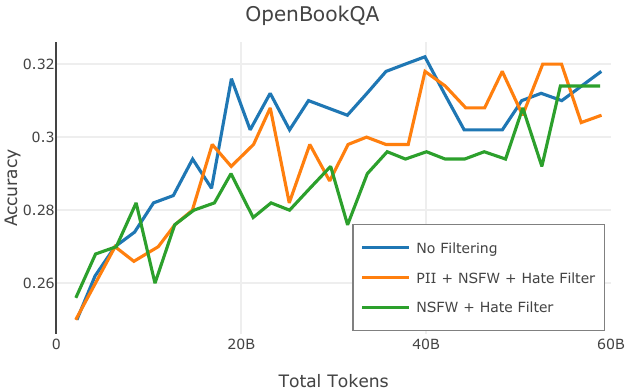}
	\end{subfigure}
	\quad
	\begin{subfigure}{0.31\textwidth}
		\includegraphics[width=\linewidth]{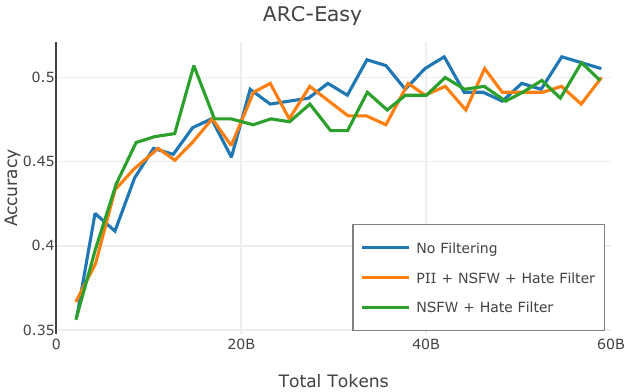}
	\end{subfigure}
	\quad
	\begin{subfigure}{0.31\textwidth}
		\includegraphics[width=\linewidth]{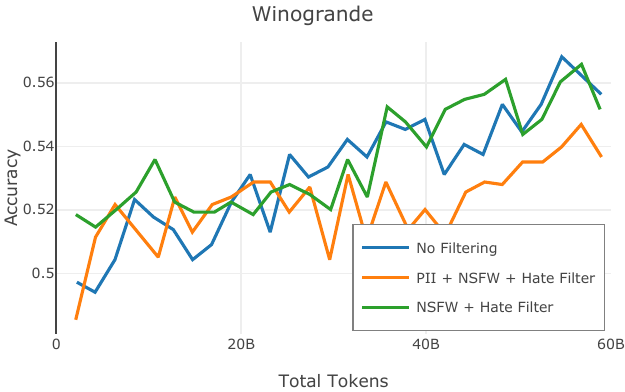}
	\end{subfigure}
	\caption{Results downstream tasks OpenBookQA~\citep{openbookQA}, ARC-E~\citep{arc}, and WinoGrande~\citep{winogrande}}
\end{figure}

\begin{figure}[h!]
	\centering
	\begin{subfigure}{0.31\textwidth}
		\includegraphics[width=\linewidth]{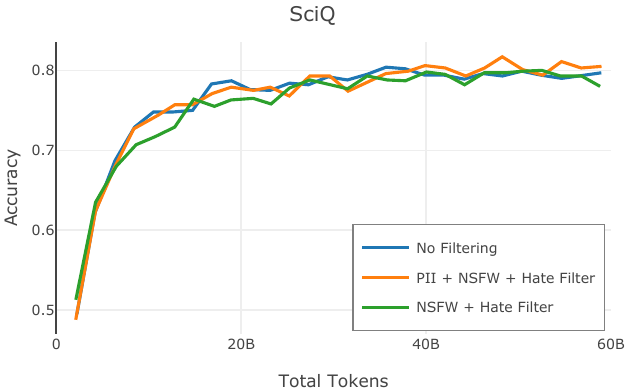}
	\end{subfigure}
	\quad
	\begin{subfigure}{0.31\textwidth}
		\includegraphics[width=\linewidth]{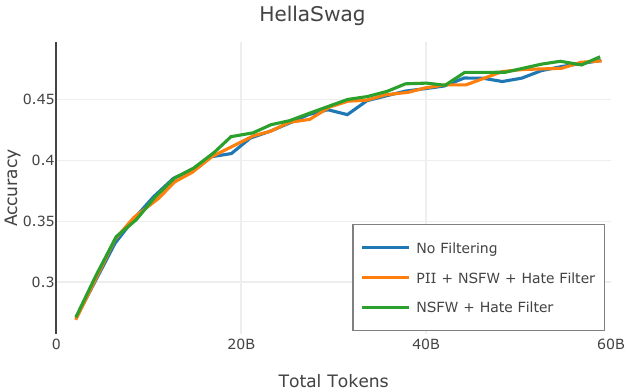}
	\end{subfigure}
	\quad
	\begin{subfigure}{0.31\textwidth}
		\includegraphics[width=\linewidth]{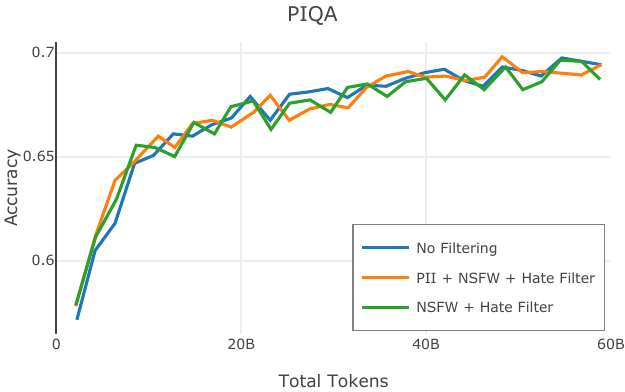}
	\end{subfigure}
	\caption{Results downstream tasks SciQ~\citep{sciq}, HellaSwag~\citep{zellers2019hellaswag}, and PIQA~\citep{piqa}}
\end{figure}

\label{sec:ablations_reddit_toxic_filtering:train}

\begin{figure}[h!]
	\centering
	\begin{subfigure}{0.31\textwidth}
		\includegraphics[width=\linewidth]{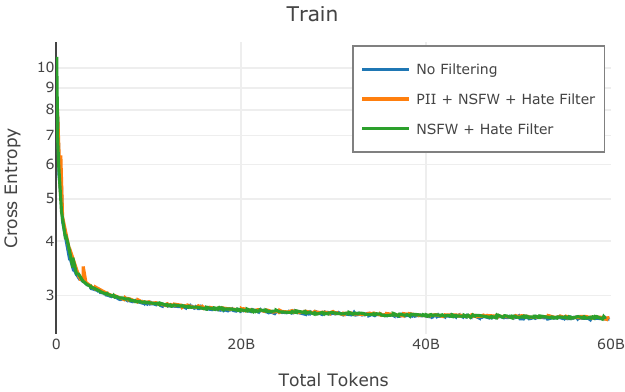}
	\end{subfigure}
	\caption{Training Cross Entropy}
\end{figure}

\FloatBarrier

\subsection{Training \OlmoTiny}
\label{sec:long_1b_run}

\label{sec:long_1b_run:ppl}

\begin{figure}[h!]
	\centering
	\begin{subfigure}{0.31\textwidth}
		\includegraphics[width=\linewidth]{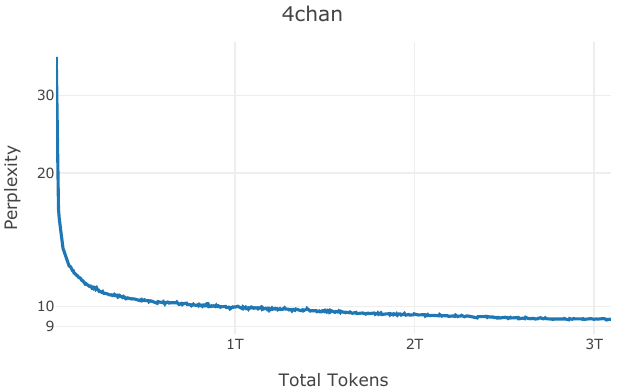}
	\end{subfigure}
	\quad
	\begin{subfigure}{0.31\textwidth}
		\includegraphics[width=\linewidth]{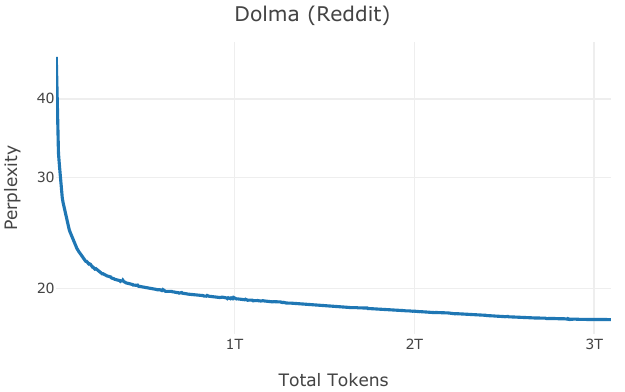}
	\end{subfigure}
	\quad
	\begin{subfigure}{0.31\textwidth}
		\includegraphics[width=\linewidth]{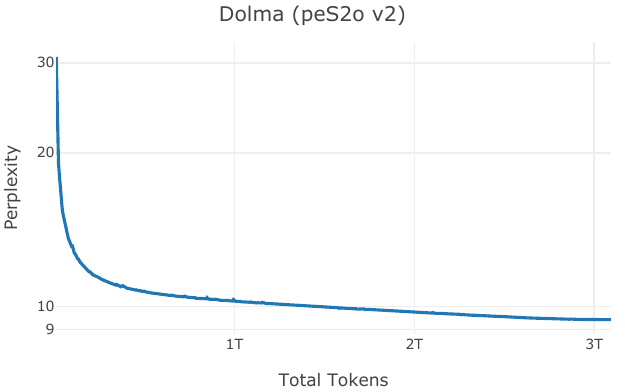}
	\end{subfigure}
	\caption{Perplexity results on Paloma~\citep{paloma}; subsets 4chan~\citep{papasavva2020raiders}, Dolma Reddit Subset, and Dolma Papers Subset}
\end{figure}

\begin{figure}[h!]
	\centering
	\begin{subfigure}{0.31\textwidth}
		\includegraphics[width=\linewidth]{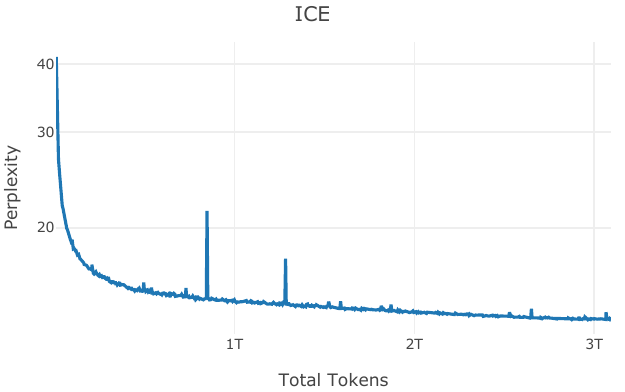}
	\end{subfigure}
	\quad
	\begin{subfigure}{0.31\textwidth}
		\includegraphics[width=\linewidth]{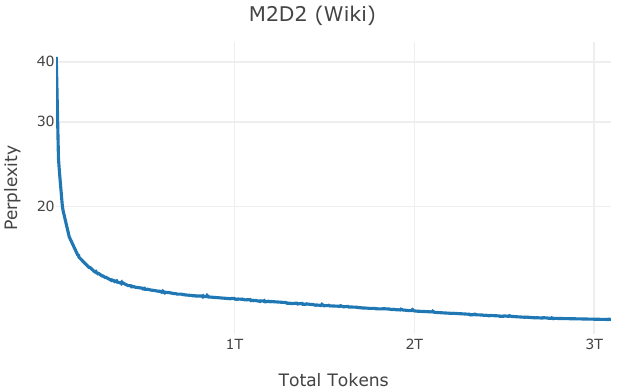}
	\end{subfigure}
	\quad
	\begin{subfigure}{0.31\textwidth}
		\includegraphics[width=\linewidth]{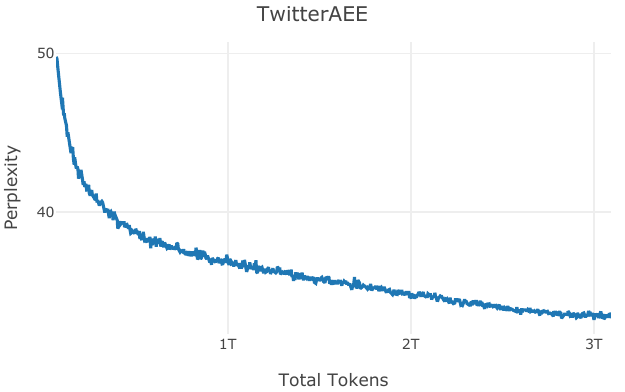}
	\end{subfigure}
	\caption{Perplexity results on Paloma~\citep{paloma}; subsets ICE~\citep{greenbaum1991ice}, M2D2~\citep{reid-etal-2022-m2d2} (Wiki), and Twitter AAE~\citep{blodgett-etal-2016-demographic}}
\end{figure}

\begin{figure}[h!]
	\centering
	\begin{subfigure}{0.31\textwidth}
		\includegraphics[width=\linewidth]{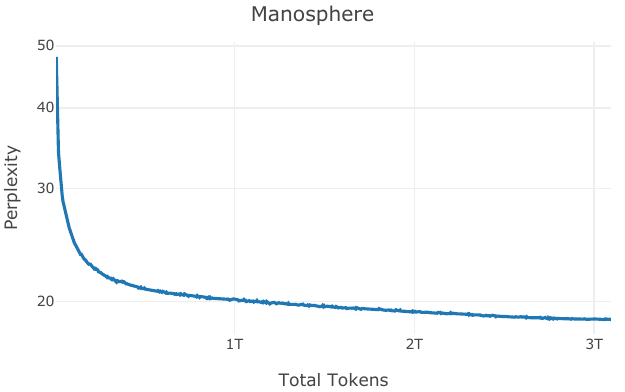}
	\end{subfigure}
	\caption{Perplexity results on Paloma~\citep{paloma}; subsets Manosphere~\citep{ribeiroevolution2021}}
\end{figure}

\begin{figure}[h!]
	\centering
	\begin{subfigure}{0.31\textwidth}
		\includegraphics[width=\linewidth]{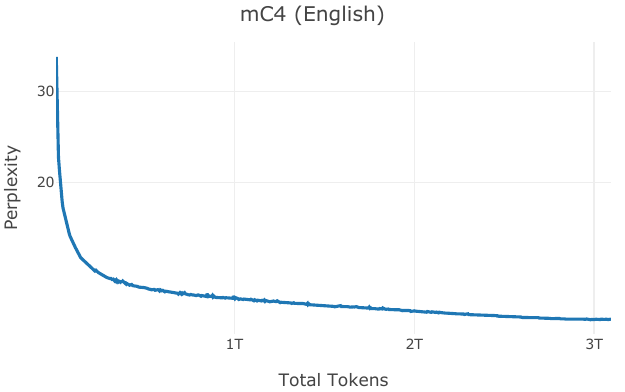}
	\end{subfigure}
	\quad
	\begin{subfigure}{0.31\textwidth}
		\includegraphics[width=\linewidth]{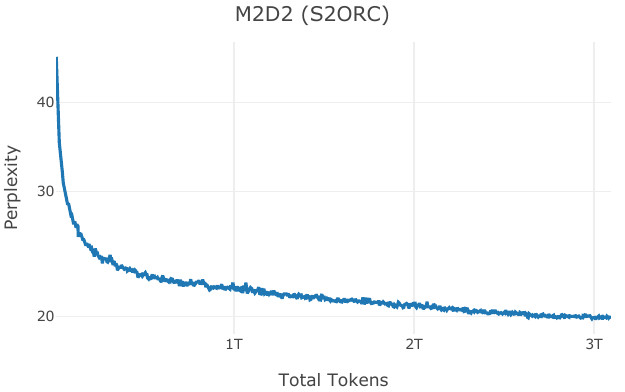}
	\end{subfigure}
	\quad
	\begin{subfigure}{0.31\textwidth}
		\includegraphics[width=\linewidth]{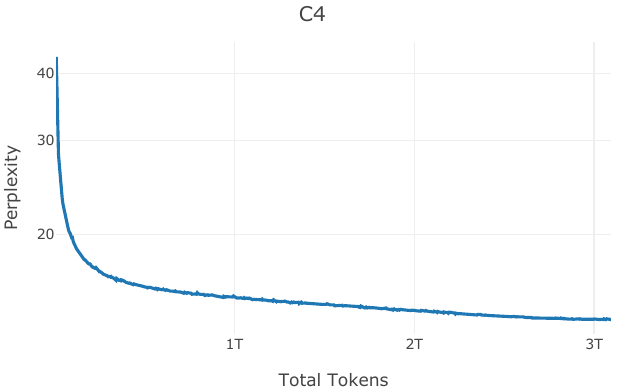}
	\end{subfigure}
	\caption{Perplexity results on Paloma~\citep{paloma}; subsets mC4~\citep{mc4} (English), M2D2~\citep{reid-etal-2022-m2d2} (S2ORC), and C4~\citep{raffel2020exploring,dodge-etal-2021-documenting}}
\end{figure}

\begin{figure}[h!]
	\centering
	\begin{subfigure}{0.31\textwidth}
		\includegraphics[width=\linewidth]{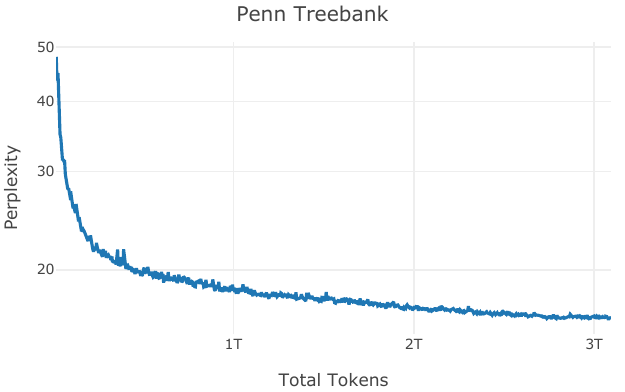}
	\end{subfigure}
	\quad
	\begin{subfigure}{0.31\textwidth}
		\includegraphics[width=\linewidth]{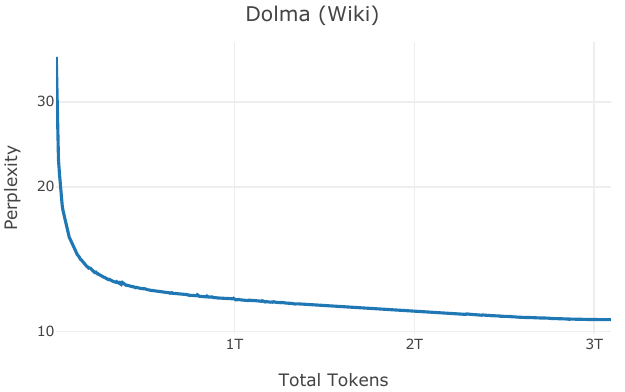}
	\end{subfigure}
	\quad
	\begin{subfigure}{0.31\textwidth}
		\includegraphics[width=\linewidth]{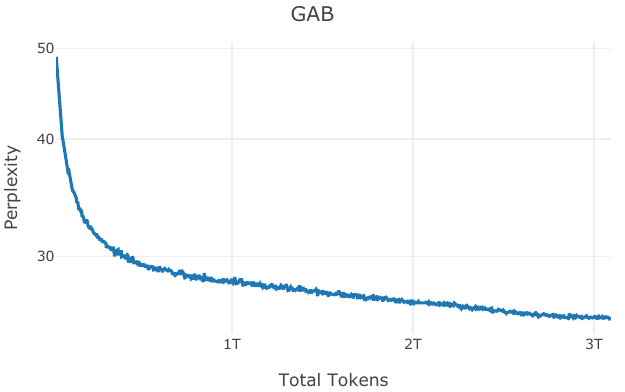}
	\end{subfigure}
	\caption{Perplexity results on Paloma~\citep{paloma}; subsets Penn Tree Bank~\citep{marcus-etal-1994-penn}, Dolma Wikipedia Subset, and Gab~\citep{zannettou2018gab}}
\end{figure}

\begin{figure}[h!]
	\centering
	\begin{subfigure}{0.31\textwidth}
		\includegraphics[width=\linewidth]{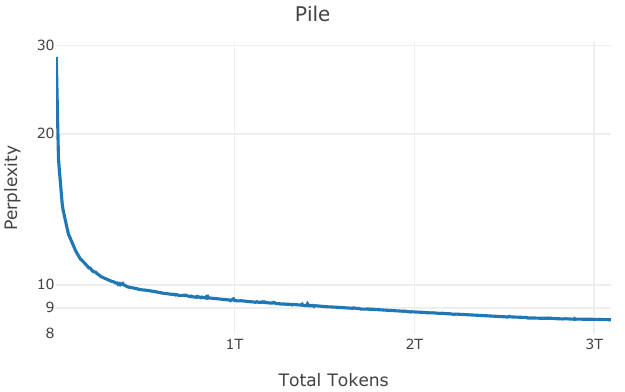}
	\end{subfigure}
	\quad
	\begin{subfigure}{0.31\textwidth}
		\includegraphics[width=\linewidth]{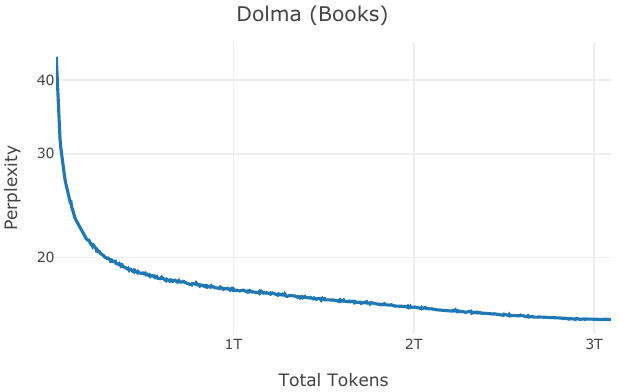}
	\end{subfigure}
	\quad
	\begin{subfigure}{0.31\textwidth}
		\includegraphics[width=\linewidth]{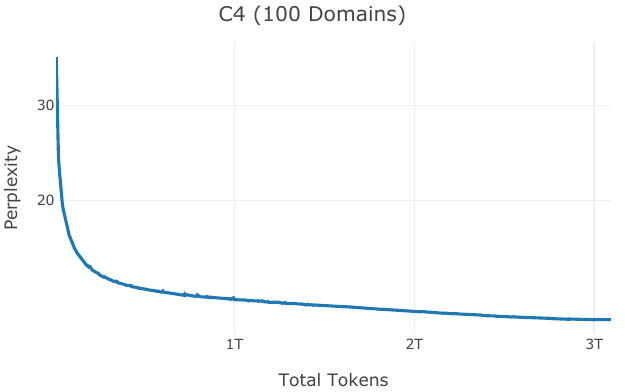}
	\end{subfigure}
	\caption{Perplexity results on Paloma~\citep{paloma}; subsets Pile~\citep{Gao2020ThePA} (Val), Dolma Books Subset, and C4 100 dom~\citep{chronopoulou-etal-2022-efficient}}
\end{figure}

\begin{figure}[h!]
	\centering
	\begin{subfigure}{0.31\textwidth}
		\includegraphics[width=\linewidth]{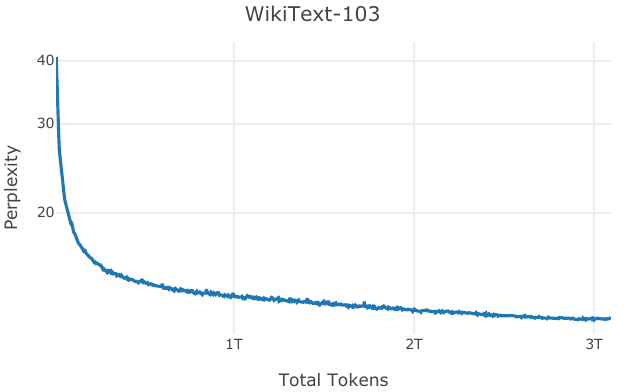}
	\end{subfigure}
	\quad
	\begin{subfigure}{0.31\textwidth}
		\includegraphics[width=\linewidth]{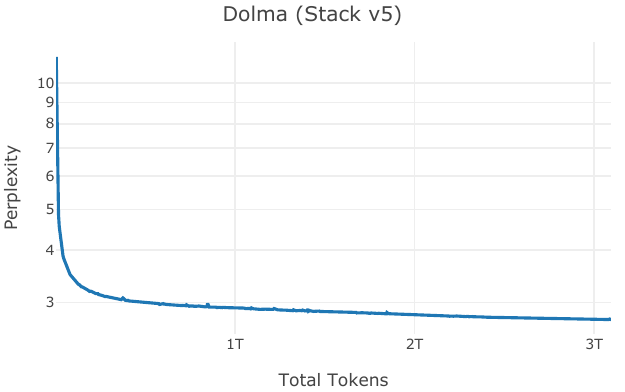}
	\end{subfigure}
	\quad
	\begin{subfigure}{0.31\textwidth}
		\includegraphics[width=\linewidth]{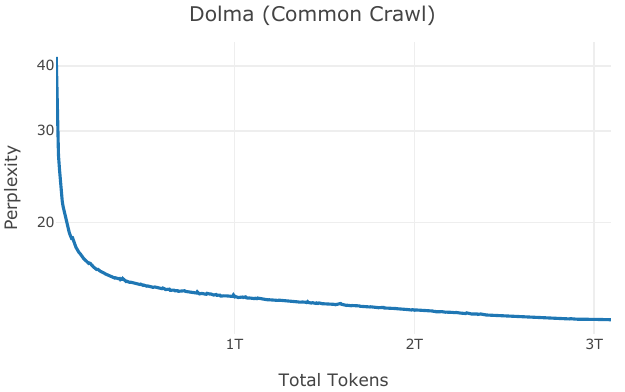}
	\end{subfigure}
	\caption{Perplexity results on Paloma~\citep{paloma}; subsets WikiText 103~\citep{merity2016pointer}, Dolma Code Subset, and Dolma Web Subset}
\end{figure}

\label{sec:long_1b_run:downstream}

\begin{figure}[h!]
	\centering
	\begin{subfigure}{0.31\textwidth}
		\includegraphics[width=\linewidth]{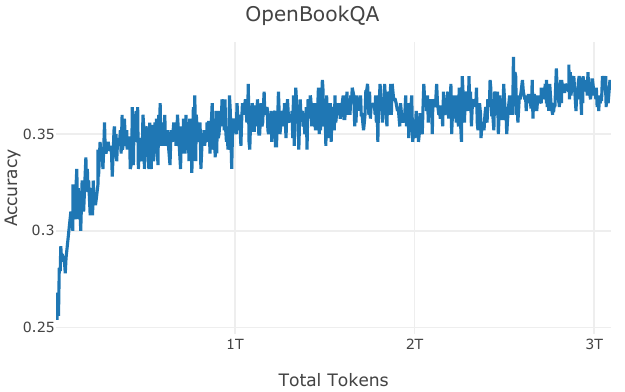}
	\end{subfigure}
	\quad
	\begin{subfigure}{0.31\textwidth}
		\includegraphics[width=\linewidth]{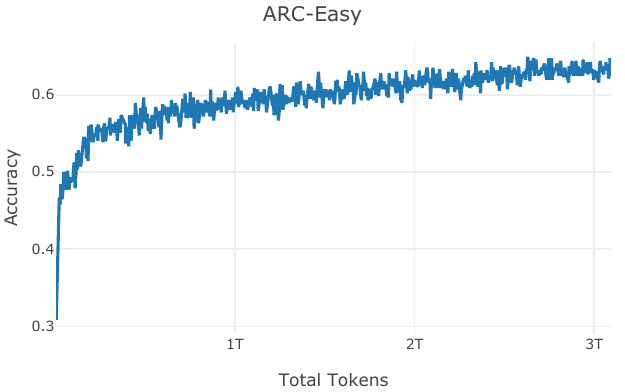}
	\end{subfigure}
	\quad
	\begin{subfigure}{0.31\textwidth}
		\includegraphics[width=\linewidth]{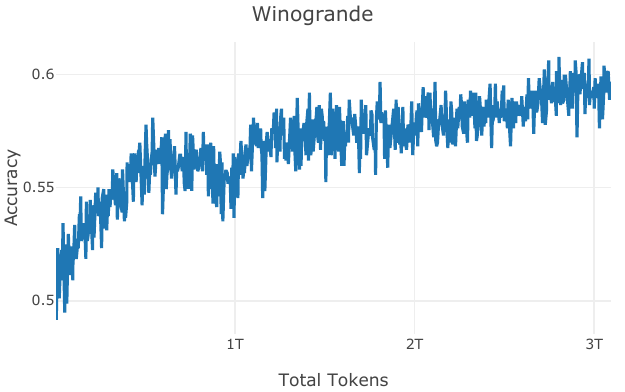}
	\end{subfigure}
	\caption{Results downstream tasks OpenBookQA~\citep{openbookQA}, ARC-E~\citep{arc}, and WinoGrande~\citep{winogrande}}
\end{figure}

\begin{figure}[h!]
	\centering
	\begin{subfigure}{0.31\textwidth}
		\includegraphics[width=\linewidth]{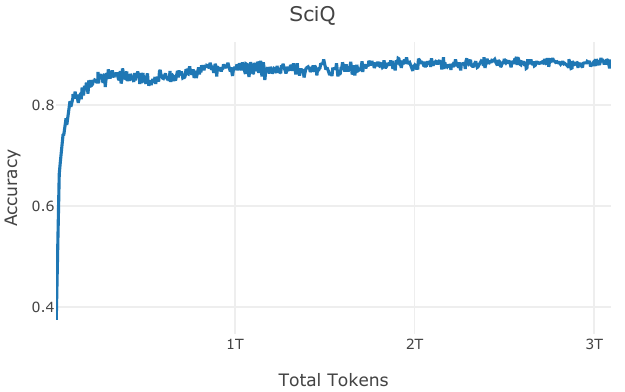}
	\end{subfigure}
	\quad
	\begin{subfigure}{0.31\textwidth}
		\includegraphics[width=\linewidth]{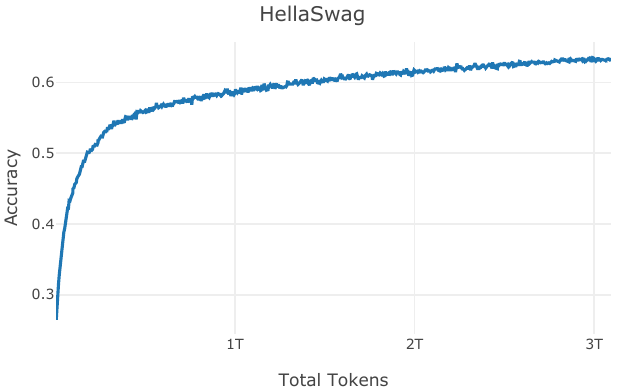}
	\end{subfigure}
	\quad
	\begin{subfigure}{0.31\textwidth}
		\includegraphics[width=\linewidth]{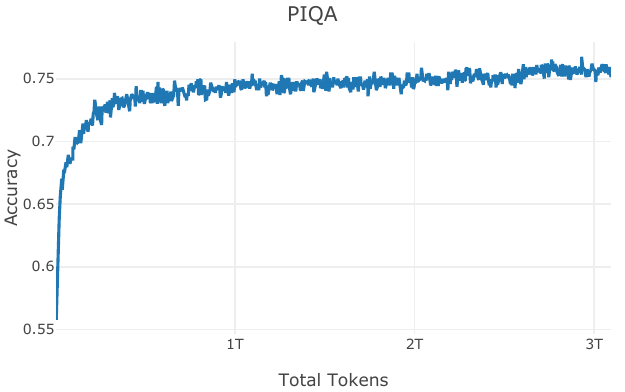}
	\end{subfigure}
	\caption{Results downstream tasks SciQ~\citep{sciq}, HellaSwag~\citep{zellers2019hellaswag}, and PIQA~\citep{piqa}}
\end{figure}

\label{sec:long_1b_run:train}

\begin{figure}[h!]
	\centering
	\begin{subfigure}{0.31\textwidth}
		\includegraphics[width=\linewidth]{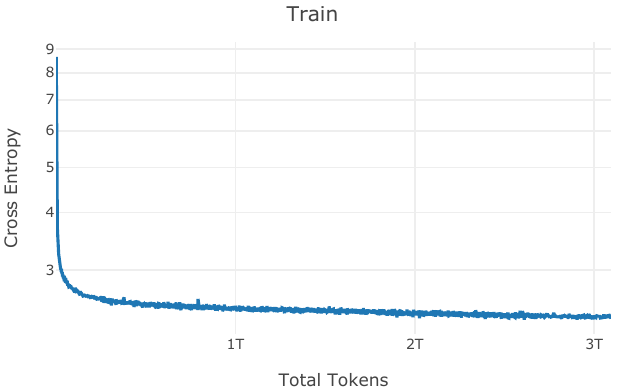}
	\end{subfigure}
	\caption{Training Cross Entropy}
\end{figure}

\FloatBarrier

\end{document}